\documentclass[]{lab}

\usepackage[utf8]{inputenc}
\usepackage[T1]{fontenc}
\usepackage{geometry}
\usepackage{amsmath,amssymb}  %
\usepackage{charter}

\usepackage[toc,page,header]{appendix}

\usepackage{minitoc}
\usepackage{cleveref} 
\usepackage{subcaption}
\usepackage{booktabs}
\usepackage{graphicx}
\usepackage{pgfplots}
\usepackage{pgfplotstable}
\usepackage{xcolor}
\usepackage{CJKutf8}
\usetikzlibrary{patterns}
\usepackage{float}
\usepackage{caption}
\usepackage{bm}

\usepackage{soul} %
\usepackage{algorithm}      %
\usepackage{algpseudocode}  %
\usepackage{parskip}

\definecolor{cvprblue}{rgb}{0.21,0.49,0.74}
\definecolor{fallbackgreen}{rgb}{130, 180, 102}
\definecolor{stopred}{rgb}{251, 225, 224}

\newcommand{\red}[1]{\textcolor{red}{#1}}

\ifdefined\final
\usepackage[disable]{todonotes}
\else
\usepackage[textsize=tiny]{todonotes}
\fi

\newcommand{\ours}{\texttt{TurboDiffusion}\xspace}

\usepackage{natbib}
\usepackage{latexsym}

\usepackage{url}
\usepackage{amssymb}
\usepackage[utf8]{inputenc}
\usepackage{microtype}
\usepackage{booktabs}
\usepackage{pifont} 
\usepackage{multirow}
\usepackage{makecell}
\usepackage{paralist}
\usepackage{xspace}
\usepackage{color}
\usepackage{xcolor}
\usepackage{colortbl}
\usepackage{adjustbox}
\usepackage{hyperref} 
\usepackage[edges]{forest}
\usepackage{tikz} 
\usepackage{caption}
\usepackage{amsfonts}

\hypersetup{
    colorlinks,
    linkcolor={blue!80!black},
    citecolor={blue!80!black},
}
\tikzset{
    root/.style =             {align=center, text width=1cm, rounded corners=3pt, line width=0.3mm, fill=gray!10, draw=gray!80, font=\small},
    demographic/.style =         {align=center, text width=1.8cm, rounded corners=3pt, line width=0.3mm, fill=blue!10, draw=blue!80, font=\footnotesize},
    demographic_work/.style =    {align=center, text width=10cm, rounded corners=3pt, line width=0.3mm, fill=blue!10, draw=blue!0, font=\footnotesize},
    character/.style =         {align=center, text width=1.8cm, rounded corners=3pt, line width=0.3mm, fill=red!10, draw=red!80, font=\footnotesize},
    character_work/.style =    {align=center, text width=10cm, rounded corners=3pt, line width=0.3mm, fill=red!10, draw=red!0, font=\footnotesize},
    personalization/.style =           {align=center, text width=1.8cm, rounded corners=3pt, line width=0.3mm, fill=cyan!10, draw=cyan!80, font=\footnotesize},
    personalization_work/.style =      {align=center, text width=10cm, rounded corners=3pt, line width=0.3mm, fill=cyan!10, draw=cyan!0, font=\footnotesize},
    risk/.style =         {align=center, text width=1.8cm, rounded corners=3pt, line width=0.3mm, fill=orange!10, draw=orange!80, font=\footnotesize},
    risk_work/.style =    {align=center, text width=10cm, rounded corners=3pt, line width=0.3mm, fill=orange!10, draw=orange!0, font=\footnotesize},
}

\usepackage{CJK}

\renewcommand{\thefootnote}{}

\newtcolorbox{promptbox}[1][]{
  enhanced,
  breakable,
  colback=promptboxlightgray,
  colframe=promptboxblue!30,
  arc=8pt,
  boxrule=0.5pt,
  left=12pt,
  right=12pt,
  top=8pt,
  bottom=8pt,
  fonttitle=\bfseries,
  fontupper=\linespread{1.2}\selectfont,
  title=#1
}

\title{TurboDiffusion: Accelerating Video Diffusion Models by 100--200 Times}

\author{Jintao Zhang$^{123*}$, Kaiwen Zheng$^{1*}$, Kai Jiang$^{12*}$, Haoxu Wang$^{1*}$, Ion Stoica$^{3}$, \\Joseph E. Gonzalez$^{3}$, Jianfei Chen$^1$, Jun Zhu$^{12\dagger}$}

\affiliation{$^1$Tsinghua University, $^2$Shengshu Technology, $^3$UC Berkeley, $^\dagger$Corresponding author\\\url{https://github.com/thu-ml/TurboDiffusion}}

\abstract{
We introduce \ours, a video generation acceleration framework that can speed up end-to-end diffusion generation by $100\text{–}200\times$ while maintaining video quality. \ours mainly relies on several components for acceleration: (1) \textbf{Attention acceleration}: \ours uses low-bit SageAttention and trainable Sparse-Linear Attention (SLA) to speed up attention computation. (2) \textbf{Step distillation}: \ours adopts rCM for efficient step distillation. (3) \textbf{W8A8 quantization}: \ours quantizes model parameters and activations to 8 bits to accelerate linear layers and compress the model. 

We conduct experiments on the \texttt{Wan2.2-I2V-A14B-720P}, \texttt{Wan2.1-T2V-1.3B-480P}, \texttt{Wan2.1-T2V-14B-720P}, and \texttt{Wan2.1-T2V-14B-480P} models. Experimental results show that \ours achieves $100\text{--}200\times$ speedup for video generation on a single RTX 5090 GPU, while maintaining comparable video quality. The GitHub repository, which contains model checkpoints, training, and inference code, is available at \url{https://github.com/thu-ml/TurboDiffusion}.
}

\begin{document}

\maketitle

\renewcommand{\thefootnote}{}
\footnotetext{*Core contributors, co-first authorship.}
\renewcommand{\thefootnote}{\arabic{footnote}}

\begin{figure}[H]
\centering
\begin{subfigure}{\textwidth}
\centering
\textbf{\large Original} ~~\textit{\large Latency: 184s}\\
\vspace{0.1cm}

\includegraphics[width=0.165\textwidth]{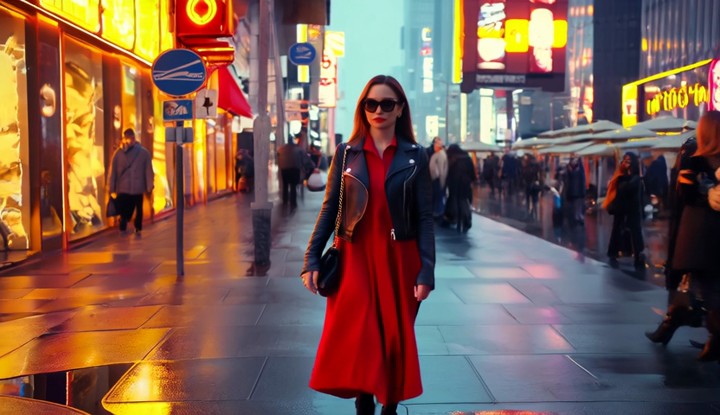}\hspace{-0.0037\textwidth}
\includegraphics[width=0.165\textwidth]{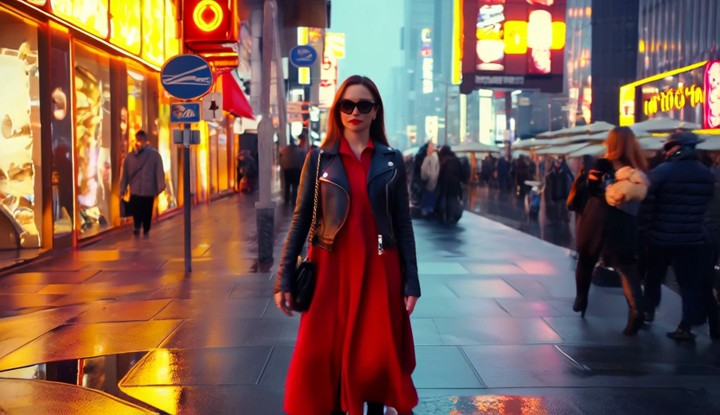}\hspace{-0.0037\textwidth}
\includegraphics[width=0.165\textwidth]{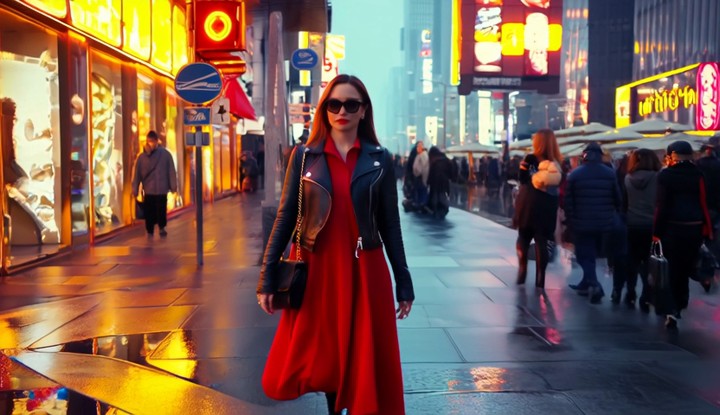}\hspace{-0.0037\textwidth}
\includegraphics[width=0.165\textwidth]{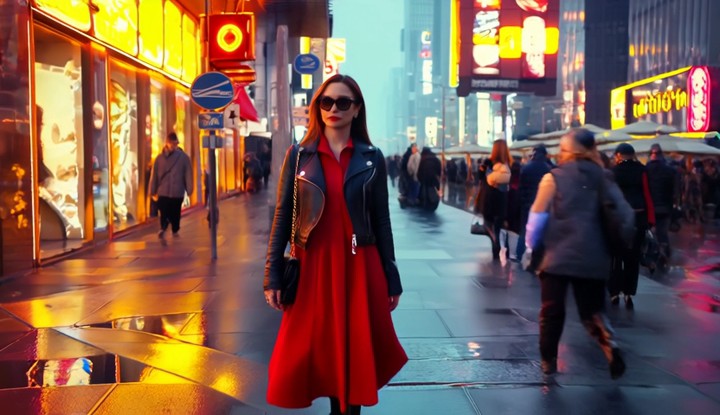}\hspace{-0.0037\textwidth}
\includegraphics[width=0.165\textwidth]{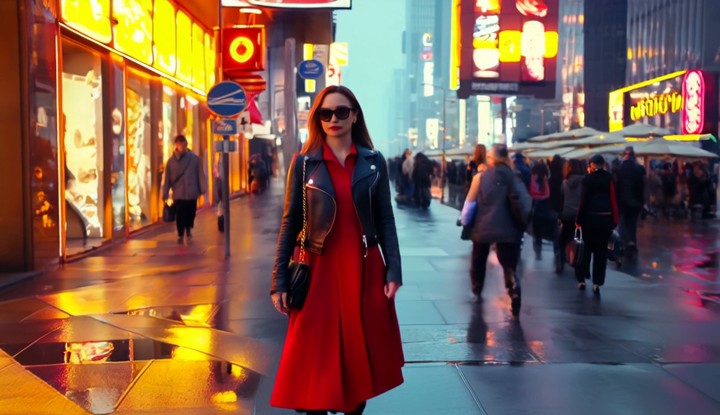}\hspace{-0.0037\textwidth}
\includegraphics[width=0.165\textwidth]{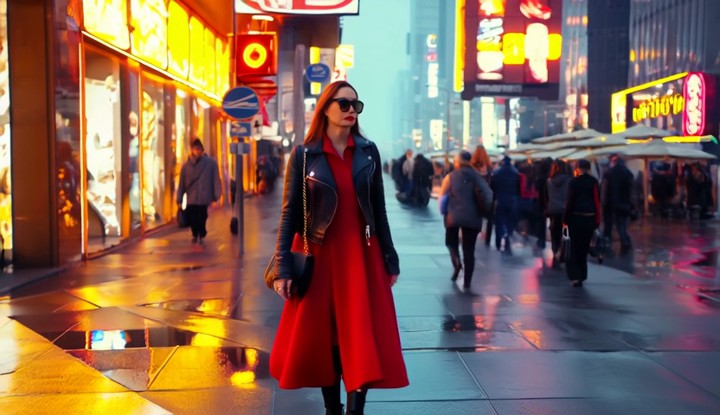}
\vspace{0.15cm}

\end{subfigure}

\vspace{0.2cm}

\begin{subfigure}{\textwidth}
\centering
\textbf{\large TurboDiffusion} ~~\textit{\large Latency: \bf \red{1.9s}}\\
\vspace{0.1cm}

\includegraphics[width=0.165\textwidth]{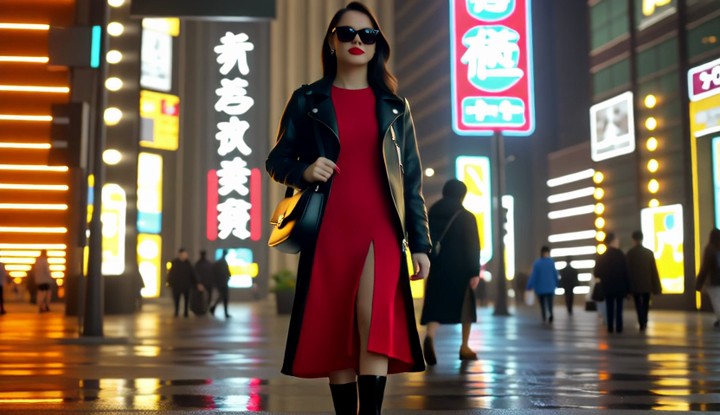}\hspace{-0.0037\textwidth}
\includegraphics[width=0.165\textwidth]{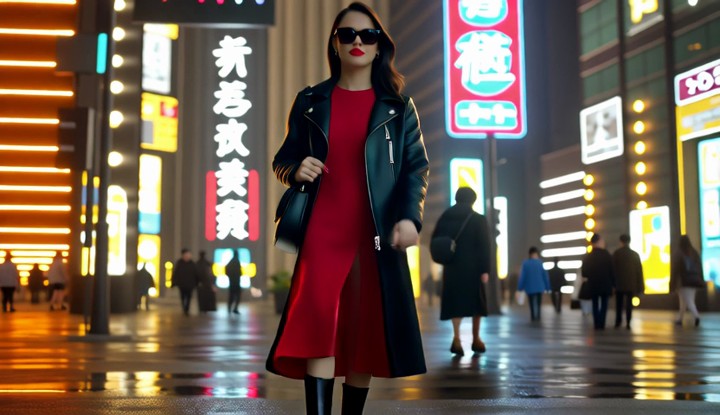}\hspace{-0.0037\textwidth}
\includegraphics[width=0.165\textwidth]{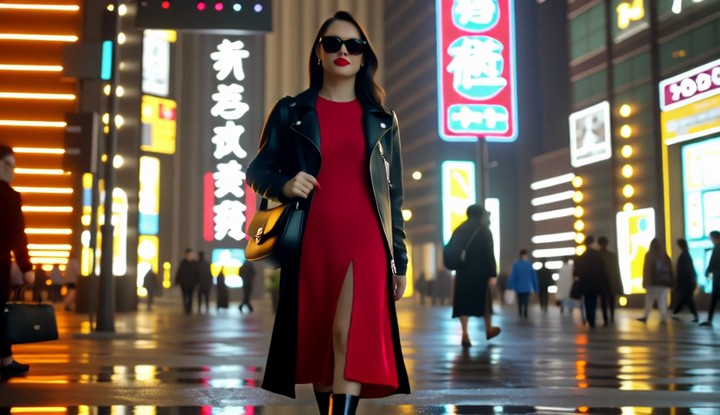}\hspace{-0.0037\textwidth}
\includegraphics[width=0.165\textwidth]{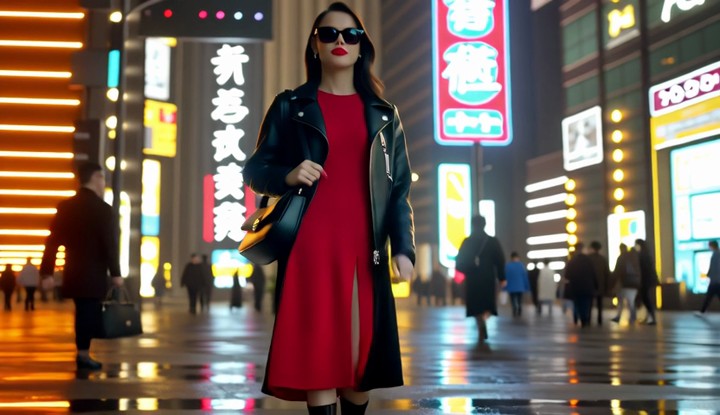}\hspace{-0.0037\textwidth}
\includegraphics[width=0.165\textwidth]{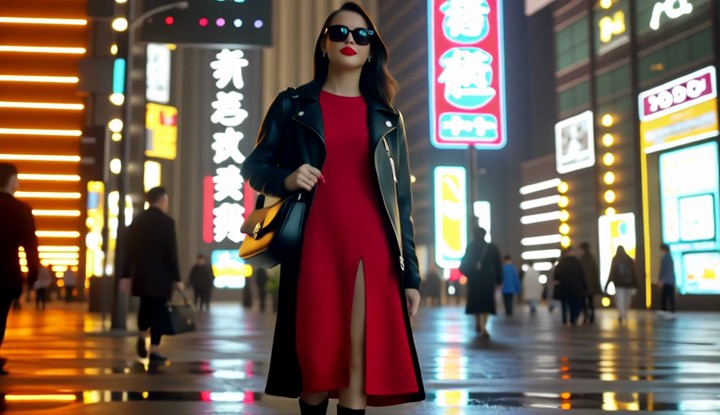}\hspace{-0.0037\textwidth}
\includegraphics[width=0.165\textwidth]{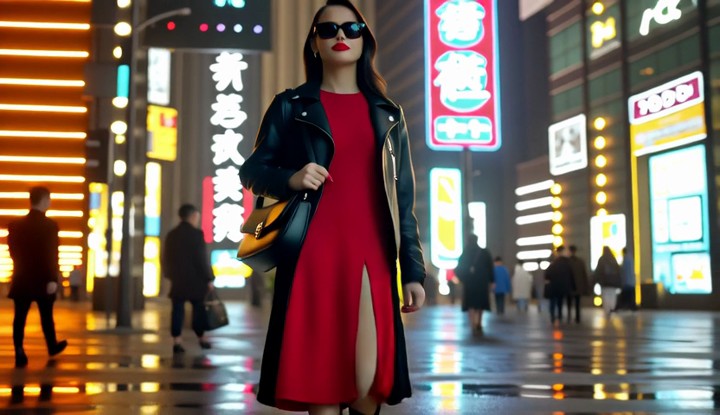}
\vspace{0.15cm}

\end{subfigure}

\vspace{-1em}
\caption{An example of a \textbf{5-second video} generated by \texttt{Wan2.1-T2V-1.3B-480P} \textbf{\red{on a single RTX 5090}}.\vspace{2em}}
\label{fig:intro}
\end{figure}

\begin{figure}[H]
\vspace{-1em}
\centering
\begin{subfigure}{\textwidth}
\centering
\textbf{\large Original} ~~\textit{\large Latency: 4549s}\\
\vspace{0.1cm}

\includegraphics[width=0.165\textwidth]{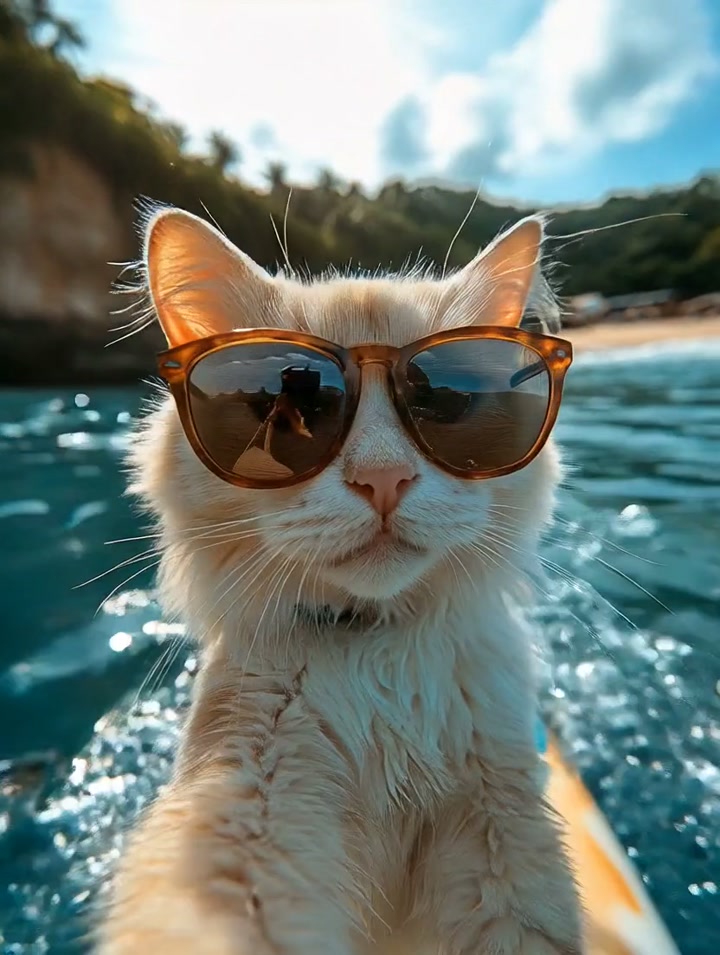}\hspace{-0.0037\textwidth}
\includegraphics[width=0.165\textwidth]{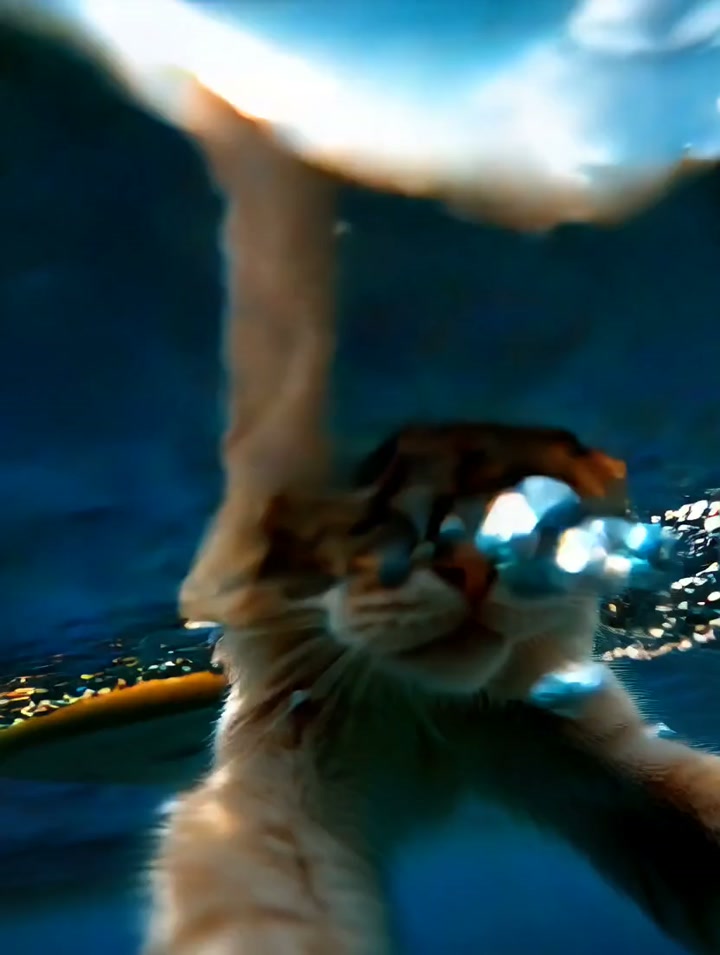}\hspace{-0.0037\textwidth}
\includegraphics[width=0.165\textwidth]{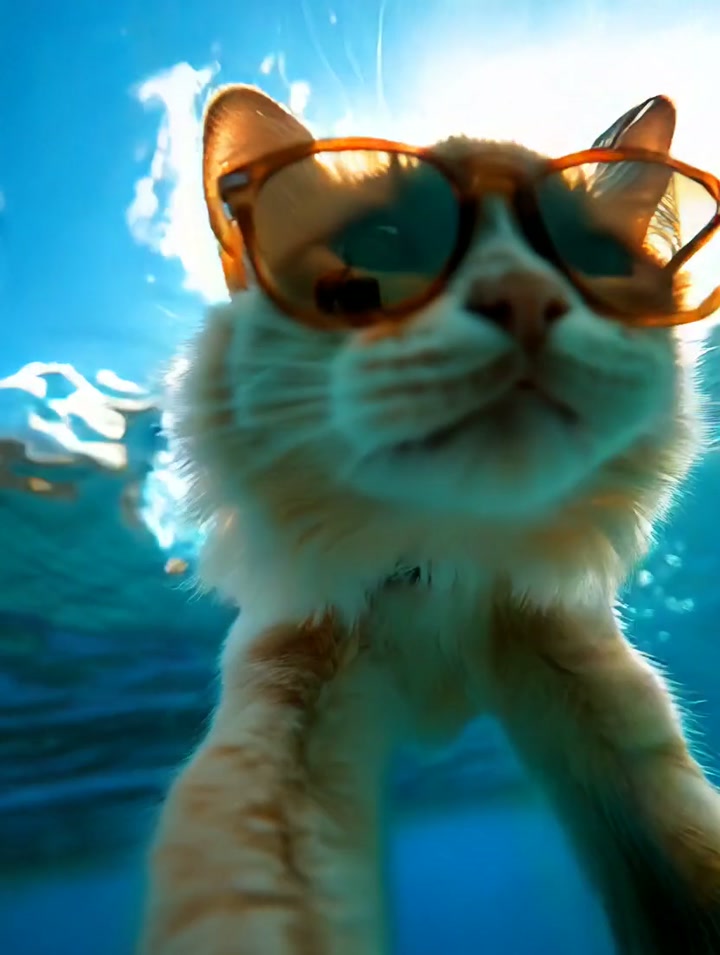}\hspace{-0.0037\textwidth}
\includegraphics[width=0.165\textwidth]{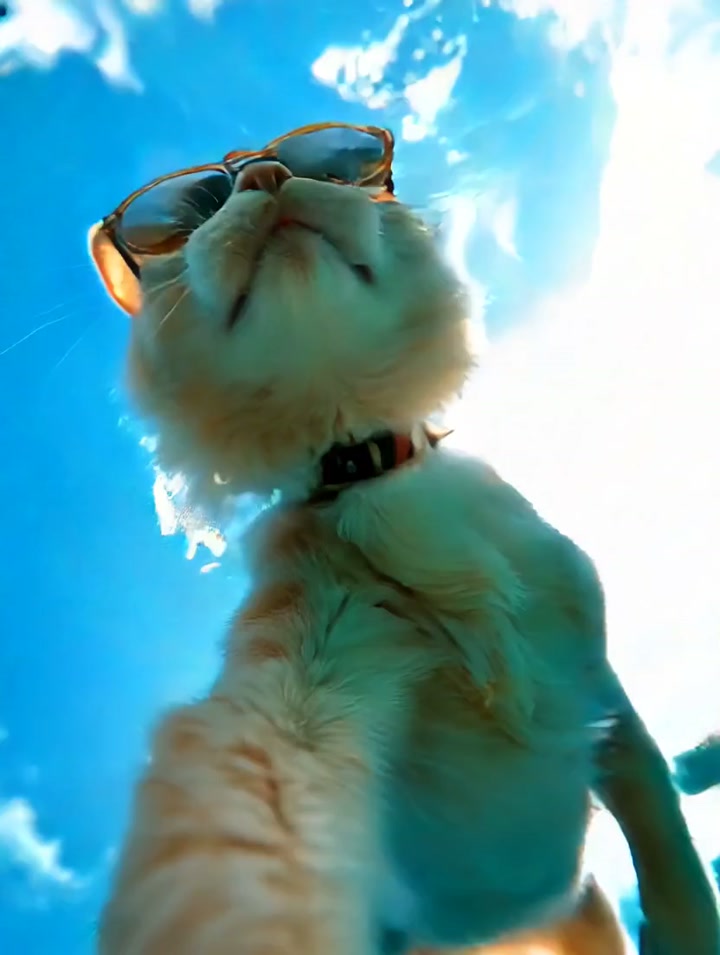}\hspace{-0.0037\textwidth}
\includegraphics[width=0.165\textwidth]{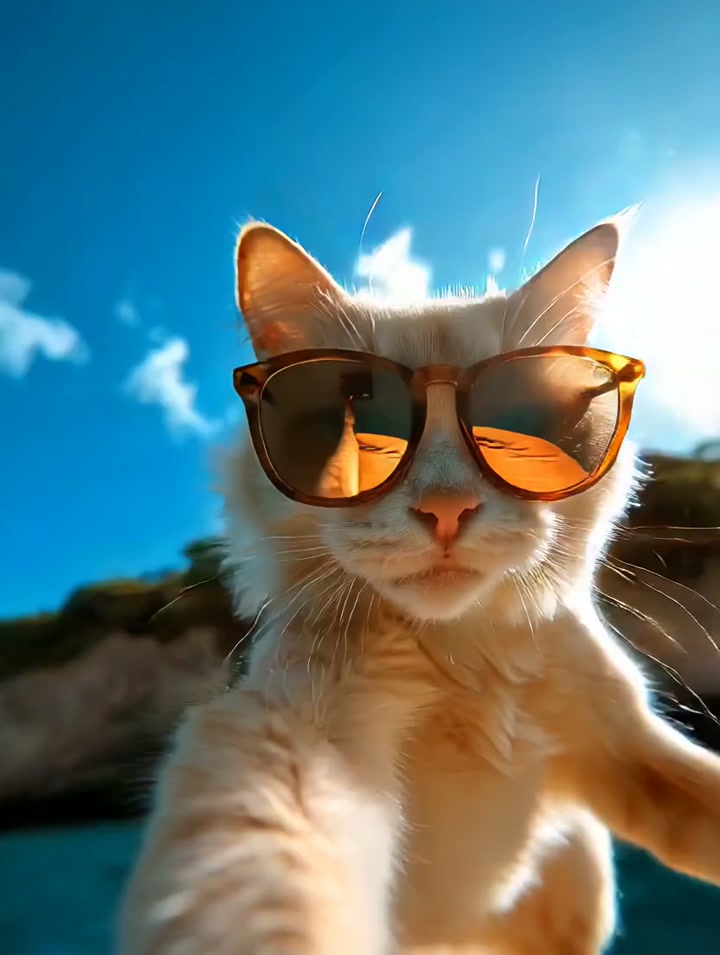}\hspace{-0.0037\textwidth}
\includegraphics[width=0.165\textwidth]{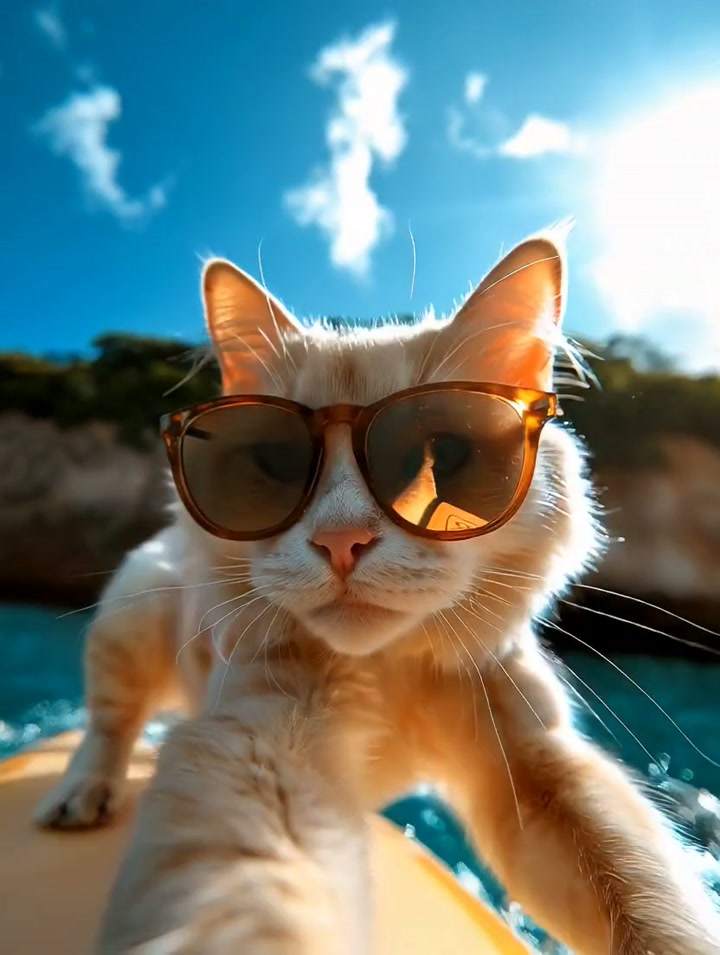}
\vspace{-0.5em}

\end{subfigure}

\vspace{0.15cm}

\begin{subfigure}{\textwidth}
\centering
\textbf{\large TurboDiffusion} ~~\textit{\large Latency: \bf \red{38s}}\\
\vspace{0.1cm}

\includegraphics[width=0.165\textwidth]{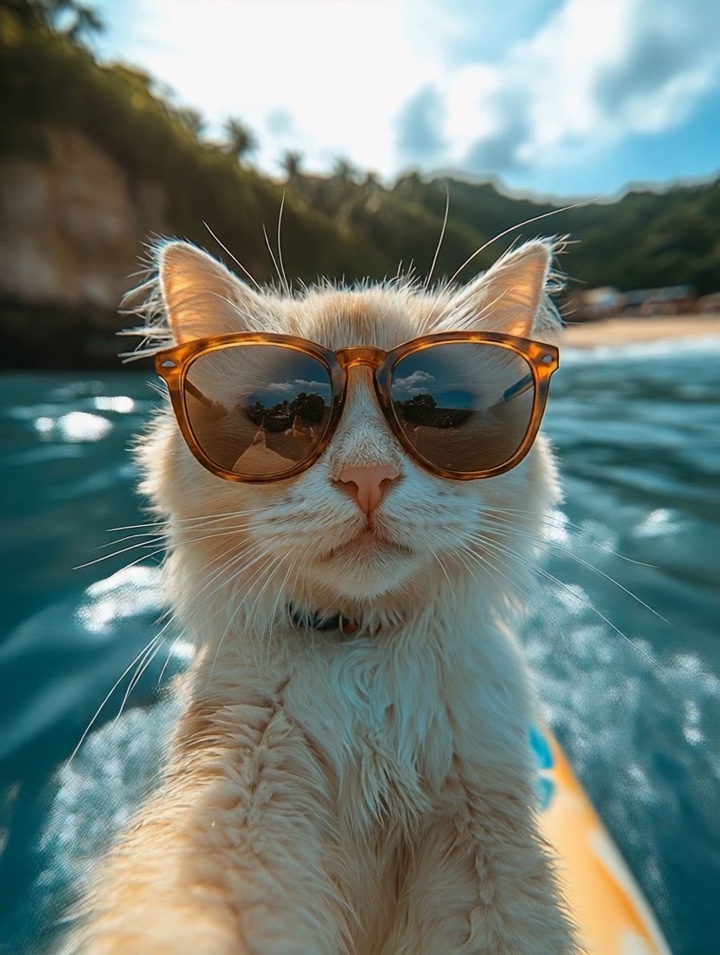}\hspace{-0.0037\textwidth}
\includegraphics[width=0.165\textwidth]{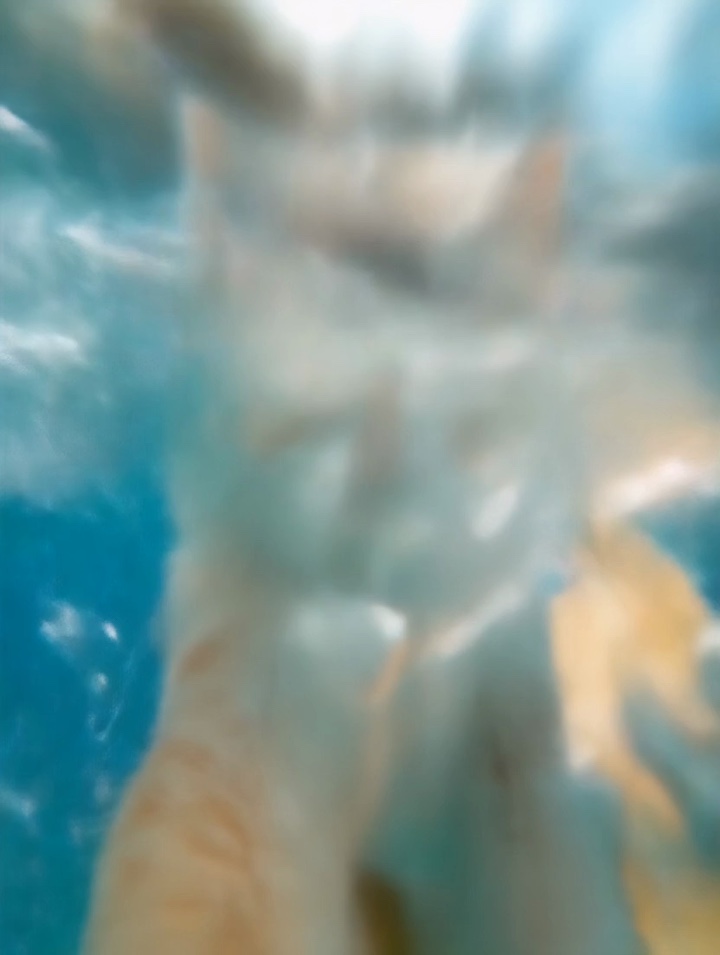}\hspace{-0.0037\textwidth}
\includegraphics[width=0.165\textwidth]{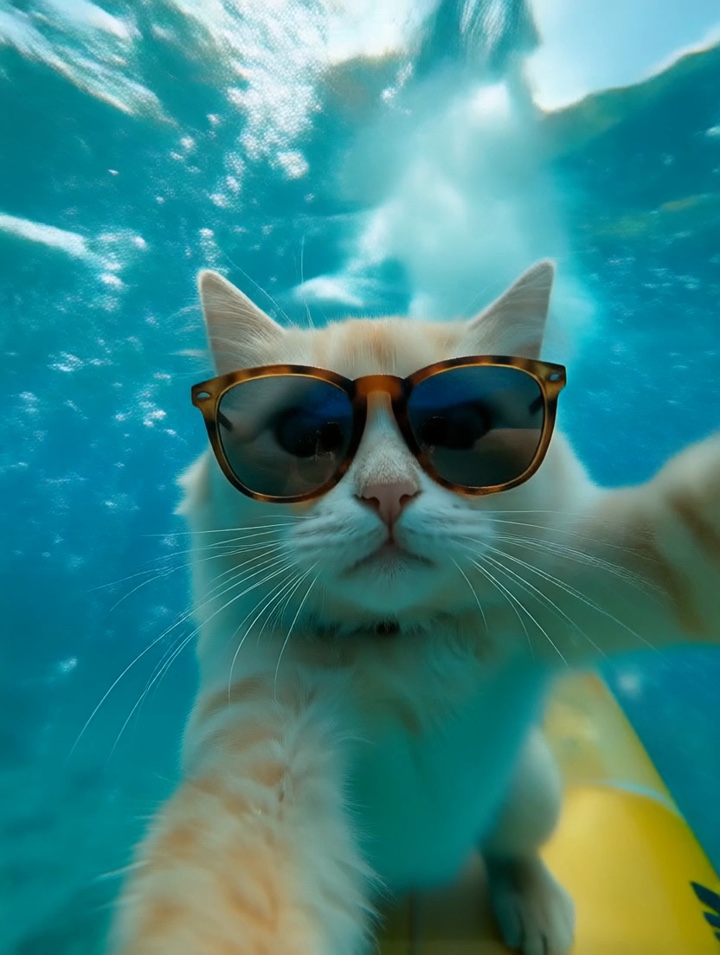}\hspace{-0.0037\textwidth}
\includegraphics[width=0.165\textwidth]{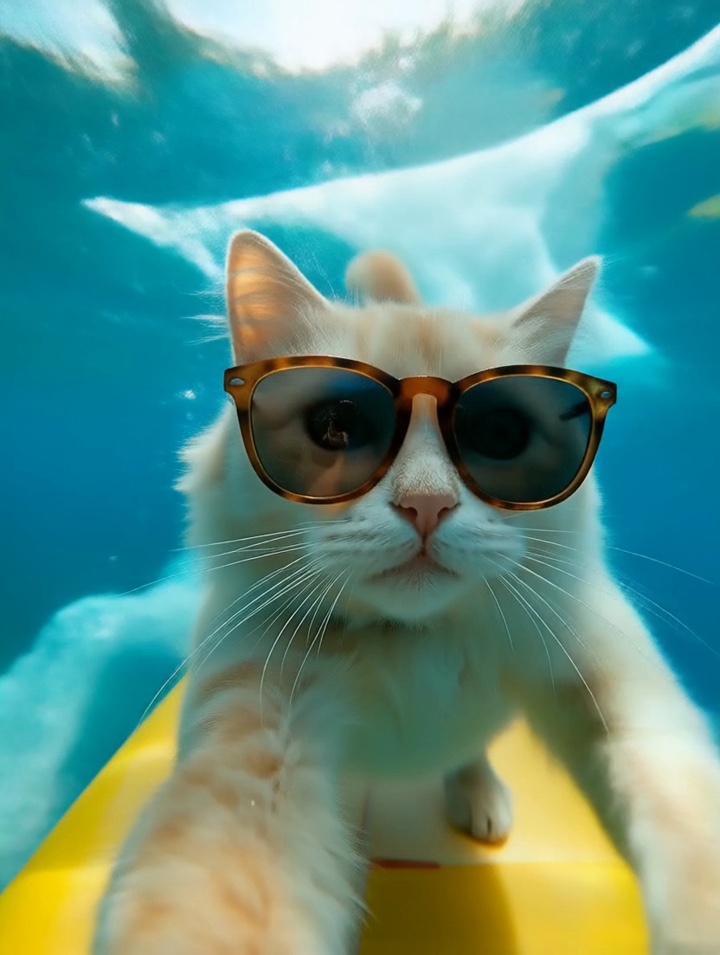}\hspace{-0.0037\textwidth}
\includegraphics[width=0.165\textwidth]{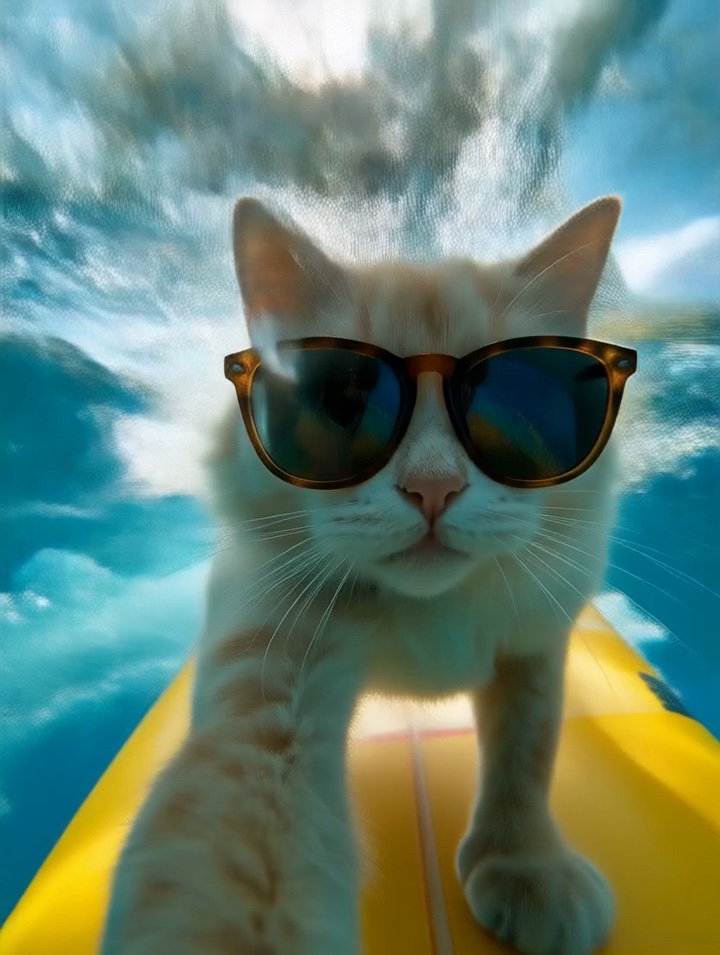}\hspace{-0.0037\textwidth}
\includegraphics[width=0.165\textwidth]{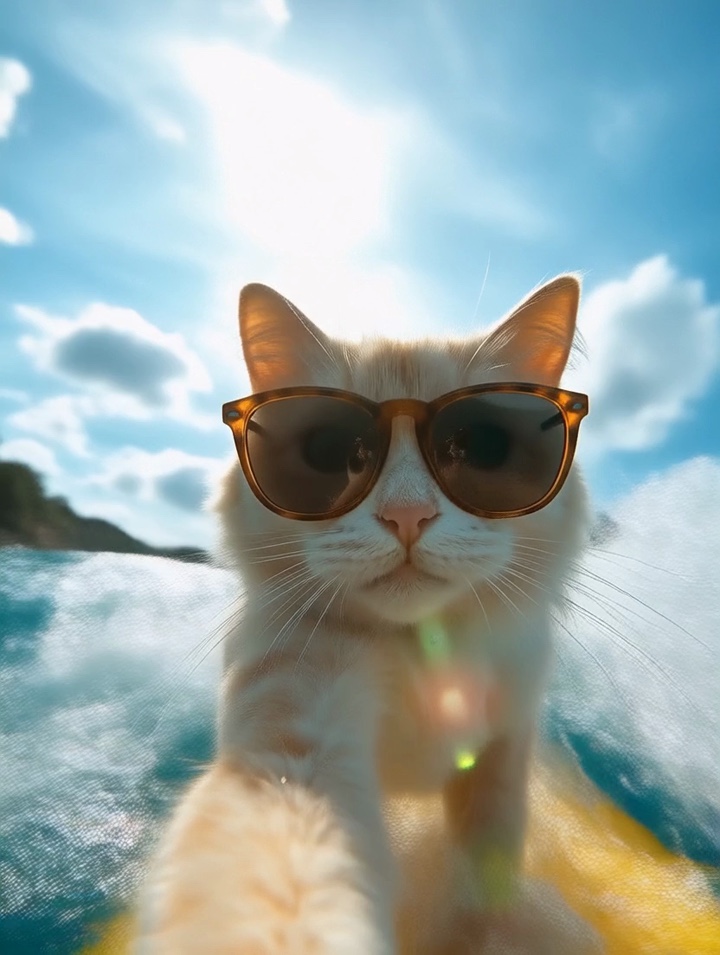}
\vspace{-0.5em}

\end{subfigure}
\vspace{-2em}
\caption{An example of a 5-second video generation on \texttt{Wan2.2-I2V-A14B-720P} \textbf{\red{using a single RTX 5090}}.}
\label{fig:intro2}
\end{figure}

\begin{figure}[H]
\vspace{-.5em}
\includegraphics[width=0.93\textwidth]{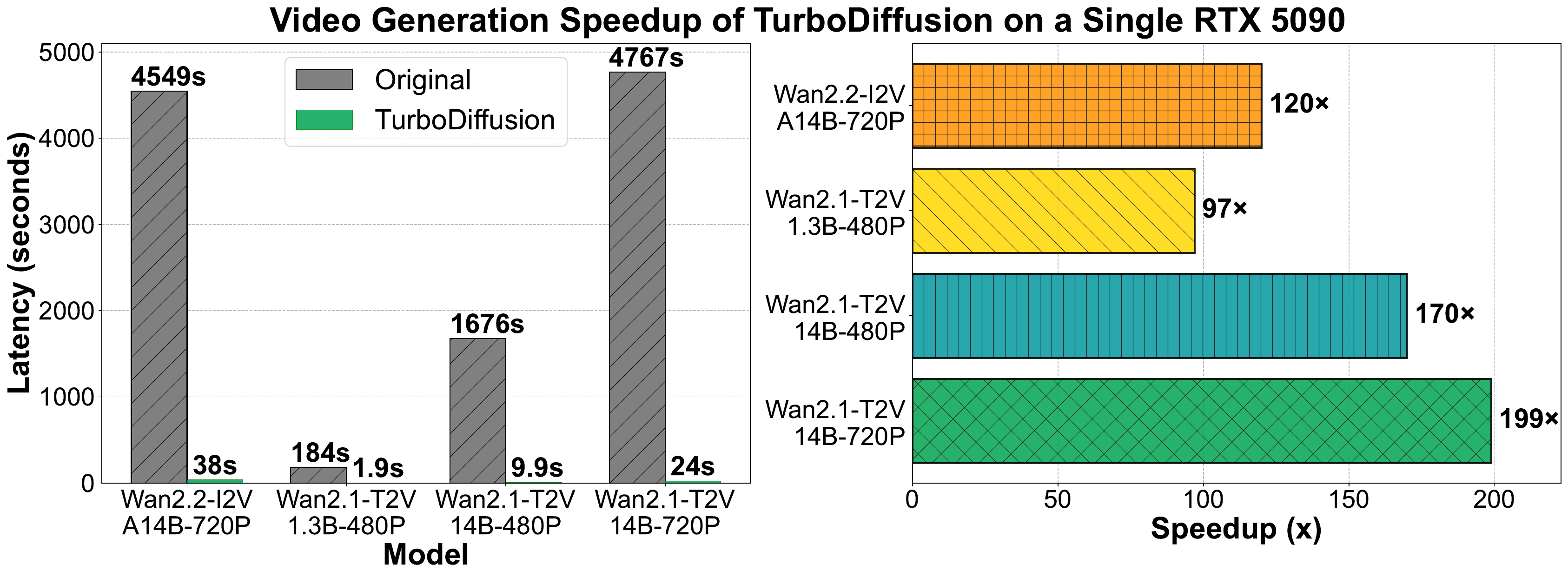}
\vspace{-.75em}
\caption{Speedup of \ours on various video generation models \textbf{\red{on a single RTX 5090}}.\\\textit{For {Wan2.2-I2V-A14B-720P}, the latency includes the switching overhead between the high-noise and low-noise models, resulting in a lower measured speedup compared to {Wan2.1-T2V-14B-720P}. In theory, the achievable speedup is identical.}}
\label{fig:all_speedup}
\end{figure}

\begin{figure}[H]
\vspace{-.5em}
\centering
\includegraphics[width=0.83\textwidth]{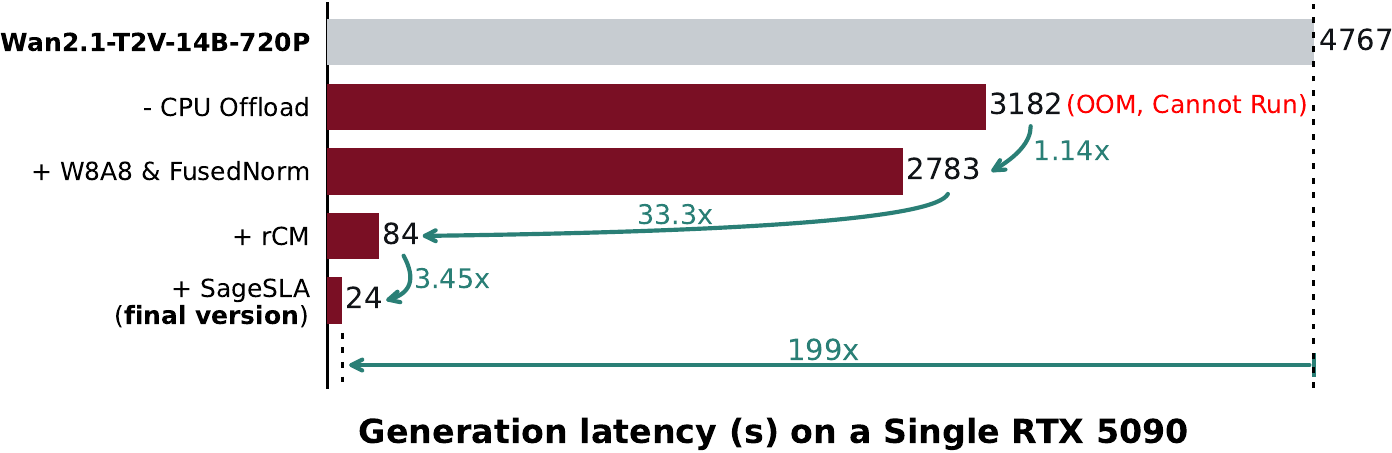}
\vspace{-.5em}
\caption{By algorithm and system co-optimization, \ours reduces the diffusion inference latency of \texttt{Wan2.1-T2V-14B-720P} by \textbf{around 200$\times$} \textbf{\red{on a single RTX 5090}.}}
\label{fig:latency_decompose}
\end{figure}

\section{Method}

We first describe the main techniques used in \ours in Section~\ref{sec:main_tech}. Then we introduce the training process and inference-time acceleration details of \ours in Section~\ref{sec:training} and Section~\ref{sec:inference}.

\subsection{Main Techniques} \label{sec:main_tech}

\ours mainly uses four techniques to accelerate diffusion models.
\uline{First}, \ours uses SageAttention~\cite{zhang2025sageattention,zhang2024sageattention2,zhang2025sageattention2++,zhang2025sageattention3} for low-bit quantized attention acceleration; specifically, \ours uses the SageAttention2++~\cite{zhang2025sageattention2++} variant.
\uline{Second}, \ours uses Sparse-Linear Attention (SLA)~\cite{zhang2025sla} for sparse attention acceleration. Since sparse computation is orthogonal to low-bit Tensor Core acceleration, SLA can build on top of SageAttention to provide cumulative speedup.
\uline{Third}, \ours uses rCM~\cite{zheng2025rcm} to reduce the number of sampling steps, which is currently a state-of-the-art diffusion distillation method. Through model weights merging, rCM naturally inherits attention-level accelerations.
\uline{Finally}, \ours uses W8A8 quantization for Linear layer acceleration. Specifically, the data type is INT8 and the quantization granularity is block-wise with a block size of $128 \times 128$.

\subsection{Training} \label{sec:training}

Given a pretrained video diffusion model, \ours performs the following training process.
First, we replace full attention with Sparse-Linear Attention (SLA)~\cite{zhang2025sla} and finetune the pretrained model to adapt to sparsity. In parallel, we use rCM~\cite{zheng2025rcm} to distill the pretrained model into a student model with fewer sampling steps. 
Second, we merge the parameter updates from both the SLA finetuning and the rCM training into a single model.
All training can utilize either real or synthetic data.

Please see our \href{https://github.com/thu-ml/TurboDiffusion}{GitHub Code} for more details.

\subsection{Inference} \label{sec:inference}

Given a video generation model trained with Sparse-Linear Attention (SLA)~\cite{zhang2025sla} and rCM~\cite{zheng2025rcm} as described in the previous section, we deploy inference-time acceleration as follows.

\textbf{\uline{Attention acceleration.}} We replace SLA with SageSLA, which is a CUDA implementation of SLA built on top of SageAttention.  

\textbf{\uline{Step distillation.}} We reduce the number of sampling steps from 100 to a much smaller value, e.g., $4$ or $3$.

\textbf{\uline{Linear layer quantization.}} First, we quantize the Linear layer parameters to INT8 with a block-wise granularity of $128 \times 128$. Second, during inference, we also quantize the activations in Linear layers to INT8 with the same block-wise granularity and use INT8 Tensor Cores to perform the Linear layer computation. In this way, we compress the model size by roughly half and achieve faster Linear layer computation. 

\textbf{\uline{Other optimizations.}} We reimplement several other operations, such as LayerNorm and RMSNorm, using Triton or CUDA for better efficiency.

Please see our \href{https://github.com/thu-ml/TurboDiffusion}{GitHub Code} for more details.
\section{Evaluations}

In this section, we evaluate the efficiency and video quality of \ours.

\subsection{Setup}

\textbf{\uline{Models and baseline.}} 
We evaluate \ours on \texttt{Wan2.2-I2V-A14B-720P}, \texttt{Wan2.1-T2V-1.3B-480P},\\ \texttt{Wan2.1-T2V-14B-720P}, and \texttt{Wan2.1-T2V-14B-480P} video diffusion models. We use the official implementation of Wan~\cite{wan2025} (denoted as \texttt{Original}) and \texttt{FastVideo}~\cite{fastvideo2024} as our main baselines.

\textbf{\uline{Hyper-parameters.}} 
We set the Top-K ratio to $0.1$, corresponding to $90\%$ attention sparsity, and use $3$ sampling steps. In practice, we recommend using a Top-K value in the range $[0.1, 0.15]$ and setting the number of steps to $4$ to consistently achieve the best video quality. For fastvideo, we use the default parameters in the official implementation ($3$ sampling steps and 0.8 sparsity in attention).

\textbf{\uline{GPU.}} 
Our primary inference experiments are conducted on a single RTX 5090 GPU. In addition, although the speedup is not as large as on the RTX 5090, we also observe substantial acceleration on other GPUs, such as RTX 4090 and H100.

\subsection{Efficiency and Quality}

We compare the video generation quality and efficiency of the \texttt{Original}, \texttt{FastVideo}, and \ours. For efficiency evaluation, we report the end-to-end diffusion generation latency, excluding the text encoding and VAE decoding stages. The following figures present visual comparisons on \texttt{Wan2.2-I2V-A14B-720P}, \texttt{Wan2.1-T2V-1.3B-480P}, \texttt{Wan2.1-T2V-14B-720P}, and \texttt{Wan2.1-T2V-14B-480P}, respectively. Since \texttt{FastVideo} does not provide an accelerated \texttt{Wan2.2-A14B-I2V-720P}, we only compare \ours with \texttt{Original} on \texttt{Wan2.2-A14B-I2V-720P}.

From the figures below, we can see that \ours not only achieves the highest efficiency but also maintains the video quality, demonstrating clear superiority to \texttt{FastVideo}.

\subsubsection{Wan2.2-I2V-A14B-720P}

\begin{figure}[H]
\centering
\begin{subfigure}{\textwidth}
\centering
\textbf{\large Original} ~~\textit{\large Latency: 4549s}\\
\vspace{0.1cm}

\includegraphics[width=0.165\textwidth]{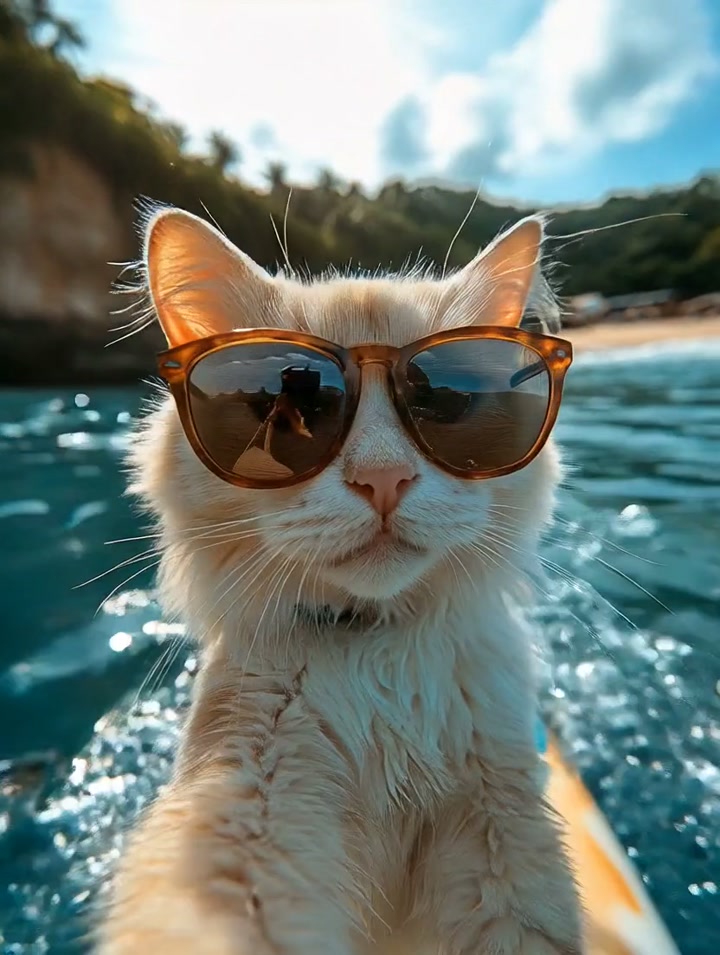}\hspace{-0.0037\textwidth}
\includegraphics[width=0.165\textwidth]{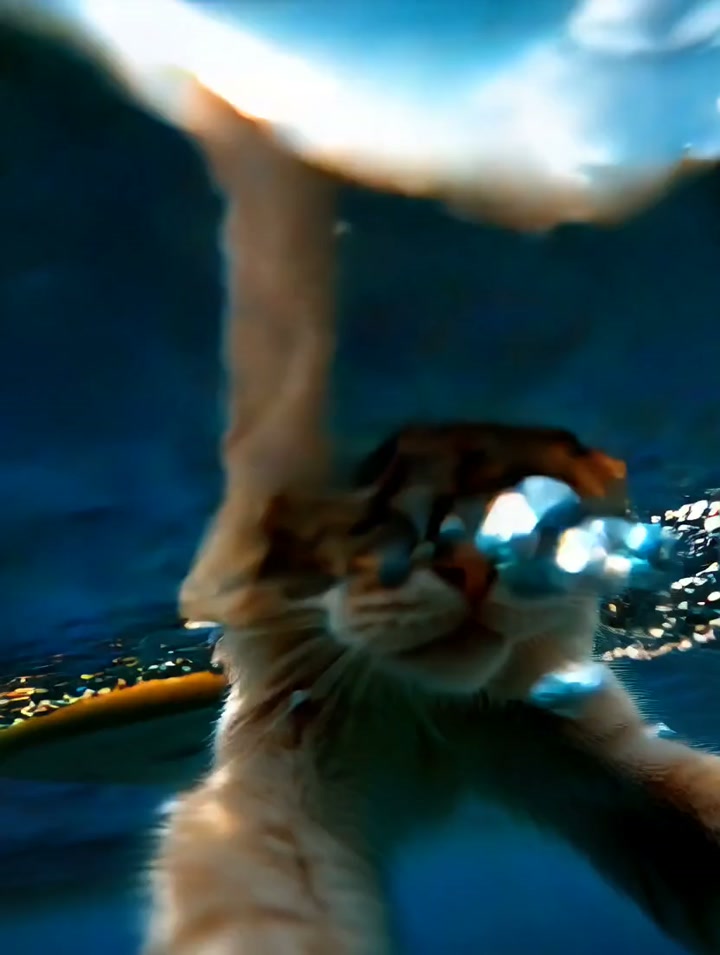}\hspace{-0.0037\textwidth}
\includegraphics[width=0.165\textwidth]{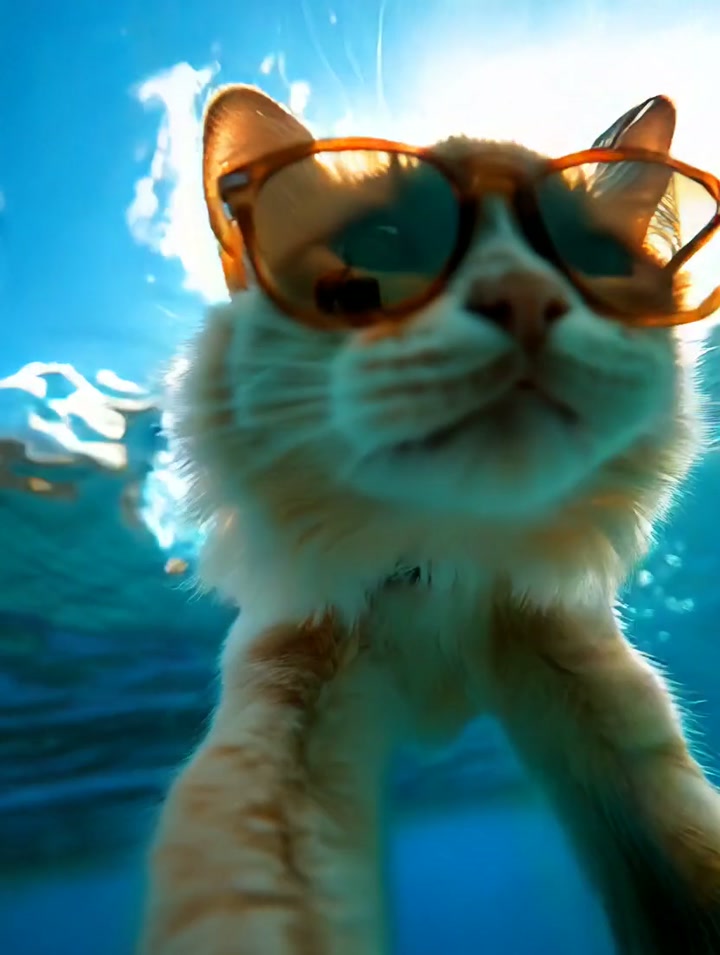}\hspace{-0.0037\textwidth}
\includegraphics[width=0.165\textwidth]{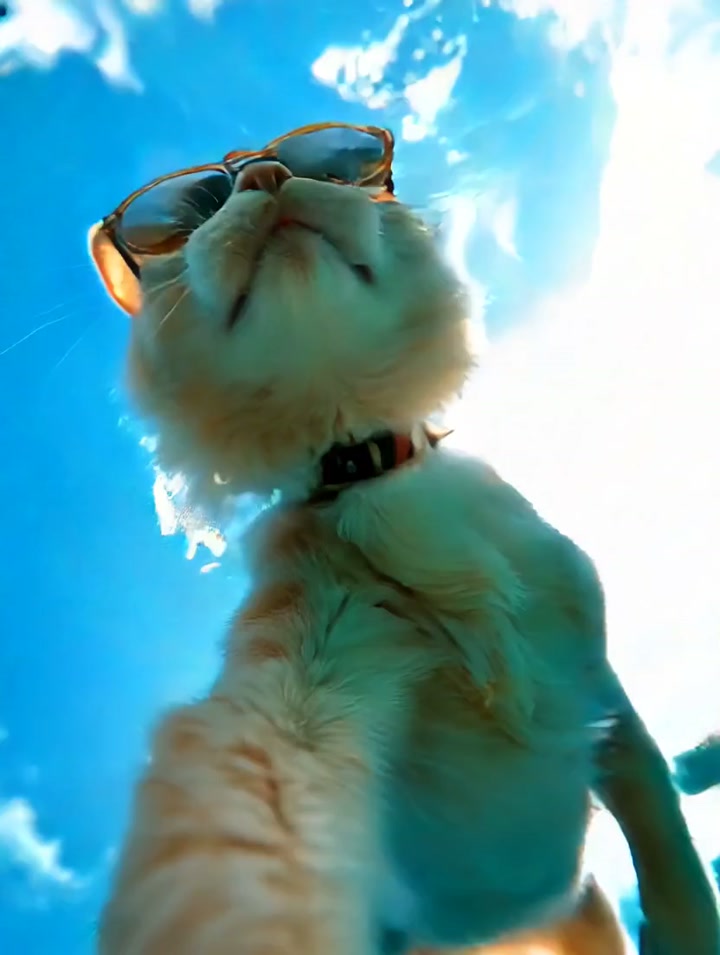}\hspace{-0.0037\textwidth}
\includegraphics[width=0.165\textwidth]{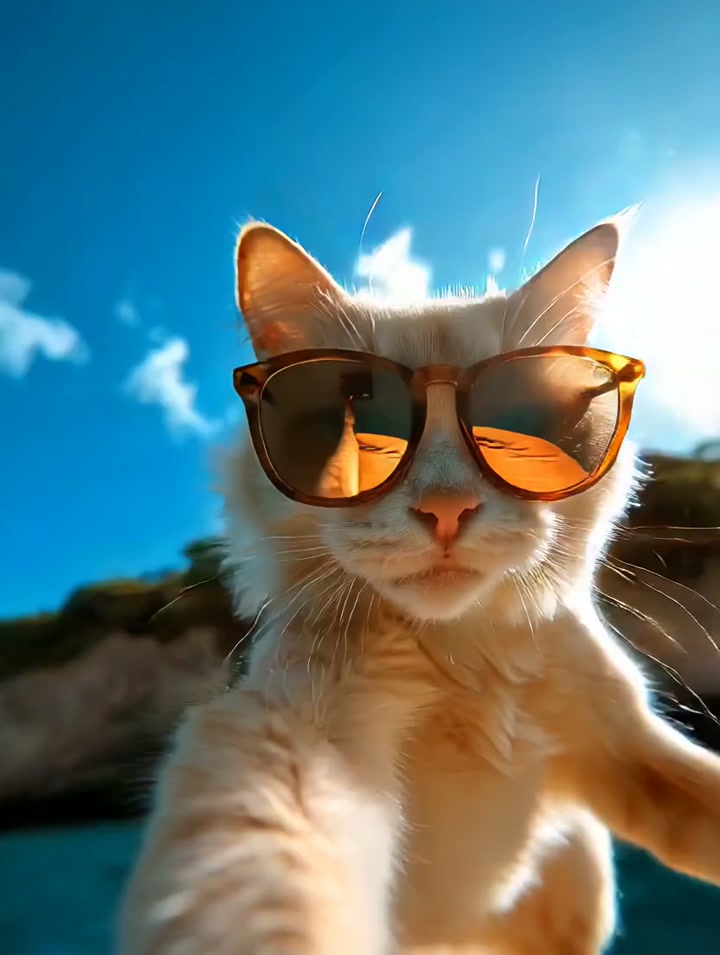}\hspace{-0.0037\textwidth}
\includegraphics[width=0.165\textwidth]{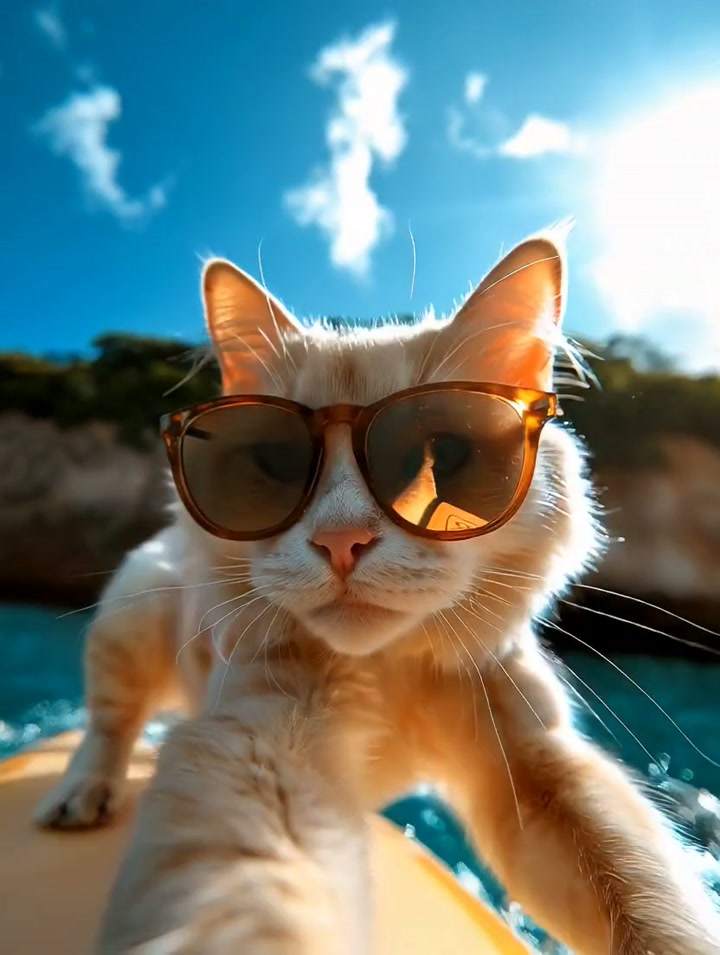}
\vspace{-0.5em}

\end{subfigure}

\vspace{0.2cm}

\begin{subfigure}{\textwidth}
\centering
\textbf{\large TurboDiffusion} ~~\textit{\large Latency: \bf \red{38s}}\\
\vspace{0.1cm}

\includegraphics[width=0.165\textwidth]{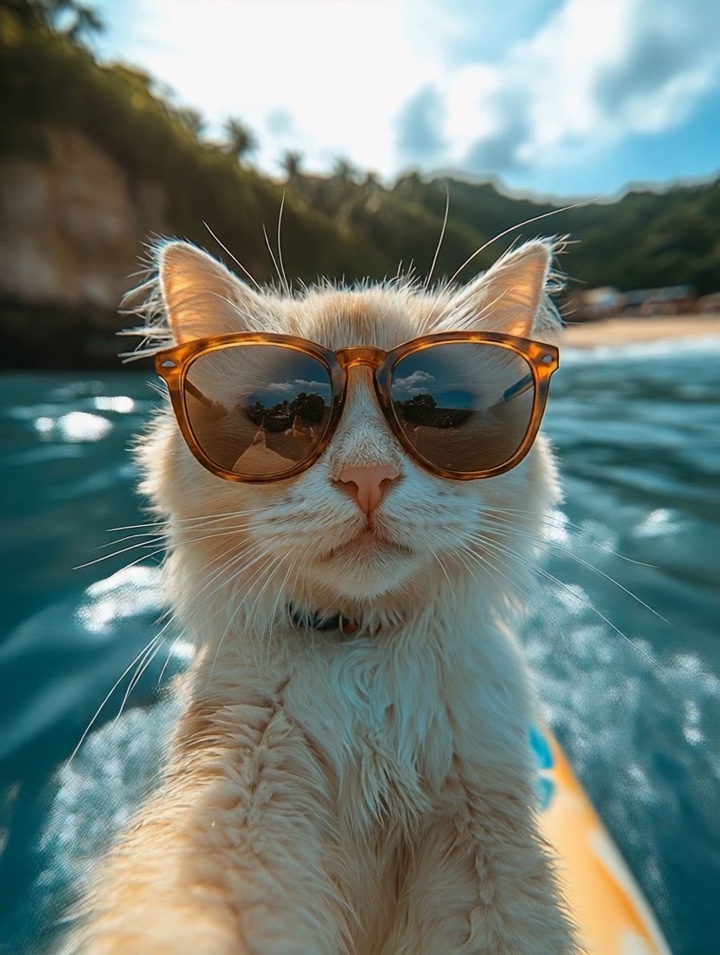}\hspace{-0.0037\textwidth}
\includegraphics[width=0.165\textwidth]{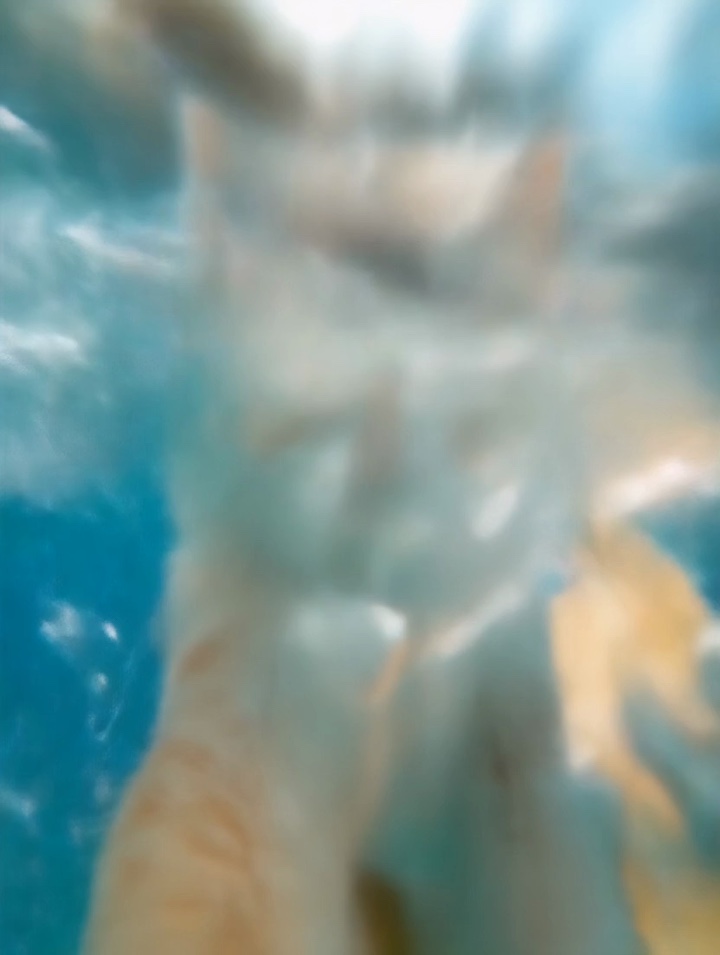}\hspace{-0.0037\textwidth}
\includegraphics[width=0.165\textwidth]{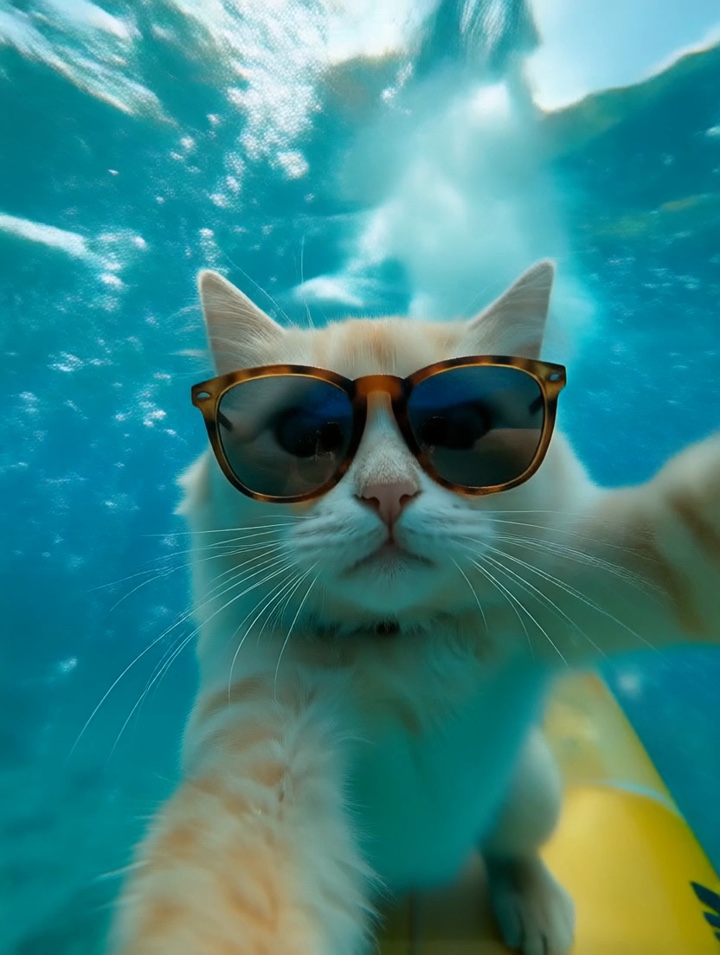}\hspace{-0.0037\textwidth}
\includegraphics[width=0.165\textwidth]{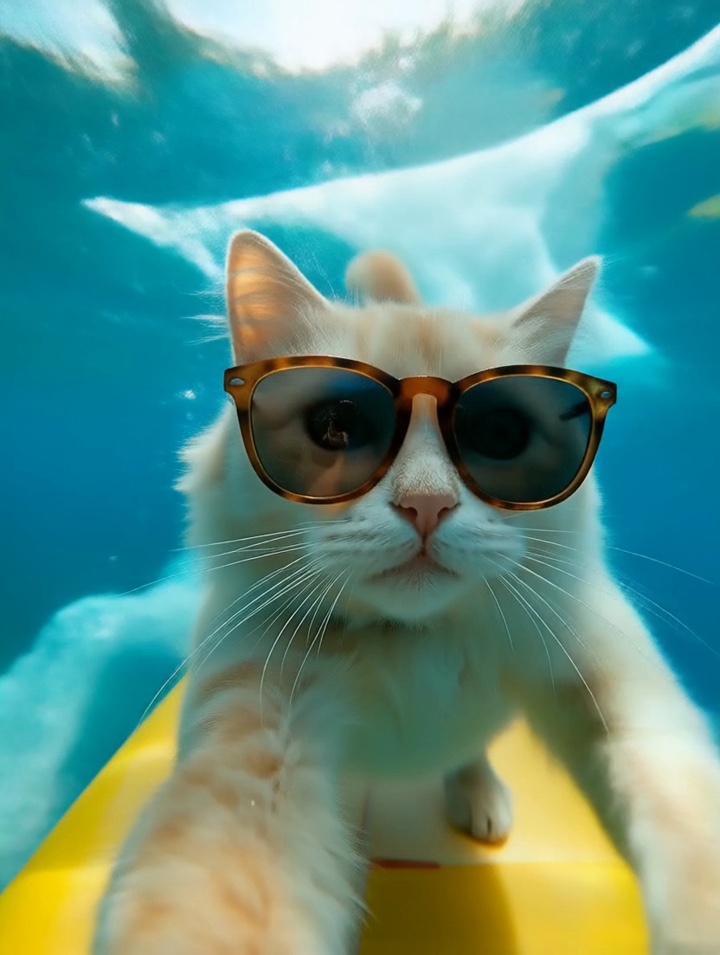}\hspace{-0.0037\textwidth}
\includegraphics[width=0.165\textwidth]{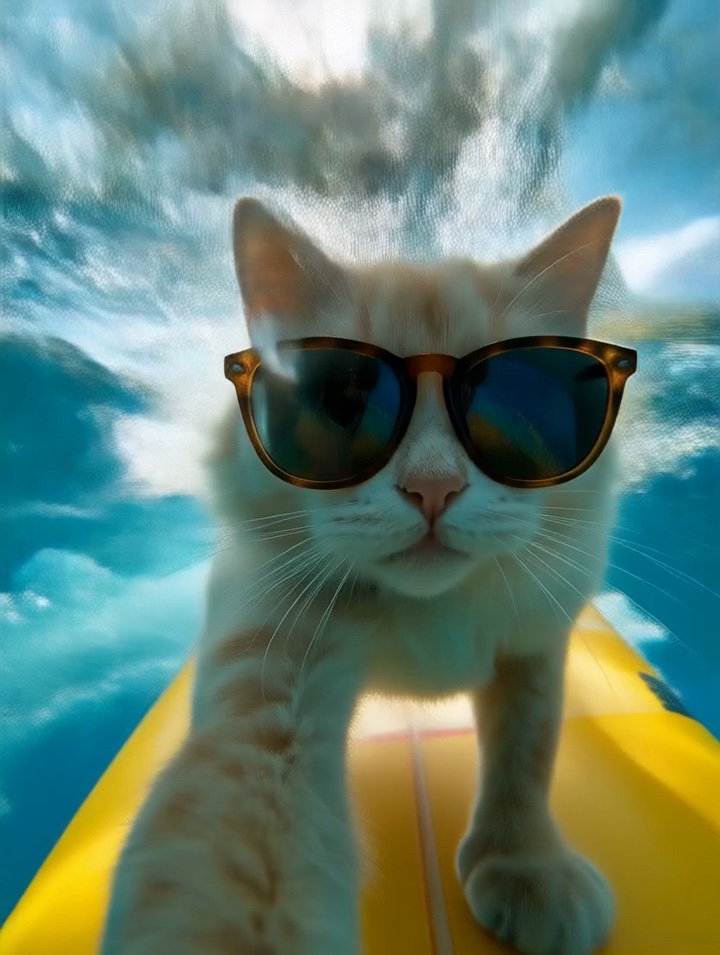}\hspace{-0.0037\textwidth}
\includegraphics[width=0.165\textwidth]{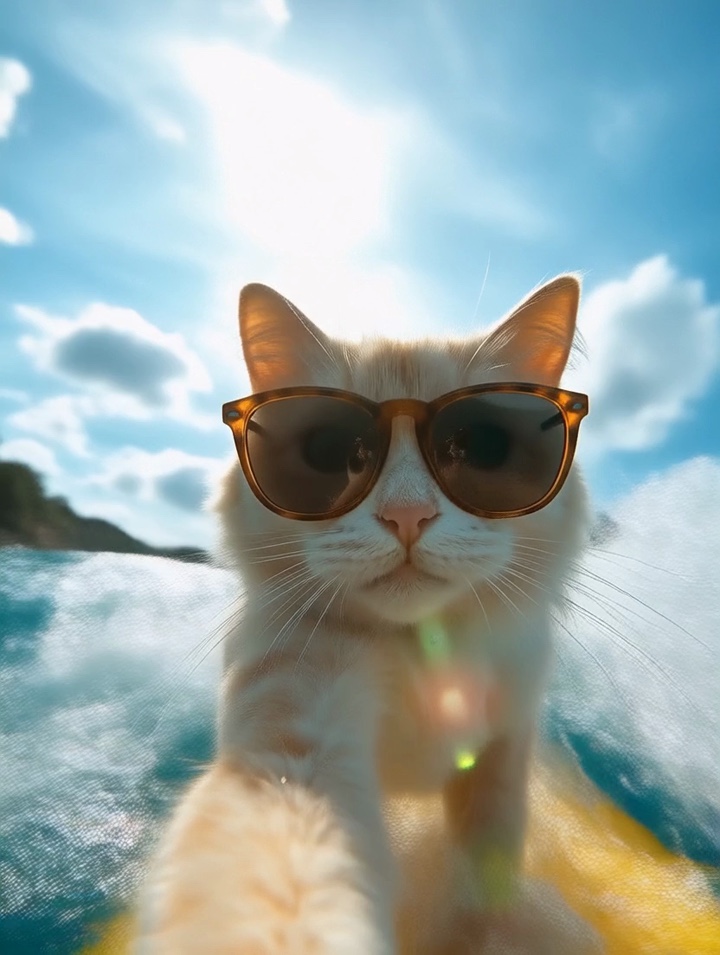}
\vspace{-0.5em}

\end{subfigure}

\vspace{-1em} \caption{5-second video generation on \texttt{Wan2.2-I2V-A14B-720P} \textbf{\red{using a single RTX 5090}}.\\\textit{Image prompt is the first frame and the text prompt is "POV selfie video, ultra-messy and extremely fast. A white cat in sunglasses stands on a surfboard with a neutral look when the board suddenly whips sideways, throwing cat and camera into the water; the frame dives sharply downward, swallowed by violent bursts of bubbles, spinning turbulence, and smeared water streaks as the camera sinks. Shadows thicken, pressure ripples distort the edges, and loose bubbles rush upward past the lens, showing the camera is still sinking. Then the cat kicks upward with explosive speed, dragging the view through churning bubbles and rapidly brightening water as sunlight floods back in; the camera races upward, water streaming off the lens, and finally breaks the surface in a sudden blast of light and spray, snapping back into a crooked, frantic selfie as the cat resurfaces."}}
\label{fig:comparison_i2v_14b_720p_video_1}
\end{figure}

\begin{figure}[H]
\centering
\begin{subfigure}{\textwidth}
\centering
\textbf{\large Original} ~~\textit{\large Latency: 4549s}\\
\vspace{0.1cm}

\includegraphics[width=0.165\textwidth]{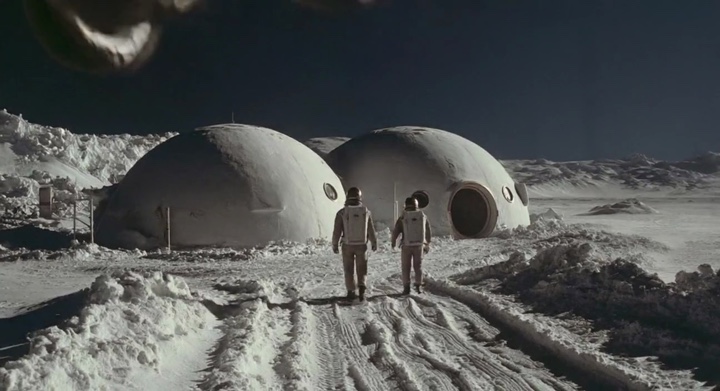}\hspace{-0.0037\textwidth}
\includegraphics[width=0.165\textwidth]{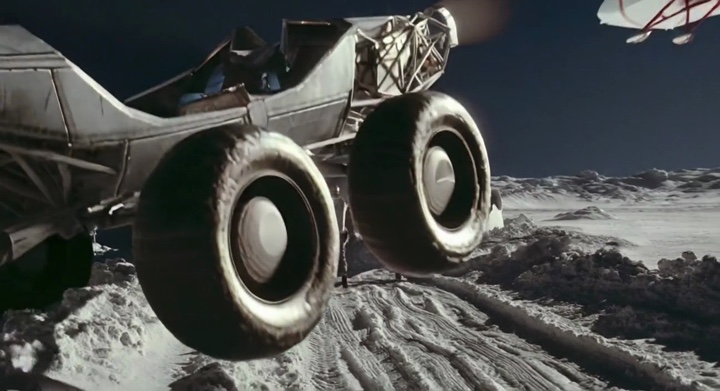}\hspace{-0.0037\textwidth}
\includegraphics[width=0.165\textwidth]{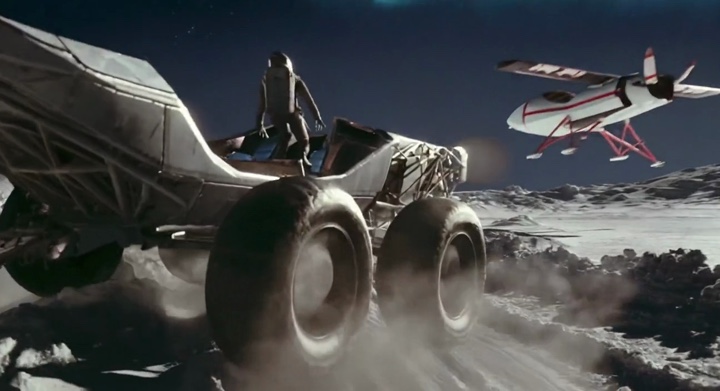}\hspace{-0.0037\textwidth}
\includegraphics[width=0.165\textwidth]{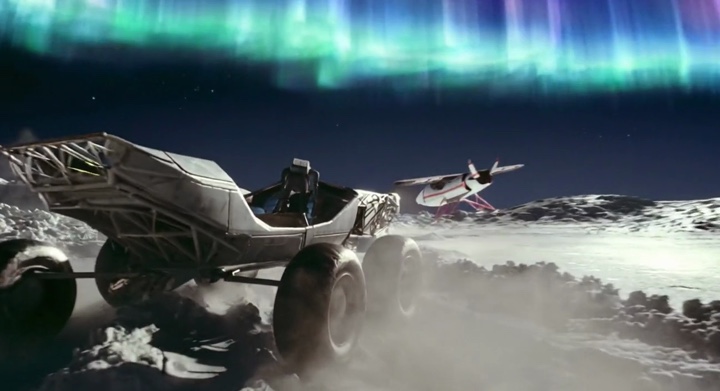}\hspace{-0.0037\textwidth}
\includegraphics[width=0.165\textwidth]{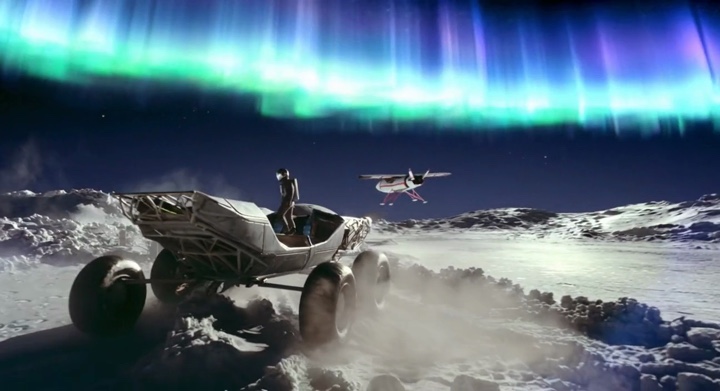}\hspace{-0.0037\textwidth}
\includegraphics[width=0.165\textwidth]{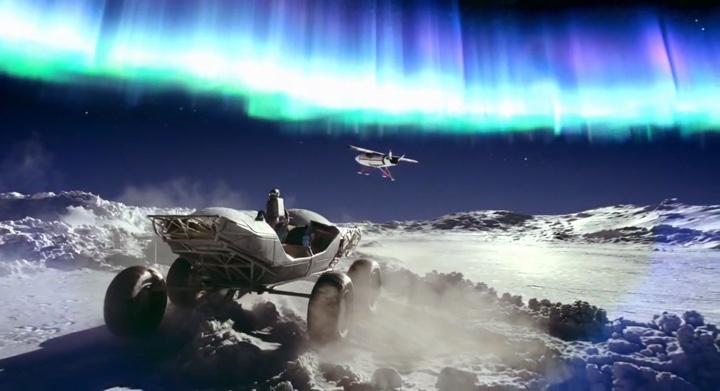}
\vspace{-0.5em}

\end{subfigure}

\vspace{0.2cm}

\begin{subfigure}{\textwidth}
\centering
\textbf{\large TurboDiffusion} ~~\textit{\large Latency: \bf \red{38s}}\\
\vspace{0.1cm}

\includegraphics[width=0.165\textwidth]{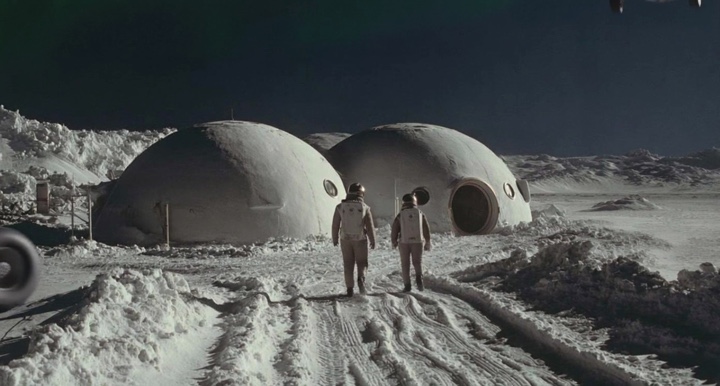}\hspace{-0.0037\textwidth}
\includegraphics[width=0.165\textwidth]{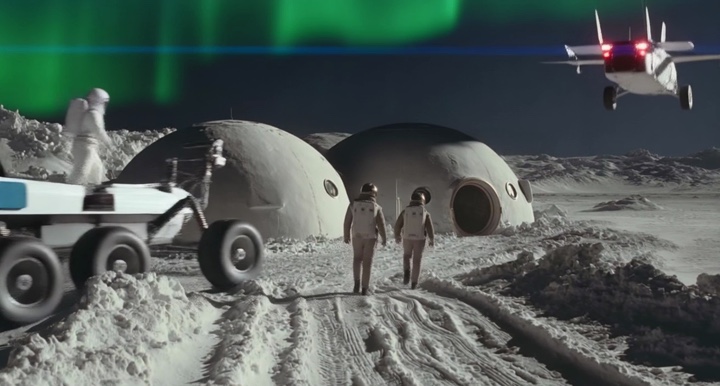}\hspace{-0.0037\textwidth}
\includegraphics[width=0.165\textwidth]{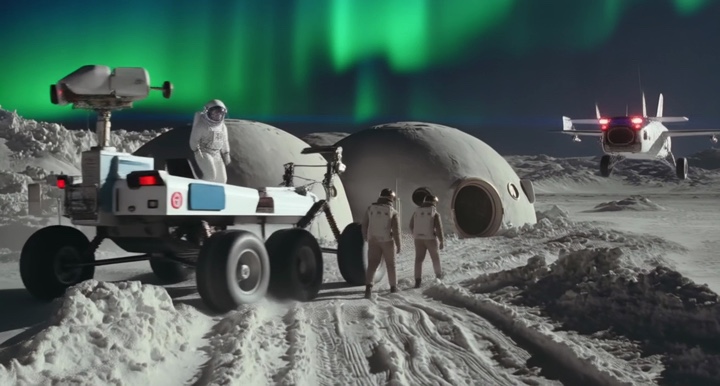}\hspace{-0.0037\textwidth}
\includegraphics[width=0.165\textwidth]{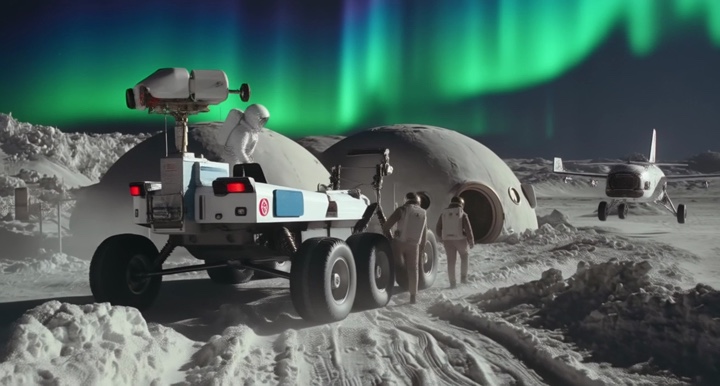}\hspace{-0.0037\textwidth}
\includegraphics[width=0.165\textwidth]{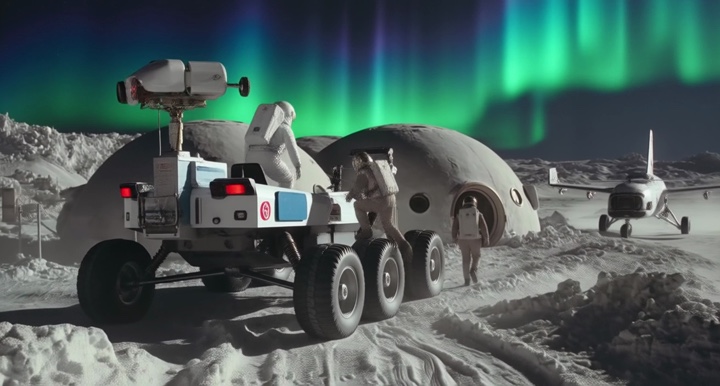}\hspace{-0.0037\textwidth}
\includegraphics[width=0.165\textwidth]{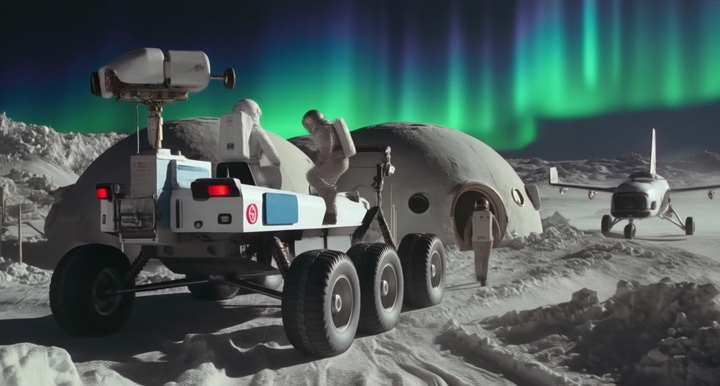}
\vspace{-0.5em}

\end{subfigure}

\vspace{-1em} \caption{5-second video generation on \texttt{Wan2.2-I2V-A14B-720P} \textbf{\red{using a single RTX 5090}}.\\\textit{Image prompt is the first frame and the text prompt is "A colorless, rugged, six-wheeled lunar rover—with exposed suspension arms, roll-cage framing, and broad low-gravity tires—glides into view from left to right, kicking up billowing plumes of moon dust that drift slowly in the vacuum. Astronauts in white spacesuits perform light, bouncing lunar strides as they hop aboard the rover’s open chassis. In the far distance, a VTOL lander with a vertical, thruster-based descent profile touches down silently on the gray surface. Above it all, vast aurora-like plasma ribbons ripple across the star-filled sky, casting shimmering green, blue, and purple light over the barren lunar plains, giving the entire scene an otherworldly, magical glow."}}
\label{fig:comparison_i2v_14b_720p_video_2}
\end{figure}

\begin{figure}[H]
\centering
\begin{subfigure}{\textwidth}
\centering
\textbf{\large Original} ~~\textit{\large Latency: 4549s}\\
\vspace{0.1cm}

\includegraphics[width=0.165\textwidth]{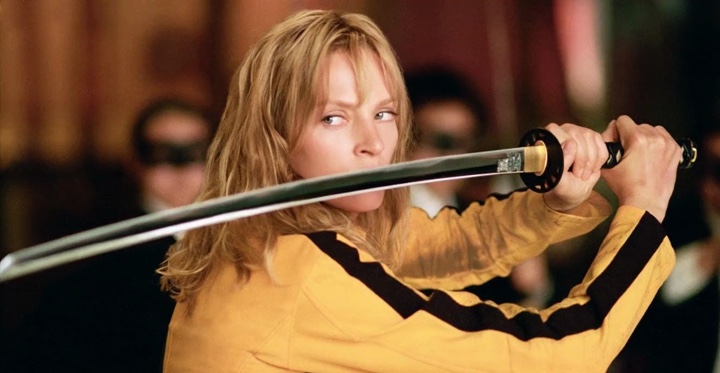}\hspace{-0.0037\textwidth}
\includegraphics[width=0.165\textwidth]{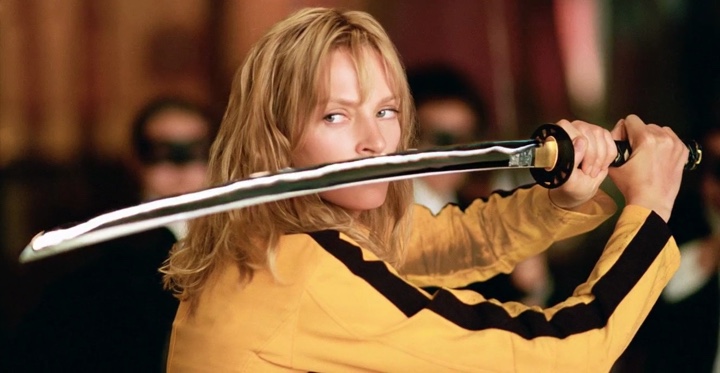}\hspace{-0.0037\textwidth}
\includegraphics[width=0.165\textwidth]{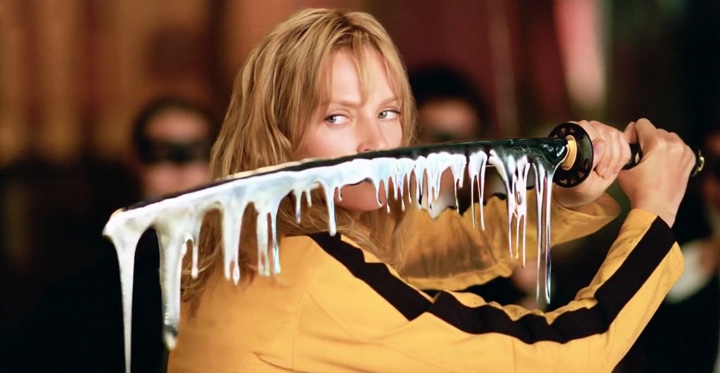}\hspace{-0.0037\textwidth}
\includegraphics[width=0.165\textwidth]{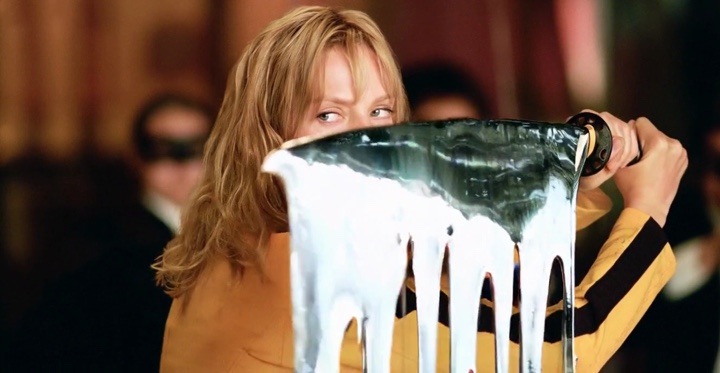}\hspace{-0.0037\textwidth}
\includegraphics[width=0.165\textwidth]{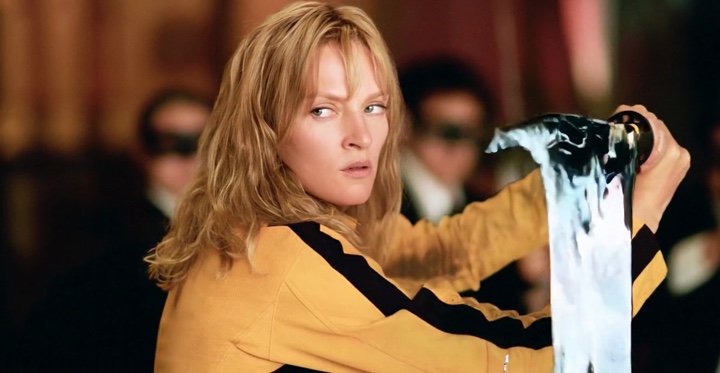}\hspace{-0.0037\textwidth}
\includegraphics[width=0.165\textwidth]{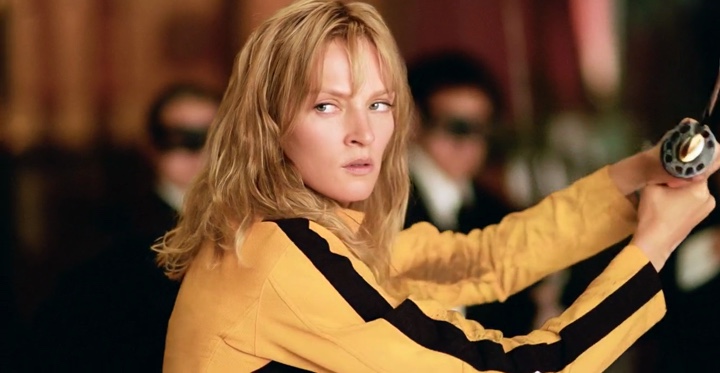}
\vspace{-0.5em}

\end{subfigure}

\vspace{0.2cm}

\begin{subfigure}{\textwidth}
\centering
\textbf{\large TurboDiffusion} ~~\textit{\large Latency: \bf \red{38s}}\\
\vspace{0.1cm}

\includegraphics[width=0.165\textwidth]{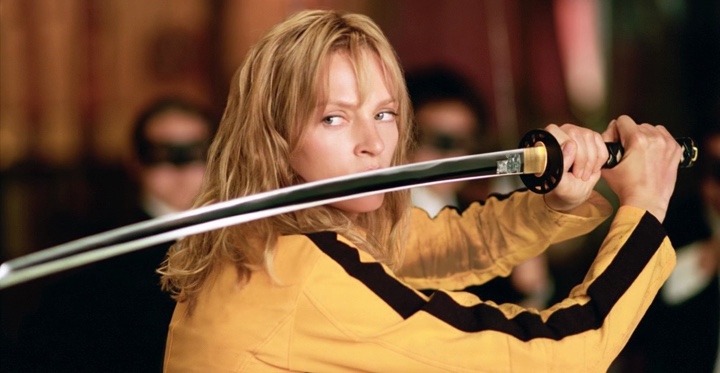}\hspace{-0.0037\textwidth}
\includegraphics[width=0.165\textwidth]{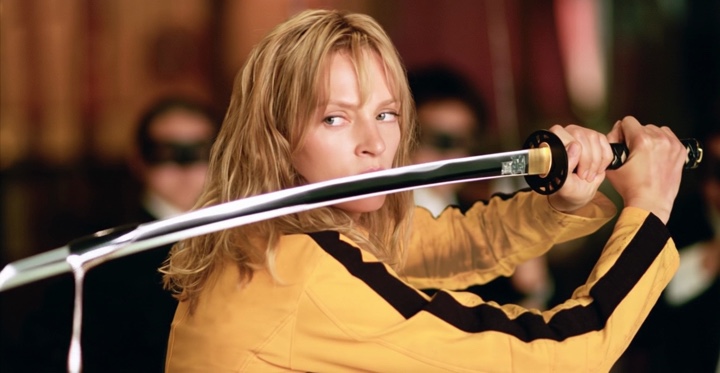}\hspace{-0.0037\textwidth}
\includegraphics[width=0.165\textwidth]{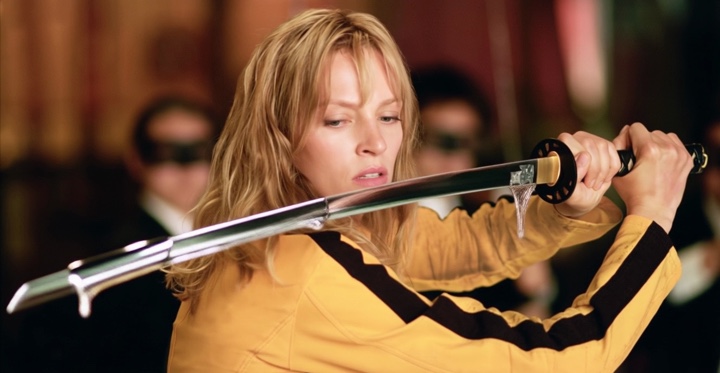}\hspace{-0.0037\textwidth}
\includegraphics[width=0.165\textwidth]{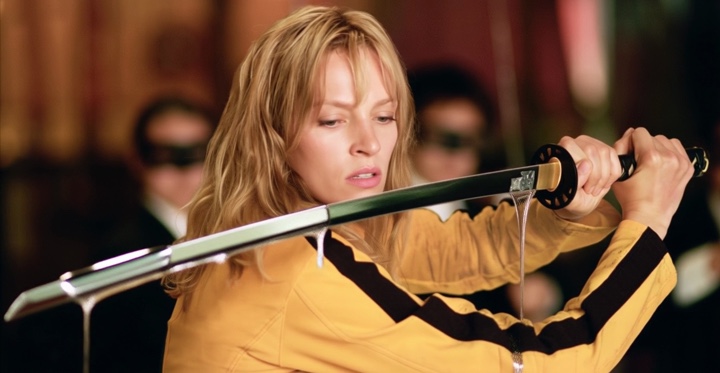}\hspace{-0.0037\textwidth}
\includegraphics[width=0.165\textwidth]{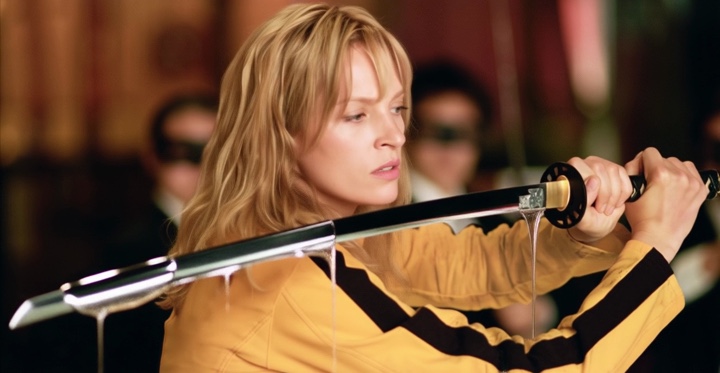}\hspace{-0.0037\textwidth}
\includegraphics[width=0.165\textwidth]{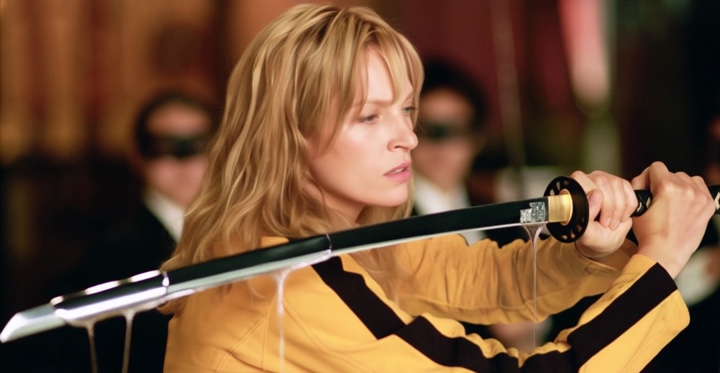}
\vspace{-0.5em}

\end{subfigure}

\vspace{-1em} \caption{5-second video generation on \texttt{Wan2.2-I2V-A14B-720P} \textbf{\red{using a single RTX 5090}}.\\\textit{Image prompt is the first frame and the text prompt is "Uma Thurman’s Beatrix Kiddo holds her razor-sharp katana blade steady in the cinematic lighting. Without warning, the entire metal piece loses rigidity at once, its material trembling like unstable liquid. The surface destabilizes completely—chunks sag off in slow folds, turning into streams of molten silver that ooze downward in drops. Within moments, the object becomes a collapsing, formless metallic mass, with no edges, and no structure remaining. Thick liquid metal spills from her grip, followed by sheets of shimmering fluid that tear away and fall to the floor. What she holds now is only a quivering blob of mercury-like liquid, constantly sagging and dripping. Her expression shifts from calm readiness to shock and confusion as the last remnants of solidity dissolve, tear apart, and pour through her fingers, leaving her defenseless and disoriented."}}
\label{fig:comparison_i2v_14b_720p_video_3}
\end{figure}

\begin{figure}[H]
\centering
\begin{subfigure}{\textwidth}
\centering
\textbf{\large Original} ~~\textit{\large Latency: 4549s}\\
\vspace{0.1cm}

\includegraphics[width=0.165\textwidth]{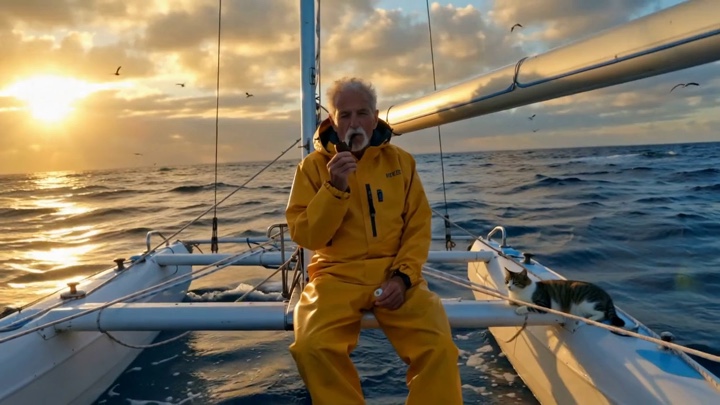}\hspace{-0.0037\textwidth}
\includegraphics[width=0.165\textwidth]{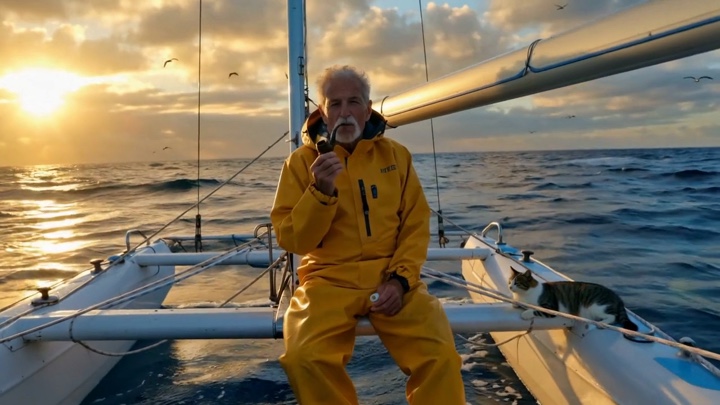}\hspace{-0.0037\textwidth}
\includegraphics[width=0.165\textwidth]{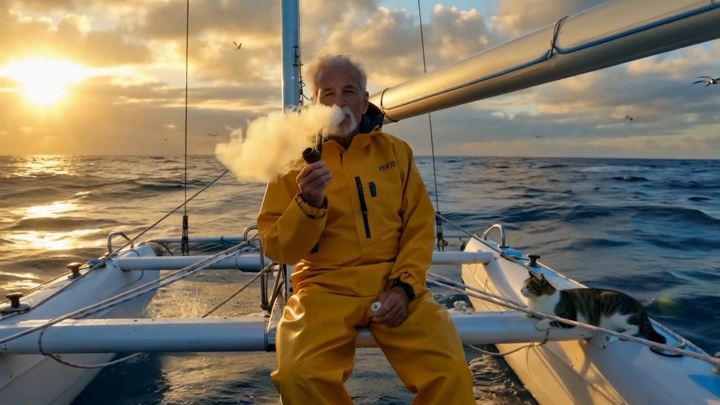}\hspace{-0.0037\textwidth}
\includegraphics[width=0.165\textwidth]{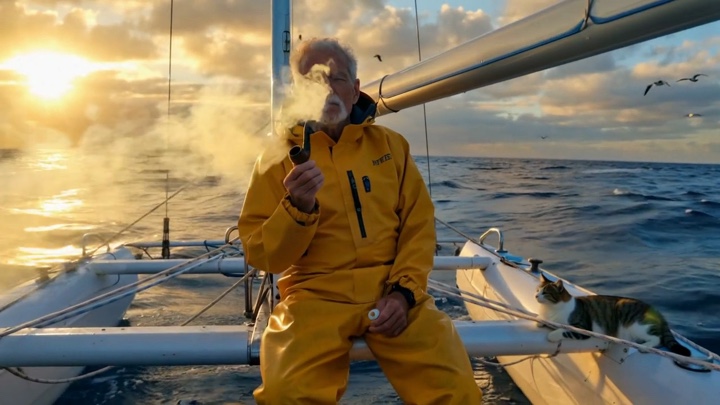}\hspace{-0.0037\textwidth}
\includegraphics[width=0.165\textwidth]{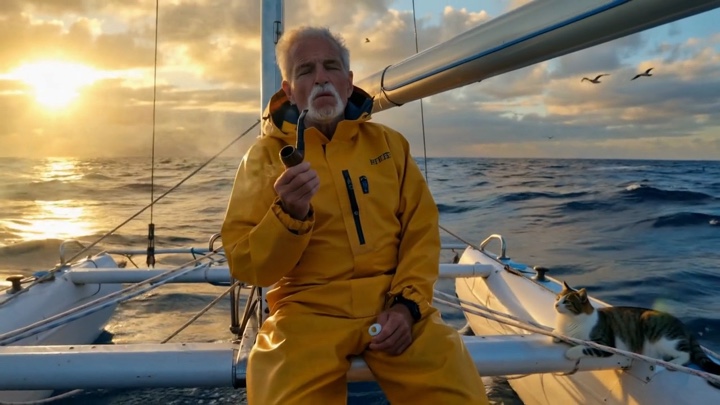}\hspace{-0.0037\textwidth}
\includegraphics[width=0.165\textwidth]{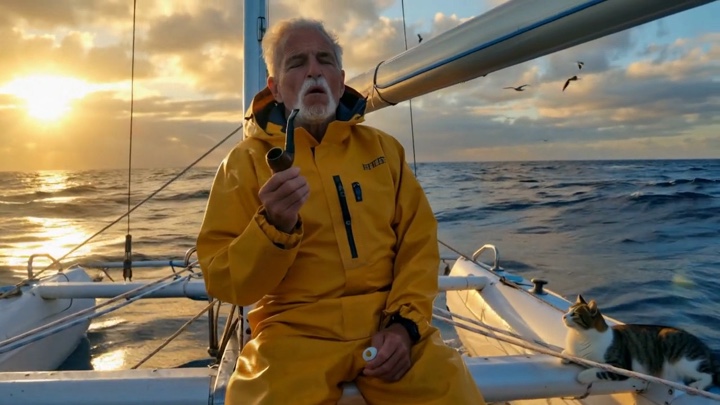}
\vspace{-0.5em}

\end{subfigure}

\vspace{0.2cm}

\begin{subfigure}{\textwidth}
\centering
\textbf{\large TurboDiffusion} ~~\textit{\large Latency: \bf \red{38s}}\\
\vspace{0.1cm}

\includegraphics[width=0.165\textwidth]{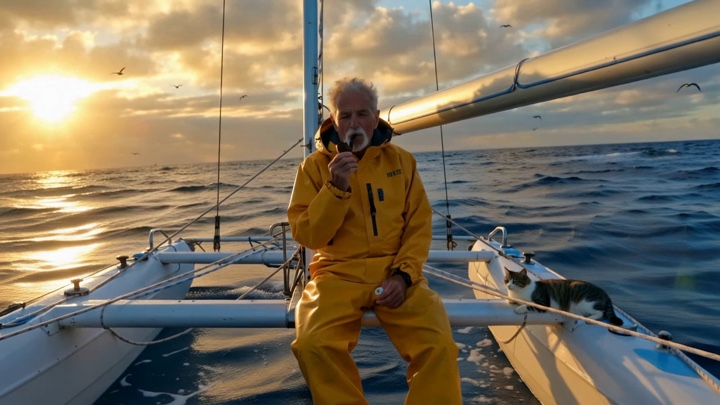}\hspace{-0.0037\textwidth}
\includegraphics[width=0.165\textwidth]{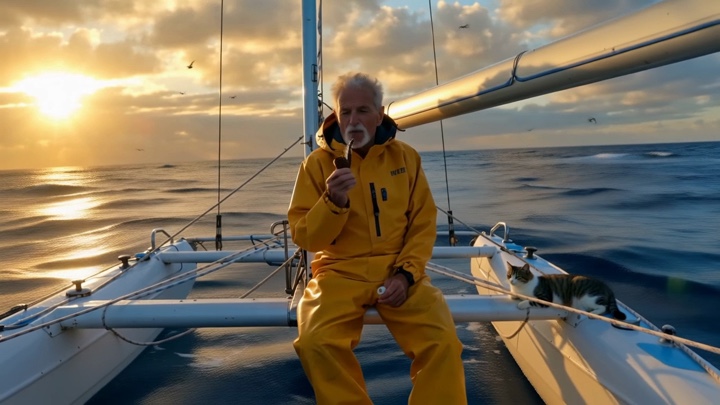}\hspace{-0.0037\textwidth}
\includegraphics[width=0.165\textwidth]{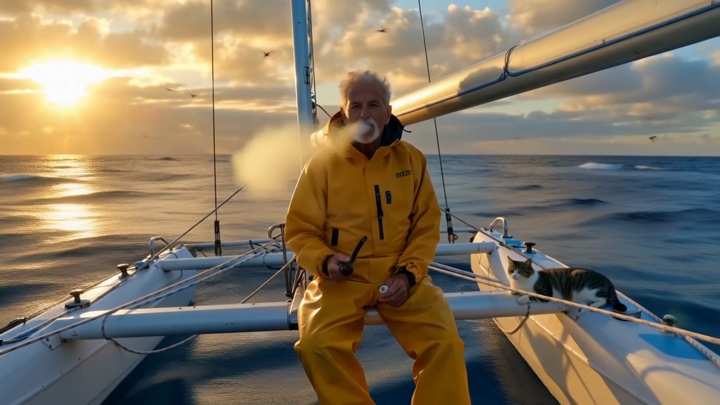}\hspace{-0.0037\textwidth}
\includegraphics[width=0.165\textwidth]{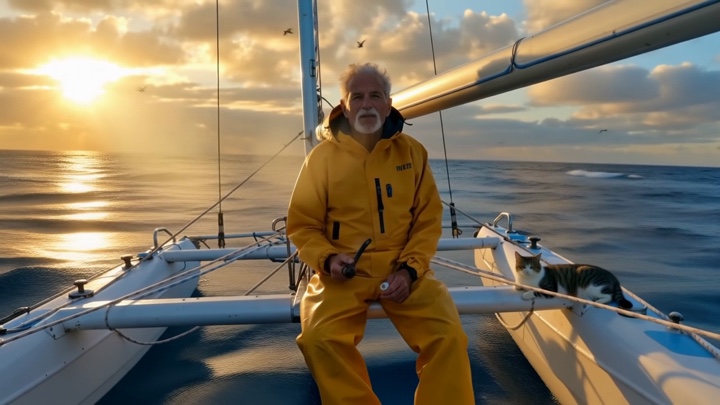}\hspace{-0.0037\textwidth}
\includegraphics[width=0.165\textwidth]{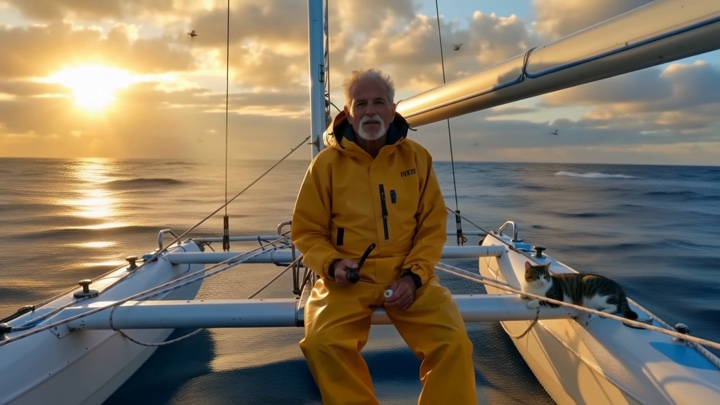}\hspace{-0.0037\textwidth}
\includegraphics[width=0.165\textwidth]{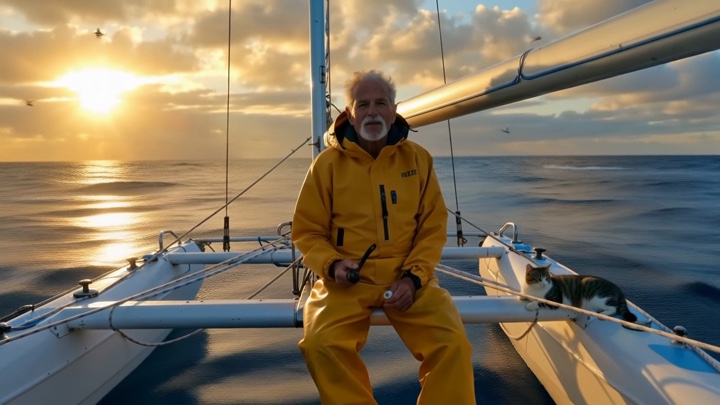}
\vspace{-0.5em}

\end{subfigure}

\vspace{-1em} \caption{5-second video generation on \texttt{Wan2.2-I2V-A14B-720P} \textbf{\red{using a single RTX 5090}}.\\\textit{Image prompt is the first frame and the text prompt is "Close-up on an elderly sailor in a weathered yellow raincoat, seated on the sun-lit deck of a gently rocking catamaran. With each small rise and dip of the hull, the shadows on his face shift subtly. He draws from his pipe, the ember brightening, and the exhale sends a thin ribbon of smoke that wavers and bends as the boat sways. His cat lies beside him, eyes half-closed, its body adjusting with soft, instinctive shifts whenever the deck tilts. Sunlight glints off the polished wood in flickering patterns as the surface of the water rolls beneath, causing brief flares of moving reflections. A pair of seabirds glide overhead, dipping slightly as the wind gusts. As the camera eases into a slow push-in, every motion becomes more pronounced—the smoke trembling, the cat’s fur fluttering, the deck creaking with each gentle sway—turning the peaceful moment into a dynamically living scene afloat at sea."}}
\label{fig:comparison_i2v_14b_720p_video_4}
\end{figure}

\begin{figure}[H]
\centering
\begin{subfigure}{\textwidth}
\centering
\textbf{\large Original} ~~\textit{\large Latency: 4549s}\\
\vspace{0.1cm}

\includegraphics[width=0.165\textwidth]{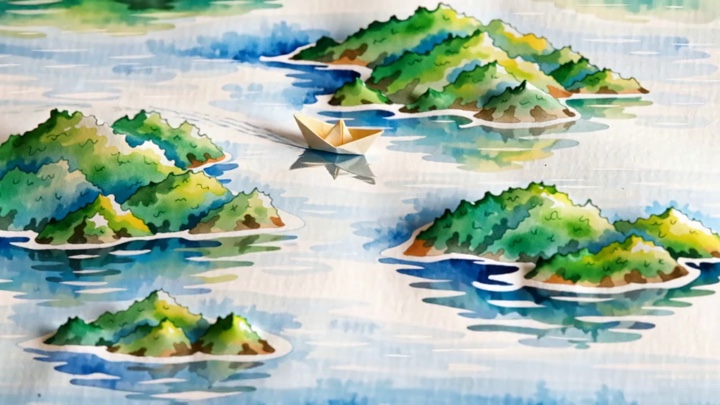}\hspace{-0.0037\textwidth}
\includegraphics[width=0.165\textwidth]{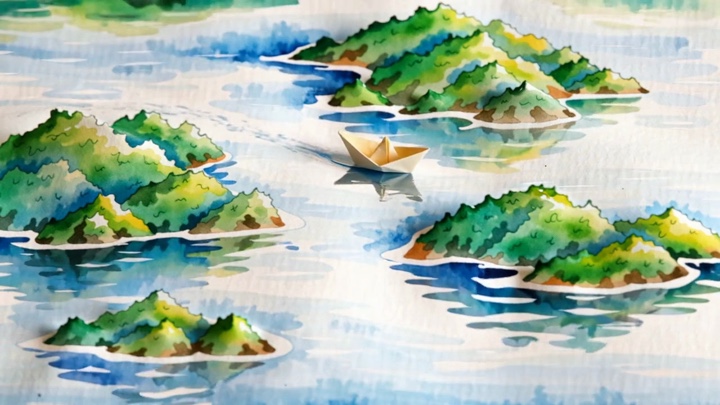}\hspace{-0.0037\textwidth}
\includegraphics[width=0.165\textwidth]{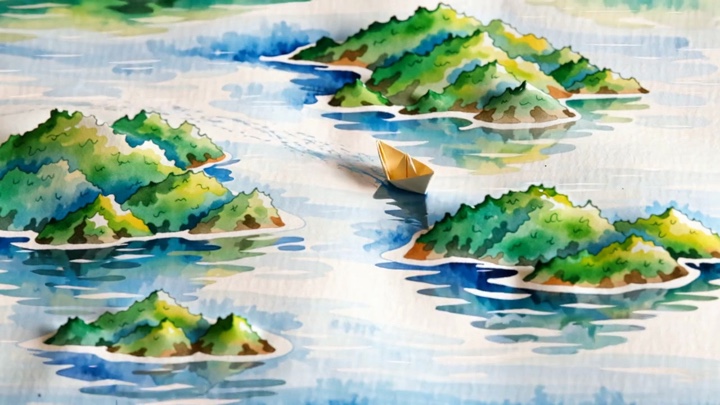}\hspace{-0.0037\textwidth}
\includegraphics[width=0.165\textwidth]{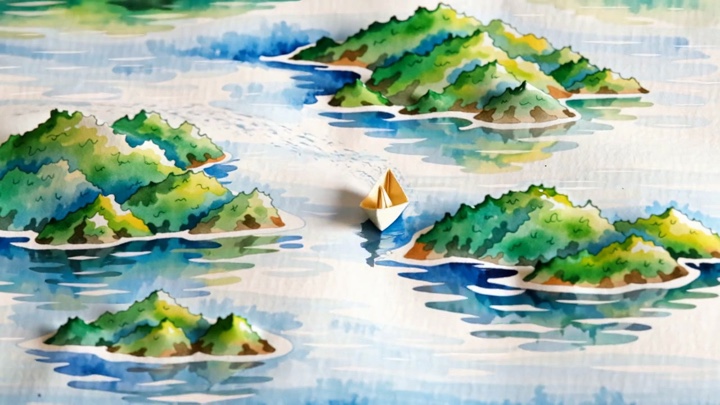}\hspace{-0.0037\textwidth}
\includegraphics[width=0.165\textwidth]{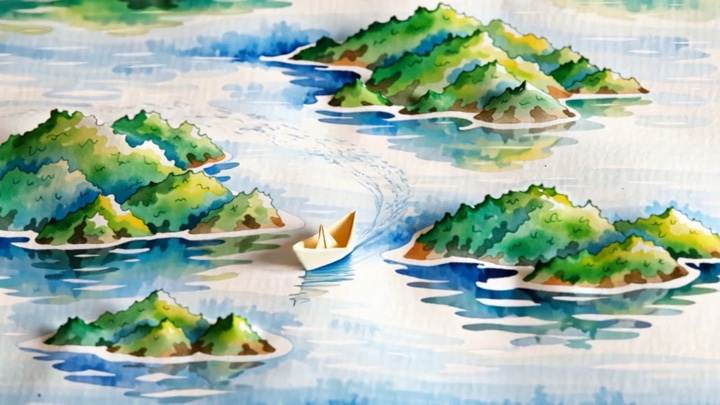}\hspace{-0.0037\textwidth}
\includegraphics[width=0.165\textwidth]{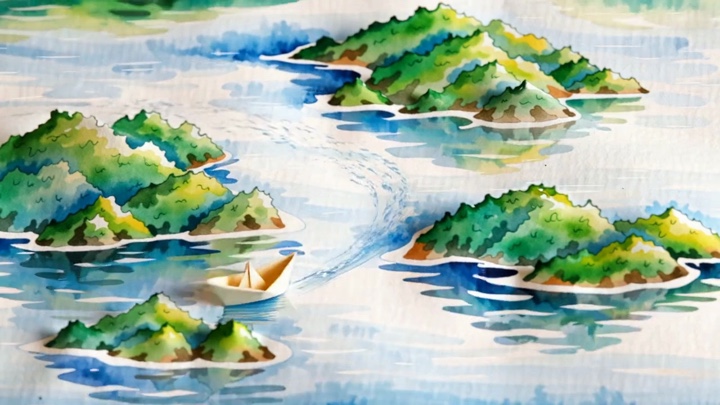}
\vspace{-0.5em}

\end{subfigure}

\vspace{0.2cm}

\begin{subfigure}{\textwidth}
\centering
\textbf{\large TurboDiffusion} ~~\textit{\large Latency: \bf \red{38s}}\\
\vspace{0.1cm}

\includegraphics[width=0.165\textwidth]{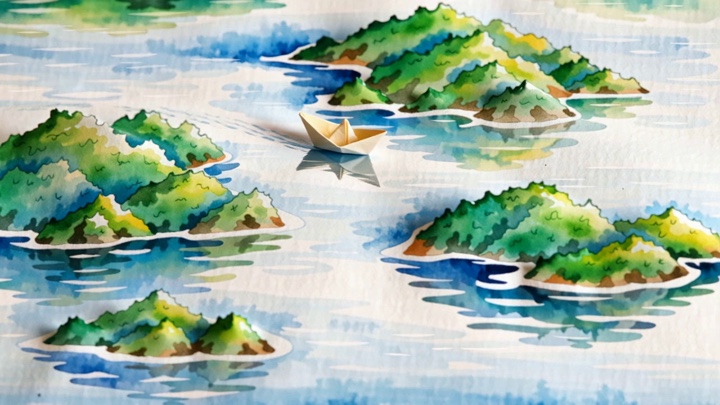}\hspace{-0.0037\textwidth}
\includegraphics[width=0.165\textwidth]{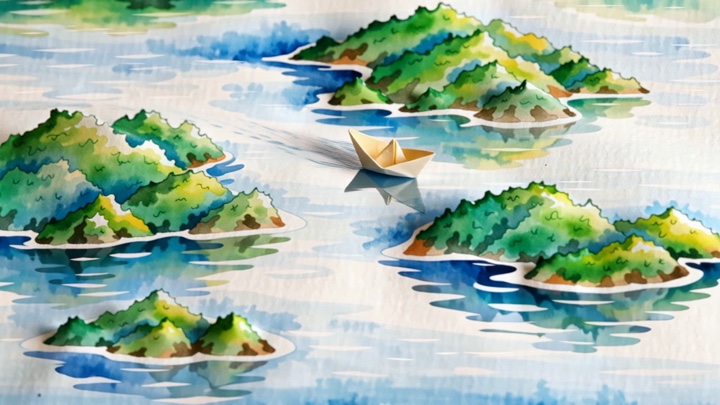}\hspace{-0.0037\textwidth}
\includegraphics[width=0.165\textwidth]{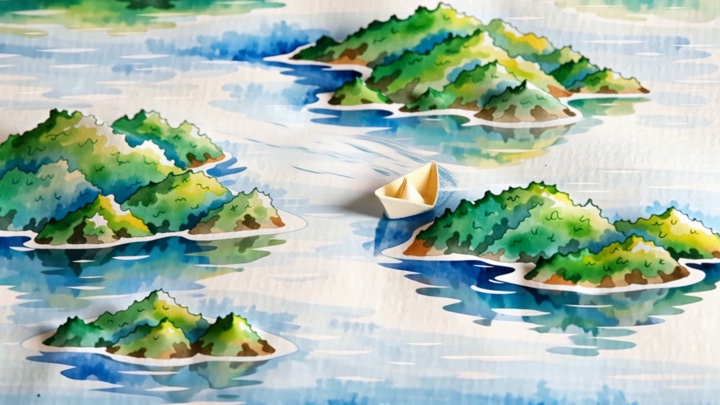}\hspace{-0.0037\textwidth}
\includegraphics[width=0.165\textwidth]{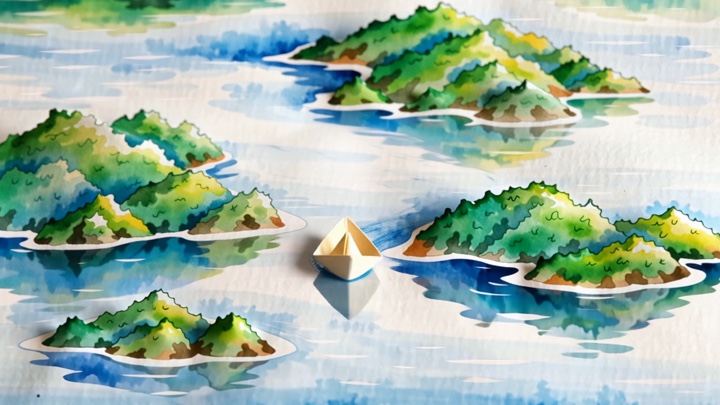}\hspace{-0.0037\textwidth}
\includegraphics[width=0.165\textwidth]{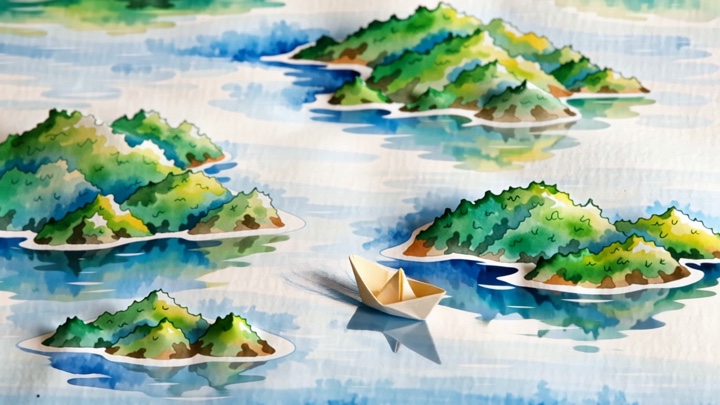}\hspace{-0.0037\textwidth}
\includegraphics[width=0.165\textwidth]{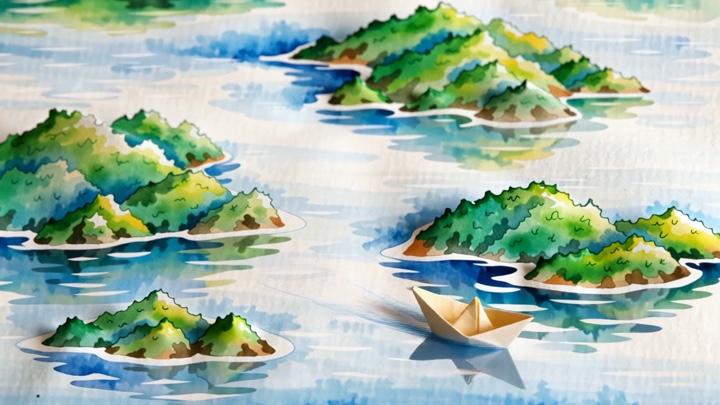}
\vspace{-0.5em}

\end{subfigure}

\vspace{-1em} \caption{5-second video generation on \texttt{Wan2.2-I2V-A14B-720P} \textbf{\red{using a single RTX 5090}}.\\\textit{Image prompt is the first frame and the text prompt is "Watercolor style. Wet suminagashi inks surge and spread rapidly across the paper, swirling outward as they form island-like shapes with actively shifting, bleeding edges. A tiny paper boat is pulled forward by a faster-moving stream of pigment, gliding swiftly toward the still-wet areas. The flow pushes it in small sudden bursts, creating sharper, overlapping ripples that distort its reflection. Ink currents twist around the boat, forming brief vortices and drifting streaks that keep redirecting its path. Soft natural side-light catches the moving sheen on the wet paper, enhancing the sense of continuous, fluid motion across the painted landscape."}}
\label{fig:comparison_i2v_14b_720p_video_5}
\end{figure}

\begin{figure}[H]
\centering
\begin{subfigure}{\textwidth}
\centering
\textbf{\large Original} ~~\textit{\large Latency: 4549s}\\
\vspace{0.1cm}

\includegraphics[width=0.165\textwidth]{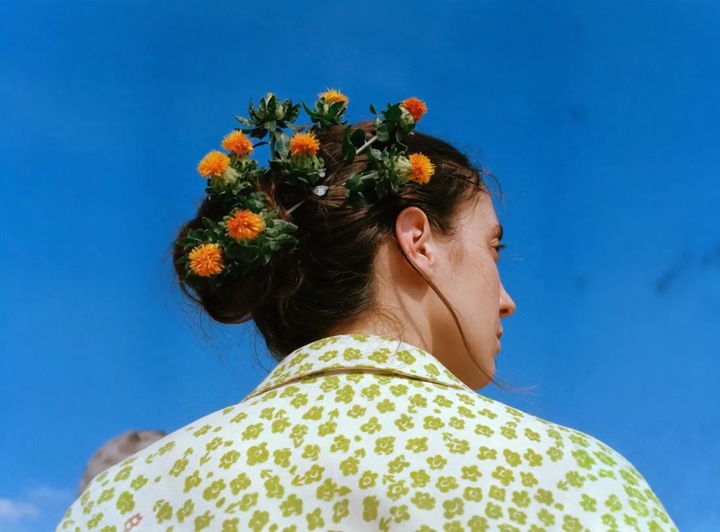}\hspace{-0.0037\textwidth}
\includegraphics[width=0.165\textwidth]{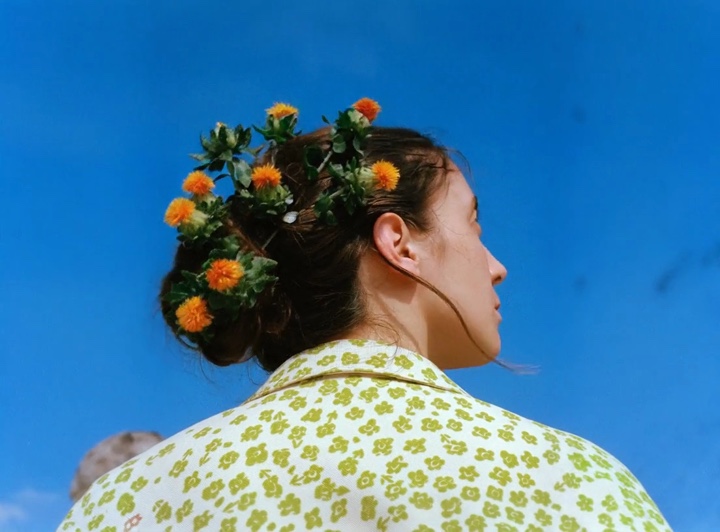}\hspace{-0.0037\textwidth}
\includegraphics[width=0.165\textwidth]{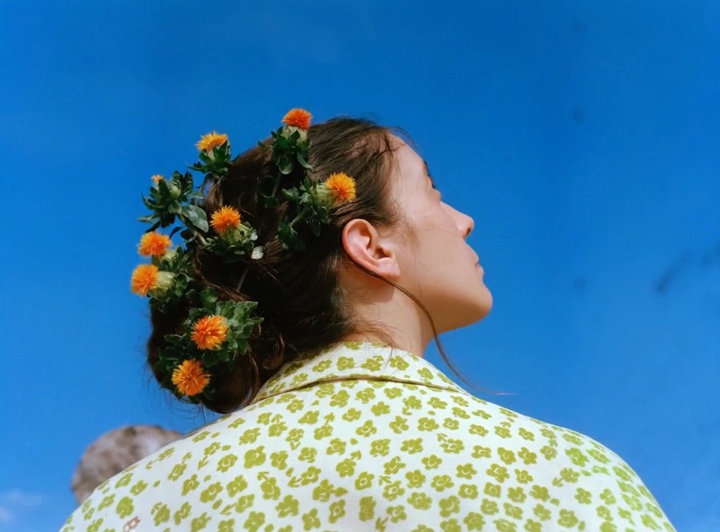}\hspace{-0.0037\textwidth}
\includegraphics[width=0.165\textwidth]{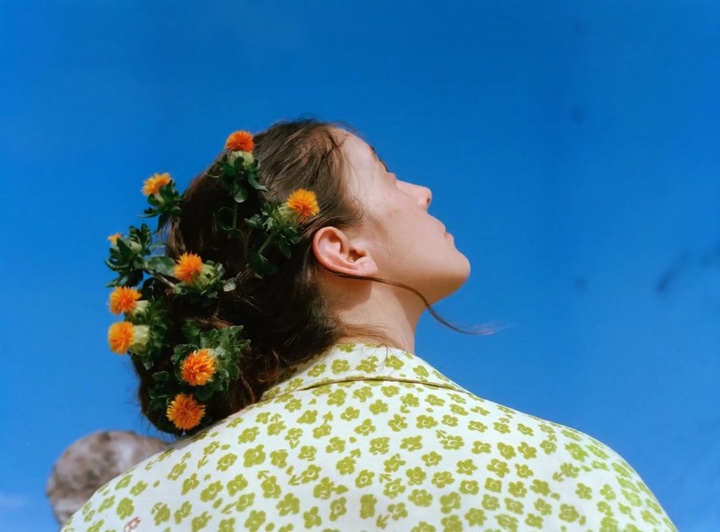}\hspace{-0.0037\textwidth}
\includegraphics[width=0.165\textwidth]{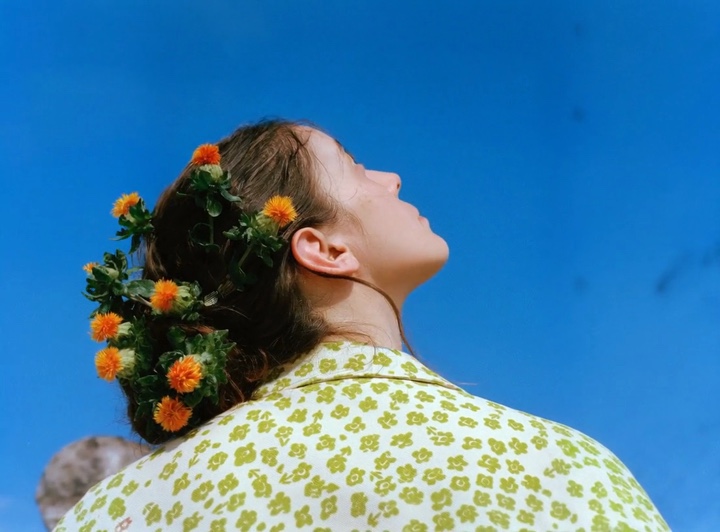}\hspace{-0.0037\textwidth}
\includegraphics[width=0.165\textwidth]{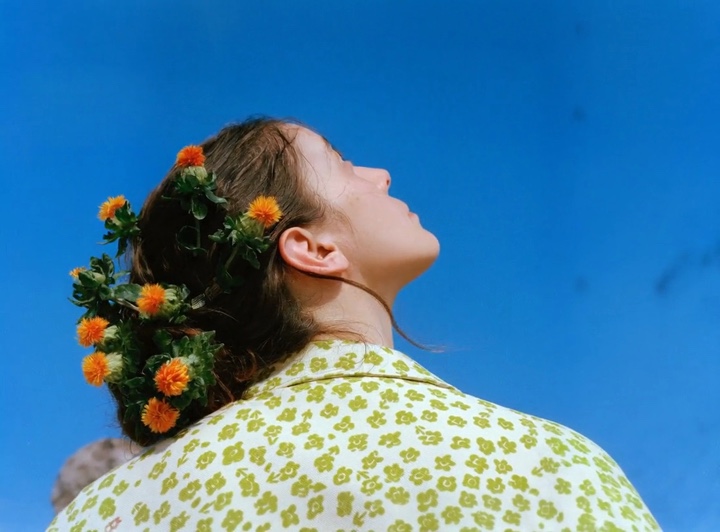}
\vspace{-0.5em}

\end{subfigure}

\vspace{0.2cm}

\begin{subfigure}{\textwidth}
\centering
\textbf{\large TurboDiffusion} ~~\textit{\large Latency: \bf \red{38s}}\\
\vspace{0.1cm}

\includegraphics[width=0.165\textwidth]{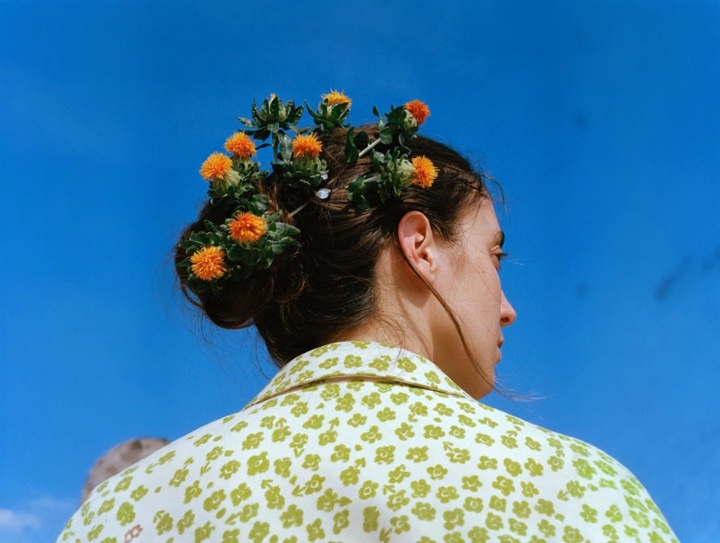}\hspace{-0.0037\textwidth}
\includegraphics[width=0.165\textwidth]{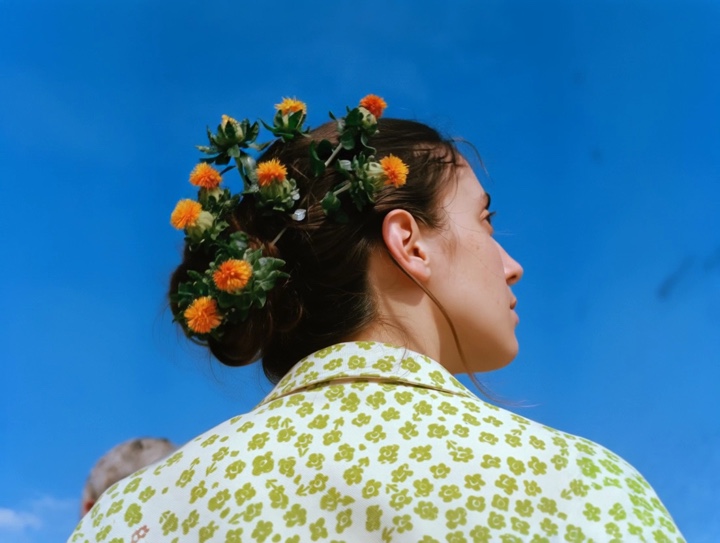}\hspace{-0.0037\textwidth}
\includegraphics[width=0.165\textwidth]{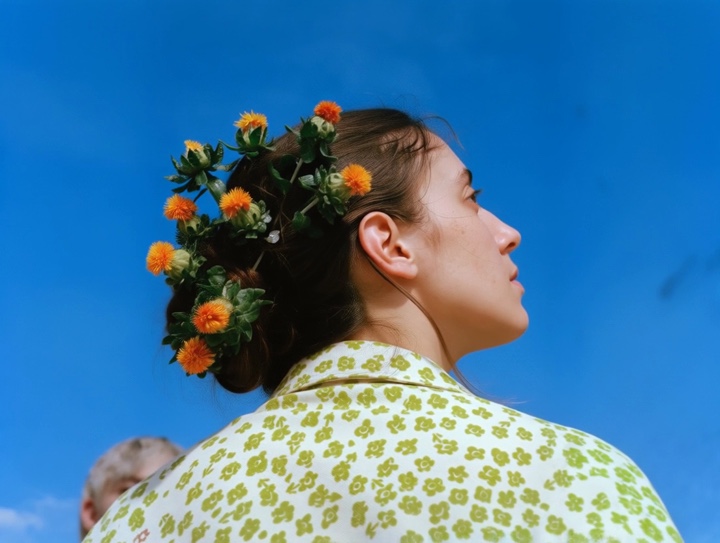}\hspace{-0.0037\textwidth}
\includegraphics[width=0.165\textwidth]{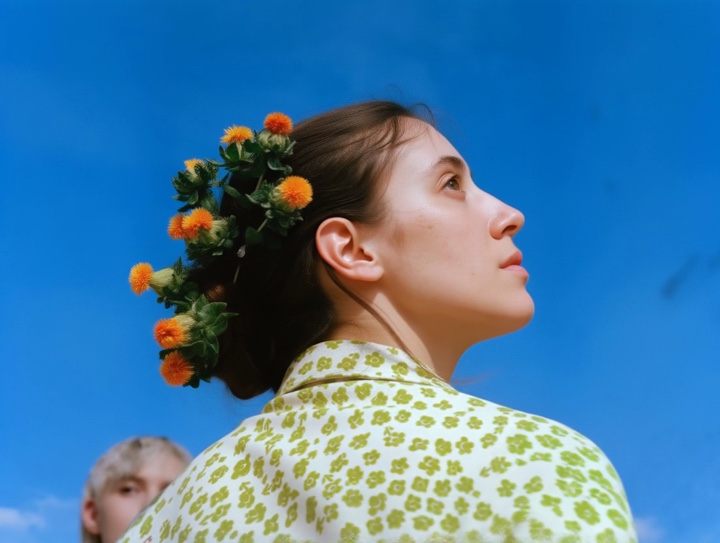}\hspace{-0.0037\textwidth}
\includegraphics[width=0.165\textwidth]{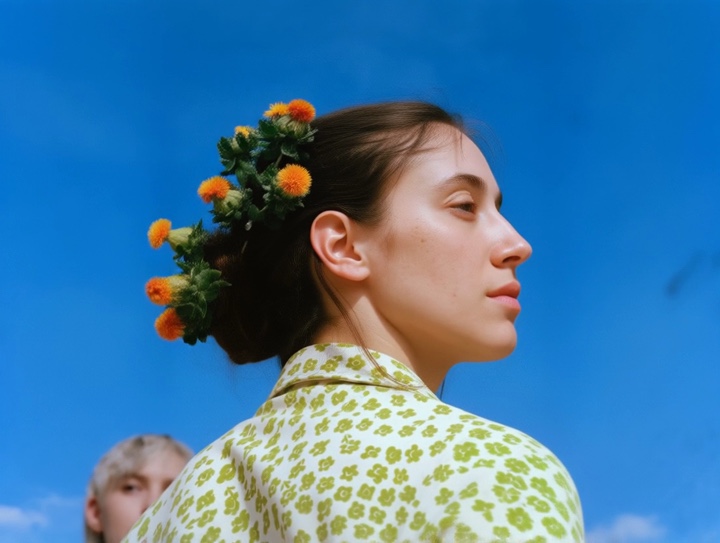}\hspace{-0.0037\textwidth}
\includegraphics[width=0.165\textwidth]{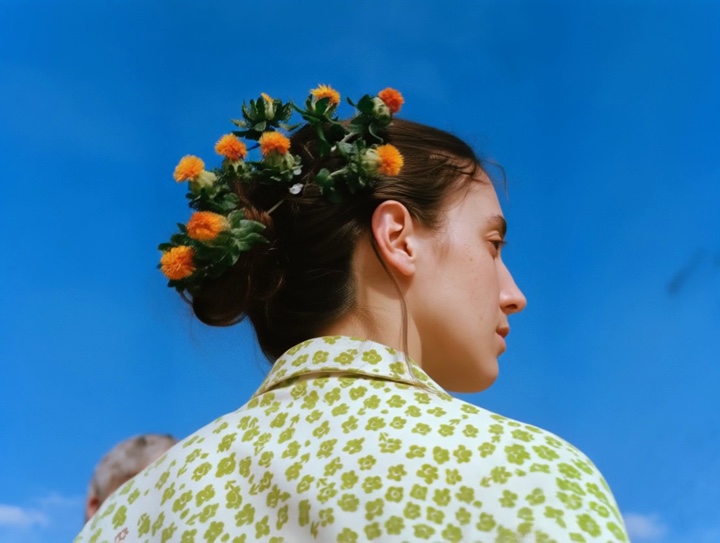}
\vspace{-0.5em}

\end{subfigure}

\vspace{-1em} \caption{5-second video generation on \texttt{Wan2.2-I2V-A14B-720P} \textbf{\red{using a single RTX 5090}}.\\\textit{Image prompt is the first frame and the text prompt is "she looks up, and then looks back"}}
\label{fig:comparison_i2v_14b_720p_video_6}
\end{figure}

\begin{figure}[H]
\centering
\begin{subfigure}{\textwidth}
\centering
\textbf{\large Original} ~~\textit{\large Latency: 4549s}\\
\vspace{0.1cm}

\includegraphics[width=0.165\textwidth]{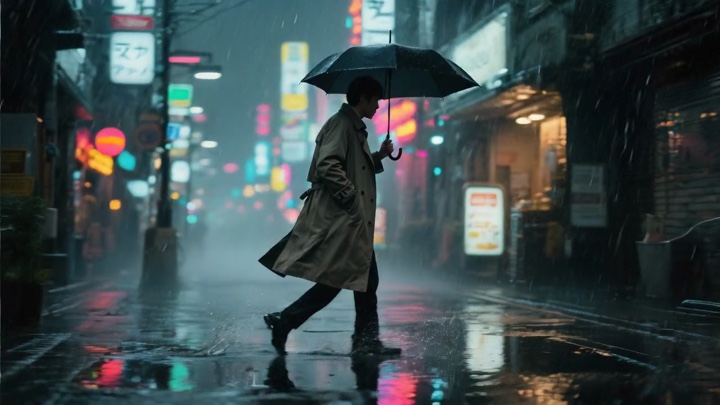}\hspace{-0.0037\textwidth}
\includegraphics[width=0.165\textwidth]{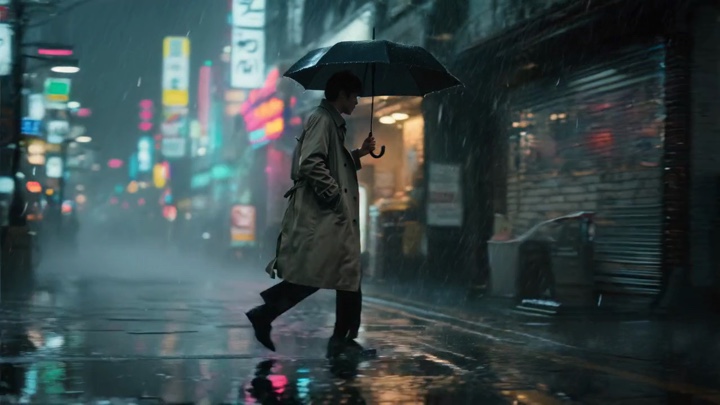}\hspace{-0.0037\textwidth}
\includegraphics[width=0.165\textwidth]{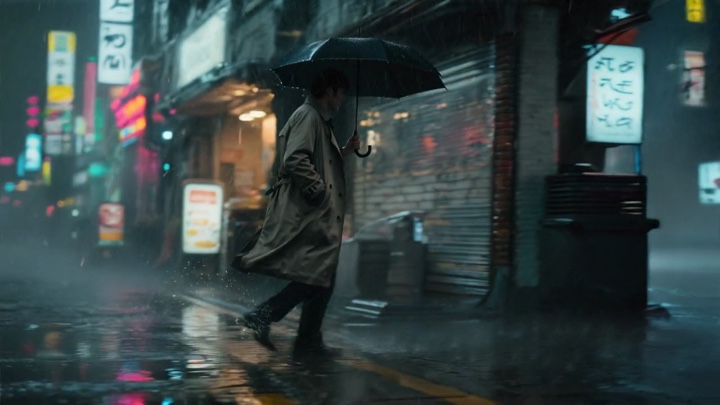}\hspace{-0.0037\textwidth}
\includegraphics[width=0.165\textwidth]{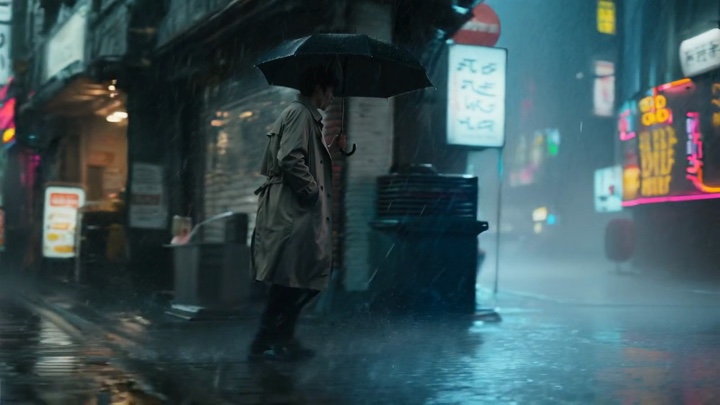}\hspace{-0.0037\textwidth}
\includegraphics[width=0.165\textwidth]{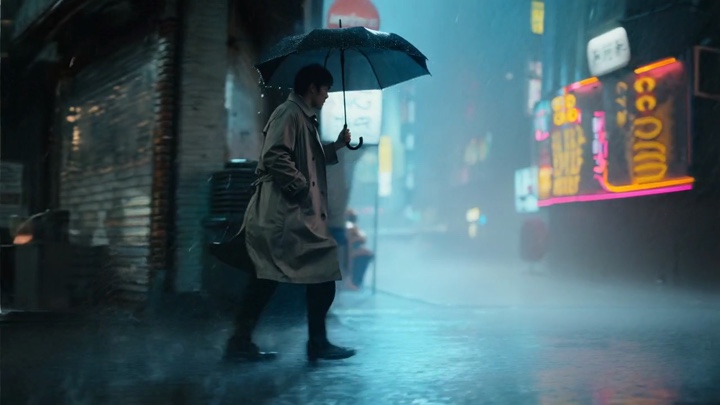}\hspace{-0.0037\textwidth}
\includegraphics[width=0.165\textwidth]{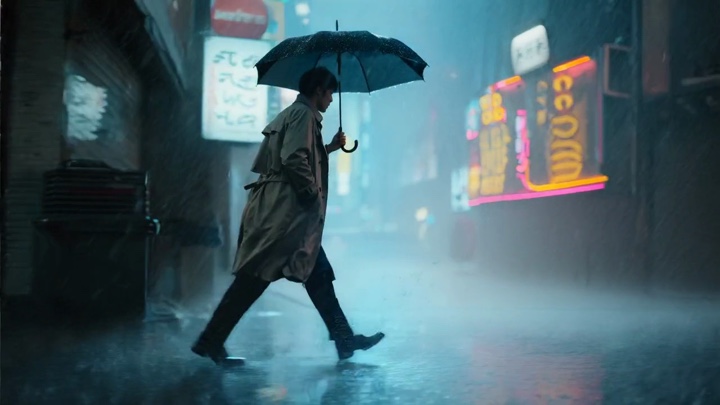}
\vspace{-0.5em}

\end{subfigure}

\vspace{0.2cm}

\begin{subfigure}{\textwidth}
\centering
\textbf{\large TurboDiffusion} ~~\textit{\large Latency: \bf \red{38s}}\\
\vspace{0.1cm}

\includegraphics[width=0.165\textwidth]{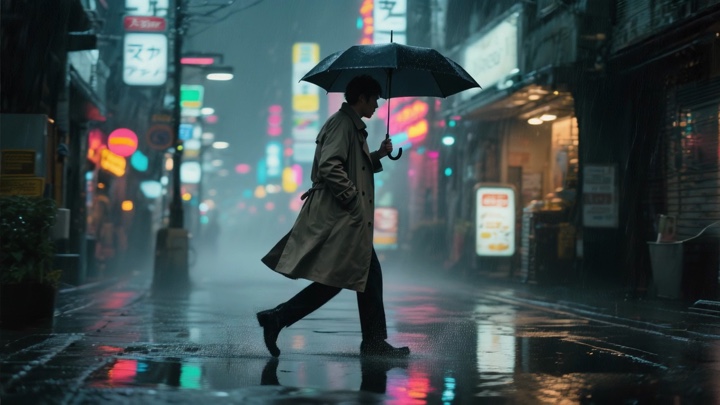}\hspace{-0.0037\textwidth}
\includegraphics[width=0.165\textwidth]{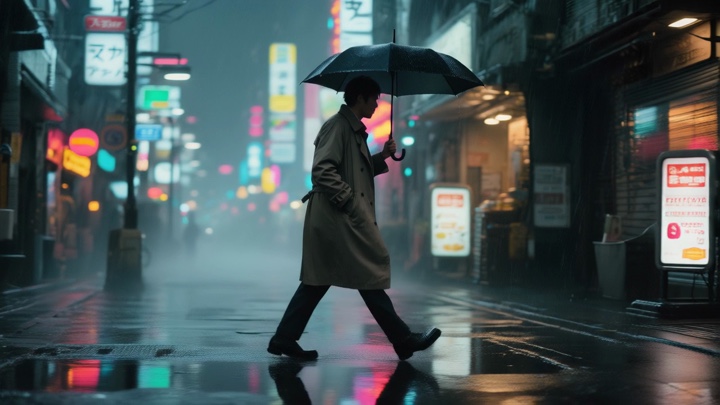}\hspace{-0.0037\textwidth}
\includegraphics[width=0.165\textwidth]{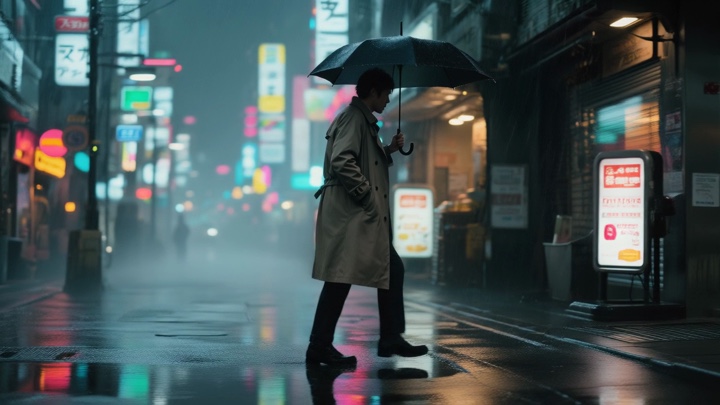}\hspace{-0.0037\textwidth}
\includegraphics[width=0.165\textwidth]{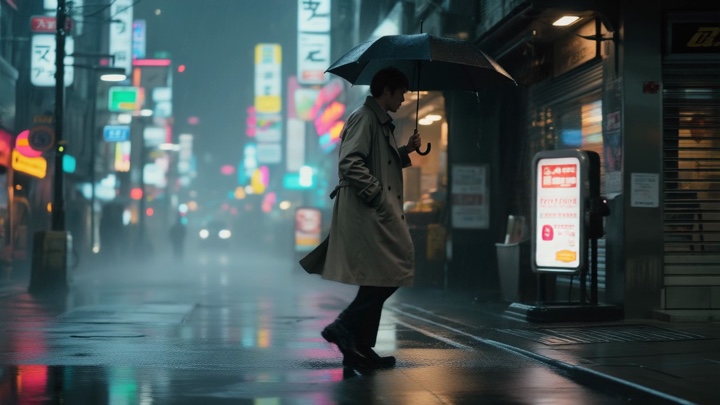}\hspace{-0.0037\textwidth}
\includegraphics[width=0.165\textwidth]{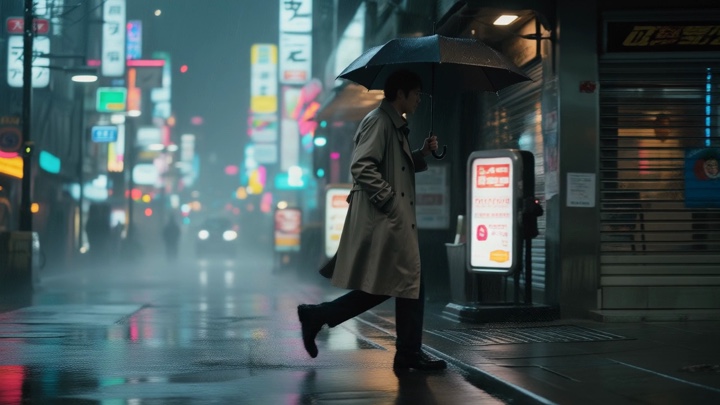}\hspace{-0.0037\textwidth}
\includegraphics[width=0.165\textwidth]{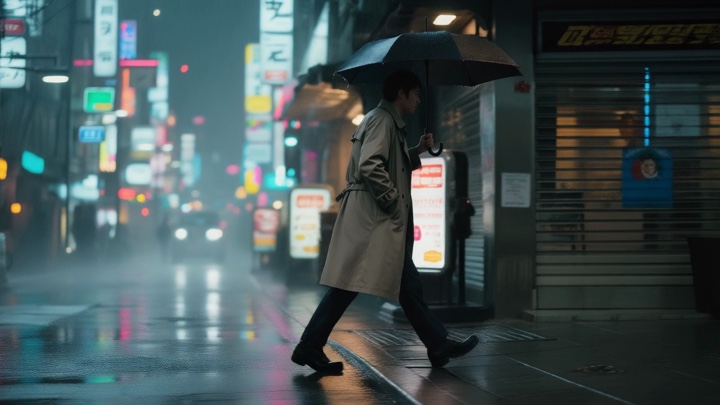}
\vspace{-0.5em}

\end{subfigure}

\vspace{-1em} \caption{5-second video generation on \texttt{Wan2.2-I2V-A14B-720P} \textbf{\red{using a single RTX 5090}}.\\\textit{Image prompt is the first frame and the text prompt is "A man in a trench coat holding a black umbrella moves at a rapid, urgent pace through the streets of Tokyo on a rainy night, splashing hard through puddles. A handheld follow-cam tracks him from the side and slightly behind with quick, jittery motion, as if struggling to keep up. The focus stays locked on the man while neon signs streak into long, colorful bokeh trails from the fast movement. The scene has a cyberpunk, film-noir mood—mysterious, lonely, and restless. The slick pavement reflects vibrant neon light; raindrops cut sharply through the frame; a thin fog shifts as the man pushes forward with fast, determined steps."}}
\label{fig:comparison_i2v_14b_720p_video_7}
\end{figure}

\subsubsection{Wan2.1-T2V-1.3B-480P}

\begin{figure}[H]
\centering
\begin{subfigure}{\textwidth}
\centering
\textbf{\large Original} ~~\textit{\large Latency: 184s}\\
\vspace{0.1cm}

\includegraphics[width=0.165\textwidth]{src/figs/original/outputs_1.3B/frames/12-1.jpg}\hspace{-0.0037\textwidth}
\includegraphics[width=0.165\textwidth]{src/figs/original/outputs_1.3B/frames/12-2.jpg}\hspace{-0.0037\textwidth}
\includegraphics[width=0.165\textwidth]{src/figs/original/outputs_1.3B/frames/12-3.jpg}\hspace{-0.0037\textwidth}
\includegraphics[width=0.165\textwidth]{src/figs/original/outputs_1.3B/frames/12-4.jpg}\hspace{-0.0037\textwidth}
\includegraphics[width=0.165\textwidth]{src/figs/original/outputs_1.3B/frames/12-5.jpg}\hspace{-0.0037\textwidth}
\includegraphics[width=0.165\textwidth]{src/figs/original/outputs_1.3B/frames/12-6.jpg}
\vspace{-0.5em}

\end{subfigure}

\vspace{0.2cm}

\begin{subfigure}{\textwidth}
\centering
\textbf{\large FastVideo} ~~\textit{\large Latency: 5.3s}\\
\vspace{0.1cm}

\includegraphics[width=0.165\textwidth]{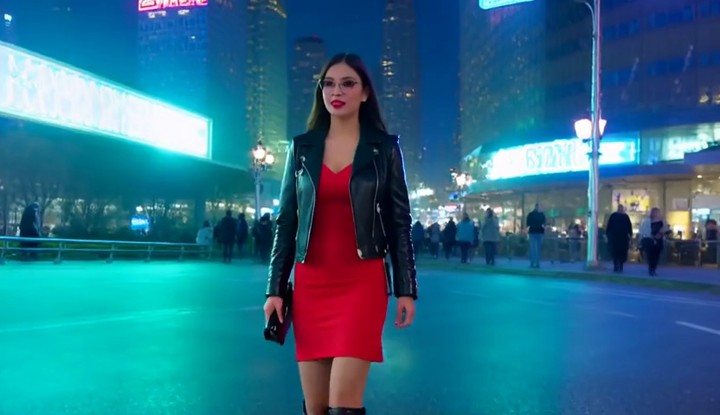}\hspace{-0.0037\textwidth}
\includegraphics[width=0.165\textwidth]{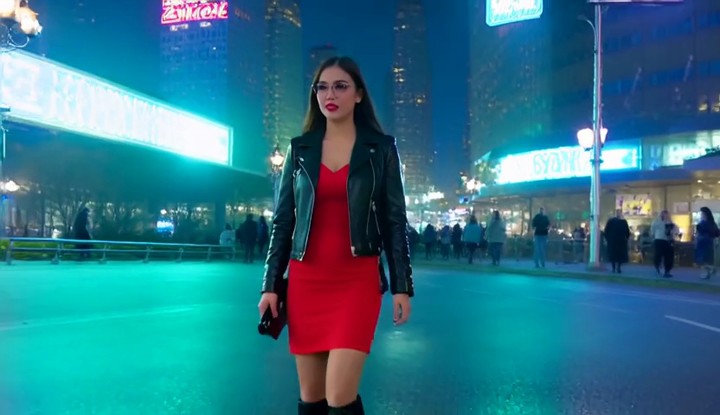}\hspace{-0.0037\textwidth}
\includegraphics[width=0.165\textwidth]{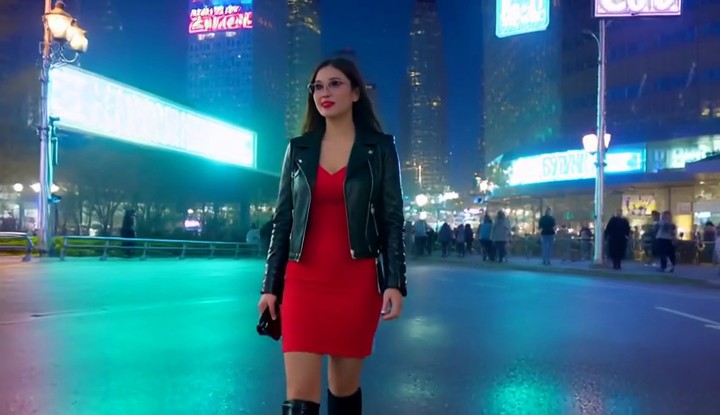}\hspace{-0.0037\textwidth}
\includegraphics[width=0.165\textwidth]{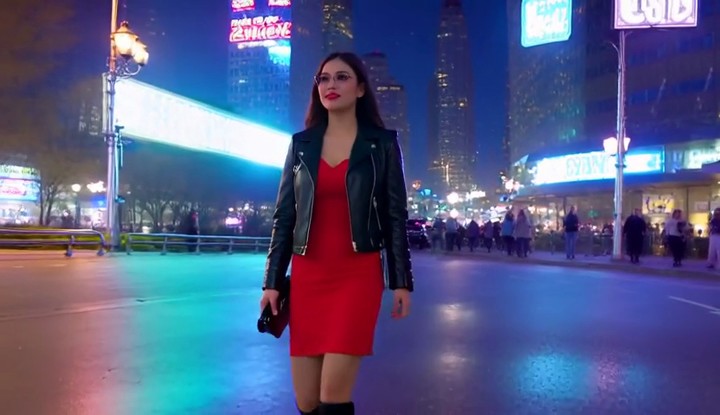}\hspace{-0.0037\textwidth}
\includegraphics[width=0.165\textwidth]{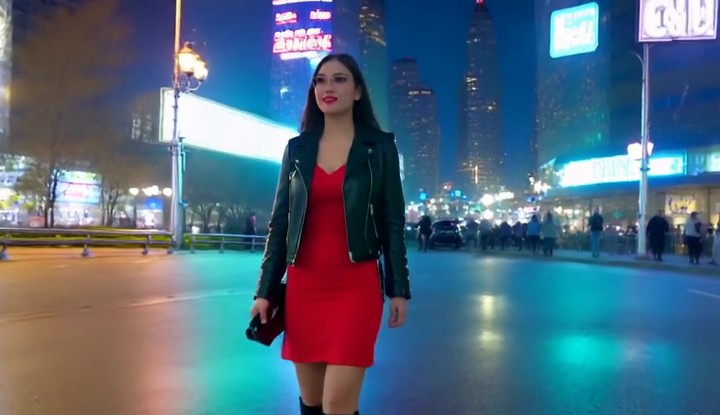}\hspace{-0.0037\textwidth}
\includegraphics[width=0.165\textwidth]{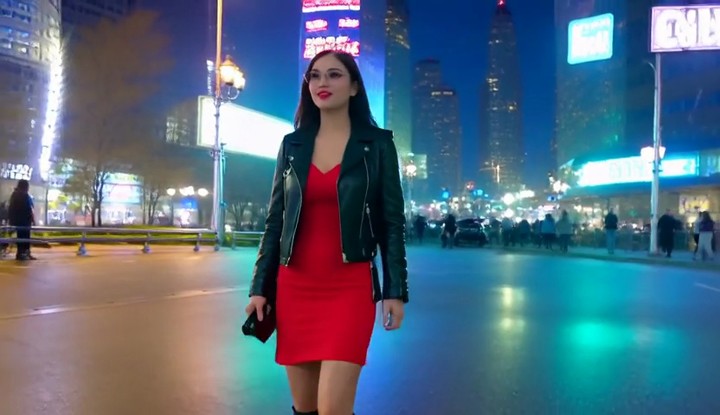}
\vspace{-0.5em}

\end{subfigure}

\vspace{0.2cm}

\begin{subfigure}{\textwidth}
\centering
\textbf{\large TurboDiffusion} ~~\textit{\large Latency: \bf \red{1.9s}}\\
\vspace{0.1cm}

\includegraphics[width=0.165\textwidth]{src/figs/turbodiffusion/outputs_1.3B/frames/12-1.jpg}\hspace{-0.0037\textwidth}
\includegraphics[width=0.165\textwidth]{src/figs/turbodiffusion/outputs_1.3B/frames/12-2.jpg}\hspace{-0.0037\textwidth}
\includegraphics[width=0.165\textwidth]{src/figs/turbodiffusion/outputs_1.3B/frames/12-3.jpg}\hspace{-0.0037\textwidth}
\includegraphics[width=0.165\textwidth]{src/figs/turbodiffusion/outputs_1.3B/frames/12-4.jpg}\hspace{-0.0037\textwidth}
\includegraphics[width=0.165\textwidth]{src/figs/turbodiffusion/outputs_1.3B/frames/12-5.jpg}\hspace{-0.0037\textwidth}
\includegraphics[width=0.165\textwidth]{src/figs/turbodiffusion/outputs_1.3B/frames/12-6.jpg}
\vspace{-0.5em}

\end{subfigure}

\vspace{-1em} \caption{5-second video generation on \texttt{Wan2.1-T2V-1.3B-480P} \textbf{\red{using a single RTX 5090}}.\\\textit{Prompt = "A stylish woman walks down a Tokyo street filled with warm glowing neon and animated city signage. She wears a black leather jacket, a long red dress, and black boots, and carries a black purse. She wears sunglasses and red lipstick. She walks confidently and casually. \red{The street is damp and reflective, creating a mirror effect of the colorful lights. Many pedestrians walk about.}"}}
\label{fig:comparison_1_3b_video_12}
\end{figure}

\begin{figure}[H]
\centering
\begin{subfigure}{\textwidth}
\centering
\textbf{\large Original} ~~\textit{\large Latency: 184s}\\
\vspace{0.1cm}

\includegraphics[width=0.165\textwidth]{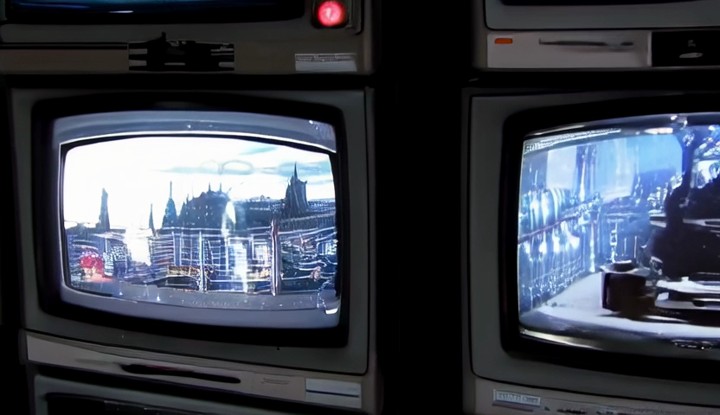}\hspace{-0.0037\textwidth}
\includegraphics[width=0.165\textwidth]{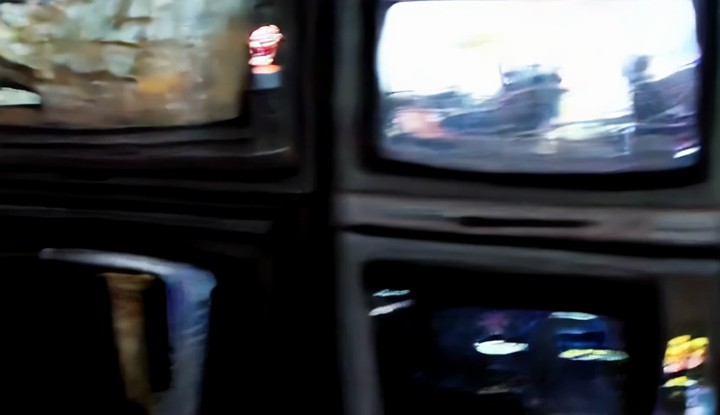}\hspace{-0.0037\textwidth}
\includegraphics[width=0.165\textwidth]{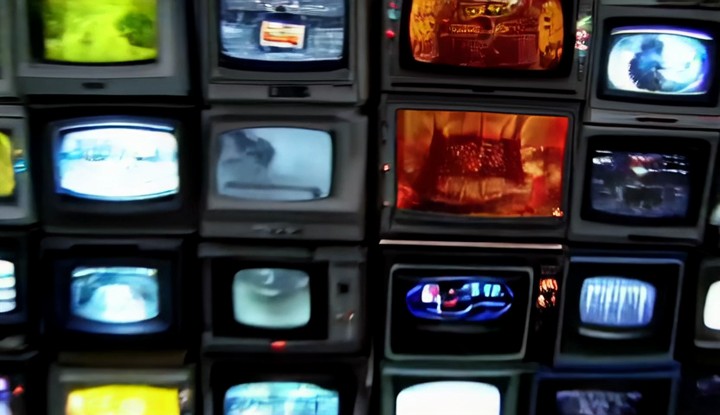}\hspace{-0.0037\textwidth}
\includegraphics[width=0.165\textwidth]{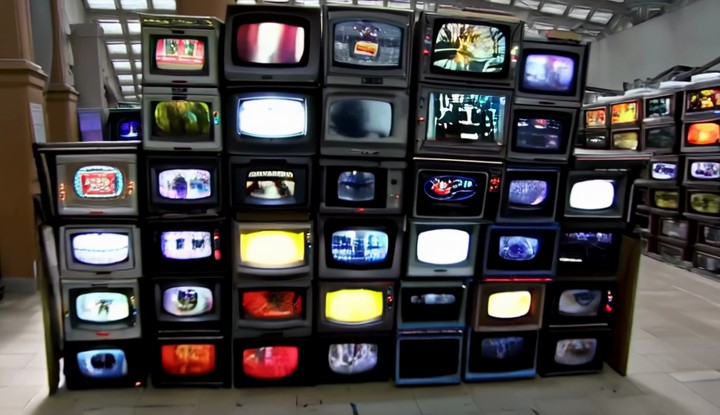}\hspace{-0.0037\textwidth}
\includegraphics[width=0.165\textwidth]{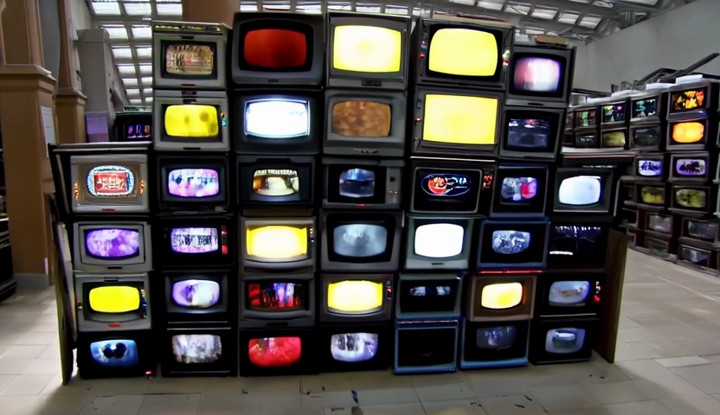}\hspace{-0.0037\textwidth}
\includegraphics[width=0.165\textwidth]{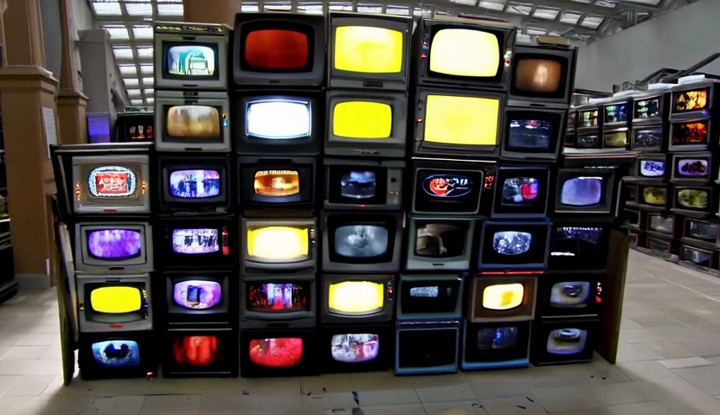}
\vspace{-0.5em}

\end{subfigure}

\vspace{0.2cm}

\begin{subfigure}{\textwidth}
\centering
\textbf{\large FastVideo} ~~\textit{\large Latency: 5.3s}\\
\vspace{0.1cm}

\includegraphics[width=0.165\textwidth]{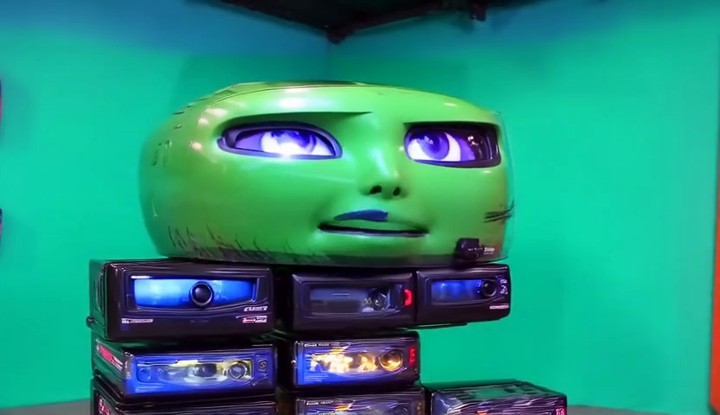}\hspace{-0.0037\textwidth}
\includegraphics[width=0.165\textwidth]{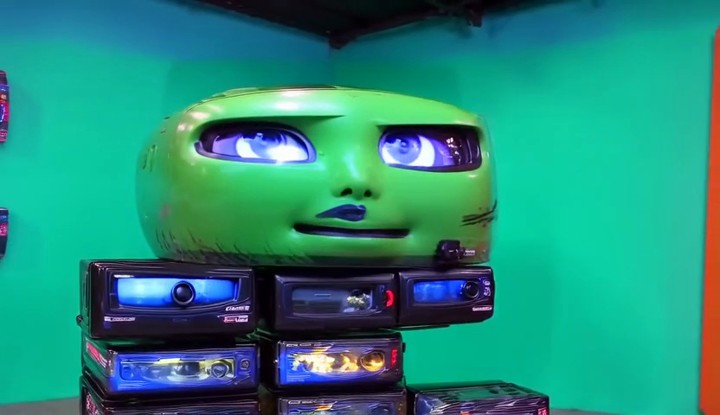}\hspace{-0.0037\textwidth}
\includegraphics[width=0.165\textwidth]{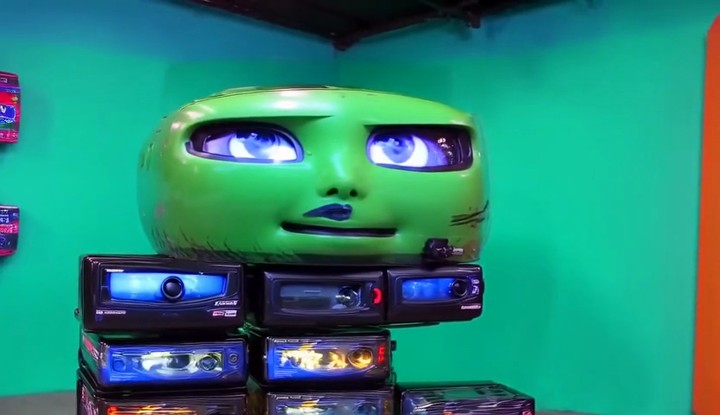}\hspace{-0.0037\textwidth}
\includegraphics[width=0.165\textwidth]{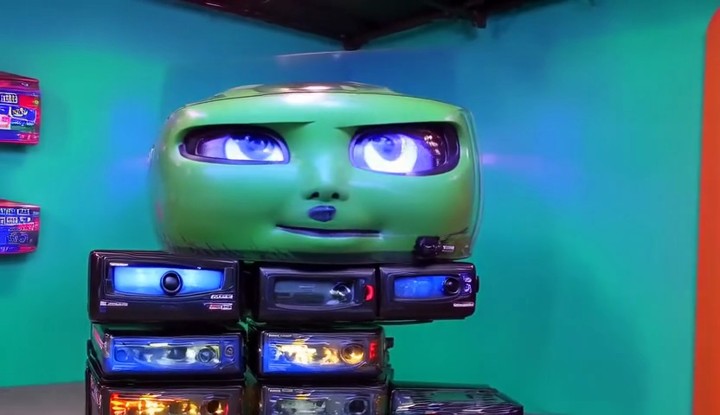}\hspace{-0.0037\textwidth}
\includegraphics[width=0.165\textwidth]{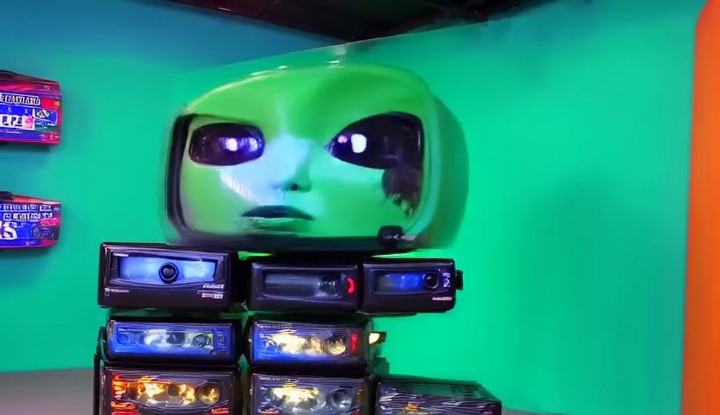}\hspace{-0.0037\textwidth}
\includegraphics[width=0.165\textwidth]{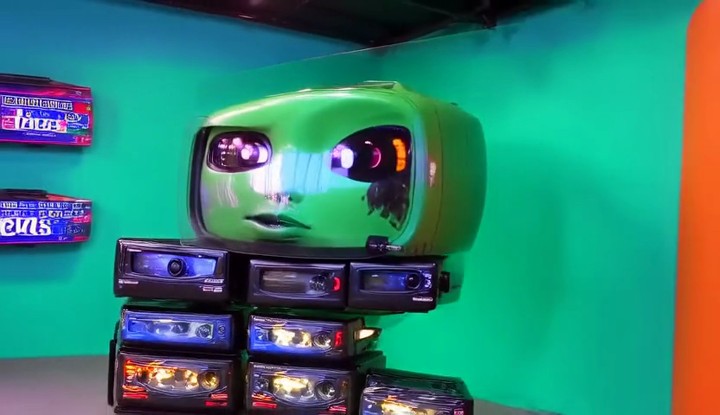}
\vspace{-0.5em}

\end{subfigure}

\vspace{0.2cm}

\begin{subfigure}{\textwidth}
\centering
\textbf{\large TurboDiffusion} ~~\textit{\large Latency: \bf \red{1.9s}}\\
\vspace{0.1cm}

\includegraphics[width=0.165\textwidth]{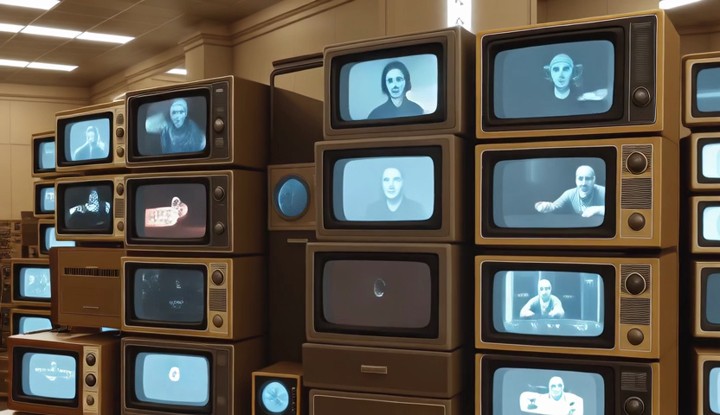}\hspace{-0.0037\textwidth}
\includegraphics[width=0.165\textwidth]{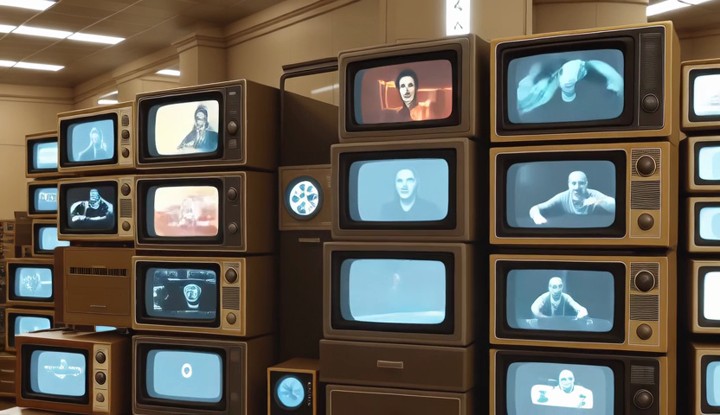}\hspace{-0.0037\textwidth}
\includegraphics[width=0.165\textwidth]{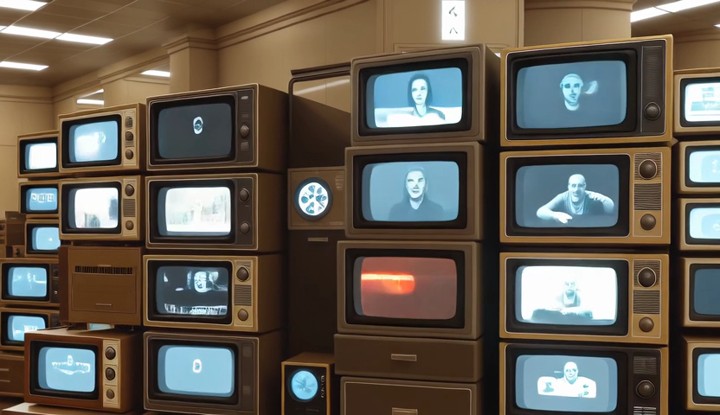}\hspace{-0.0037\textwidth}
\includegraphics[width=0.165\textwidth]{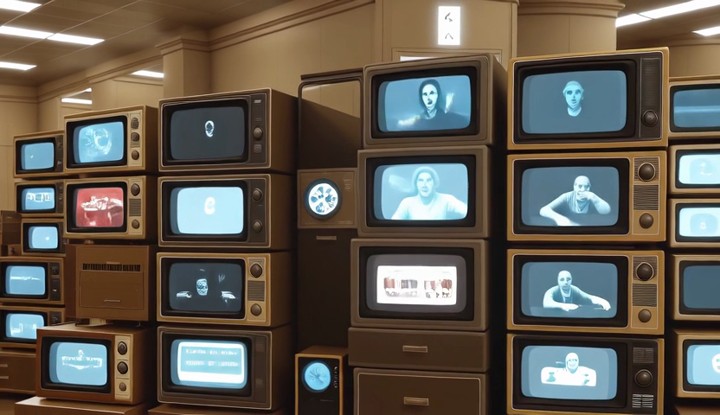}\hspace{-0.0037\textwidth}
\includegraphics[width=0.165\textwidth]{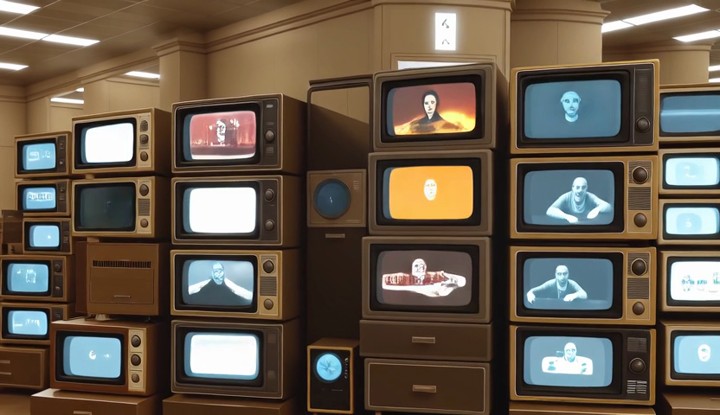}\hspace{-0.0037\textwidth}
\includegraphics[width=0.165\textwidth]{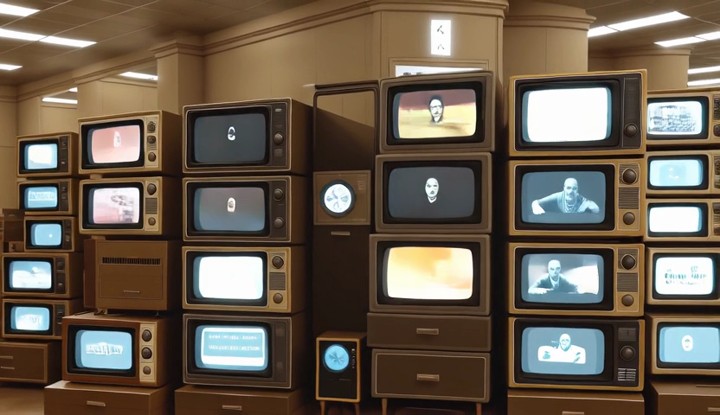}
\vspace{-0.5em}

\end{subfigure}

\vspace{-1em} \caption{5-second video generation on \texttt{Wan2.1-T2V-1.3B-480P} \textbf{\red{using a single RTX 5090}}.\\\textit{Prompt = "The camera rotates around a large stack of vintage televisions all showing different programs — 1950s sci-fi movies, horror movies, news, static, a 1970s sitcom, etc, set inside a large New York museum gallery."}}
\label{fig:comparison_1_3b_video_15}
\end{figure}

\begin{figure}[H]
\centering
\begin{subfigure}{\textwidth}
\centering
\textbf{\large Original} ~~\textit{\large Latency: 184s}\\
\vspace{0.1cm}

\includegraphics[width=0.165\textwidth]{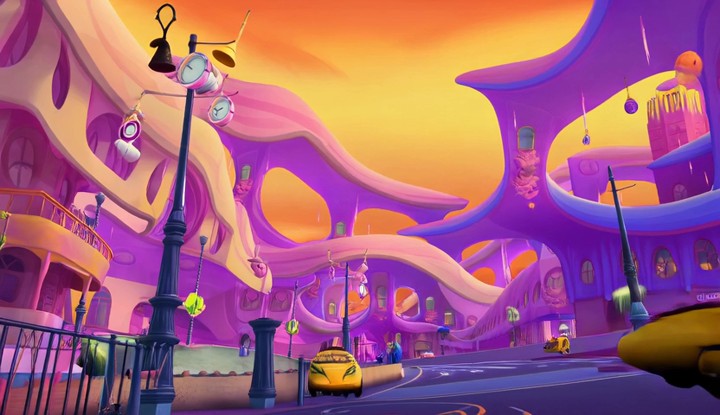}\hspace{-0.0037\textwidth}
\includegraphics[width=0.165\textwidth]{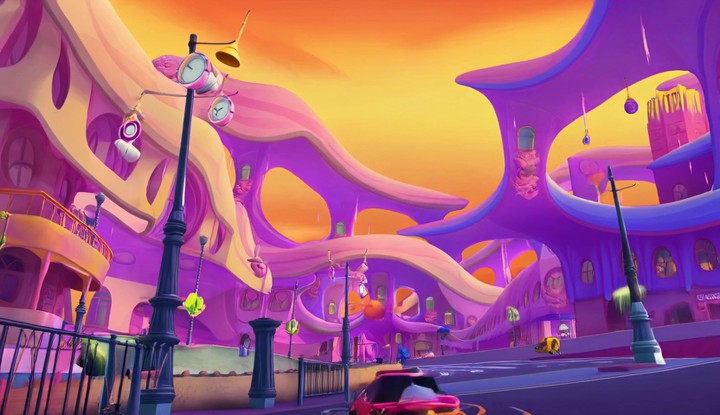}\hspace{-0.0037\textwidth}
\includegraphics[width=0.165\textwidth]{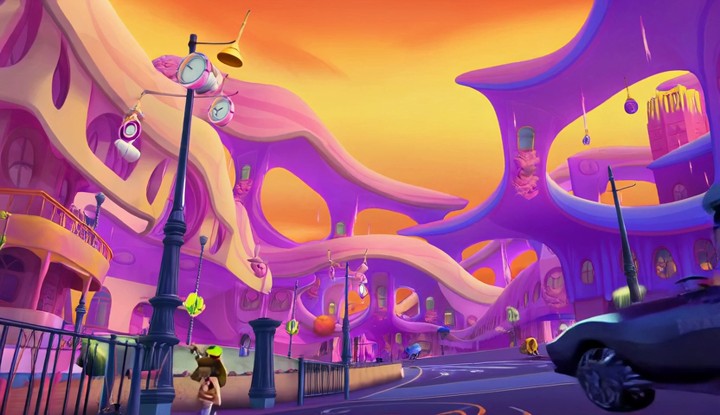}\hspace{-0.0037\textwidth}
\includegraphics[width=0.165\textwidth]{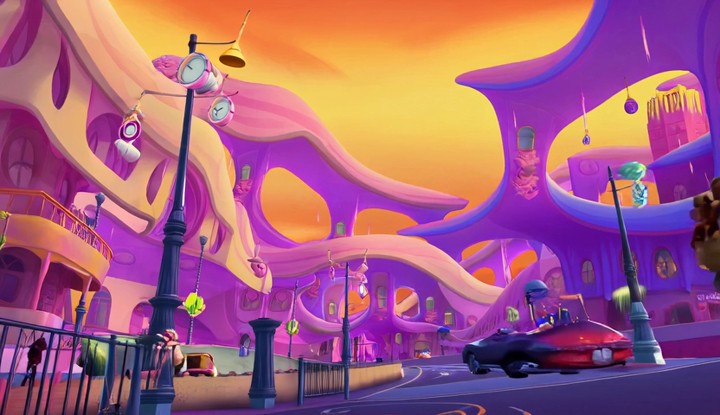}\hspace{-0.0037\textwidth}
\includegraphics[width=0.165\textwidth]{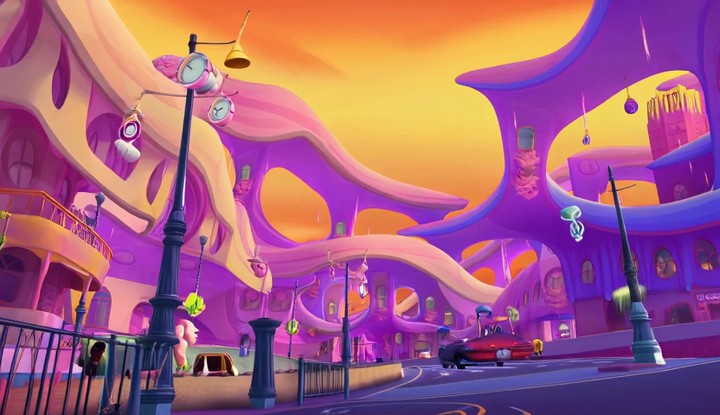}\hspace{-0.0037\textwidth}
\includegraphics[width=0.165\textwidth]{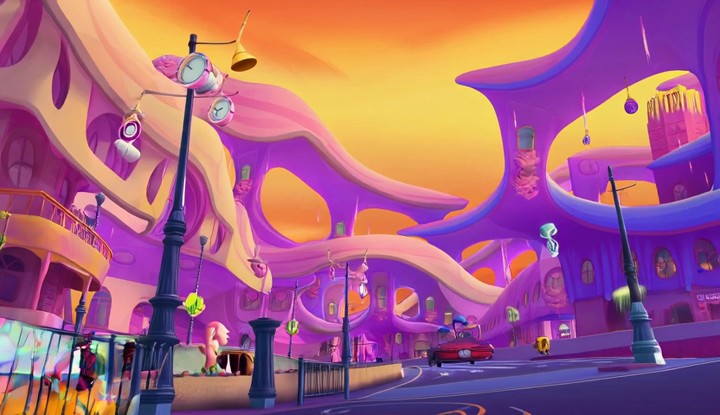}
\vspace{-0.5em}

\end{subfigure}

\vspace{0.2cm}

\begin{subfigure}{\textwidth}
\centering
\textbf{\large FastVideo} ~~\textit{\large Latency: 5.3s}\\
\vspace{0.1cm}

\includegraphics[width=0.165\textwidth]{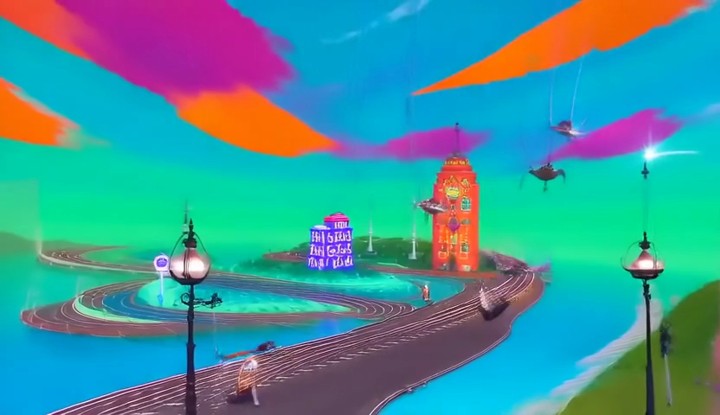}\hspace{-0.0037\textwidth}
\includegraphics[width=0.165\textwidth]{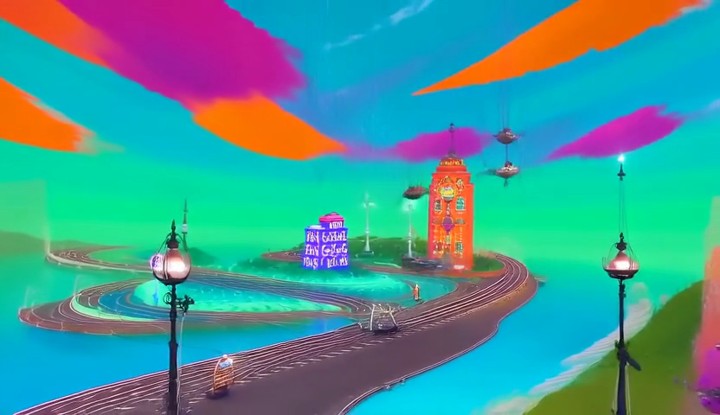}\hspace{-0.0037\textwidth}
\includegraphics[width=0.165\textwidth]{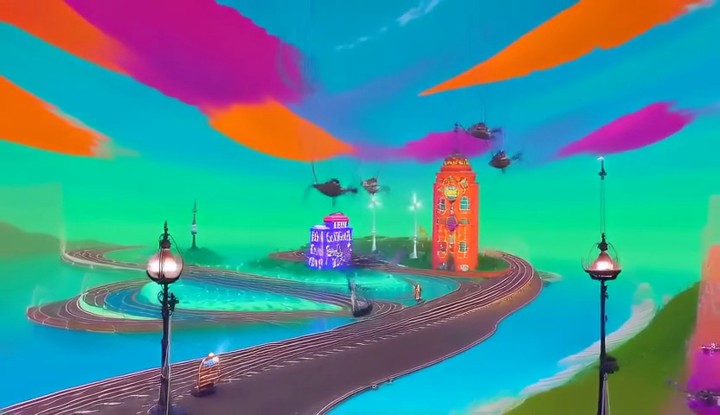}\hspace{-0.0037\textwidth}
\includegraphics[width=0.165\textwidth]{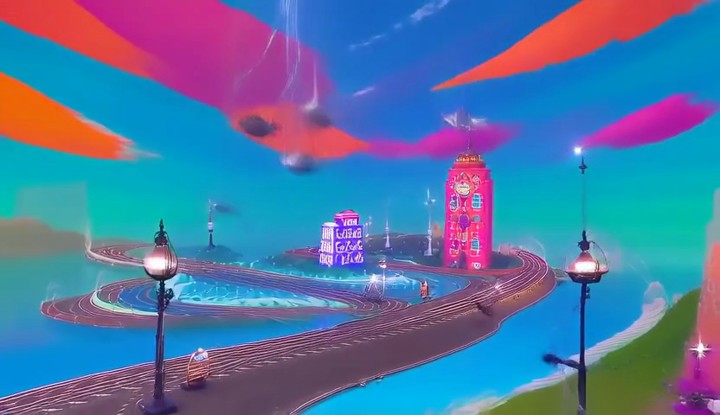}\hspace{-0.0037\textwidth}
\includegraphics[width=0.165\textwidth]{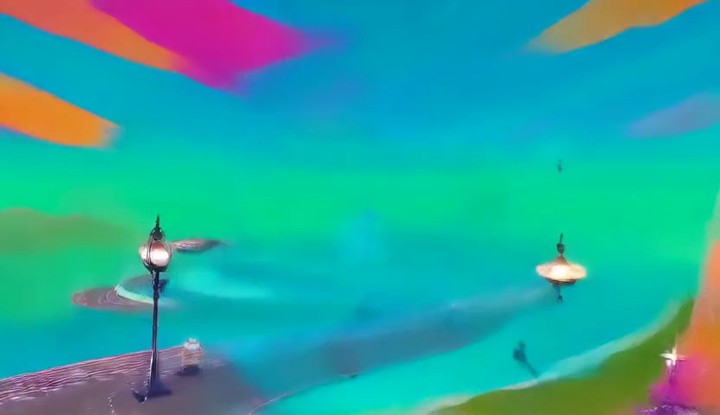}\hspace{-0.0037\textwidth}
\includegraphics[width=0.165\textwidth]{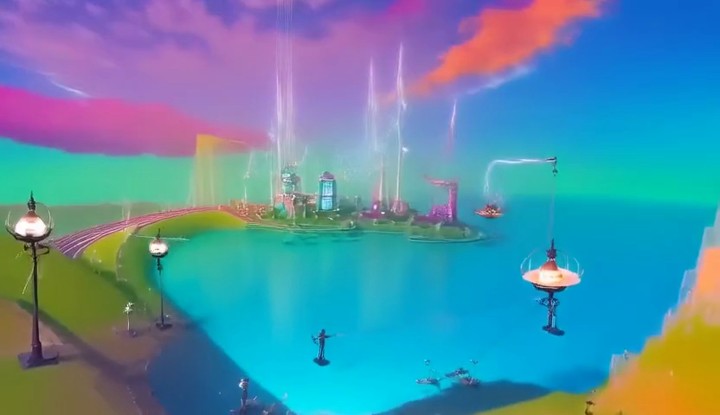}
\vspace{-0.5em}

\end{subfigure}

\vspace{0.2cm}

\begin{subfigure}{\textwidth}
\centering
\textbf{\large TurboDiffusion} ~~\textit{\large Latency: \bf \red{1.9s}}\\
\vspace{0.1cm}

\includegraphics[width=0.165\textwidth]{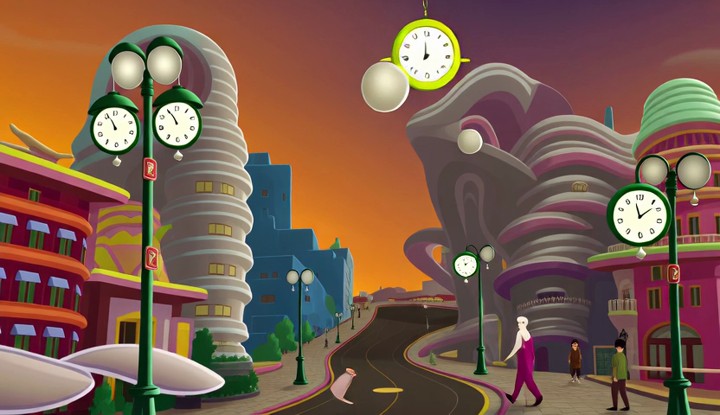}\hspace{-0.0037\textwidth}
\includegraphics[width=0.165\textwidth]{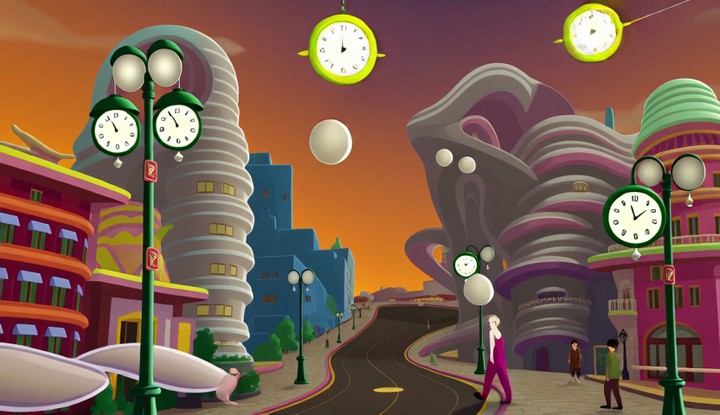}\hspace{-0.0037\textwidth}
\includegraphics[width=0.165\textwidth]{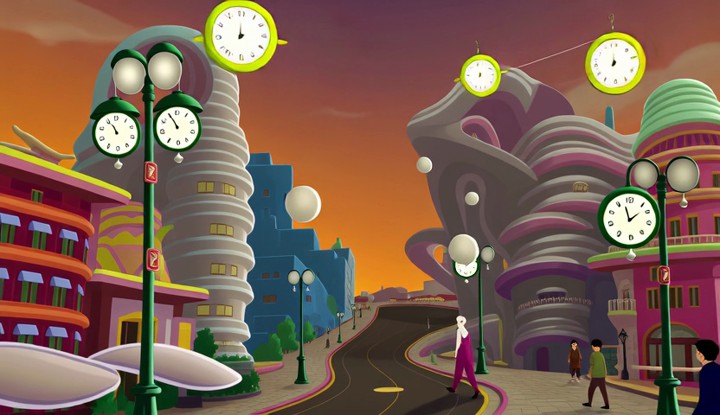}\hspace{-0.0037\textwidth}
\includegraphics[width=0.165\textwidth]{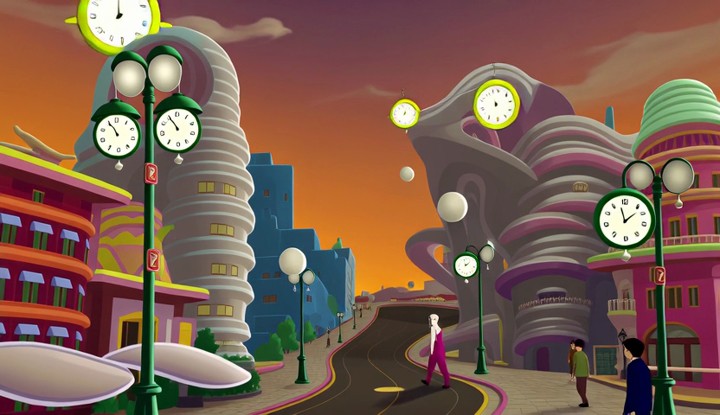}\hspace{-0.0037\textwidth}
\includegraphics[width=0.165\textwidth]{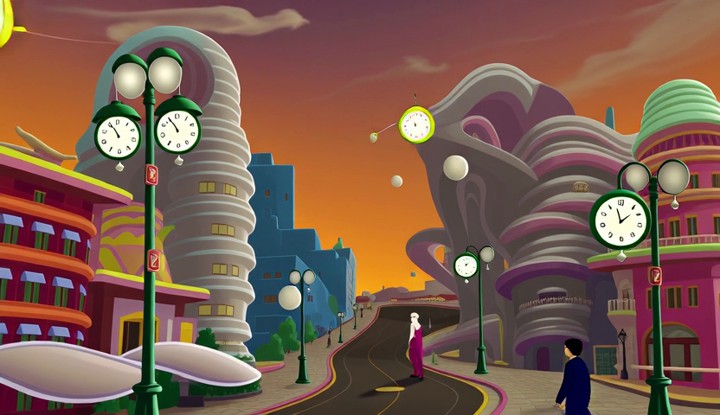}\hspace{-0.0037\textwidth}
\includegraphics[width=0.165\textwidth]{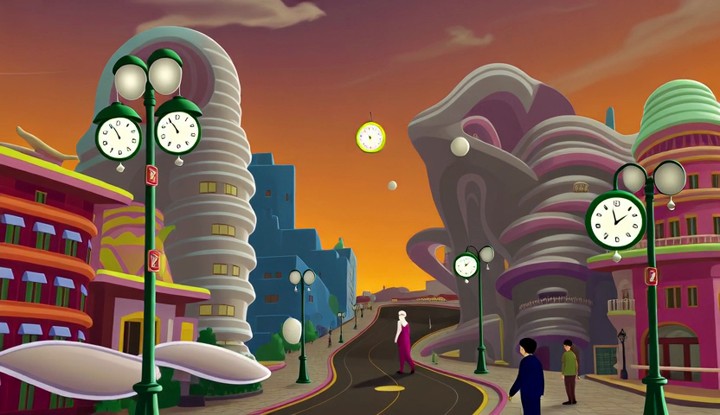}
\vspace{-0.5em}

\end{subfigure}

\vspace{-1em} \caption{5-second video generation on \texttt{Wan2.1-T2V-1.3B-480P} \textbf{\red{using a single RTX 5090}}.\\\textit{Prompt = "In the surreal world reminiscent of Salvador Dali's style, downtown Clowntown morphs into a mesmerizing dreamscape. The streets twist and bend in impossible ways, with melting clocks hanging from lampposts and floating objects suspended mid-air. Giant distorted buildings with soft, flowing edges loom over the scene, blending seamlessly into each other. The sky is a vivid blend of oranges, pinks, and purples, casting an ethereal glow over the bizarre landscape. Characters move slowly and dreamily through the streets, their expressions detached and contemplative. The scene is captured as a wide-angle view, emphasizing the surreal and dreamlike atmosphere."}}
\label{fig:comparison_1_3b_video_14}
\end{figure}

\begin{figure}[H]
\centering
\begin{subfigure}{\textwidth}
\centering
\textbf{\large Original} ~~\textit{\large Latency: 184s}\\
\vspace{0.1cm}

\includegraphics[width=0.165\textwidth]{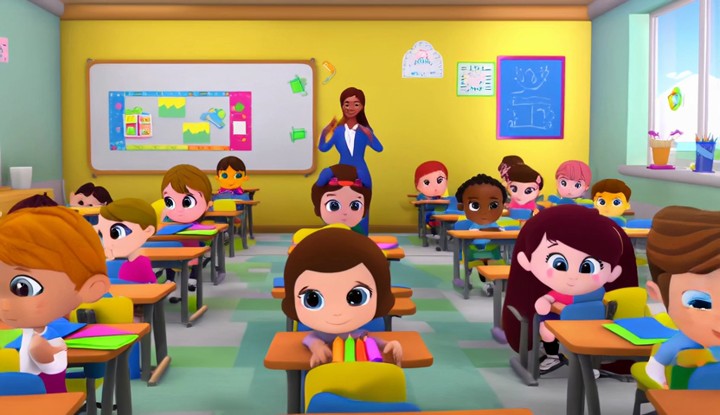}\hspace{-0.0037\textwidth}
\includegraphics[width=0.165\textwidth]{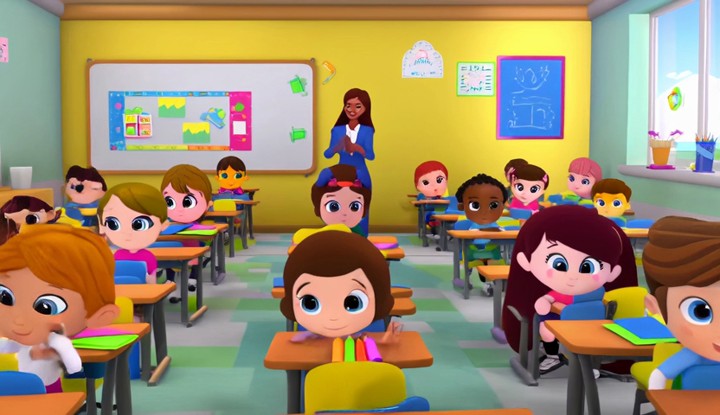}\hspace{-0.0037\textwidth}
\includegraphics[width=0.165\textwidth]{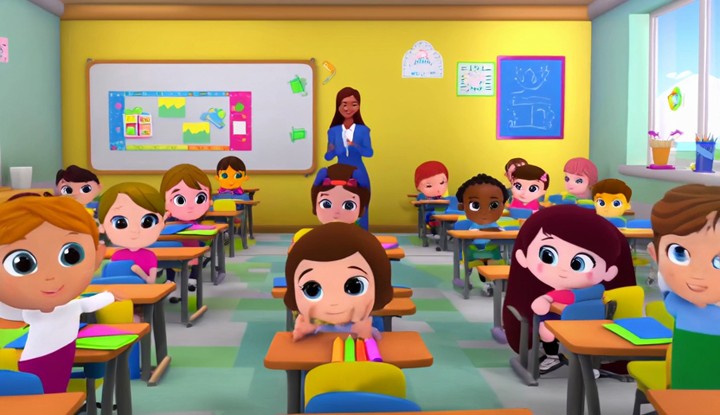}\hspace{-0.0037\textwidth}
\includegraphics[width=0.165\textwidth]{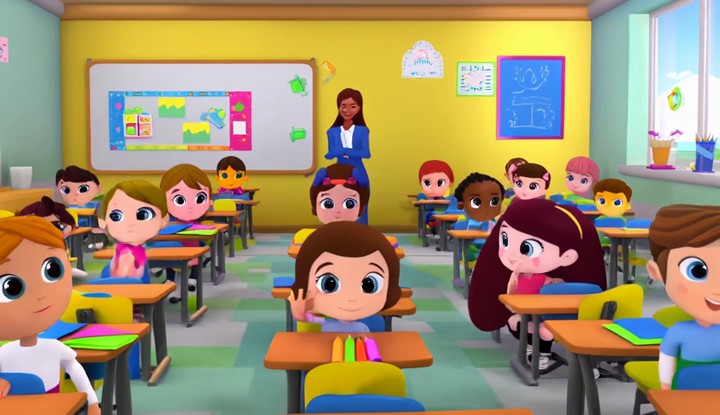}\hspace{-0.0037\textwidth}
\includegraphics[width=0.165\textwidth]{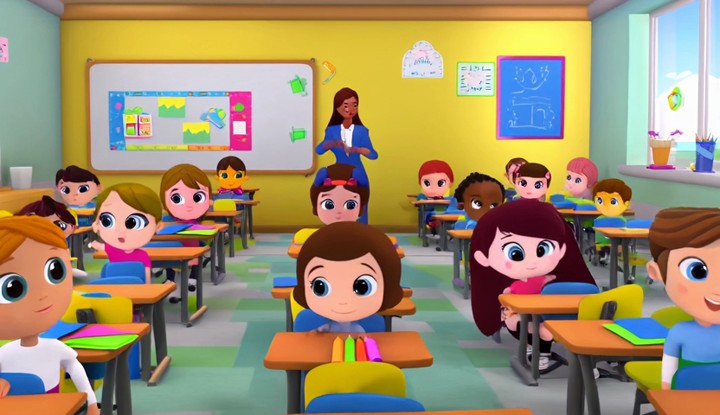}\hspace{-0.0037\textwidth}
\includegraphics[width=0.165\textwidth]{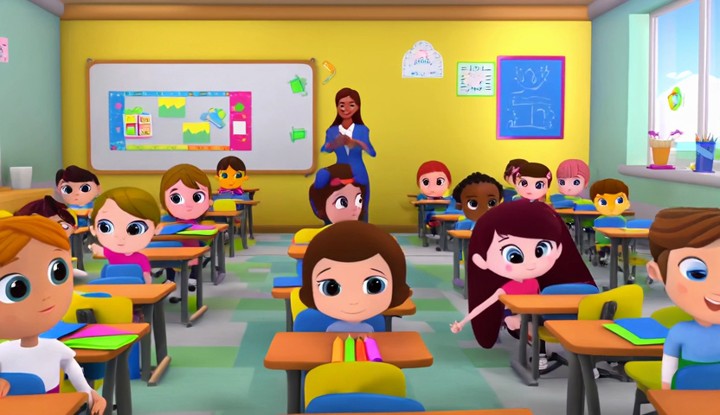}
\vspace{-0.5em}

\end{subfigure}

\vspace{0.2cm}

\begin{subfigure}{\textwidth}
\centering
\textbf{\large FastVideo} ~~\textit{\large Latency: 5.3s}\\
\vspace{0.1cm}

\includegraphics[width=0.165\textwidth]{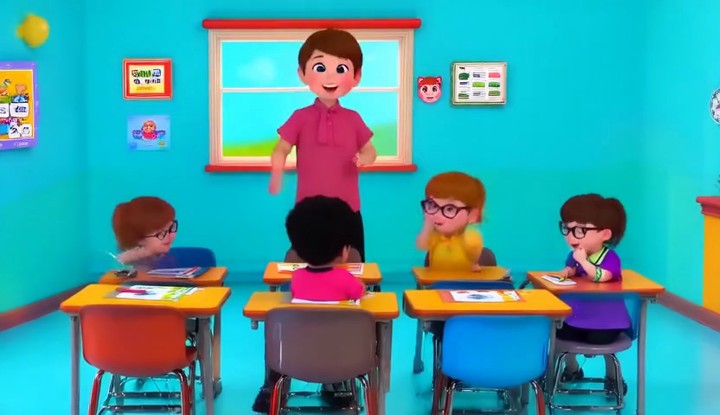}\hspace{-0.0037\textwidth}
\includegraphics[width=0.165\textwidth]{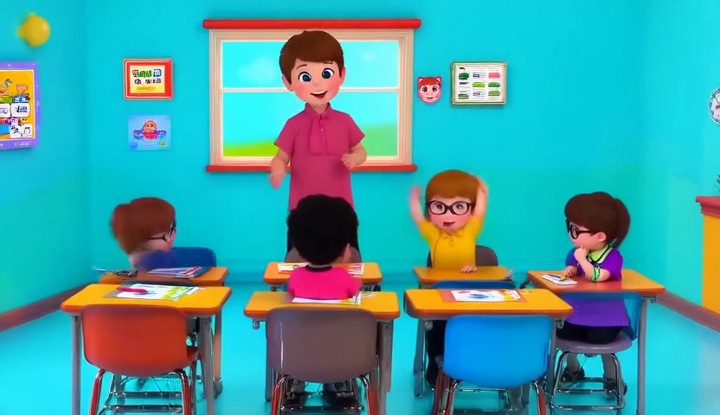}\hspace{-0.0037\textwidth}
\includegraphics[width=0.165\textwidth]{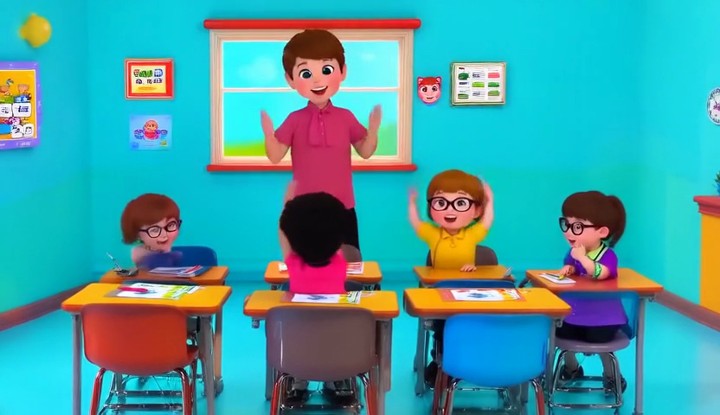}\hspace{-0.0037\textwidth}
\includegraphics[width=0.165\textwidth]{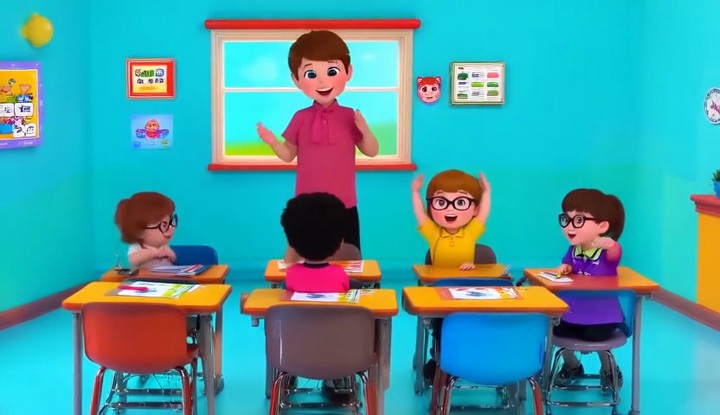}\hspace{-0.0037\textwidth}
\includegraphics[width=0.165\textwidth]{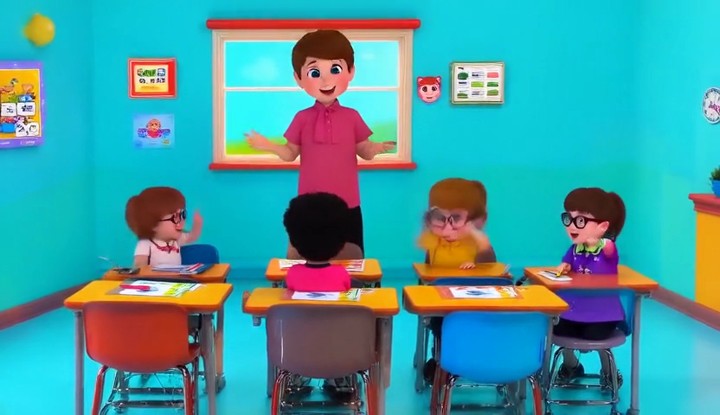}\hspace{-0.0037\textwidth}
\includegraphics[width=0.165\textwidth]{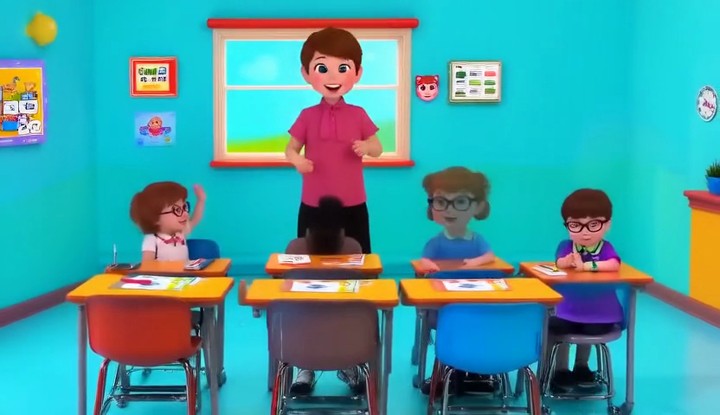}
\vspace{-0.5em}

\end{subfigure}

\vspace{0.2cm}

\begin{subfigure}{\textwidth}
\centering
\textbf{\large TurboDiffusion} ~~\textit{\large Latency: \bf \red{1.9s}}\\
\vspace{0.1cm}

\includegraphics[width=0.165\textwidth]{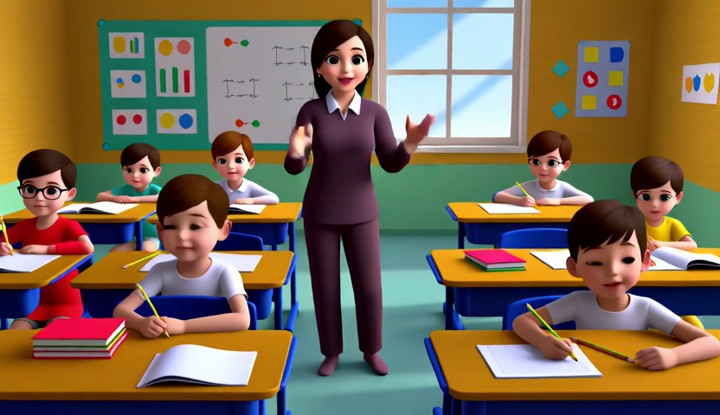}\hspace{-0.0037\textwidth}
\includegraphics[width=0.165\textwidth]{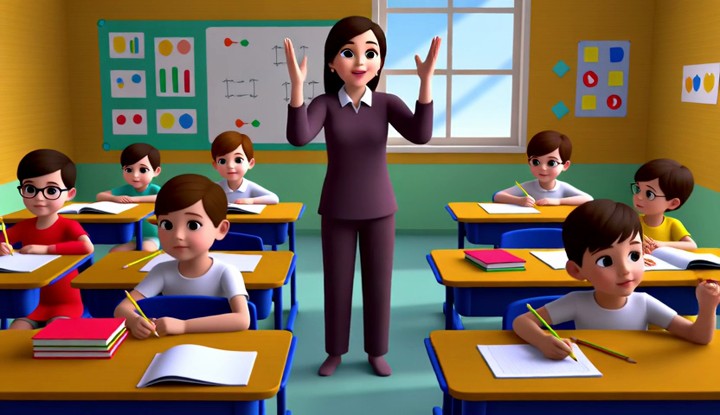}\hspace{-0.0037\textwidth}
\includegraphics[width=0.165\textwidth]{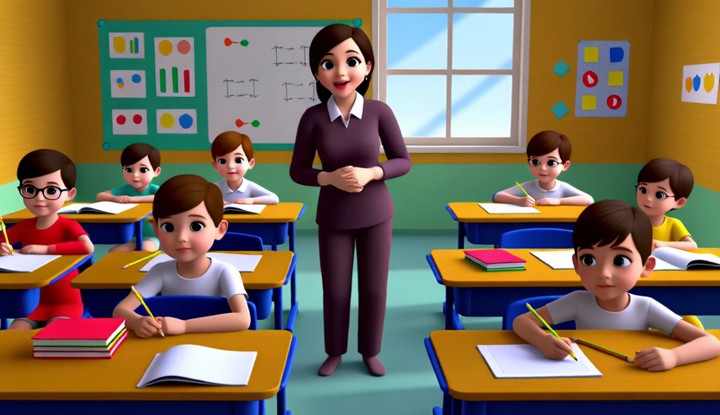}\hspace{-0.0037\textwidth}
\includegraphics[width=0.165\textwidth]{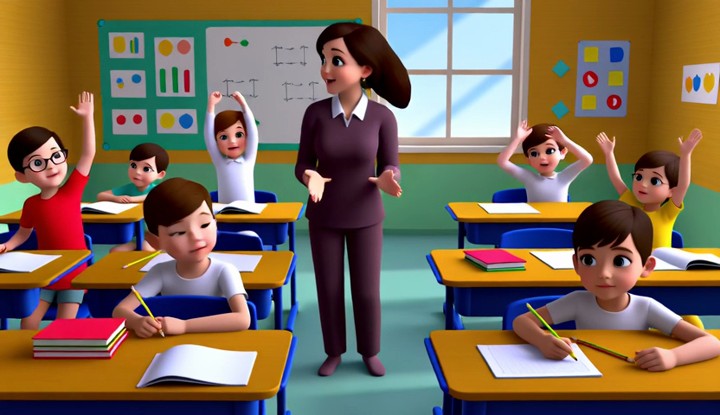}\hspace{-0.0037\textwidth}
\includegraphics[width=0.165\textwidth]{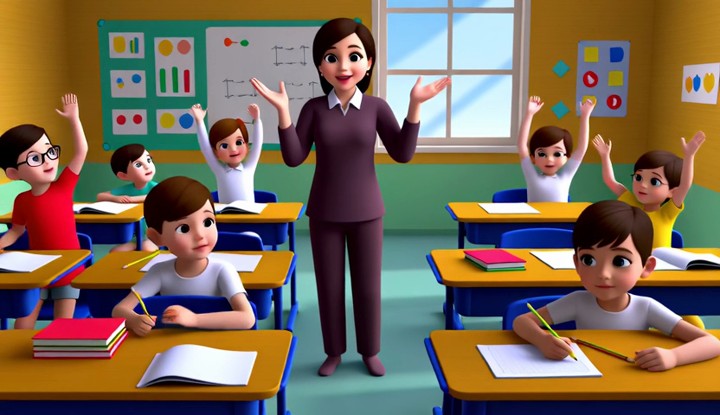}\hspace{-0.0037\textwidth}
\includegraphics[width=0.165\textwidth]{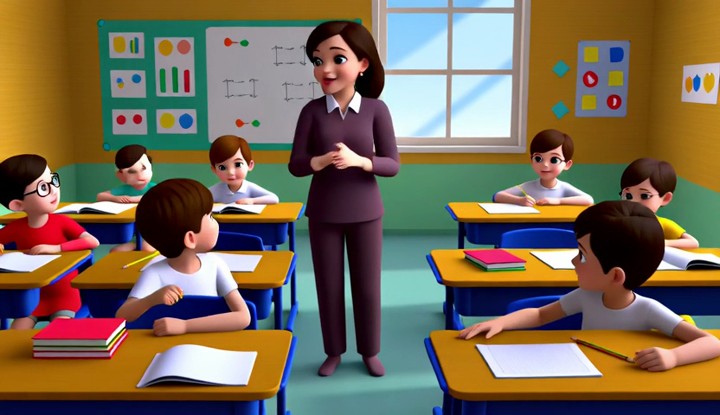}
\vspace{-0.5em}

\end{subfigure}

\vspace{-1em} \caption{5-second video generation on \texttt{Wan2.1-T2V-1.3B-480P} \textbf{\red{using a single RTX 5090}}.\\\textit{Prompt = "A 3D classroom scene filled with small children aged between 6 and 8 years old, all attentively sitting at their desks. A dedicated teacher stands in front of the class, engaged in lively interaction, using gestures and facial expressions to convey enthusiasm. \red{The classroom is brightly lit, with colorful posters and charts adorning the walls. Each child has a focused expression, some nodding along, others raising their hands to ask questions. The desks are neatly arranged in rows, with books and pencils organized on each desk. The background showcases a window with sunlight streaming in, casting gentle shadows across the room. Wide shot, capturing the entire classroom from a frontal perspective.}"}}
\label{fig:comparison_1_3b_video_1}
\end{figure}

\begin{figure}[H]
\centering
\begin{subfigure}{\textwidth}
\centering
\textbf{\large Original} ~~\textit{\large Latency: 184s}\\
\vspace{0.1cm}

\includegraphics[width=0.165\textwidth]{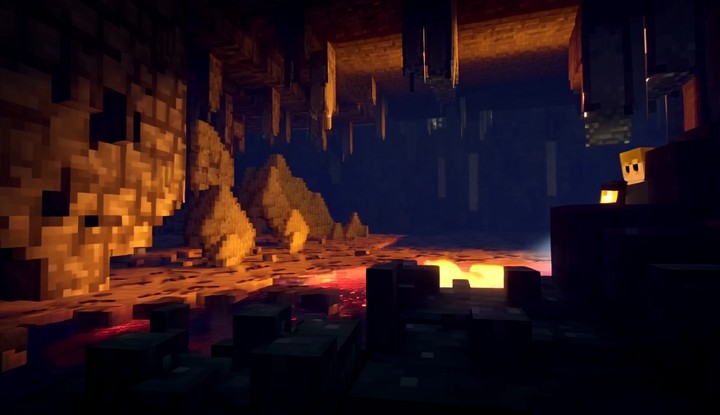}\hspace{-0.0037\textwidth}
\includegraphics[width=0.165\textwidth]{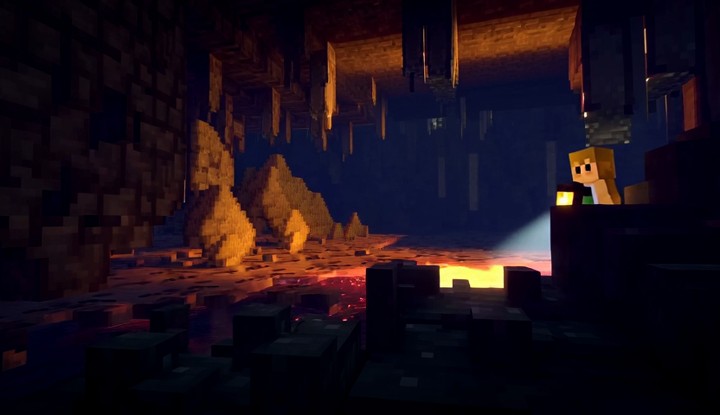}\hspace{-0.0037\textwidth}
\includegraphics[width=0.165\textwidth]{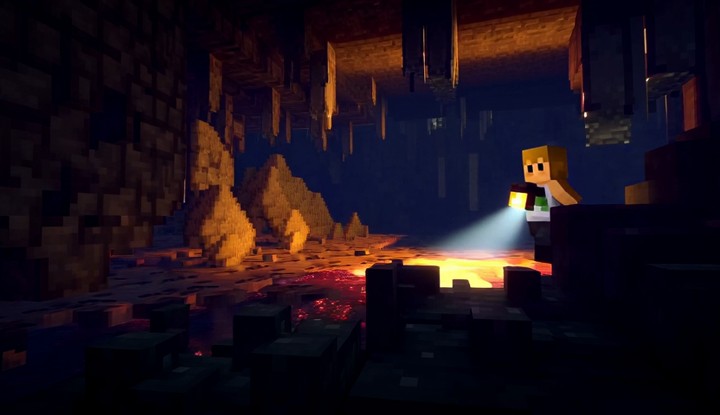}\hspace{-0.0037\textwidth}
\includegraphics[width=0.165\textwidth]{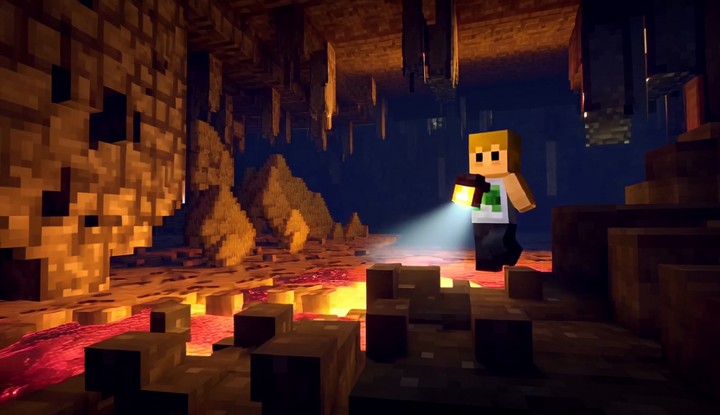}\hspace{-0.0037\textwidth}
\includegraphics[width=0.165\textwidth]{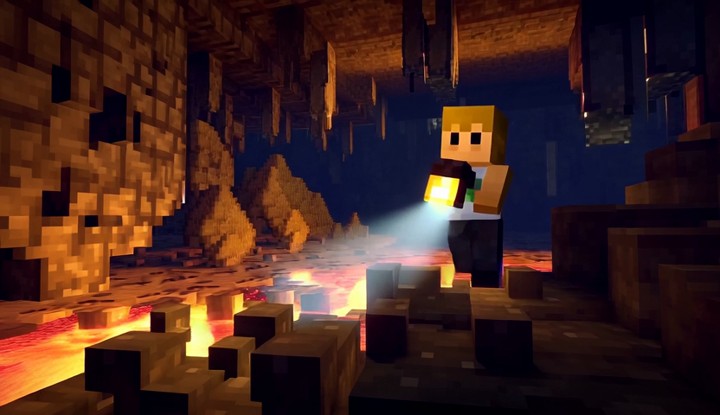}\hspace{-0.0037\textwidth}
\includegraphics[width=0.165\textwidth]{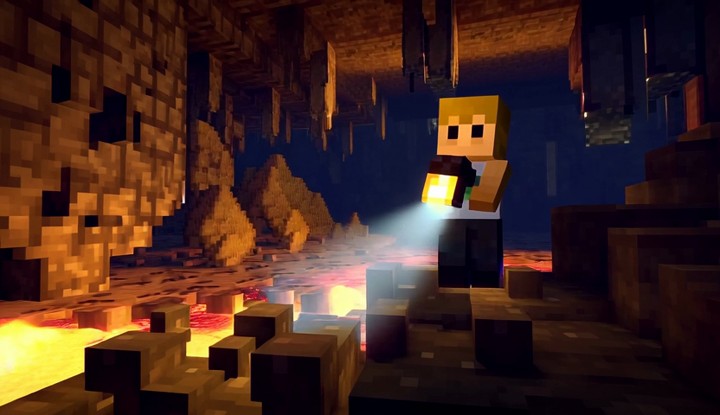}
\vspace{-0.5em}

\end{subfigure}

\vspace{0.2cm}

\begin{subfigure}{\textwidth}
\centering
\textbf{\large FastVideo} ~~\textit{\large Latency: 5.3s}\\
\vspace{0.1cm}

\includegraphics[width=0.165\textwidth]{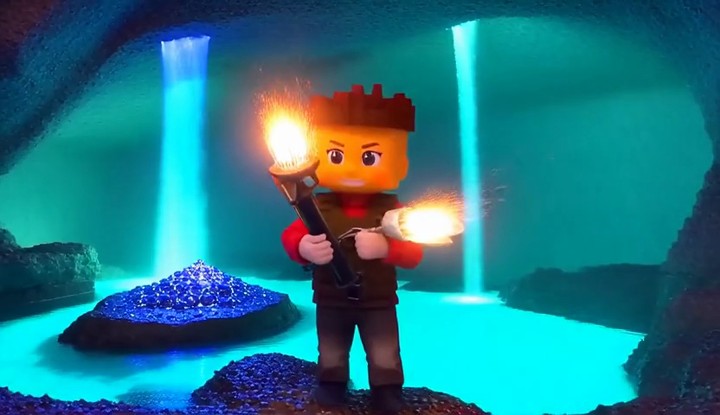}\hspace{-0.0037\textwidth}
\includegraphics[width=0.165\textwidth]{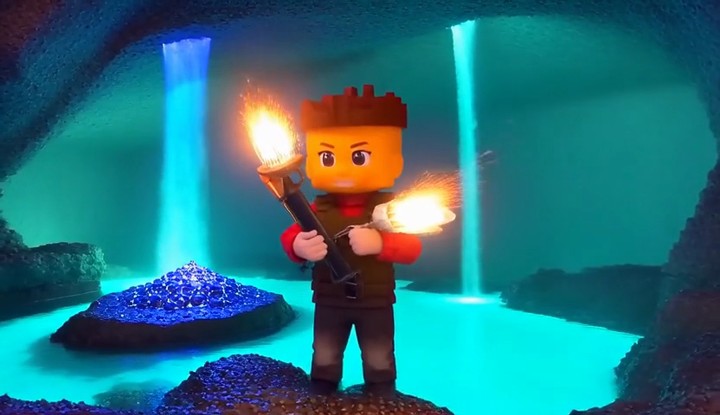}\hspace{-0.0037\textwidth}
\includegraphics[width=0.165\textwidth]{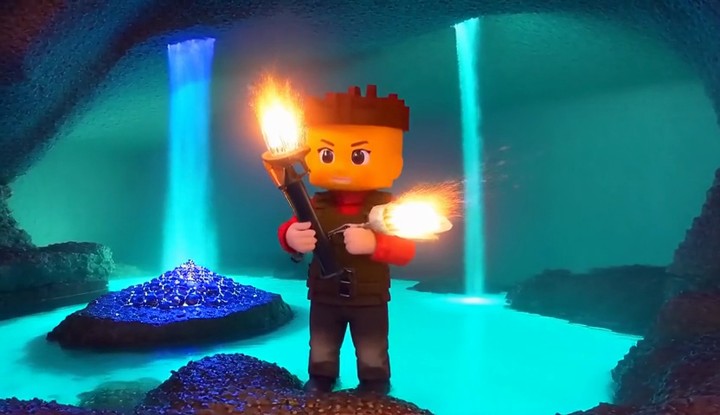}\hspace{-0.0037\textwidth}
\includegraphics[width=0.165\textwidth]{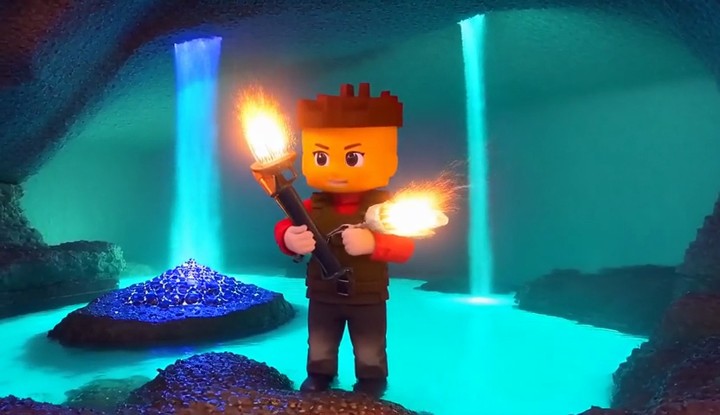}\hspace{-0.0037\textwidth}
\includegraphics[width=0.165\textwidth]{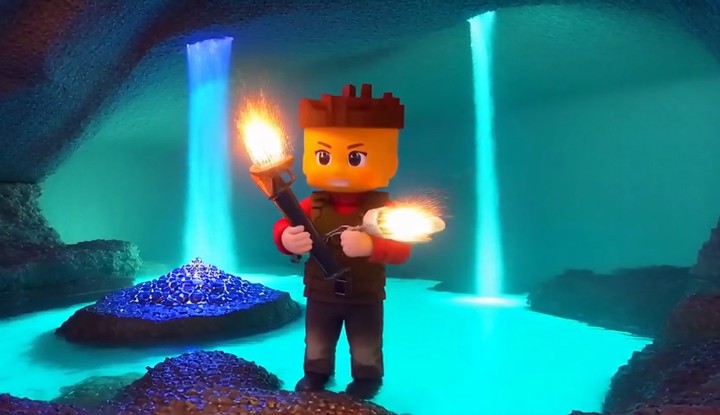}\hspace{-0.0037\textwidth}
\includegraphics[width=0.165\textwidth]{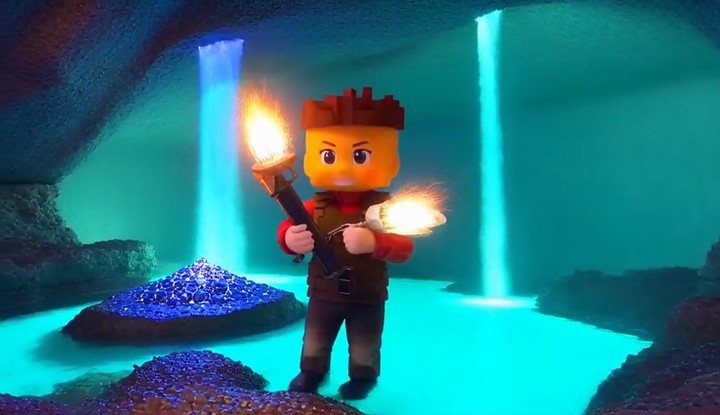}
\vspace{-0.5em}

\end{subfigure}

\vspace{0.2cm}

\begin{subfigure}{\textwidth}
\centering
\textbf{\large TurboDiffusion} ~~\textit{\large Latency: \bf \red{1.9s}}\\
\vspace{0.1cm}

\includegraphics[width=0.165\textwidth]{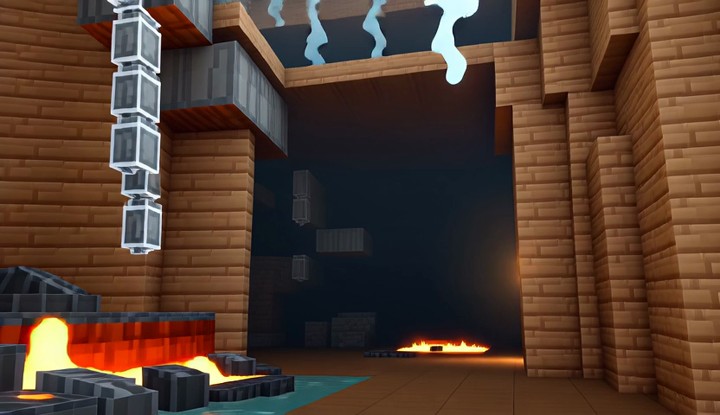}\hspace{-0.0037\textwidth}
\includegraphics[width=0.165\textwidth]{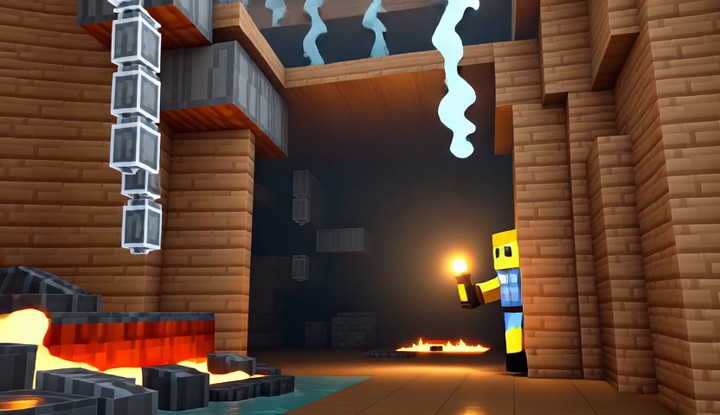}\hspace{-0.0037\textwidth}
\includegraphics[width=0.165\textwidth]{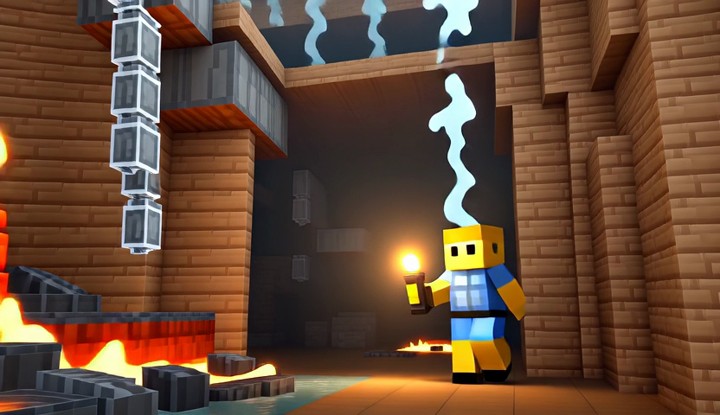}\hspace{-0.0037\textwidth}
\includegraphics[width=0.165\textwidth]{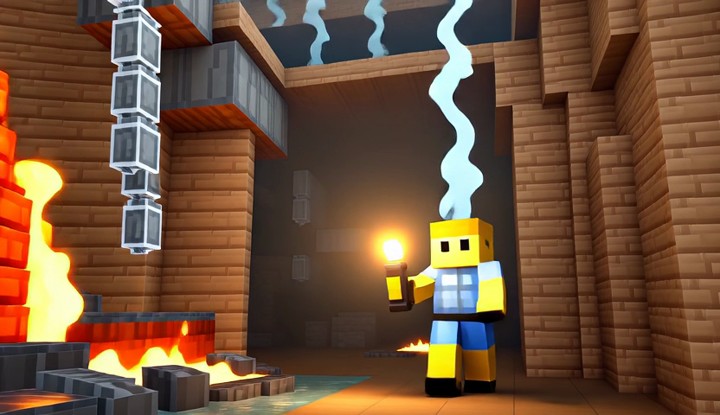}\hspace{-0.0037\textwidth}
\includegraphics[width=0.165\textwidth]{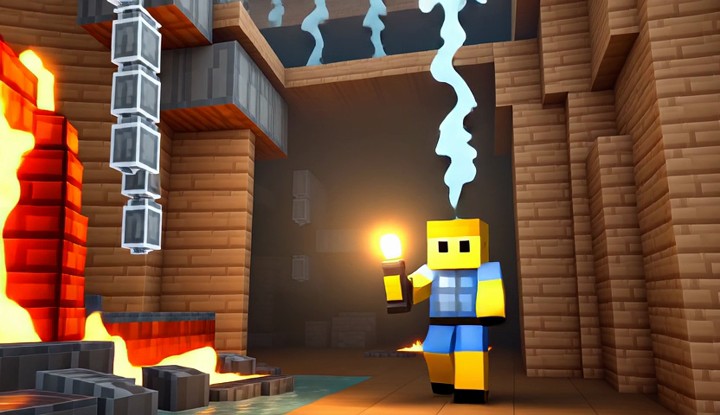}\hspace{-0.0037\textwidth}
\includegraphics[width=0.165\textwidth]{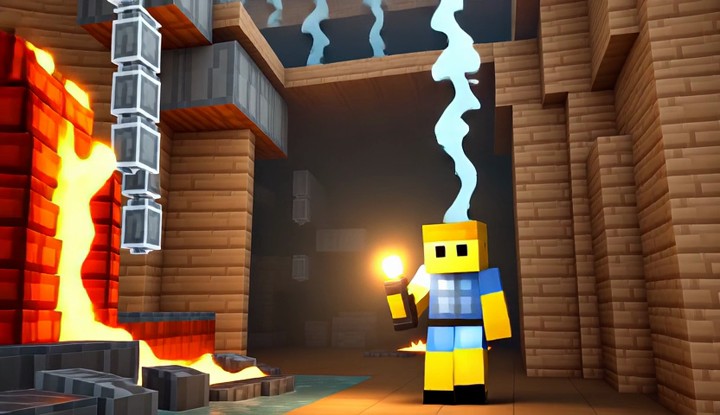}
\vspace{-0.5em}

\end{subfigure}

\vspace{-1em} \caption{5-second video generation on \texttt{Wan2.1-T2V-1.3B-480P} \textbf{\red{using a single RTX 5090}}.\\\textit{Prompt = "A Minecraft player character holding a torch enters a massive underground cave. \red{The torchlight flickers against jagged stone walls}, illuminating patches of iron and diamond ores embedded in the rock. Stalactites hang from the ceiling, \red{lava flows in glowing streams nearby}, and the faint sound of water dripping echoes through the cavern."}}
\label{fig:comparison_1_3b_video_2}
\end{figure}

\begin{figure}[H]
\centering
\begin{subfigure}{\textwidth}
\centering
\textbf{\large Original} ~~\textit{\large Latency: 184s}\\
\vspace{0.1cm}

\includegraphics[width=0.165\textwidth]{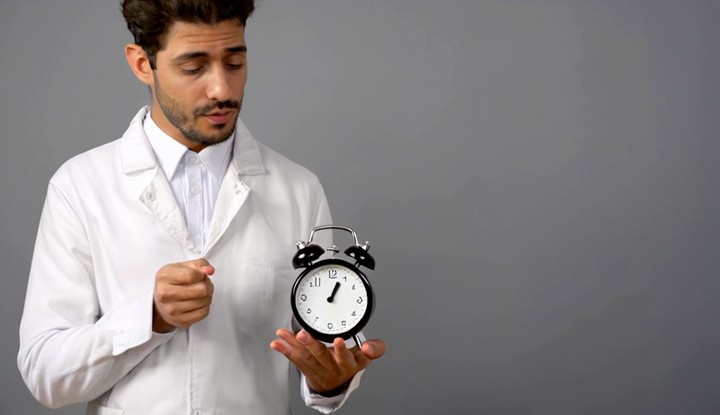}\hspace{-0.0037\textwidth}
\includegraphics[width=0.165\textwidth]{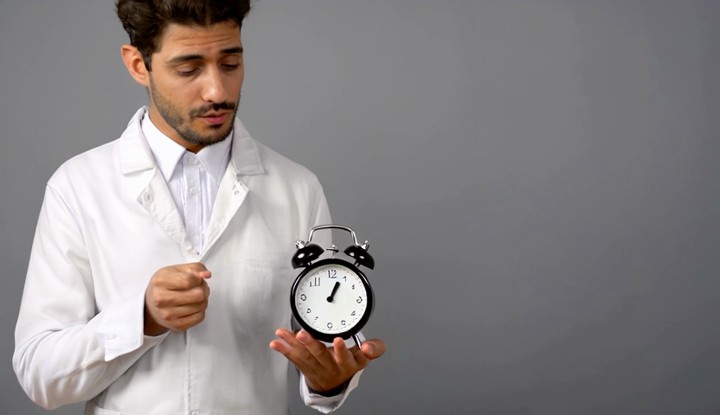}\hspace{-0.0037\textwidth}
\includegraphics[width=0.165\textwidth]{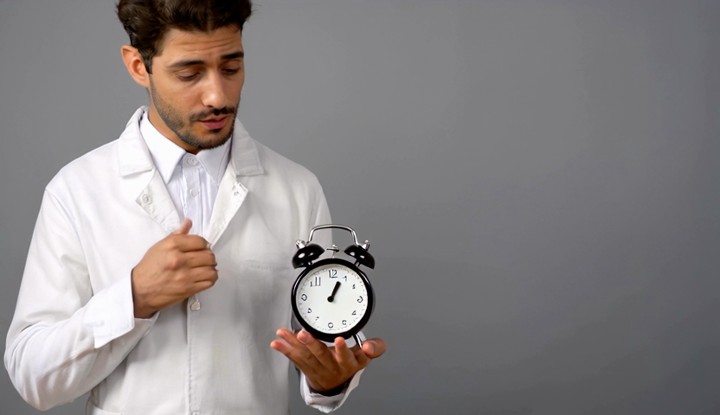}\hspace{-0.0037\textwidth}
\includegraphics[width=0.165\textwidth]{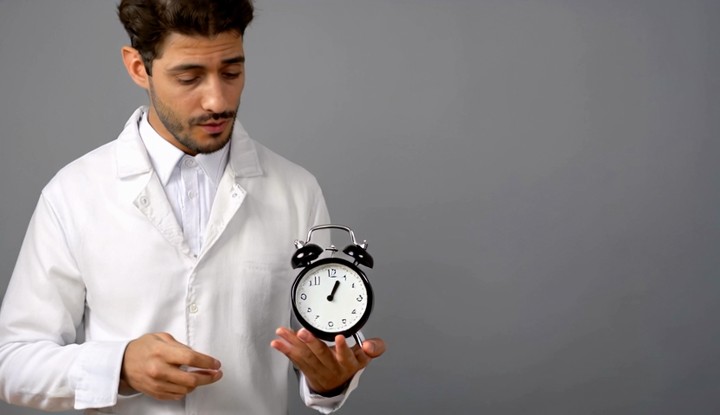}\hspace{-0.0037\textwidth}
\includegraphics[width=0.165\textwidth]{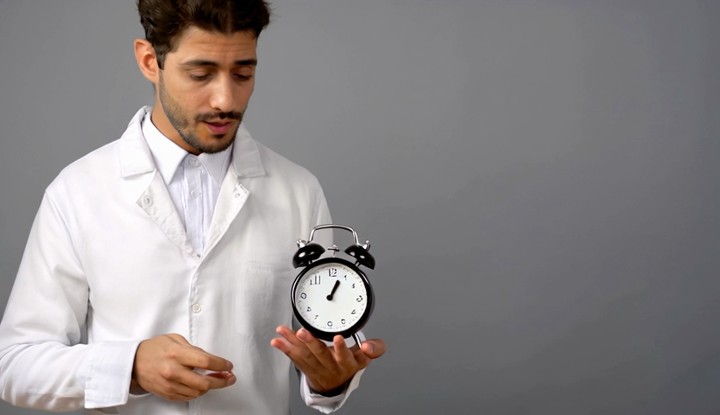}\hspace{-0.0037\textwidth}
\includegraphics[width=0.165\textwidth]{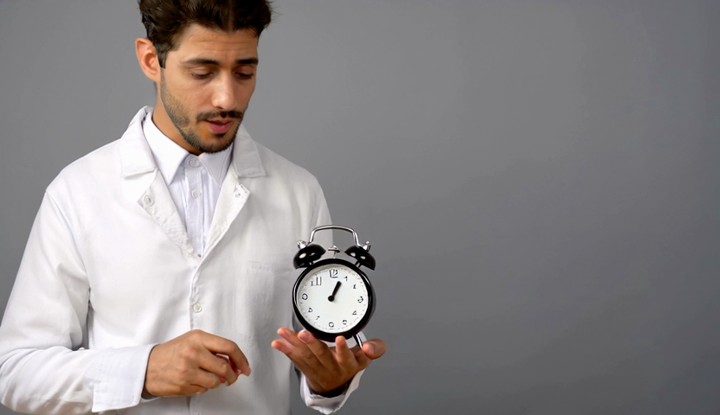}
\vspace{-0.5em}

\end{subfigure}

\vspace{0.2cm}

\begin{subfigure}{\textwidth}
\centering
\textbf{\large FastVideo} ~~\textit{\large Latency: 5.3s}\\
\vspace{0.1cm}

\includegraphics[width=0.165\textwidth]{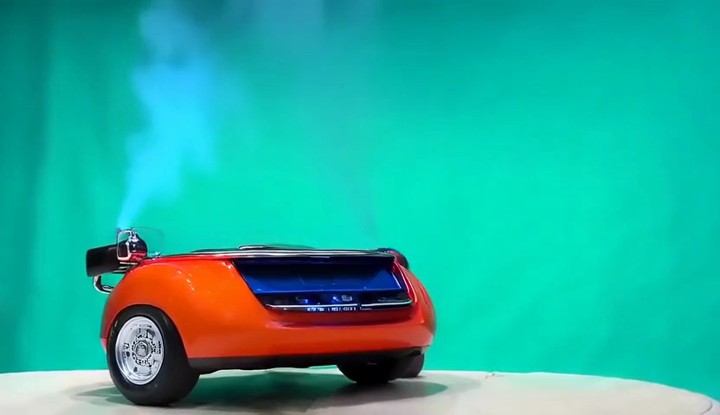}\hspace{-0.0037\textwidth}
\includegraphics[width=0.165\textwidth]{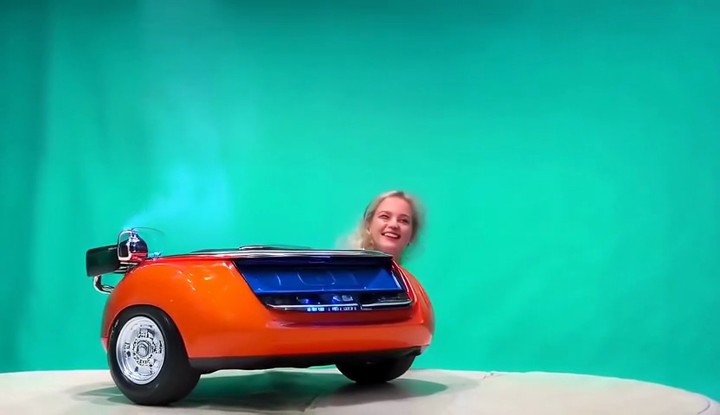}\hspace{-0.0037\textwidth}
\includegraphics[width=0.165\textwidth]{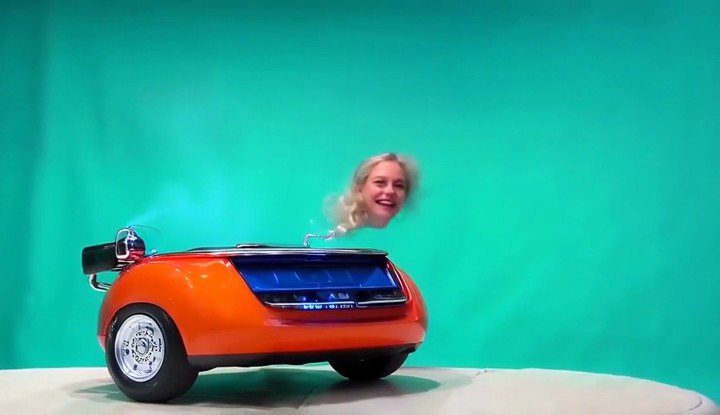}\hspace{-0.0037\textwidth}
\includegraphics[width=0.165\textwidth]{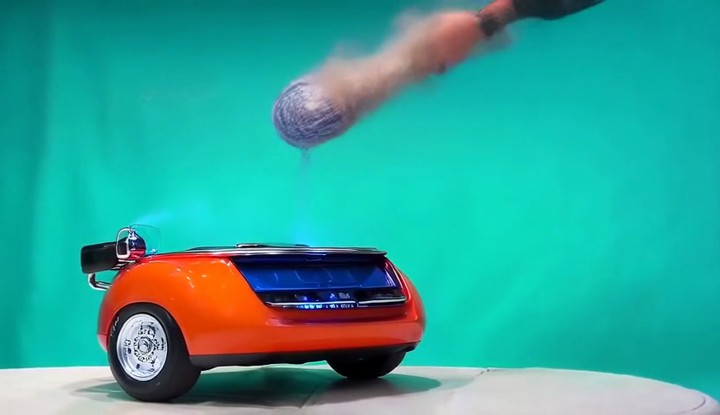}\hspace{-0.0037\textwidth}
\includegraphics[width=0.165\textwidth]{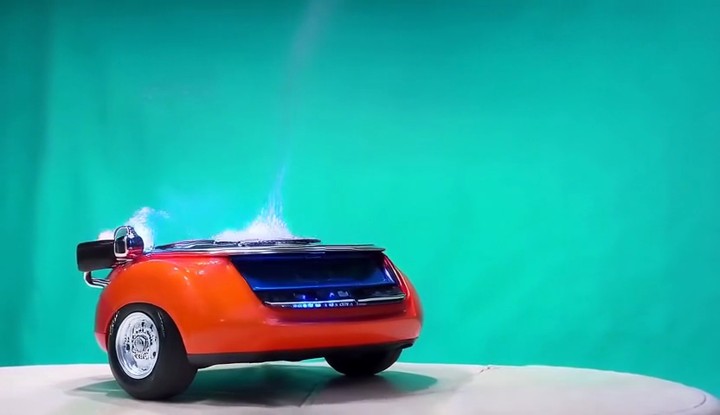}\hspace{-0.0037\textwidth}
\includegraphics[width=0.165\textwidth]{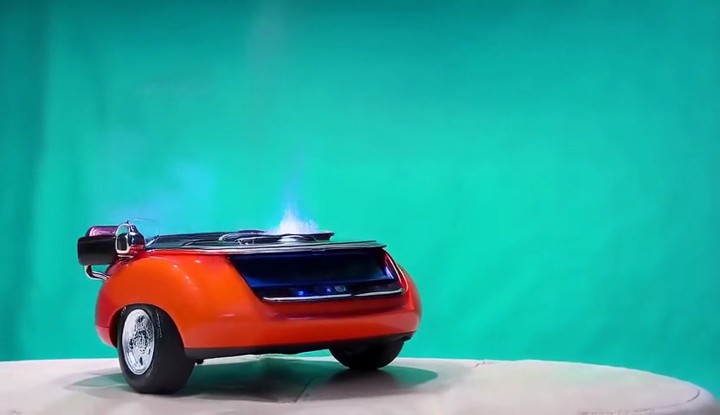}
\vspace{-0.5em}

\end{subfigure}

\vspace{0.2cm}

\begin{subfigure}{\textwidth}
\centering
\textbf{\large TurboDiffusion} ~~\textit{\large Latency: \bf \red{1.9s}}\\
\vspace{0.1cm}

\includegraphics[width=0.165\textwidth]{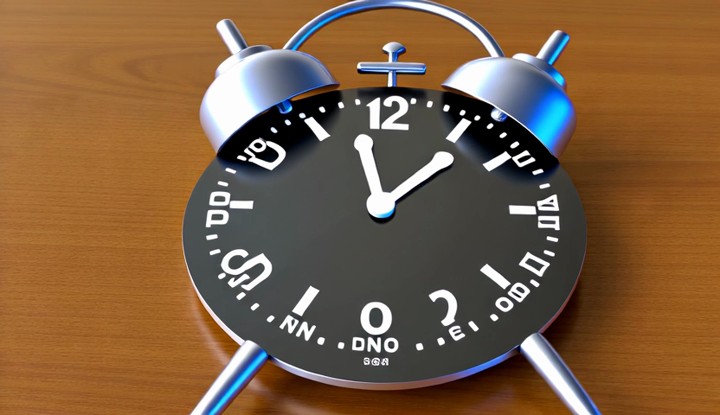}\hspace{-0.0037\textwidth}
\includegraphics[width=0.165\textwidth]{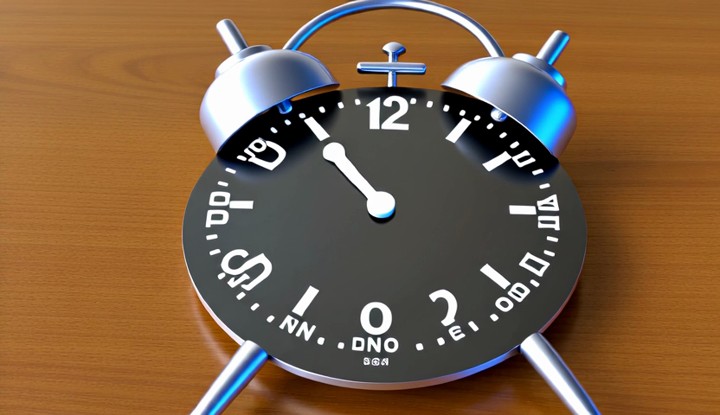}\hspace{-0.0037\textwidth}
\includegraphics[width=0.165\textwidth]{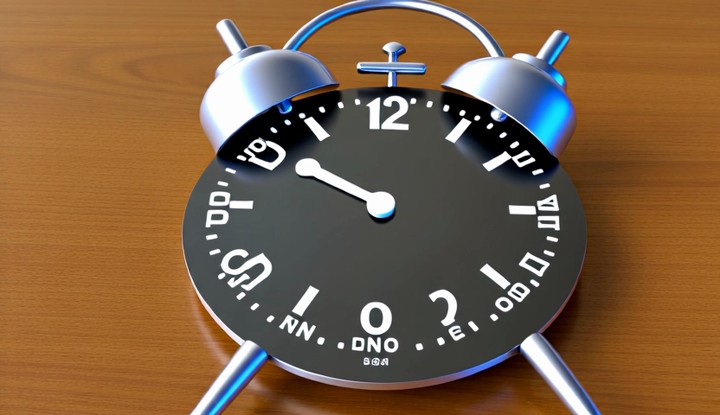}\hspace{-0.0037\textwidth}
\includegraphics[width=0.165\textwidth]{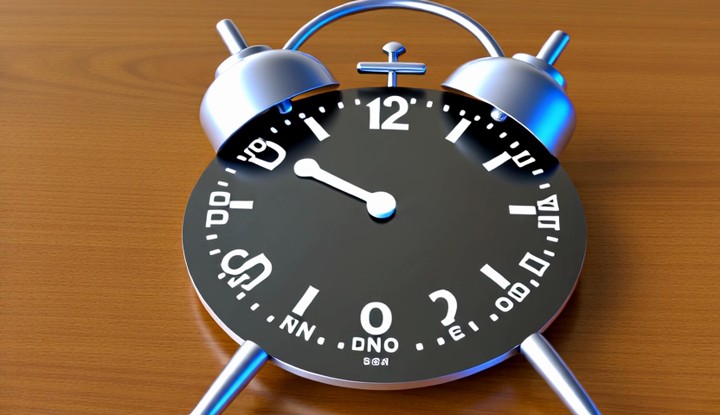}\hspace{-0.0037\textwidth}
\includegraphics[width=0.165\textwidth]{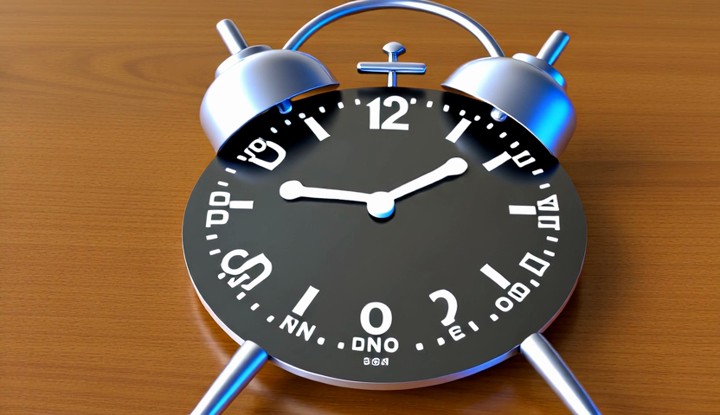}\hspace{-0.0037\textwidth}
\includegraphics[width=0.165\textwidth]{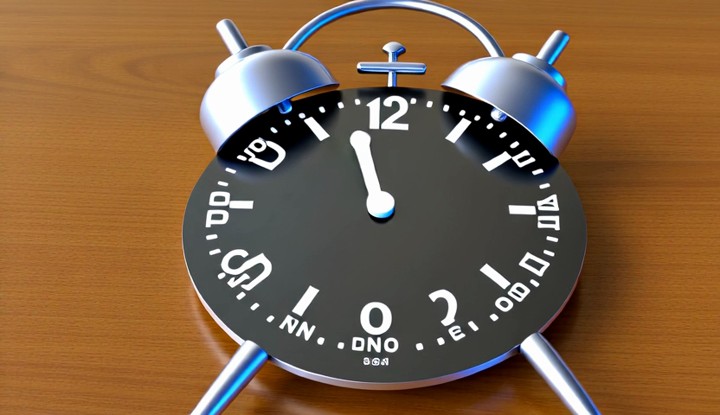}
\vspace{-0.5em}

\end{subfigure}

\vspace{-1em} \caption{5-second video generation on \texttt{Wan2.1-T2V-1.3B-480P} \textbf{\red{using a single RTX 5090}}.\\\textit{Prompt = "alarm clock"}.}
\label{fig:comparison_1_3b_video_13}
\end{figure}

\begin{figure}[H]
\centering
\begin{subfigure}{\textwidth}
\centering
\textbf{\large Original} ~~\textit{\large Latency: 184s}\\
\vspace{0.1cm}

\includegraphics[width=0.165\textwidth]{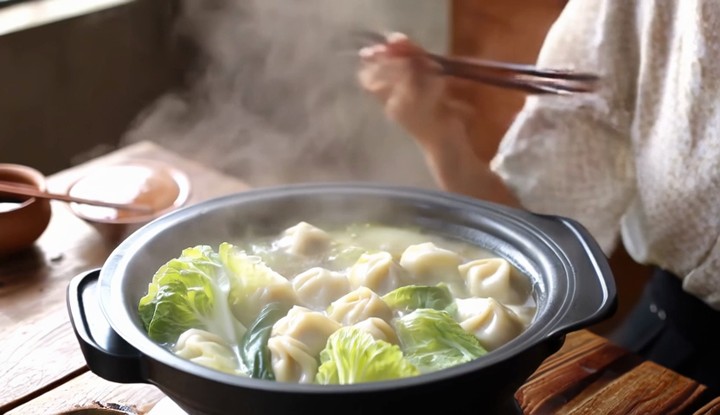}\hspace{-0.0037\textwidth}
\includegraphics[width=0.165\textwidth]{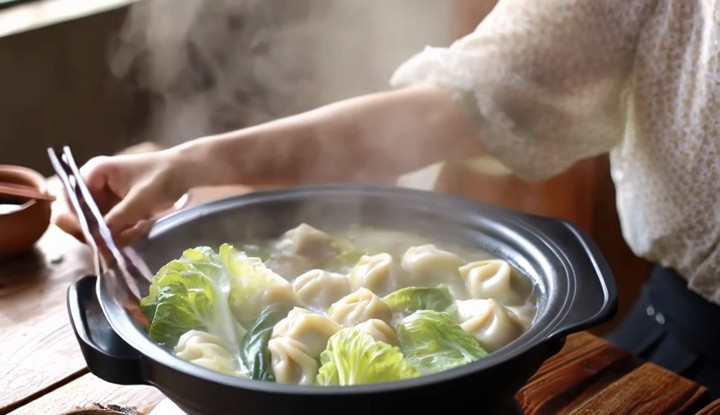}\hspace{-0.0037\textwidth}
\includegraphics[width=0.165\textwidth]{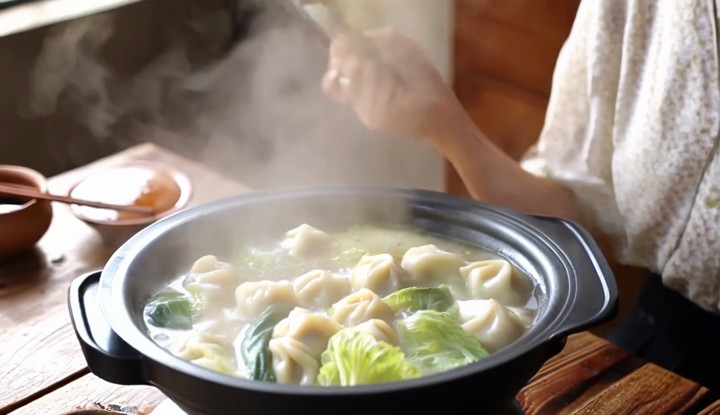}\hspace{-0.0037\textwidth}
\includegraphics[width=0.165\textwidth]{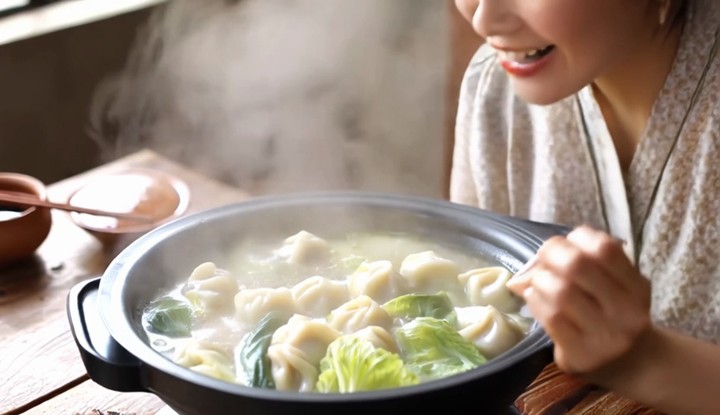}\hspace{-0.0037\textwidth}
\includegraphics[width=0.165\textwidth]{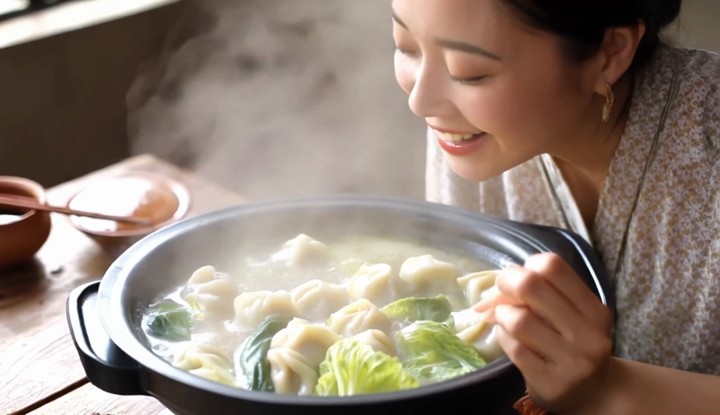}\hspace{-0.0037\textwidth}
\includegraphics[width=0.165\textwidth]{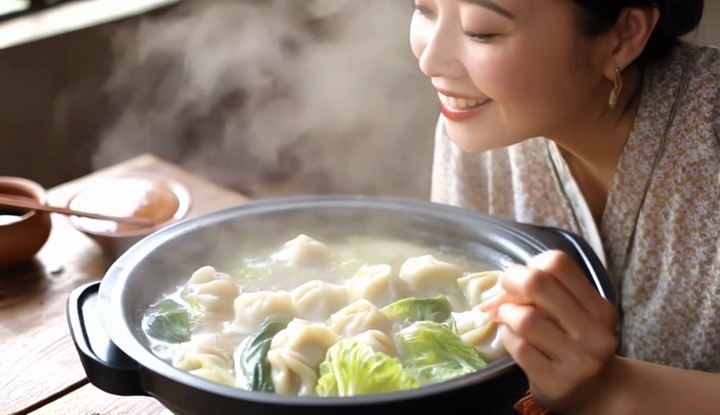}
\vspace{-0.5em}

\end{subfigure}

\vspace{0.2cm}

\begin{subfigure}{\textwidth}
\centering
\textbf{\large FastVideo} ~~\textit{\large Latency: 5.3s}\\
\vspace{0.1cm}

\includegraphics[width=0.165\textwidth]{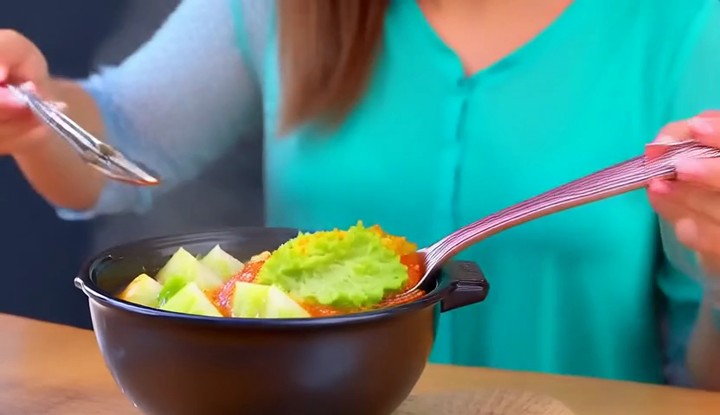}\hspace{-0.0037\textwidth}
\includegraphics[width=0.165\textwidth]{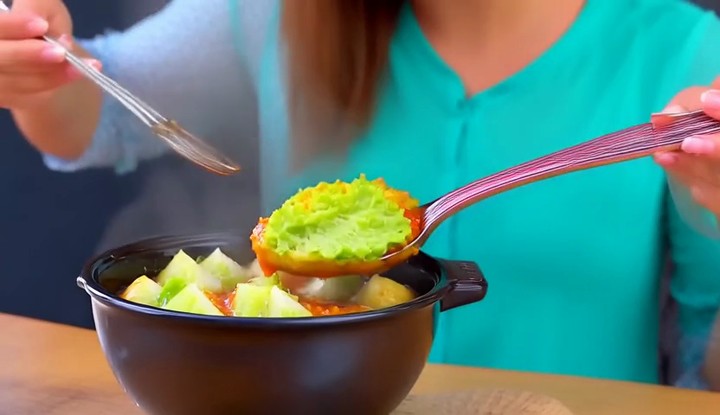}\hspace{-0.0037\textwidth}
\includegraphics[width=0.165\textwidth]{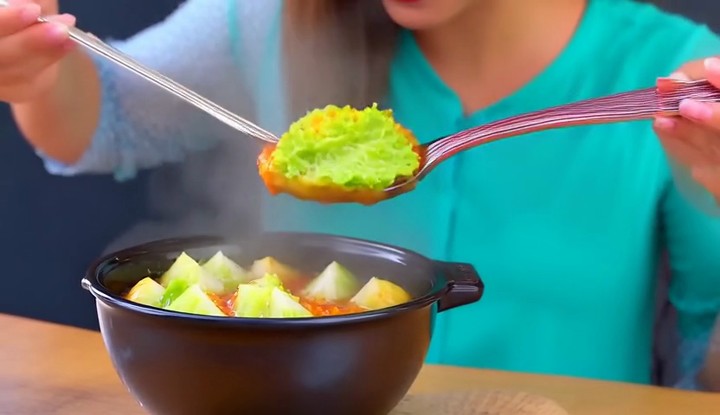}\hspace{-0.0037\textwidth}
\includegraphics[width=0.165\textwidth]{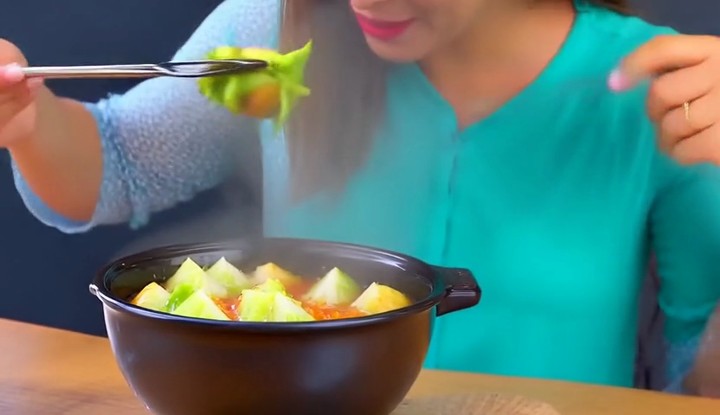}\hspace{-0.0037\textwidth}
\includegraphics[width=0.165\textwidth]{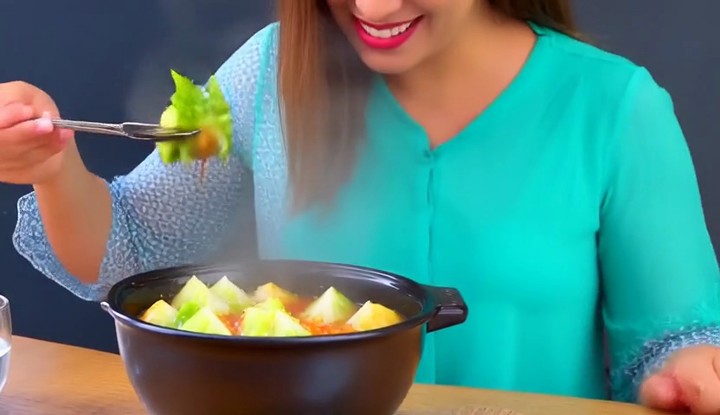}\hspace{-0.0037\textwidth}
\includegraphics[width=0.165\textwidth]{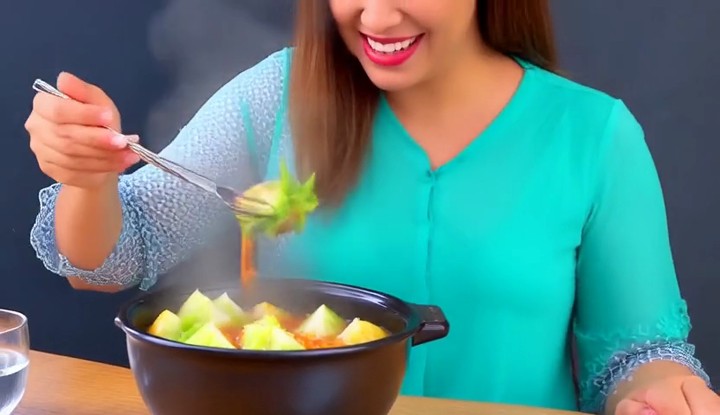}
\vspace{-0.5em}

\end{subfigure}

\vspace{0.2cm}

\begin{subfigure}{\textwidth}
\centering
\textbf{\large TurboDiffusion} ~~\textit{\large Latency: \bf \red{1.9s}}\\
\vspace{0.1cm}

\includegraphics[width=0.165\textwidth]{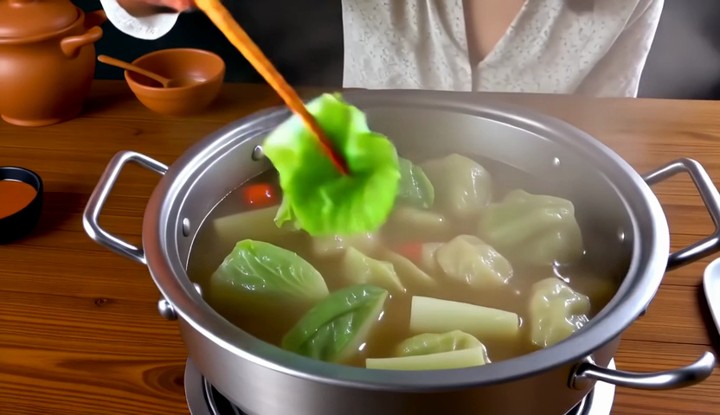}\hspace{-0.0037\textwidth}
\includegraphics[width=0.165\textwidth]{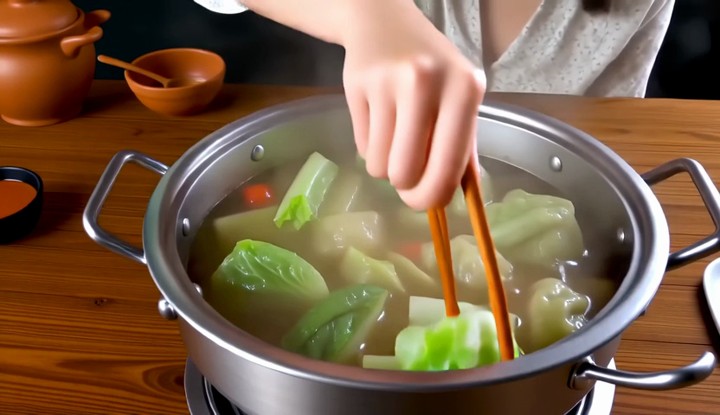}\hspace{-0.0037\textwidth}
\includegraphics[width=0.165\textwidth]{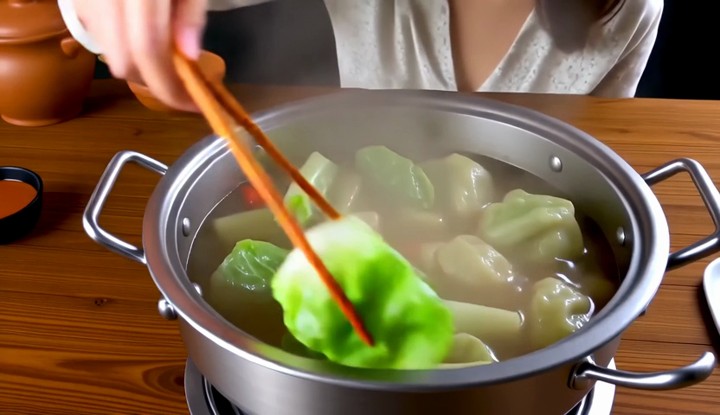}\hspace{-0.0037\textwidth}
\includegraphics[width=0.165\textwidth]{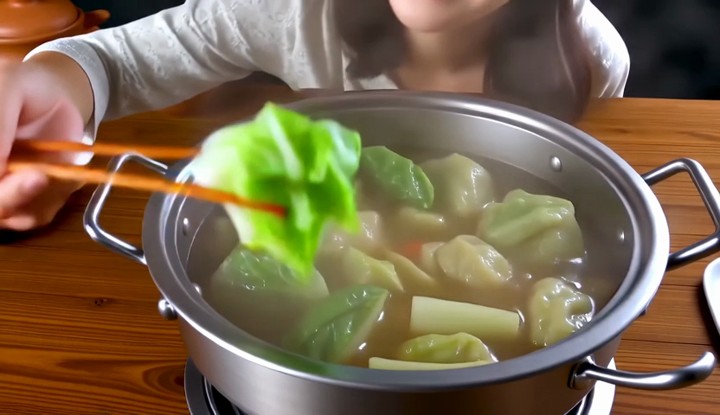}\hspace{-0.0037\textwidth}
\includegraphics[width=0.165\textwidth]{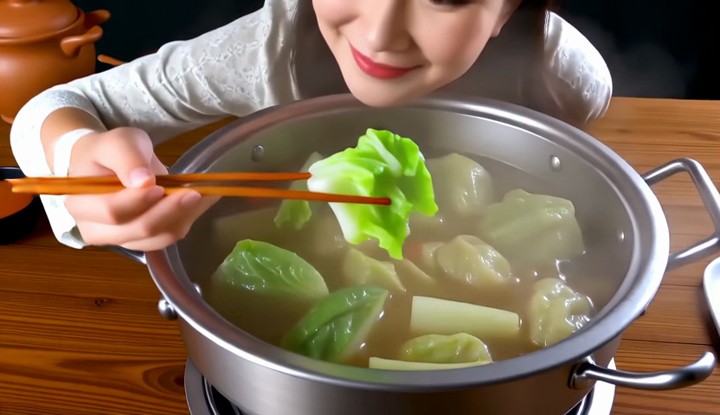}\hspace{-0.0037\textwidth}
\includegraphics[width=0.165\textwidth]{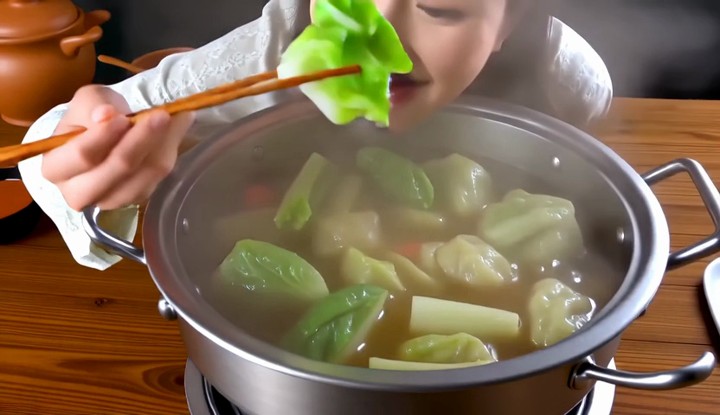}
\vspace{-0.5em}

\end{subfigure}

\vspace{-1em} \caption{5-second video generation on \texttt{Wan2.1-T2V-1.3B-480P} \textbf{\red{using a single RTX 5090}}.\\\textit{Prompt = "A close-up shot captures a steaming hot pot brimming with \red{vegetables and dumplings}, set on a rustic wooden table. The camera focuses on the bubbling broth as a woman, dressed in a light, patterned blouse, reaches in with chopsticks to lift a tender leaf of cabbage from the simmering mixture. Steam rises around her as she leans back slightly, her warm smile reflecting satisfaction and joy. Her movements are smooth and deliberate, showcasing her comfort and familiarity with the dining process. \red{The background includes a small bowl of dipping sauce and a clay pot}, adding to the cozy, communal dining atmosphere."}}
\label{fig:comparison_1_3b_video_3}
\end{figure}

\begin{figure}[H]
\centering
\begin{subfigure}{\textwidth}
\centering
\textbf{\large Original} ~~\textit{\large Latency: 184s}\\
\vspace{0.1cm}

\includegraphics[width=0.165\textwidth]{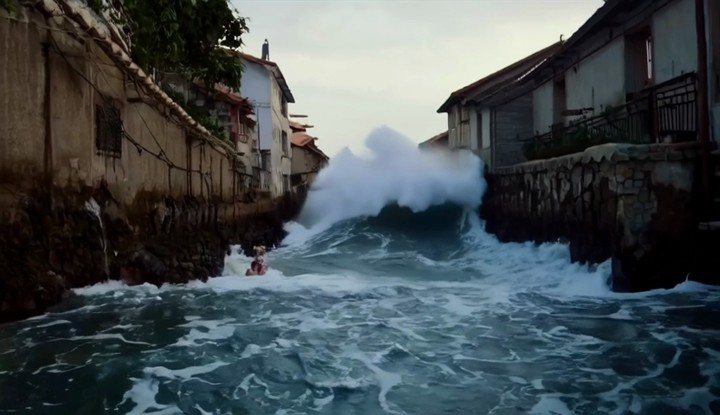}\hspace{-0.0037\textwidth}
\includegraphics[width=0.165\textwidth]{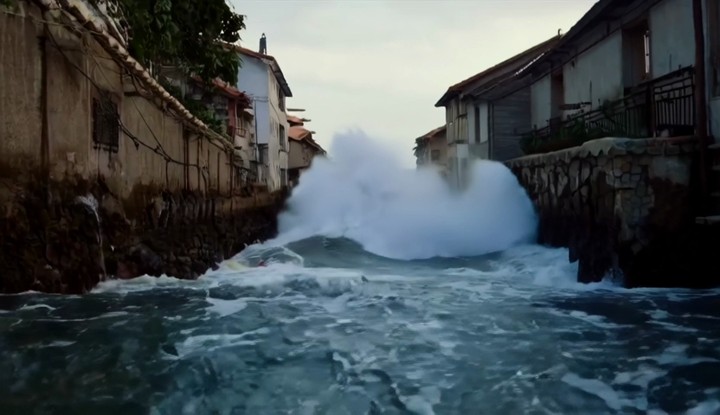}\hspace{-0.0037\textwidth}
\includegraphics[width=0.165\textwidth]{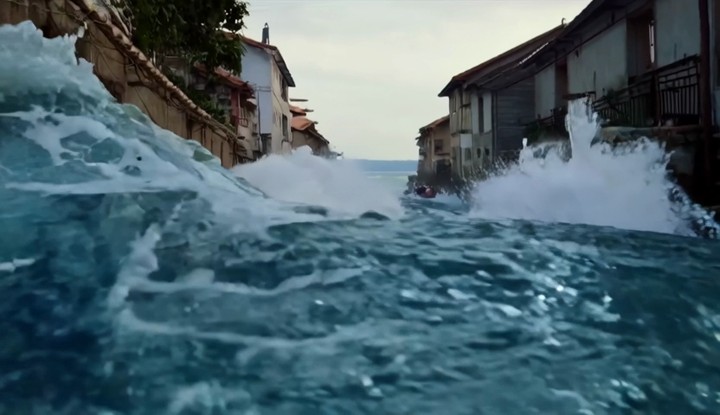}\hspace{-0.0037\textwidth}
\includegraphics[width=0.165\textwidth]{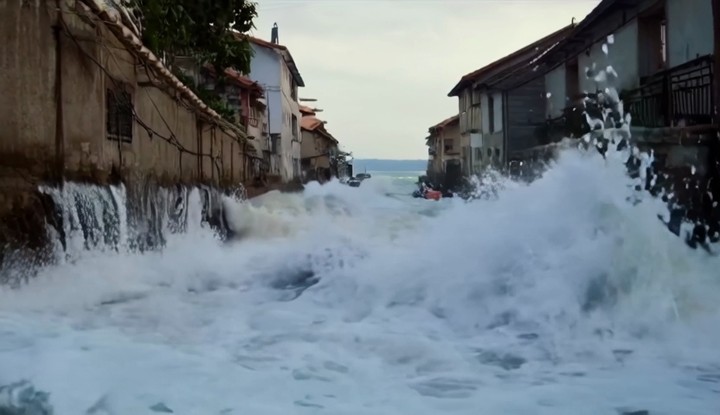}\hspace{-0.0037\textwidth}
\includegraphics[width=0.165\textwidth]{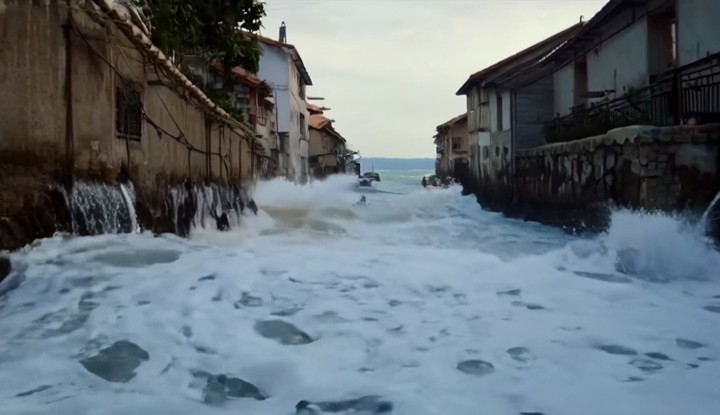}\hspace{-0.0037\textwidth}
\includegraphics[width=0.165\textwidth]{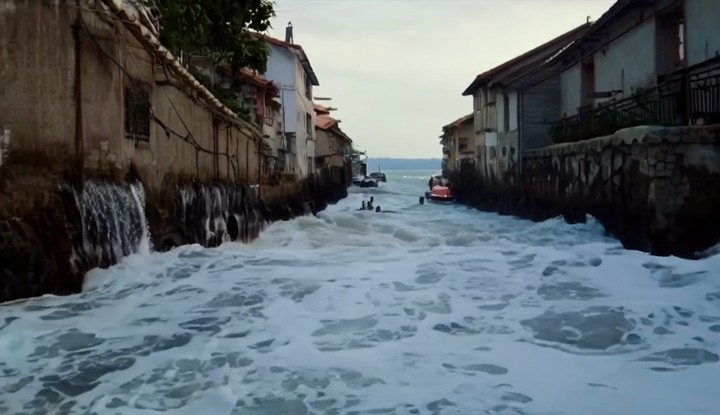}
\vspace{-0.5em}

\end{subfigure}

\vspace{0.2cm}

\begin{subfigure}{\textwidth}
\centering
\textbf{\large FastVideo} ~~\textit{\large Latency: 5.3s}\\
\vspace{0.1cm}

\includegraphics[width=0.165\textwidth]{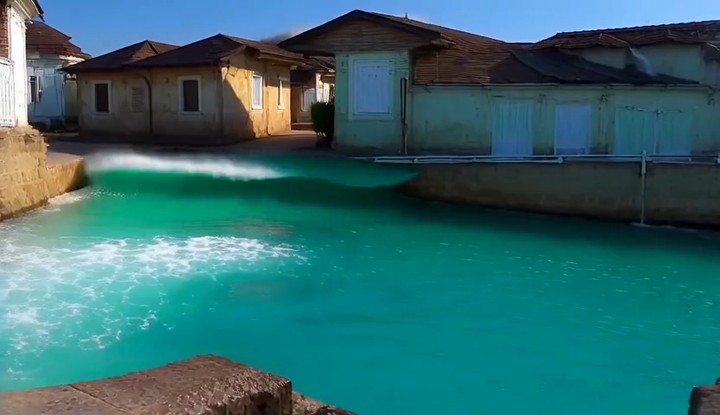}\hspace{-0.0037\textwidth}
\includegraphics[width=0.165\textwidth]{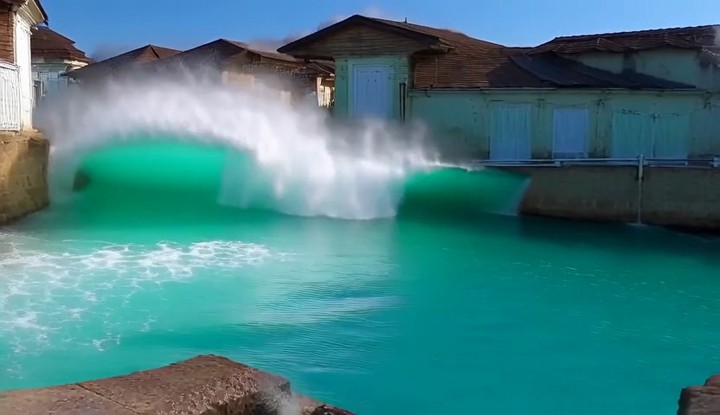}\hspace{-0.0037\textwidth}
\includegraphics[width=0.165\textwidth]{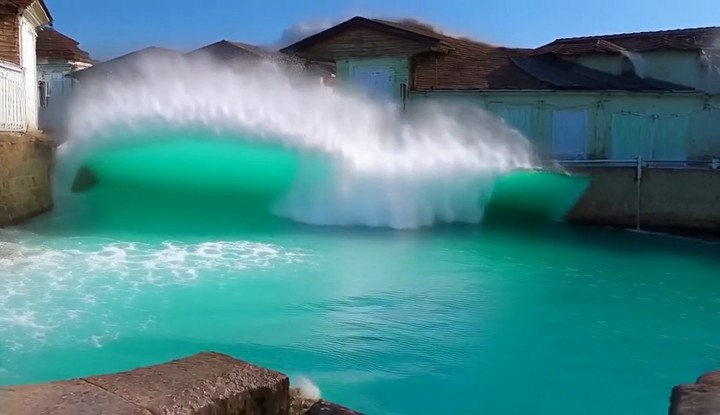}\hspace{-0.0037\textwidth}
\includegraphics[width=0.165\textwidth]{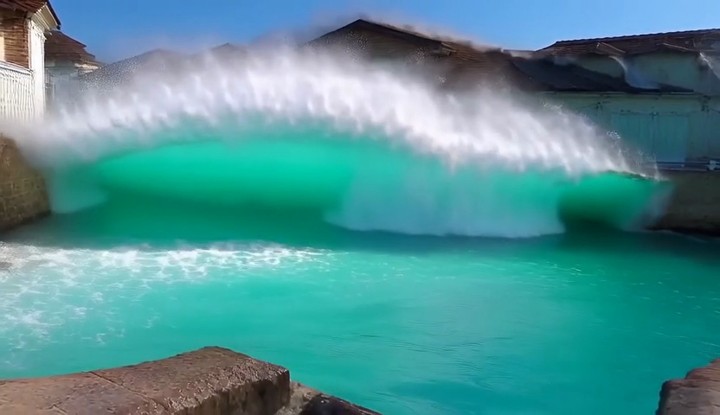}\hspace{-0.0037\textwidth}
\includegraphics[width=0.165\textwidth]{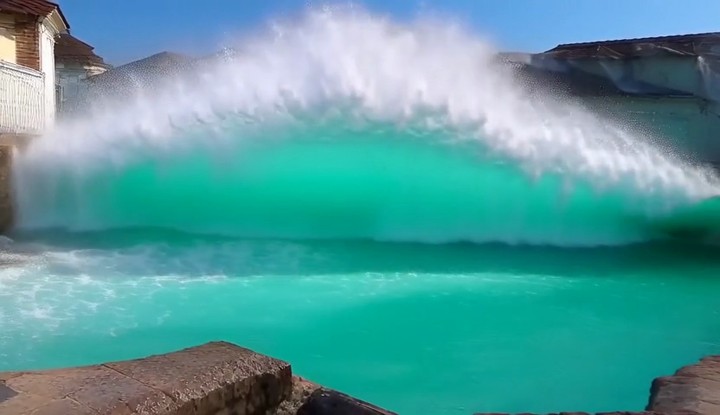}\hspace{-0.0037\textwidth}
\includegraphics[width=0.165\textwidth]{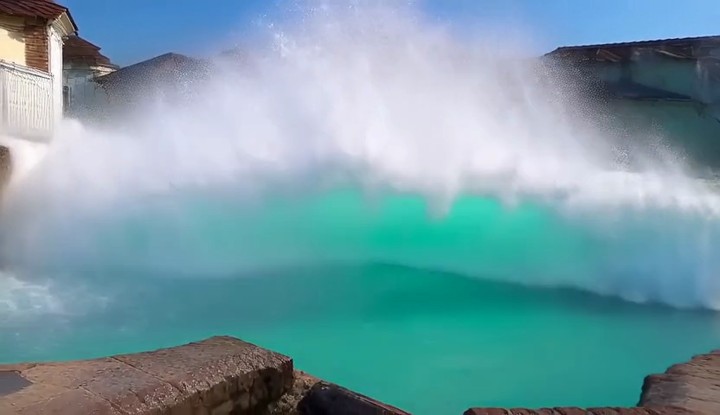}
\vspace{-0.5em}

\end{subfigure}

\vspace{0.2cm}

\begin{subfigure}{\textwidth}
\centering
\textbf{\large TurboDiffusion} ~~\textit{\large Latency: \bf \red{1.9s}}\\
\vspace{0.1cm}

\includegraphics[width=0.165\textwidth]{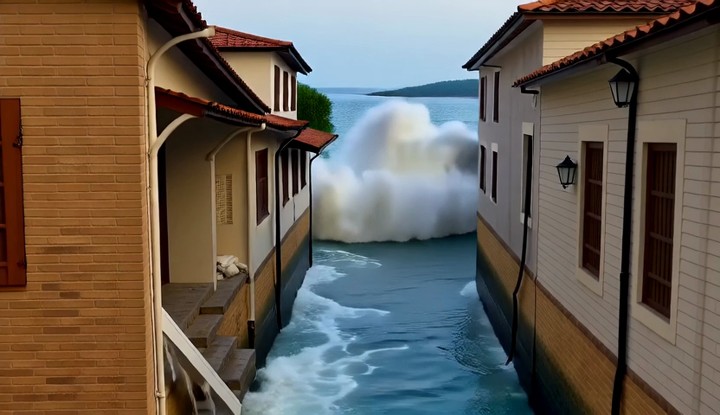}\hspace{-0.0037\textwidth}
\includegraphics[width=0.165\textwidth]{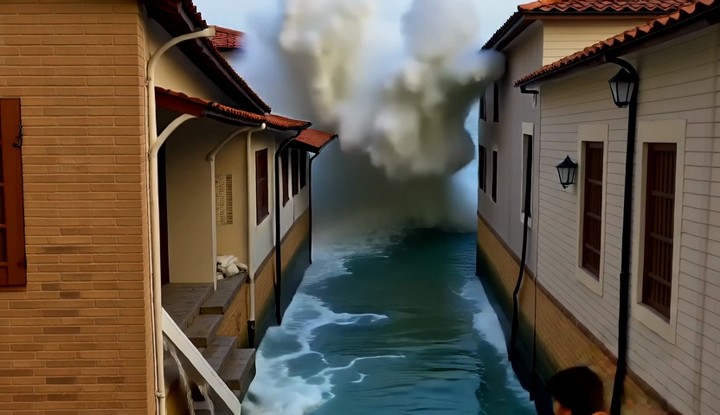}\hspace{-0.0037\textwidth}
\includegraphics[width=0.165\textwidth]{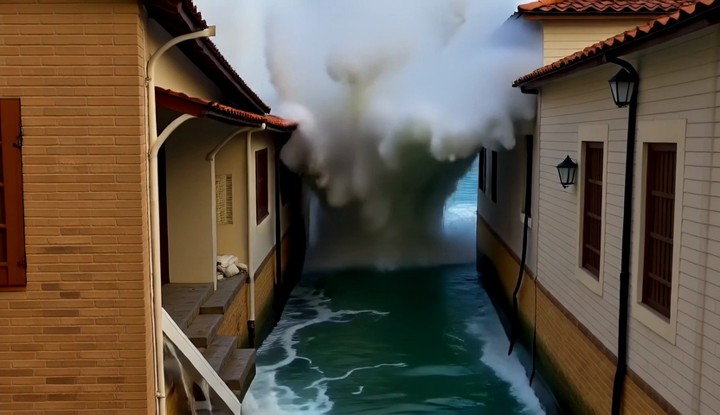}\hspace{-0.0037\textwidth}
\includegraphics[width=0.165\textwidth]{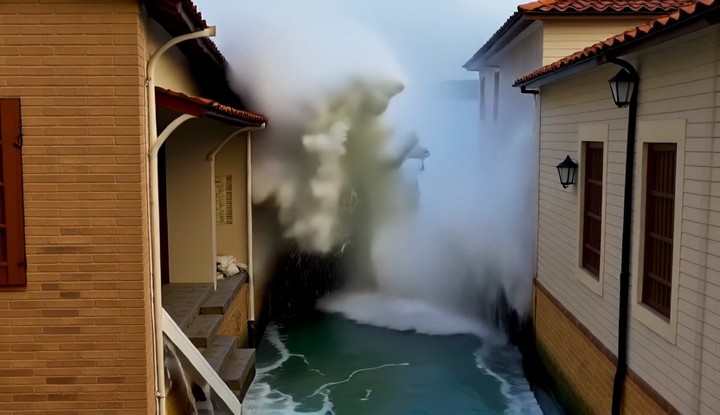}\hspace{-0.0037\textwidth}
\includegraphics[width=0.165\textwidth]{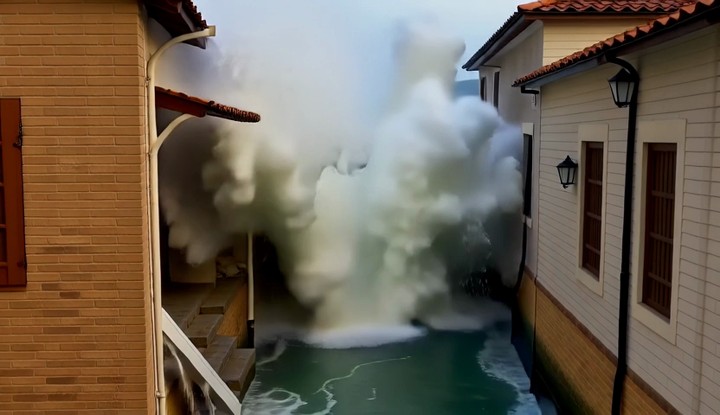}\hspace{-0.0037\textwidth}
\includegraphics[width=0.165\textwidth]{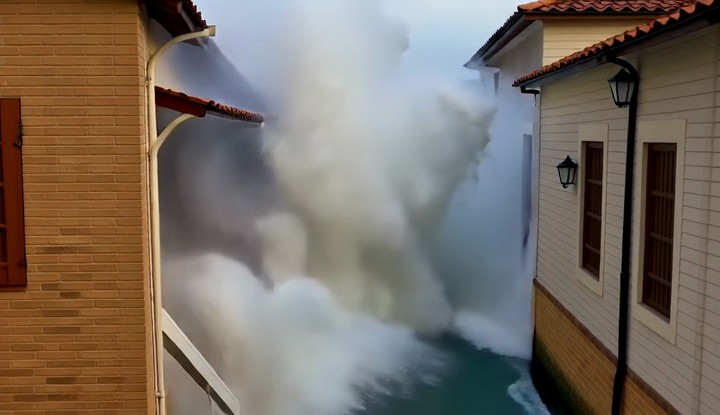}
\vspace{-0.5em}

\end{subfigure}

\vspace{-1em} \caption{5-second video generation on \texttt{Wan2.1-T2V-1.3B-480P} \textbf{\red{using a single RTX 5090}}.\\Prompt = \textit{"A dramatic and dynamic scene in the style of a disaster movie, depicting a powerful tsunami rushing through \red{a narrow alley} in Bulgaria. The water is turbulent and chaotic, with \red{waves crashing violently against the walls and buildings on either side. The alley is lined with old, weathered houses, their facades partially submerged} and splintered. The camera angle is low, capturing the full force of the tsunami as it surges forward, creating a sense of urgency and danger. People can be seen running frantically, adding to the chaos. The background features a distant horizon, hinting at the larger scale of the tsunami. A dynamic, sweeping shot from a low-angle perspective, emphasizing the movement and intensity of the event."}}
\label{fig:comparison_1_3b_video_5}
\end{figure}

\begin{figure}[H]
\centering
\begin{subfigure}{\textwidth}
\centering
\textbf{\large Original} ~~\textit{\large Latency: 184s}\\
\vspace{0.1cm}

\includegraphics[width=0.165\textwidth]{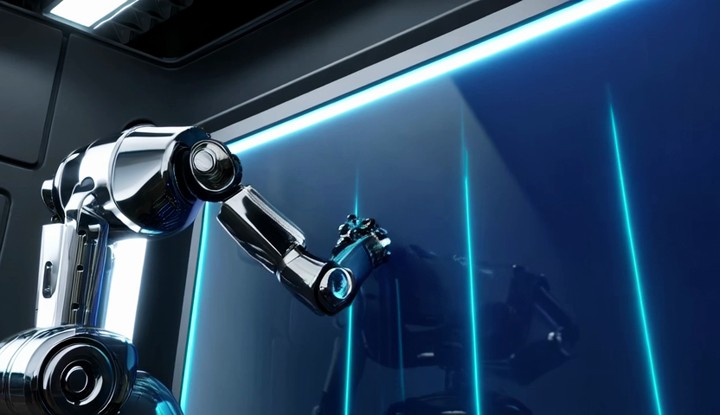}\hspace{-0.0037\textwidth}
\includegraphics[width=0.165\textwidth]{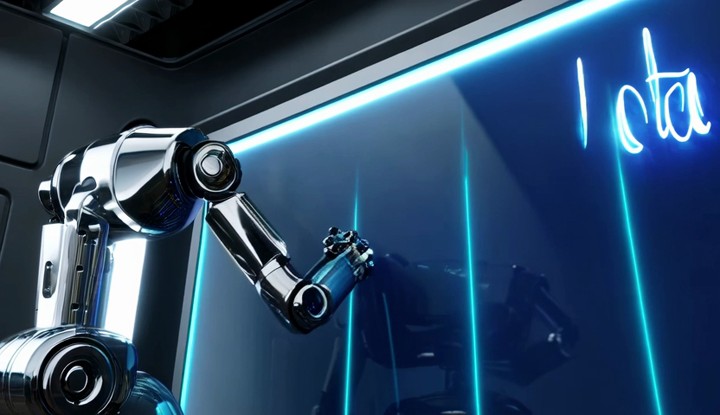}\hspace{-0.0037\textwidth}
\includegraphics[width=0.165\textwidth]{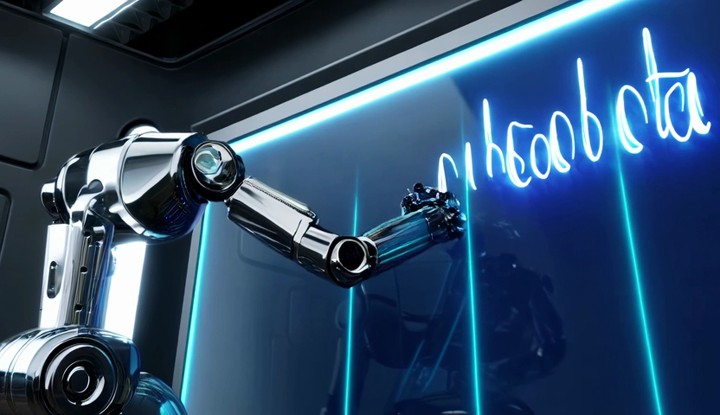}\hspace{-0.0037\textwidth}
\includegraphics[width=0.165\textwidth]{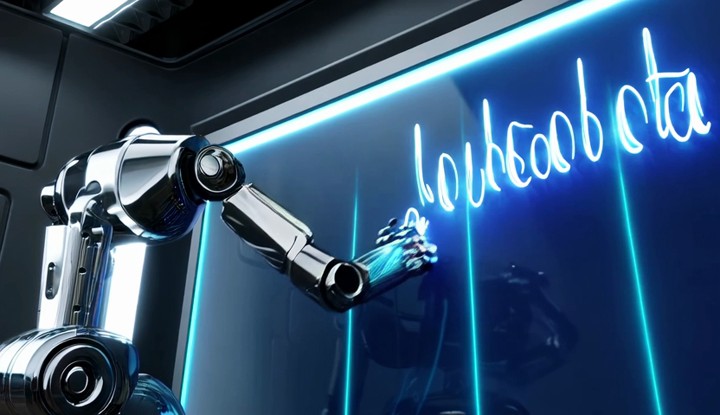}\hspace{-0.0037\textwidth}
\includegraphics[width=0.165\textwidth]{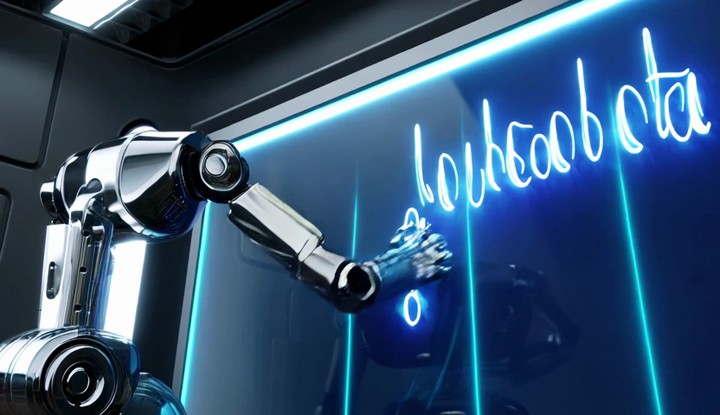}\hspace{-0.0037\textwidth}
\includegraphics[width=0.165\textwidth]{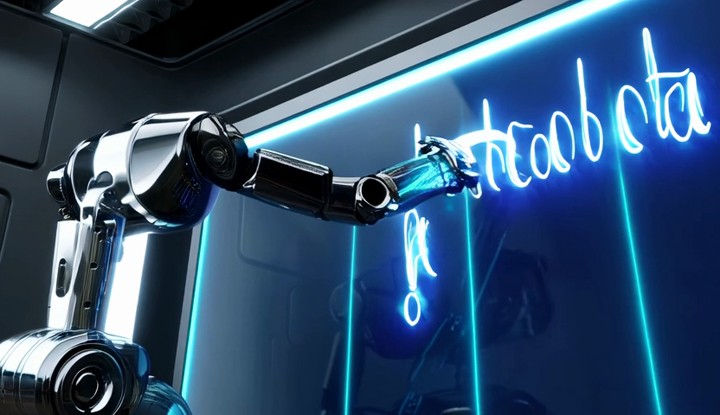}
\vspace{-0.5em}

\end{subfigure}

\vspace{0.2cm}

\begin{subfigure}{\textwidth}
\centering
\textbf{\large FastVideo} ~~\textit{\large Latency: 5.3s}\\
\vspace{0.1cm}

\includegraphics[width=0.165\textwidth]{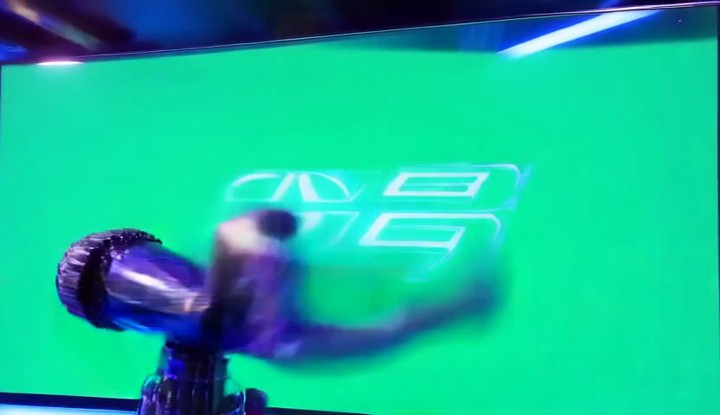}\hspace{-0.0037\textwidth}
\includegraphics[width=0.165\textwidth]{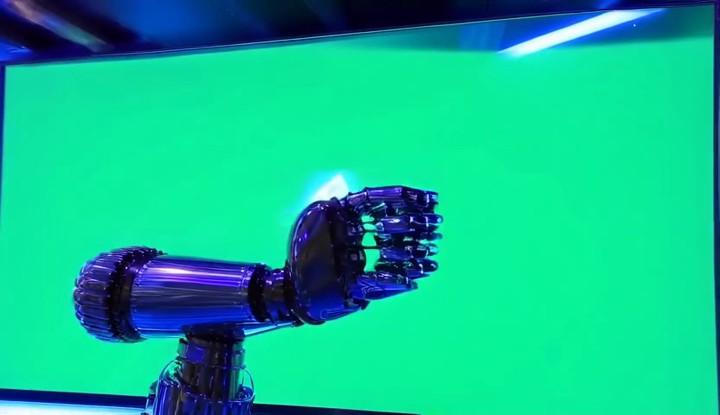}\hspace{-0.0037\textwidth}
\includegraphics[width=0.165\textwidth]{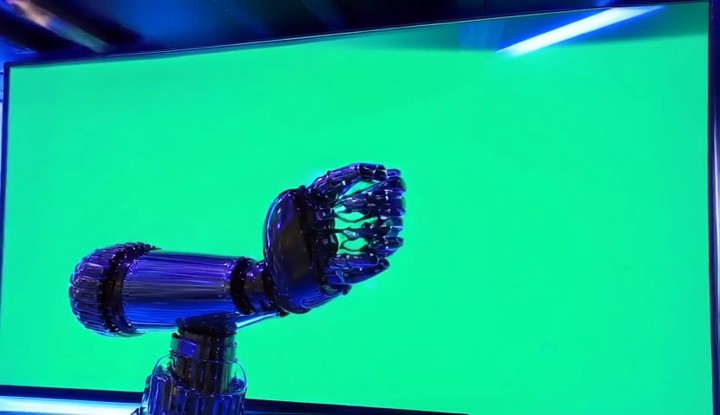}\hspace{-0.0037\textwidth}
\includegraphics[width=0.165\textwidth]{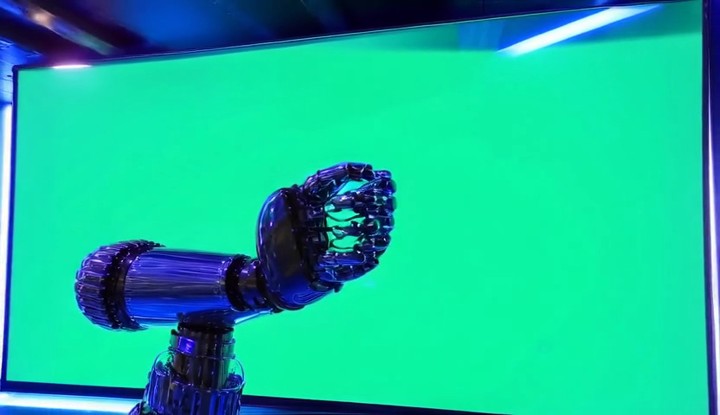}\hspace{-0.0037\textwidth}
\includegraphics[width=0.165\textwidth]{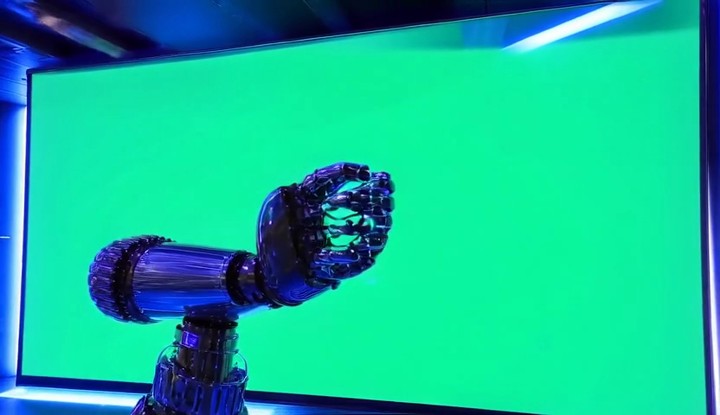}\hspace{-0.0037\textwidth}
\includegraphics[width=0.165\textwidth]{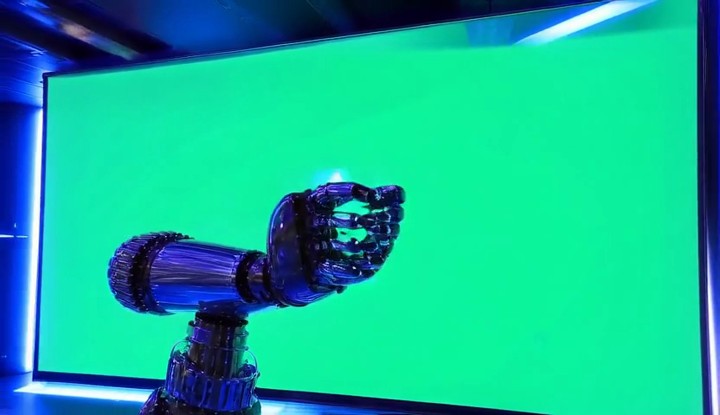}
\vspace{-0.5em}

\end{subfigure}

\vspace{0.2cm}

\begin{subfigure}{\textwidth}
\centering
\textbf{\large TurboDiffusion} ~~\textit{\large Latency: \bf \red{1.9s}}\\
\vspace{0.1cm}

\includegraphics[width=0.165\textwidth]{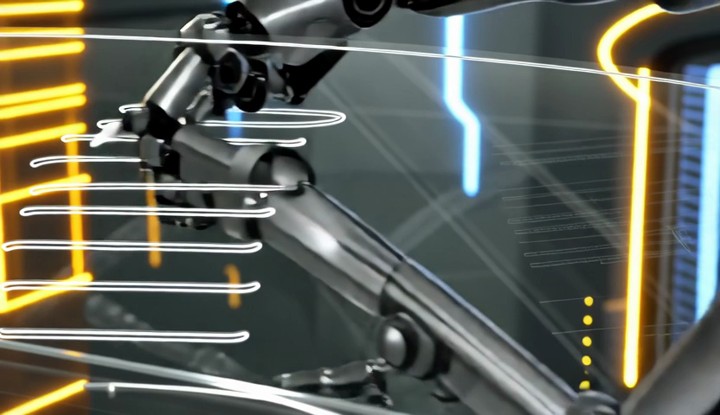}\hspace{-0.0037\textwidth}
\includegraphics[width=0.165\textwidth]{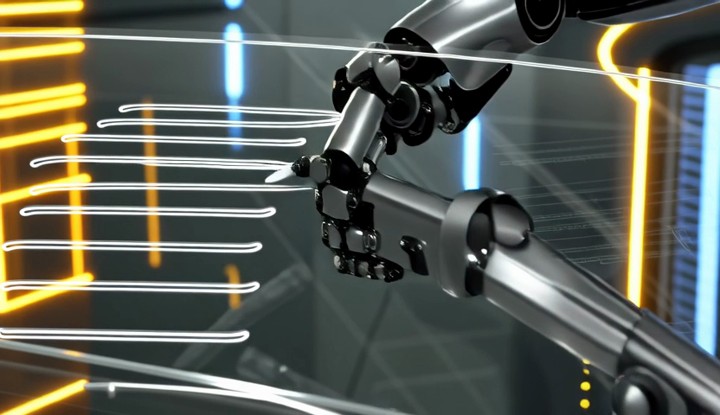}\hspace{-0.0037\textwidth}
\includegraphics[width=0.165\textwidth]{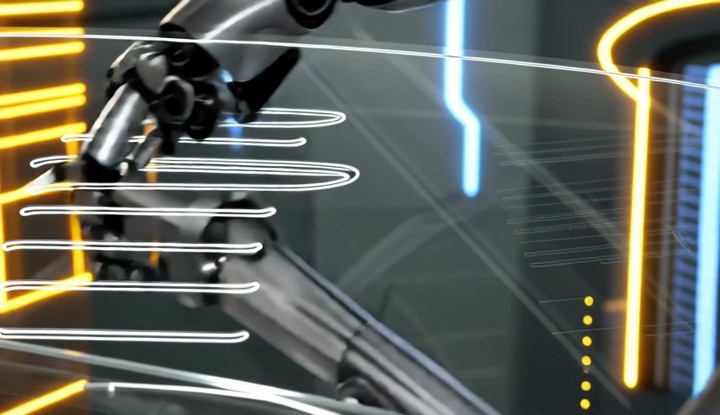}\hspace{-0.0037\textwidth}
\includegraphics[width=0.165\textwidth]{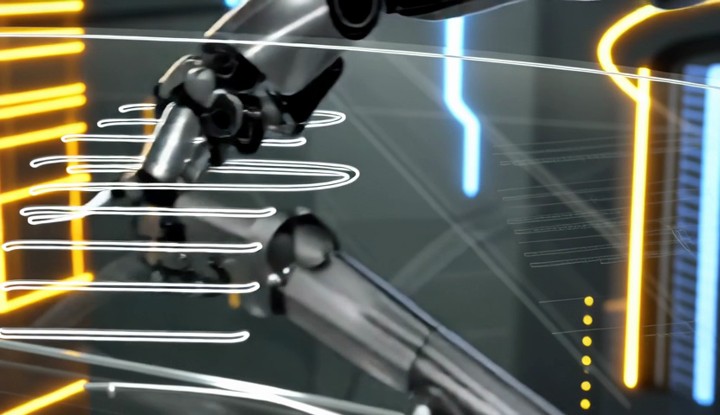}\hspace{-0.0037\textwidth}
\includegraphics[width=0.165\textwidth]{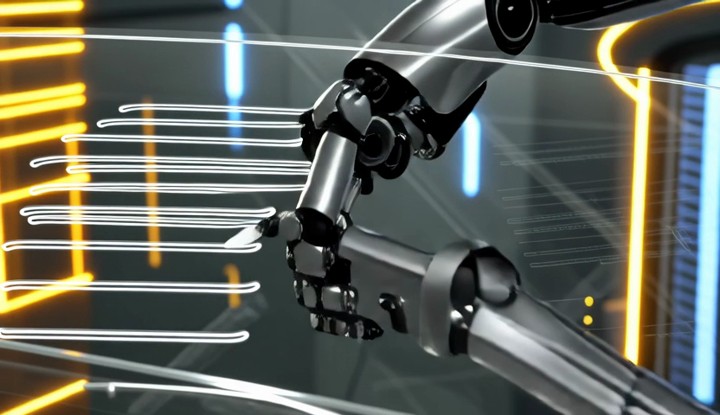}\hspace{-0.0037\textwidth}
\includegraphics[width=0.165\textwidth]{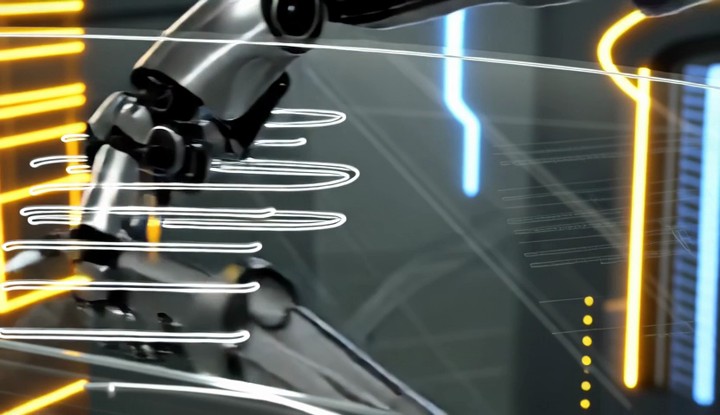}
\vspace{-0.5em}

\end{subfigure}

\vspace{-1em} \caption{5-second video generation on \texttt{Wan2.1-T2V-1.3B-480P} \textbf{\red{using a single RTX 5090}}.\\\textit{Prompt = "\red{A futuristic robotic arm writing on a large glass screen.} The robot has sleek metallic joints and a precise, articulated hand. It moves smoothly across the screen, leaving behind elegant, flowing text. The lighting highlights the reflective surfaces, creating dynamic shadows and highlights. The background is a high-tech lab with neon accents and advanced machinery. The scene focuses on the robotic arm as it writes, emphasizing the fluid motion and intricate design. Medium close-up shot, static camera."}}
\label{fig:comparison_1_3b_video_7}
\end{figure}

\begin{figure}[H]
\centering
\begin{subfigure}{\textwidth}
\centering
\textbf{\large Original} ~~\textit{\large Latency: 184s}\\
\vspace{0.1cm}

\includegraphics[width=0.165\textwidth]{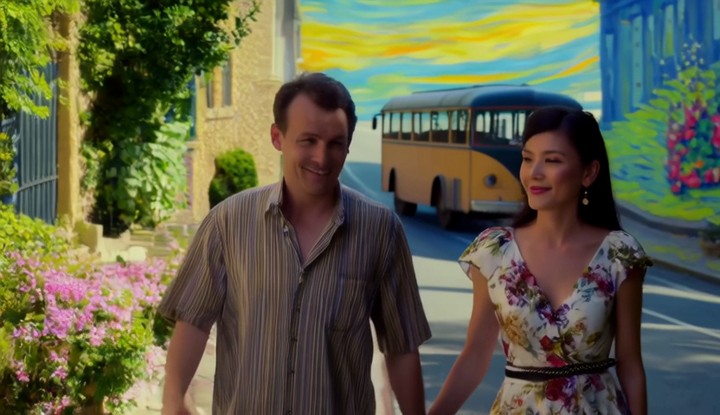}\hspace{-0.0037\textwidth}
\includegraphics[width=0.165\textwidth]{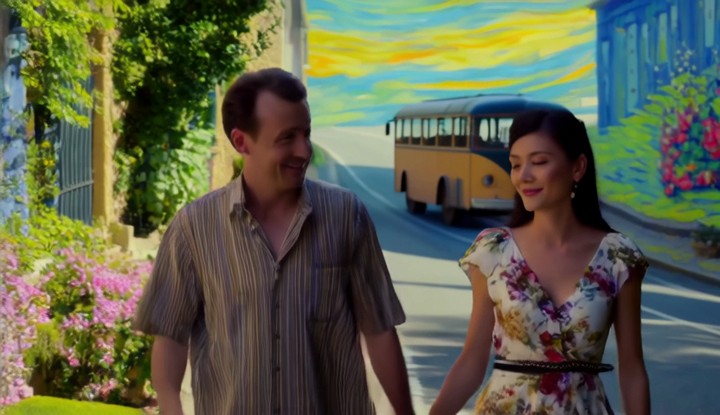}\hspace{-0.0037\textwidth}
\includegraphics[width=0.165\textwidth]{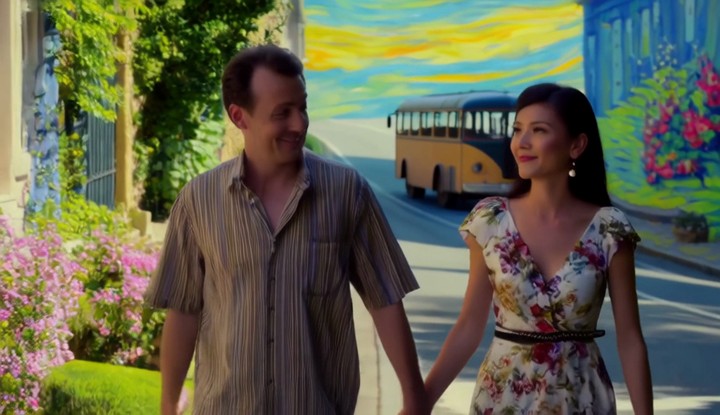}\hspace{-0.0037\textwidth}
\includegraphics[width=0.165\textwidth]{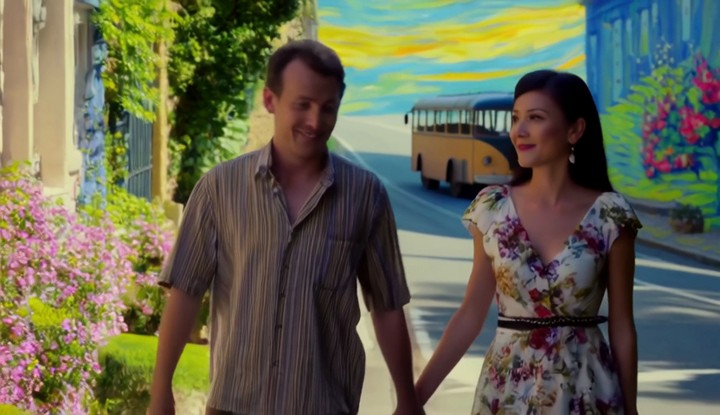}\hspace{-0.0037\textwidth}
\includegraphics[width=0.165\textwidth]{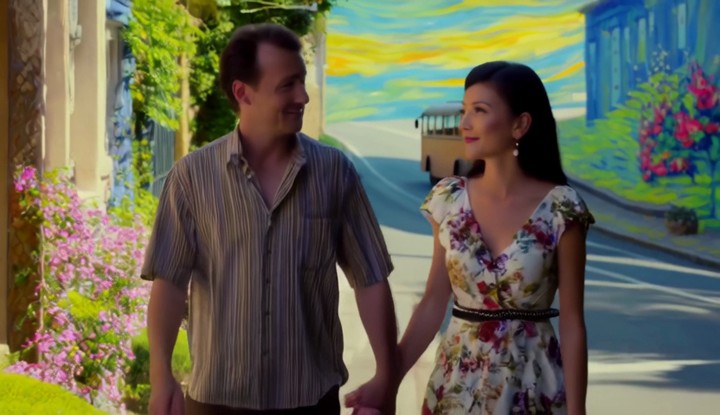}\hspace{-0.0037\textwidth}
\includegraphics[width=0.165\textwidth]{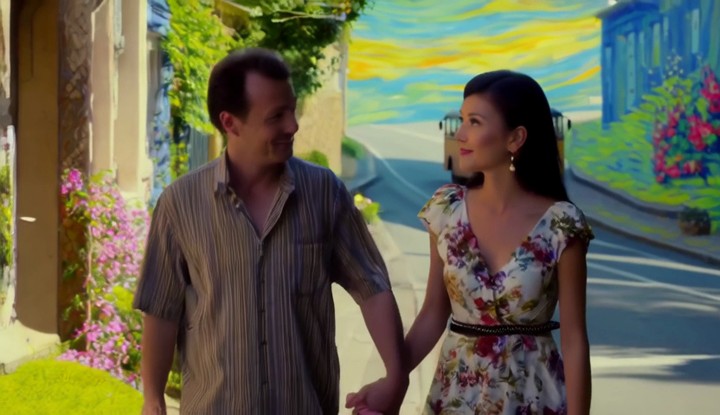}
\vspace{-0.5em}

\end{subfigure}

\vspace{0.2cm}

\begin{subfigure}{\textwidth}
\centering
\textbf{\large FastVideo} ~~\textit{\large Latency: 5.3s}\\
\vspace{0.1cm}

\includegraphics[width=0.165\textwidth]{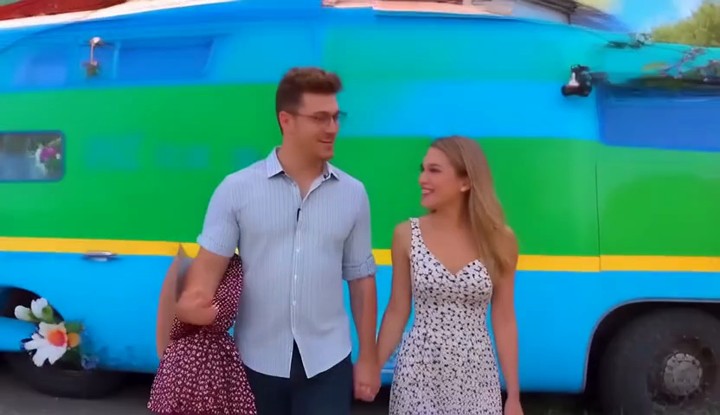}\hspace{-0.0037\textwidth}
\includegraphics[width=0.165\textwidth]{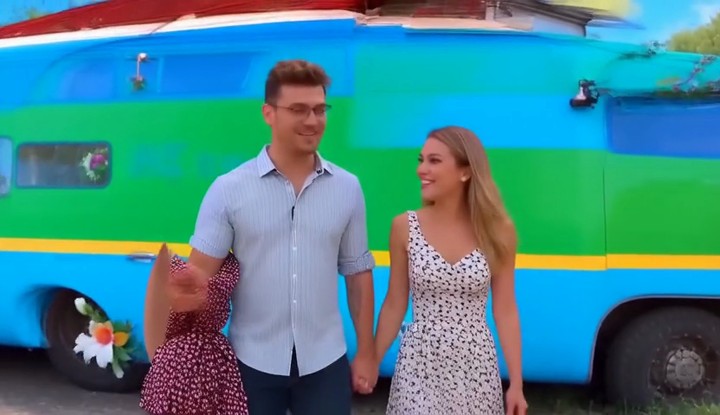}\hspace{-0.0037\textwidth}
\includegraphics[width=0.165\textwidth]{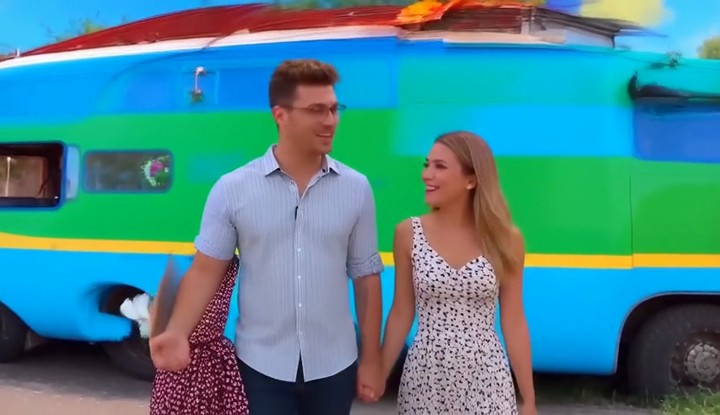}\hspace{-0.0037\textwidth}
\includegraphics[width=0.165\textwidth]{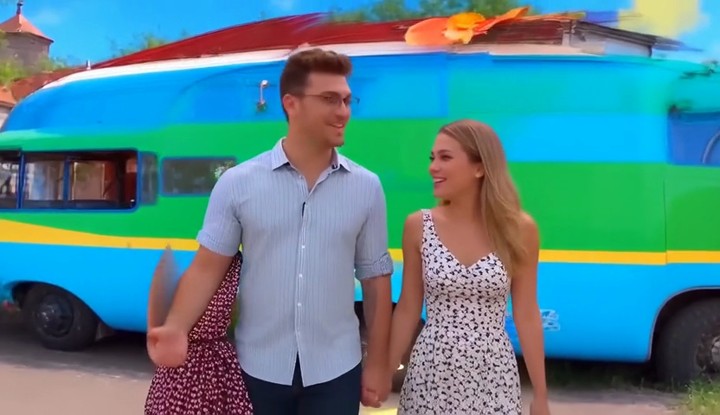}\hspace{-0.0037\textwidth}
\includegraphics[width=0.165\textwidth]{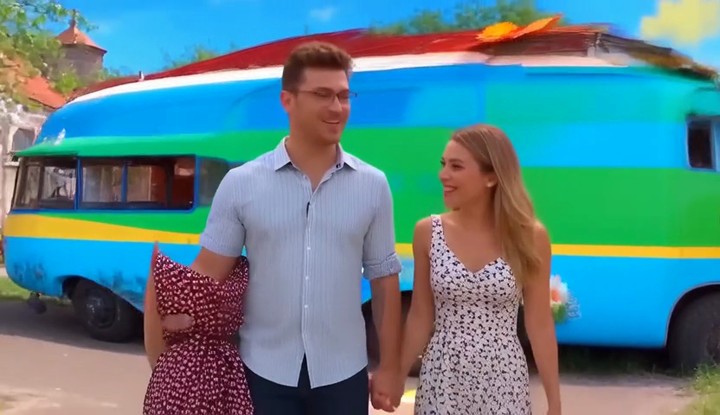}\hspace{-0.0037\textwidth}
\includegraphics[width=0.165\textwidth]{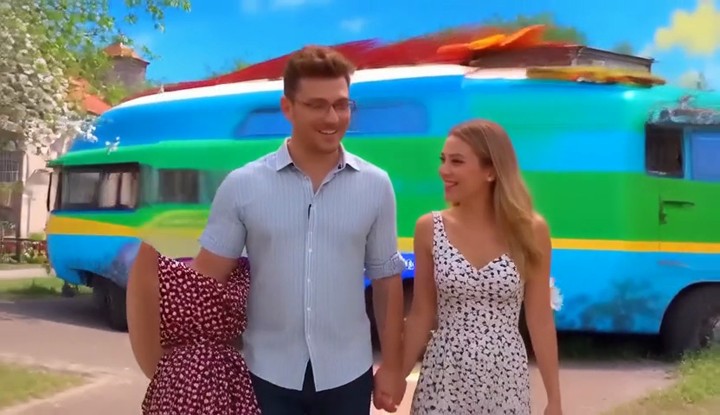}
\vspace{-0.5em}

\end{subfigure}

\vspace{0.2cm}

\begin{subfigure}{\textwidth}
\centering
\textbf{\large TurboDiffusion} ~~\textit{\large Latency: \bf \red{1.9s}}\\
\vspace{0.1cm}

\includegraphics[width=0.165\textwidth]{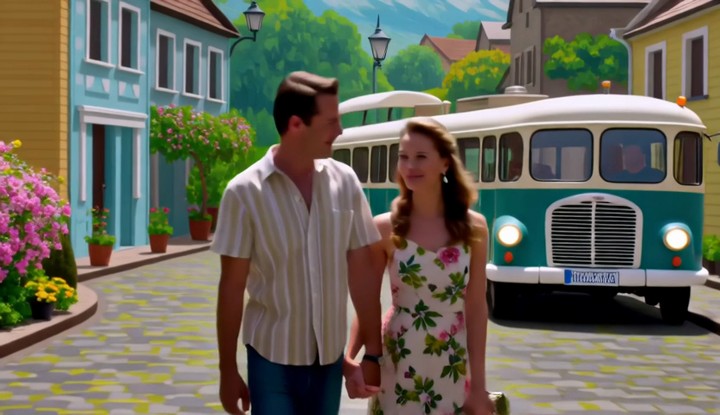}\hspace{-0.0037\textwidth}
\includegraphics[width=0.165\textwidth]{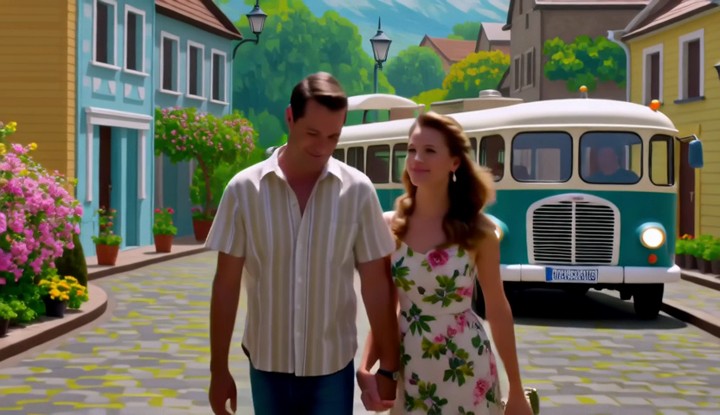}\hspace{-0.0037\textwidth}
\includegraphics[width=0.165\textwidth]{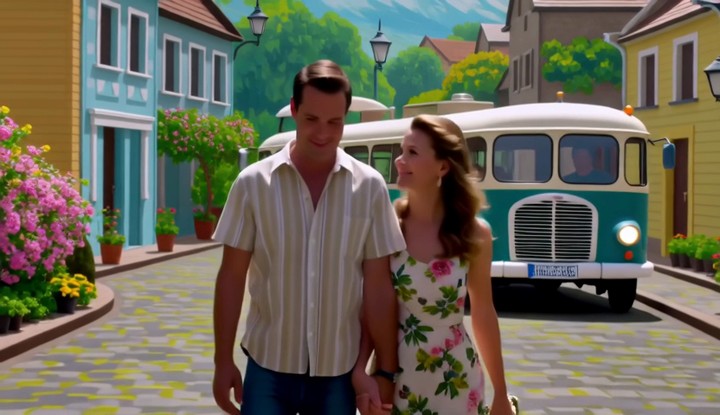}\hspace{-0.0037\textwidth}
\includegraphics[width=0.165\textwidth]{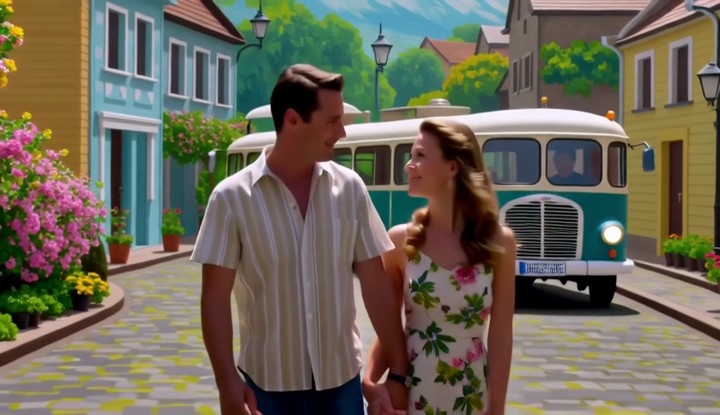}\hspace{-0.0037\textwidth}
\includegraphics[width=0.165\textwidth]{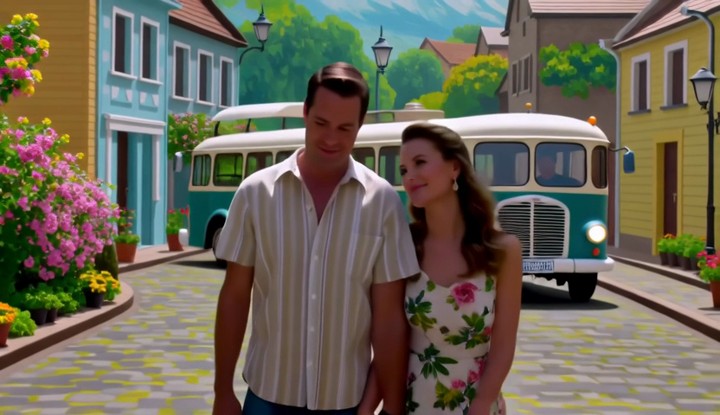}\hspace{-0.0037\textwidth}
\includegraphics[width=0.165\textwidth]{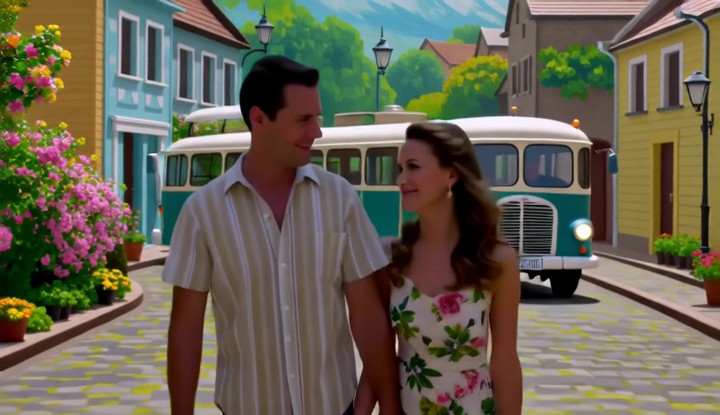}
\vspace{-0.5em}

\end{subfigure}

\vspace{-1em} \caption{5-second video generation on \texttt{Wan2.1-T2V-1.3B-480P} \textbf{\red{using a single RTX 5090}}.\\\textit{Prompt = "\red{A man and woman}, walking hand in hand down a vibrant, winding street, their figures illuminated against a backdrop reminiscent of a Van Gogh painting. They are strolling leisurely, with the man wearing a casual, striped shirt and the woman in a floral dress, both with warm, content expressions. Behind them, a vintage bus passes slowly, adding to the lively atmosphere of the scene. The background features swirling brushstrokes of blues, yellows, and greens, typical of Van Gogh\u2019s style, with a charming village street filled with blooming flowers and quaint buildings. Medium shot, capturing the couple from mid-torso up, with the bus visible in the background."}}
\label{fig:comparison_1_3b_video_8}
\end{figure}

\begin{figure}[H]
\centering
\begin{subfigure}{\textwidth}
\centering
\textbf{\large Original} ~~\textit{\large Latency: 184s}\\
\vspace{0.1cm}

\includegraphics[width=0.165\textwidth]{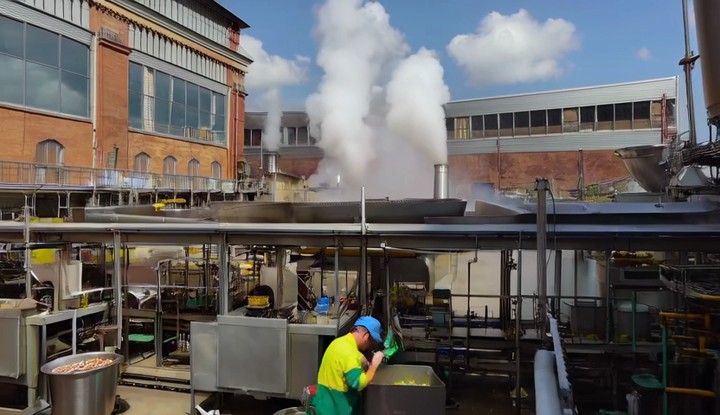}\hspace{-0.0037\textwidth}
\includegraphics[width=0.165\textwidth]{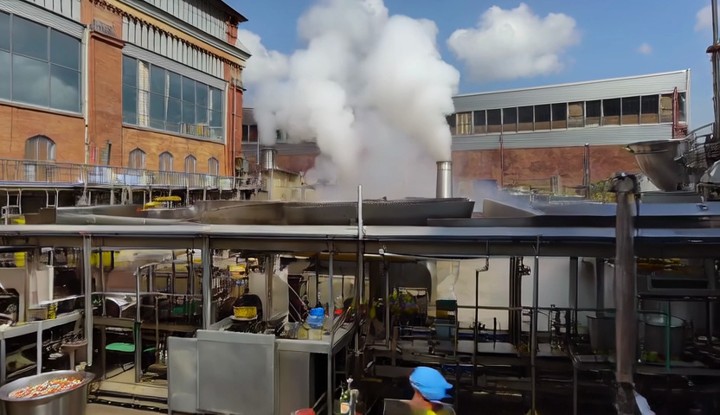}\hspace{-0.0037\textwidth}
\includegraphics[width=0.165\textwidth]{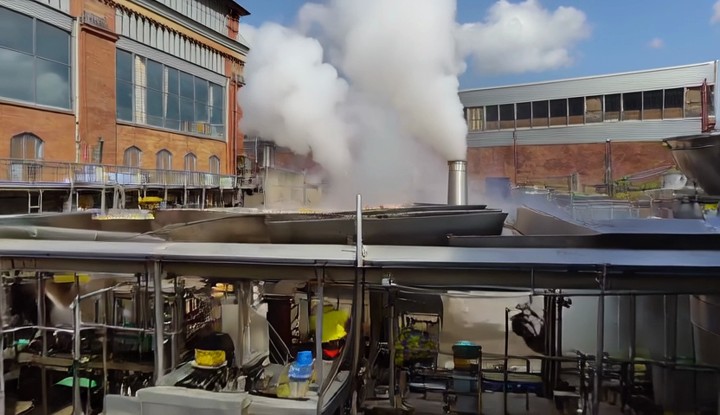}\hspace{-0.0037\textwidth}
\includegraphics[width=0.165\textwidth]{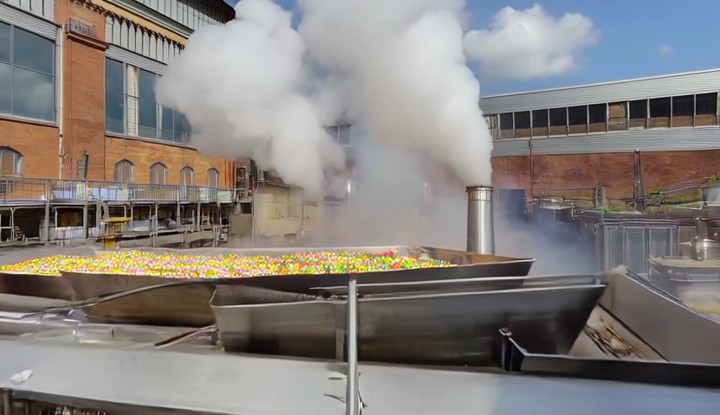}\hspace{-0.0037\textwidth}
\includegraphics[width=0.165\textwidth]{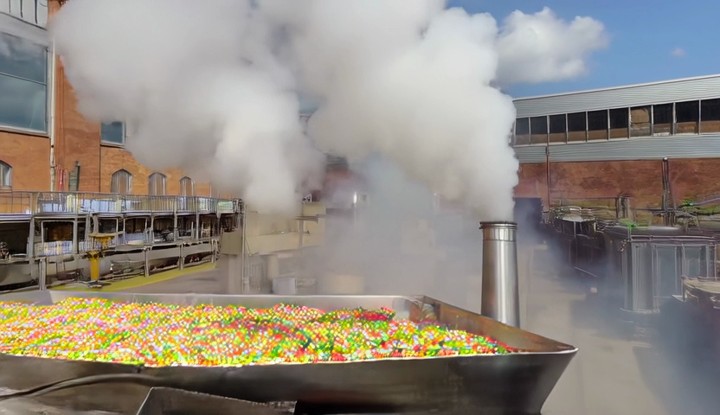}\hspace{-0.0037\textwidth}
\includegraphics[width=0.165\textwidth]{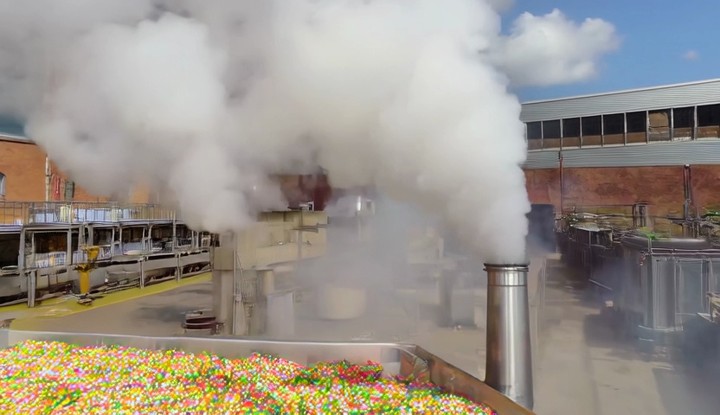}
\vspace{-0.5em}

\end{subfigure}

\vspace{0.2cm}

\begin{subfigure}{\textwidth}
\centering
\textbf{\large FastVideo} ~~\textit{\large Latency: 5.3s}\\
\vspace{0.1cm}

\includegraphics[width=0.165\textwidth]{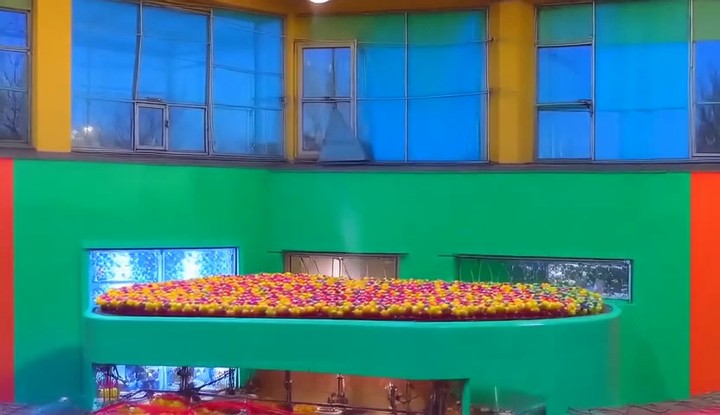}\hspace{-0.0037\textwidth}
\includegraphics[width=0.165\textwidth]{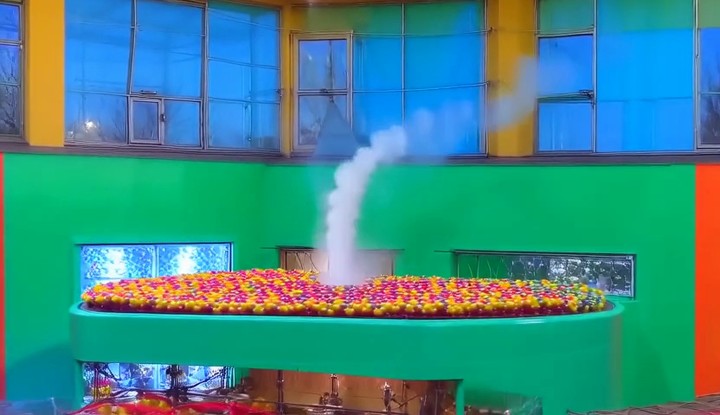}\hspace{-0.0037\textwidth}
\includegraphics[width=0.165\textwidth]{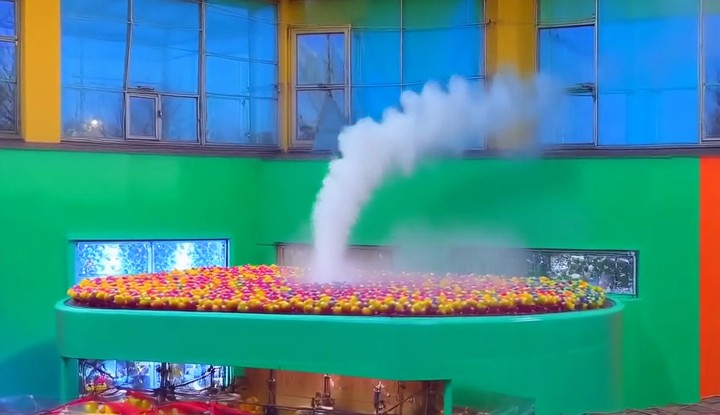}\hspace{-0.0037\textwidth}
\includegraphics[width=0.165\textwidth]{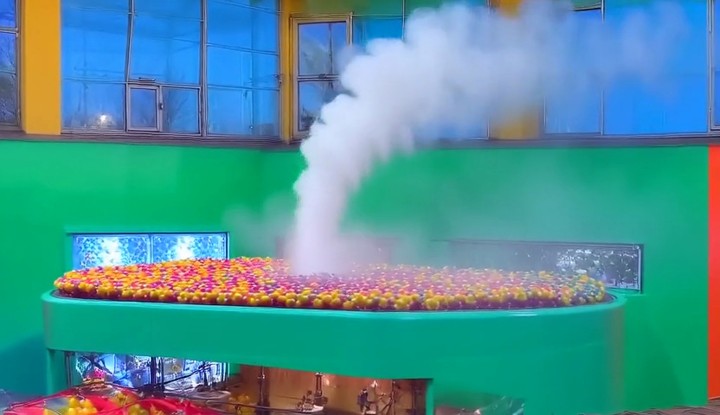}\hspace{-0.0037\textwidth}
\includegraphics[width=0.165\textwidth]{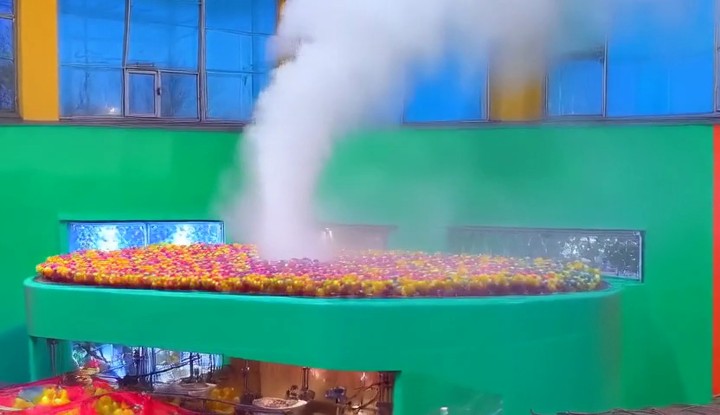}\hspace{-0.0037\textwidth}
\includegraphics[width=0.165\textwidth]{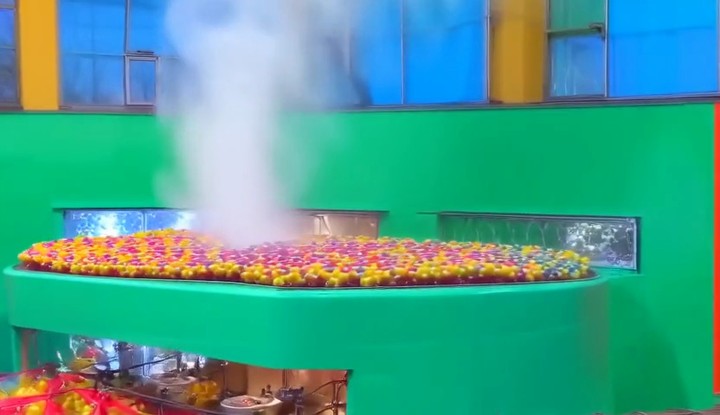}
\vspace{-0.5em}

\end{subfigure}

\vspace{0.2cm}

\begin{subfigure}{\textwidth}
\centering
\textbf{\large TurboDiffusion} ~~\textit{\large Latency: \bf \red{1.9s}}\\
\vspace{0.1cm}

\includegraphics[width=0.165\textwidth]{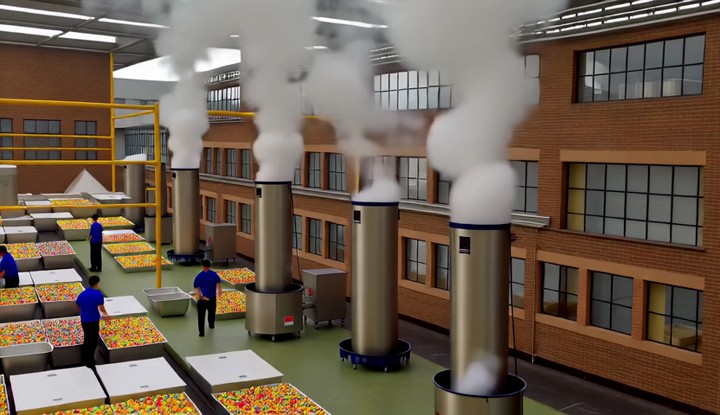}\hspace{-0.0037\textwidth}
\includegraphics[width=0.165\textwidth]{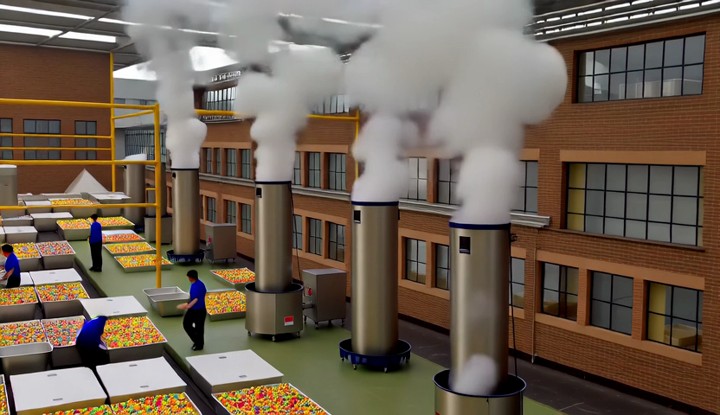}\hspace{-0.0037\textwidth}
\includegraphics[width=0.165\textwidth]{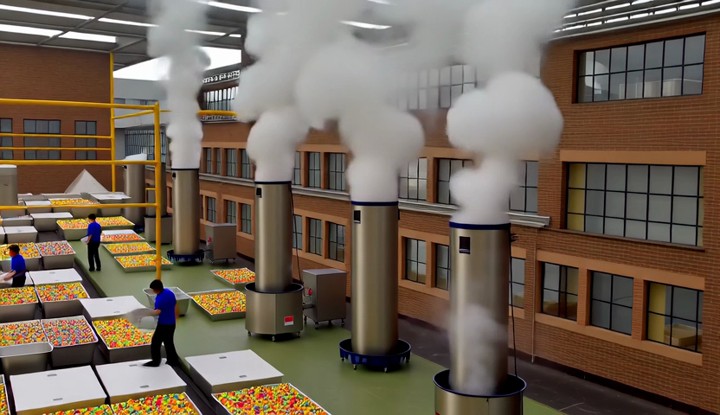}\hspace{-0.0037\textwidth}
\includegraphics[width=0.165\textwidth]{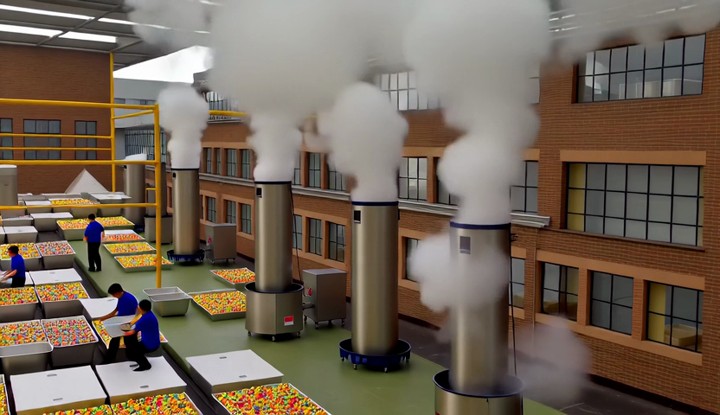}\hspace{-0.0037\textwidth}
\includegraphics[width=0.165\textwidth]{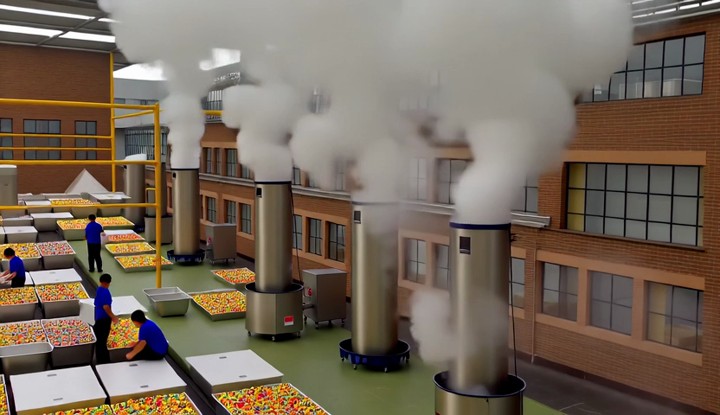}\hspace{-0.0037\textwidth}
\includegraphics[width=0.165\textwidth]{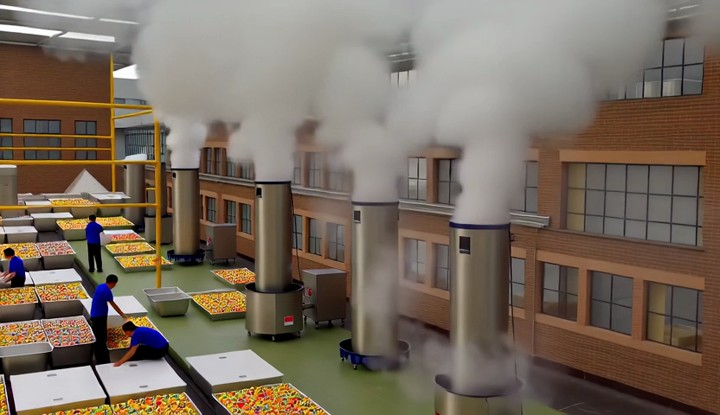}
\vspace{-0.5em}

\end{subfigure}

\vspace{-1em} \caption{5-second video generation on \texttt{Wan2.1-T2V-1.3B-480P} \textbf{\red{using a single RTX 5090}}.\\\textit{Prompt = "A slow zoom into a bustling candy factory during operation, with smoke gently rising from multiple chimneys. The scene showcases workers actively engaged in various tasks such as mixing ingredients, operating machinery, and packaging candies. The interior is bright and colorful, filled with vibrant hues of the candies being produced. The exterior features a mix of old and modern industrial architecture, with large windows allowing glimpses of the production lines inside. The focus gradually narrows from an overview of the factory grounds to a detailed close-up of the chimneys emitting soft plumes of smoke. Wide shot transitioning to medium shot with a slow zoom effect."}}
\label{fig:comparison_1_3b_video_11}
\end{figure}

\begin{figure}[H]
\centering
\begin{subfigure}{\textwidth}
\centering
\textbf{\large Original} ~~\textit{\large Latency: 184s}\\
\vspace{0.1cm}

\includegraphics[width=0.165\textwidth]{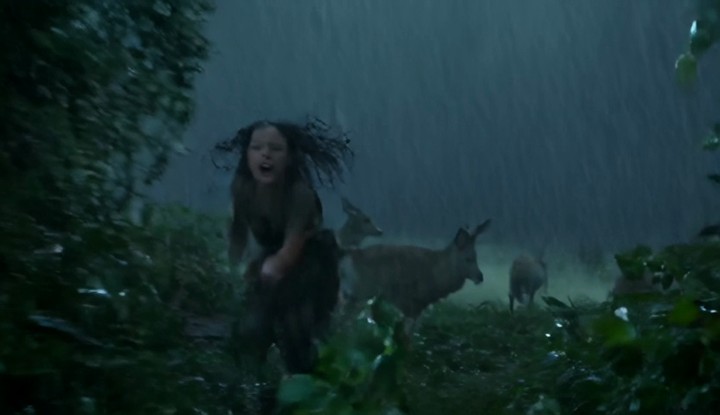}\hspace{-0.0037\textwidth}
\includegraphics[width=0.165\textwidth]{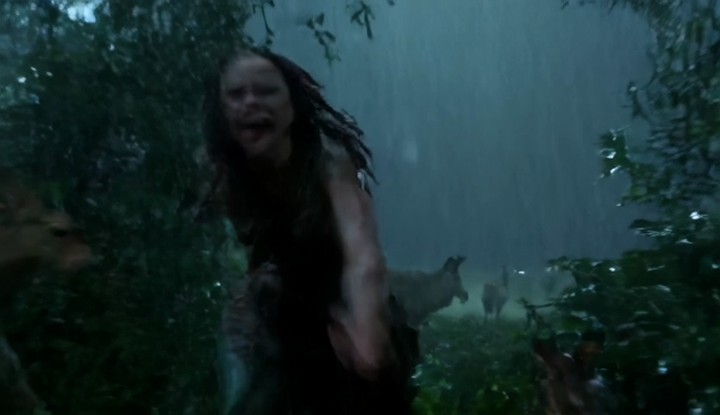}\hspace{-0.0037\textwidth}
\includegraphics[width=0.165\textwidth]{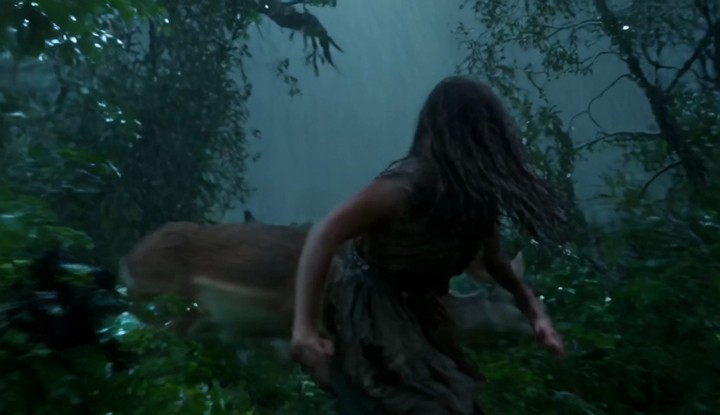}\hspace{-0.0037\textwidth}
\includegraphics[width=0.165\textwidth]{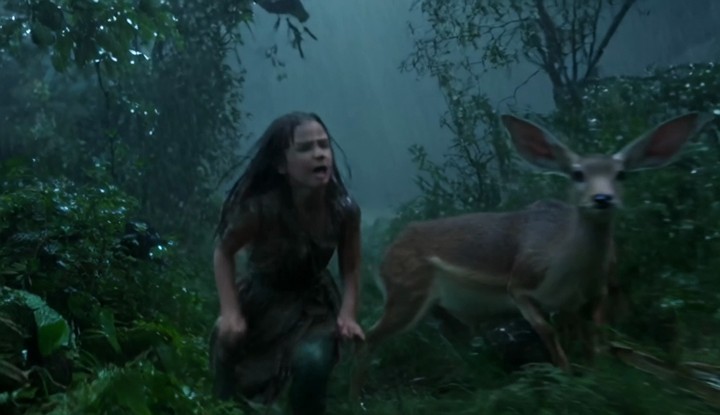}\hspace{-0.0037\textwidth}
\includegraphics[width=0.165\textwidth]{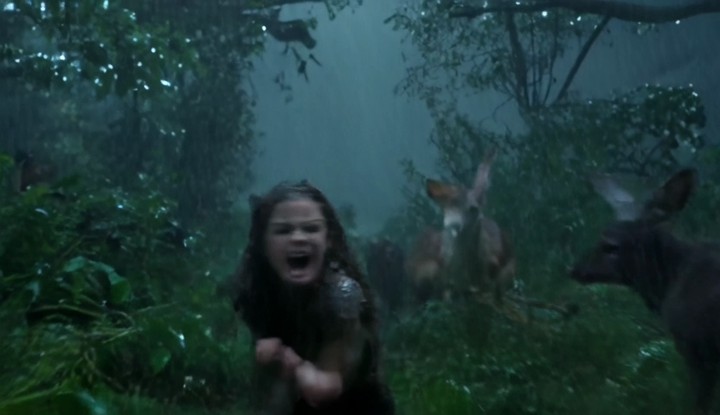}\hspace{-0.0037\textwidth}
\includegraphics[width=0.165\textwidth]{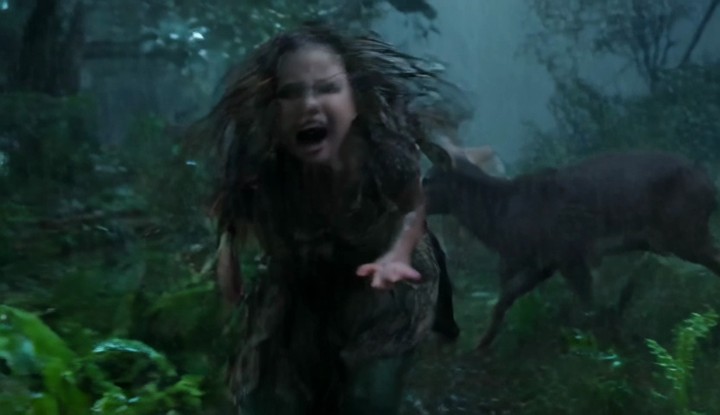}
\vspace{-0.5em}

\end{subfigure}

\vspace{0.2cm}

\begin{subfigure}{\textwidth}
\centering
\textbf{\large FastVideo} ~~\textit{\large Latency: 5.3s}\\
\vspace{0.1cm}

\includegraphics[width=0.165\textwidth]{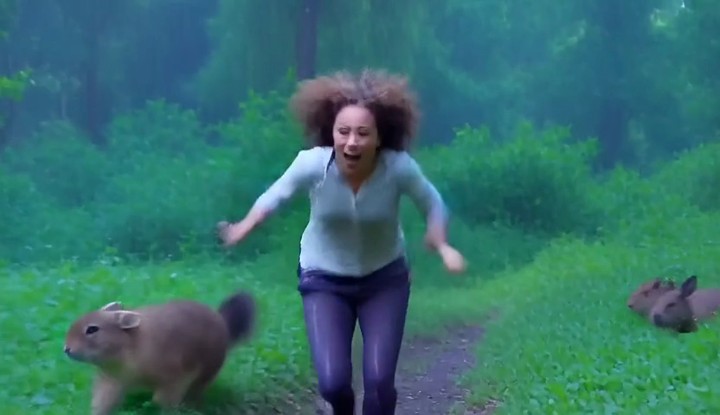}\hspace{-0.0037\textwidth}
\includegraphics[width=0.165\textwidth]{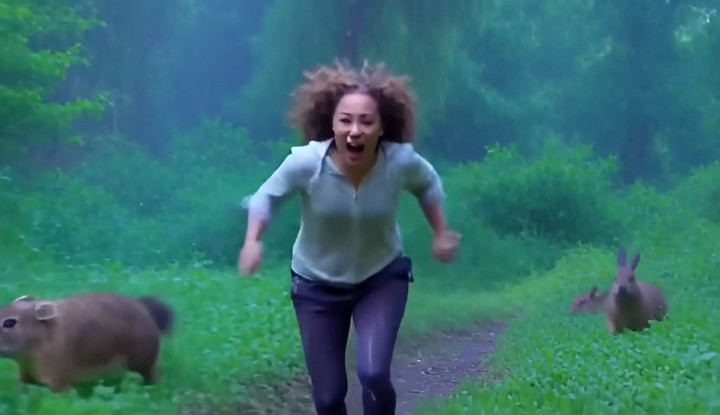}\hspace{-0.0037\textwidth}
\includegraphics[width=0.165\textwidth]{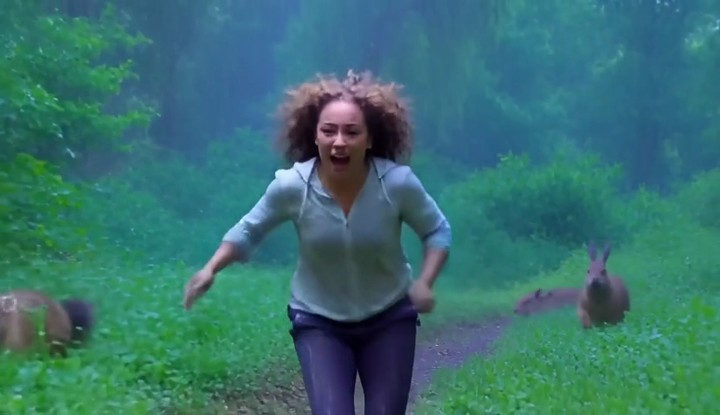}\hspace{-0.0037\textwidth}
\includegraphics[width=0.165\textwidth]{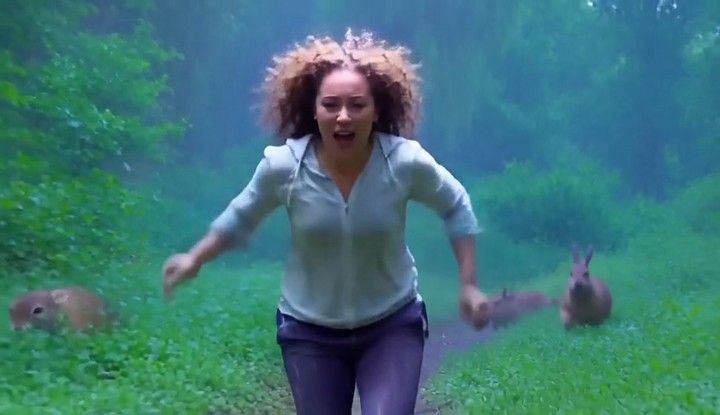}\hspace{-0.0037\textwidth}
\includegraphics[width=0.165\textwidth]{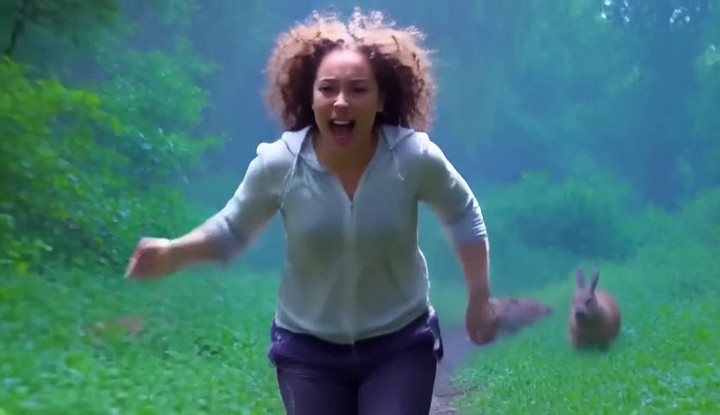}\hspace{-0.0037\textwidth}
\includegraphics[width=0.165\textwidth]{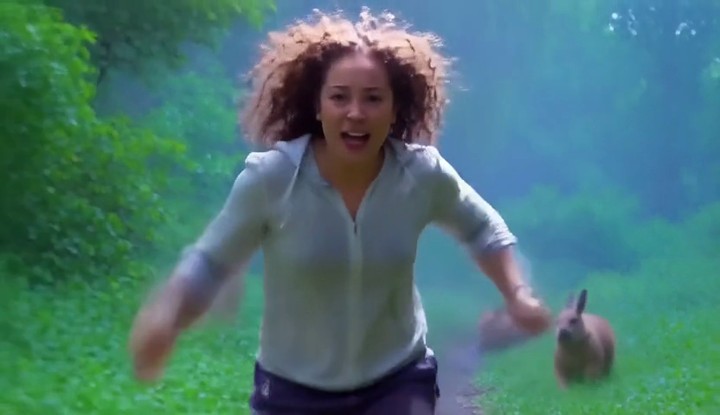}
\vspace{-0.5em}

\end{subfigure}

\vspace{0.2cm}

\begin{subfigure}{\textwidth}
\centering
\textbf{\large TurboDiffusion} ~~\textit{\large Latency: \bf \red{1.9s}}\\
\vspace{0.1cm}

\includegraphics[width=0.165\textwidth]{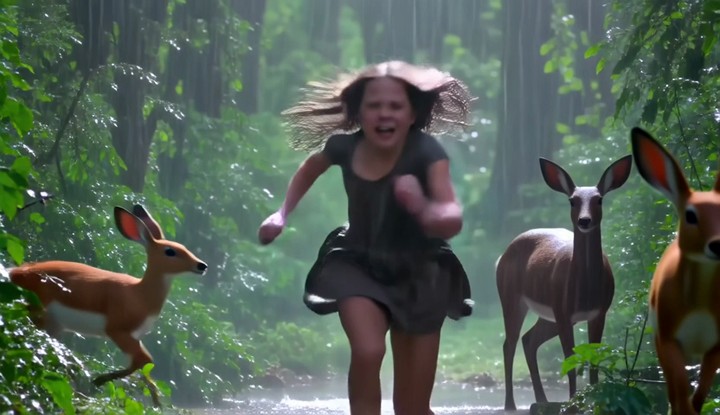}\hspace{-0.0037\textwidth}
\includegraphics[width=0.165\textwidth]{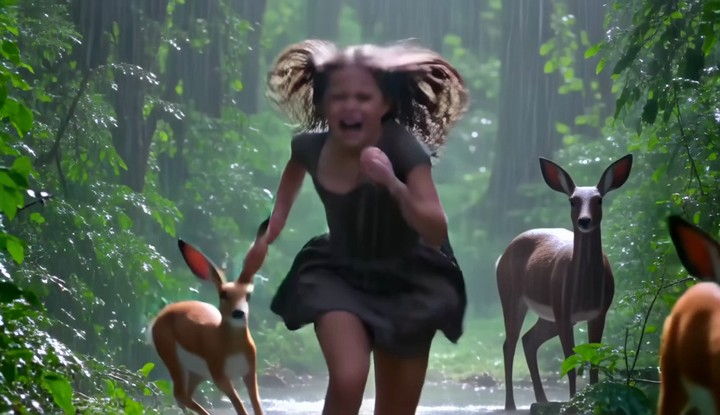}\hspace{-0.0037\textwidth}
\includegraphics[width=0.165\textwidth]{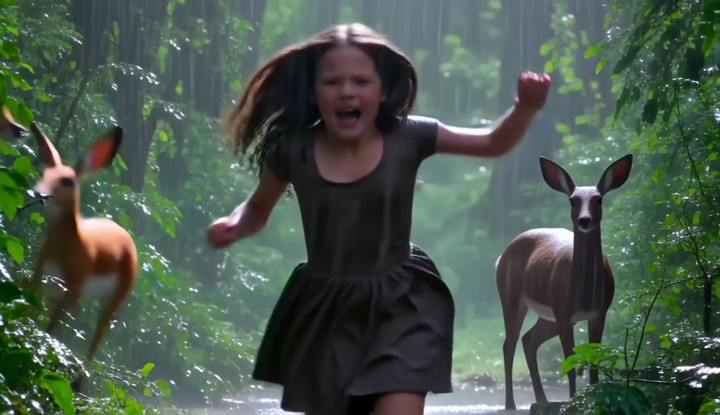}\hspace{-0.0037\textwidth}
\includegraphics[width=0.165\textwidth]{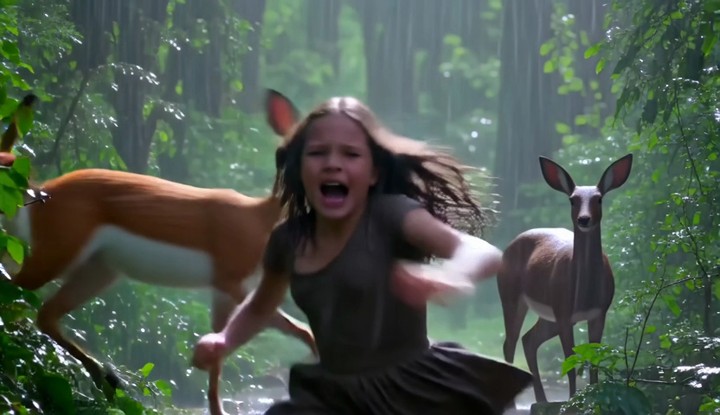}\hspace{-0.0037\textwidth}
\includegraphics[width=0.165\textwidth]{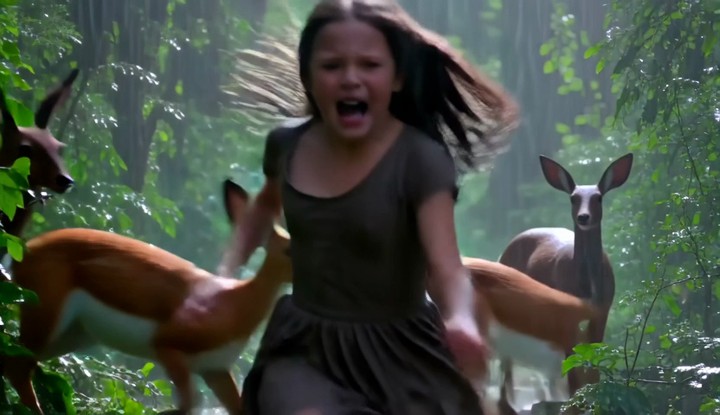}\hspace{-0.0037\textwidth}
\includegraphics[width=0.165\textwidth]{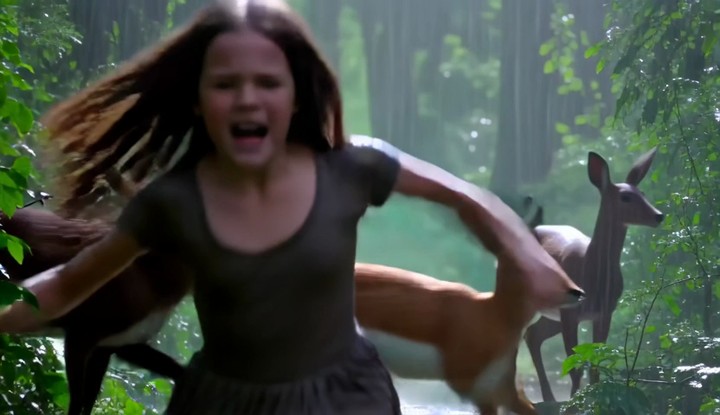}
\vspace{-0.5em}

\end{subfigure}

\vspace{-1em} \caption{5-second video generation on \texttt{Wan2.1-T2V-1.3B-480P} \textbf{\red{using a single RTX 5090}}.\\Prompt = "A dynamic and chaotic scene in a dense forest during a heavy rainstorm, capturing a real girl frantically running through the foliage. Her wild hair flows behind her as she sprints, her arms flailing and her face contorted in fear and desperation. Behind her, various \red{animals—rabbits, deer}, and birds—are also running, creating a frenzied atmosphere. The girl's clothes are soaked, clinging to her body, and she is screaming and shouting as she tries to escape. The background is a blur of greenery and rain-drenched trees, with occasional glimpses of the darkening sky. A wide-angle shot from a low angle, emphasizing the urgency and chaos of the moment."}
\label{fig:comparison_1_3b_video_6}
\end{figure}

\subsubsection{Wan2.1-T2V-14B-720P}

\begin{figure}[H]
\centering
\begin{subfigure}{\textwidth}
\centering
\textbf{\large Original} ~~\textit{\large Latency: 4767s}\\
\vspace{0.1cm}

\includegraphics[width=0.165\textwidth]{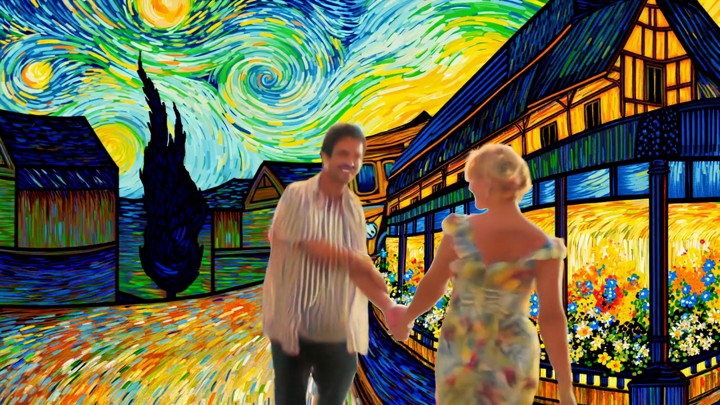}\hspace{-0.0037\textwidth}
\includegraphics[width=0.165\textwidth]{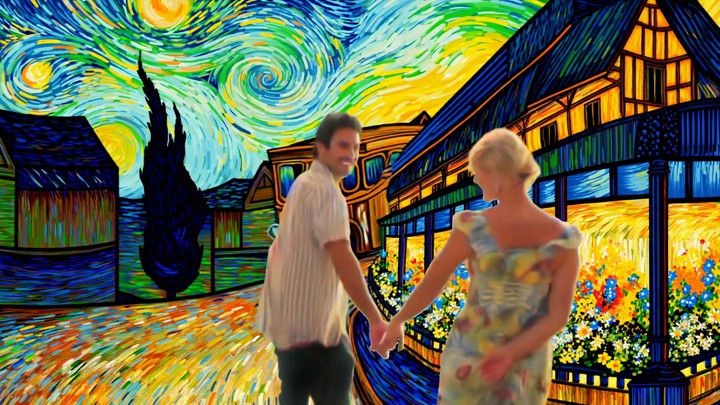}\hspace{-0.0037\textwidth}
\includegraphics[width=0.165\textwidth]{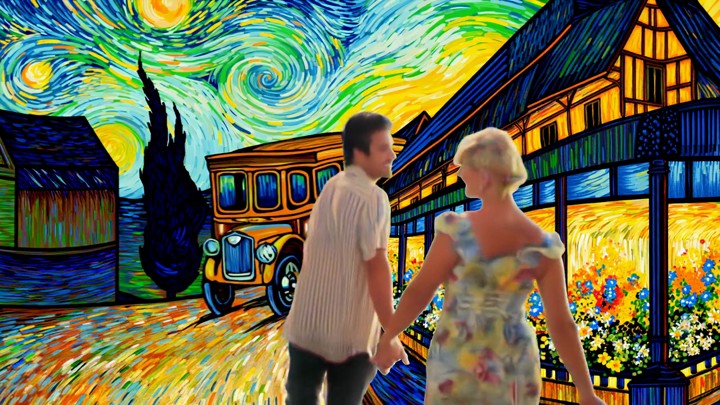}\hspace{-0.0037\textwidth}
\includegraphics[width=0.165\textwidth]{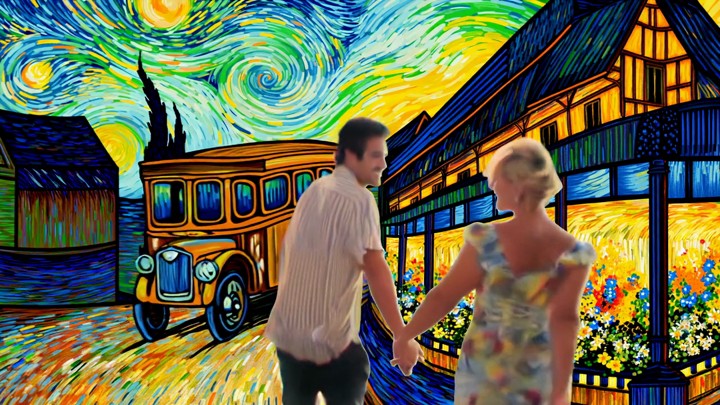}\hspace{-0.0037\textwidth}
\includegraphics[width=0.165\textwidth]{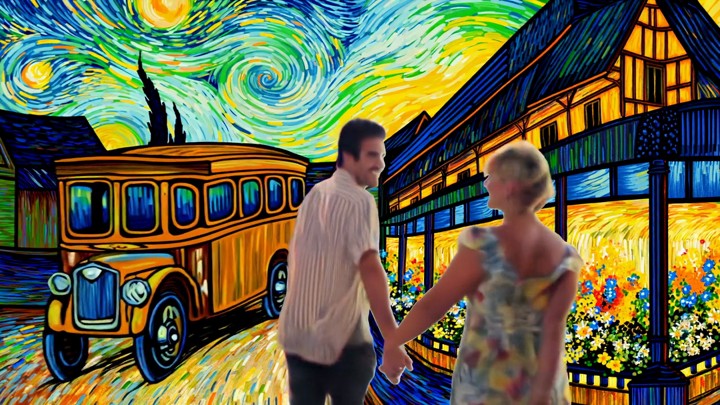}\hspace{-0.0037\textwidth}
\includegraphics[width=0.165\textwidth]{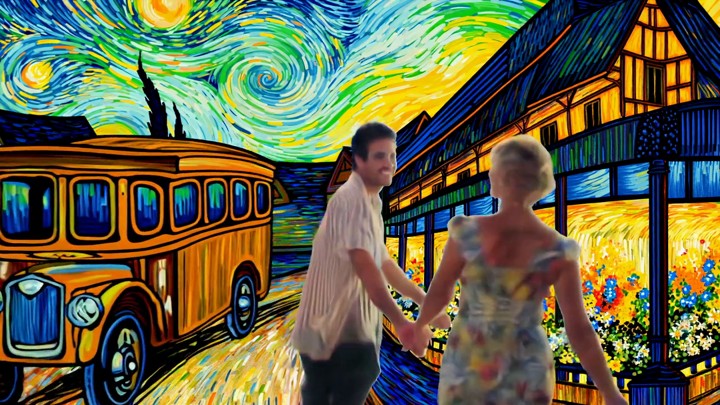}
\vspace{-0.5em}

\end{subfigure}

\vspace{0.2cm}

\begin{subfigure}{\textwidth}
\centering
\textbf{\large FastVideo} ~~\textit{\large Latency: 72.6s}\\
\vspace{0.1cm}

\includegraphics[width=0.165\textwidth]{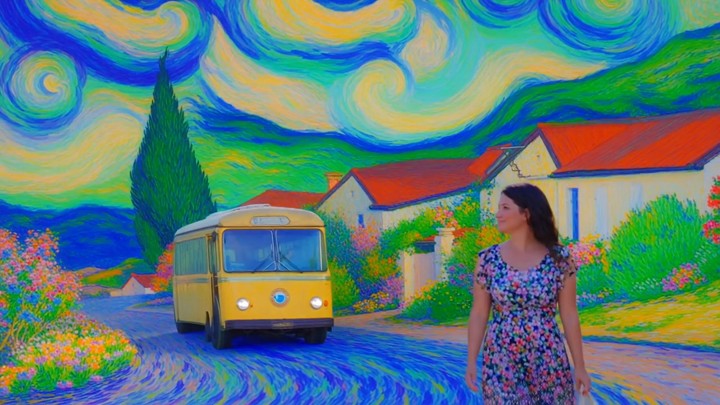}\hspace{-0.0037\textwidth}
\includegraphics[width=0.165\textwidth]{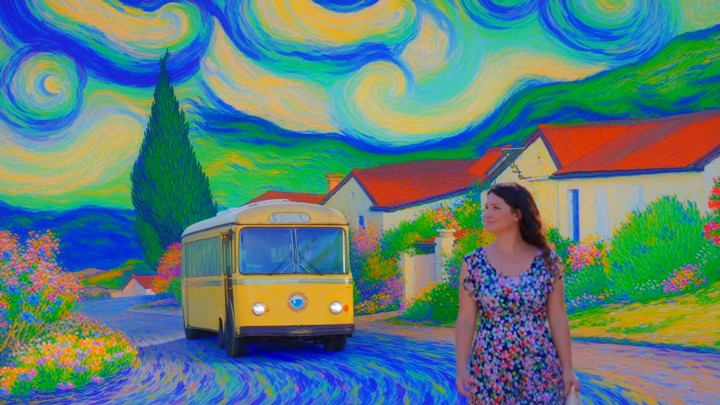}\hspace{-0.0037\textwidth}
\includegraphics[width=0.165\textwidth]{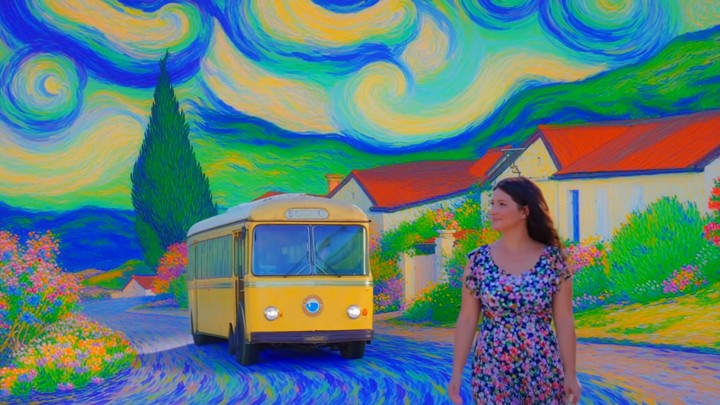}\hspace{-0.0037\textwidth}
\includegraphics[width=0.165\textwidth]{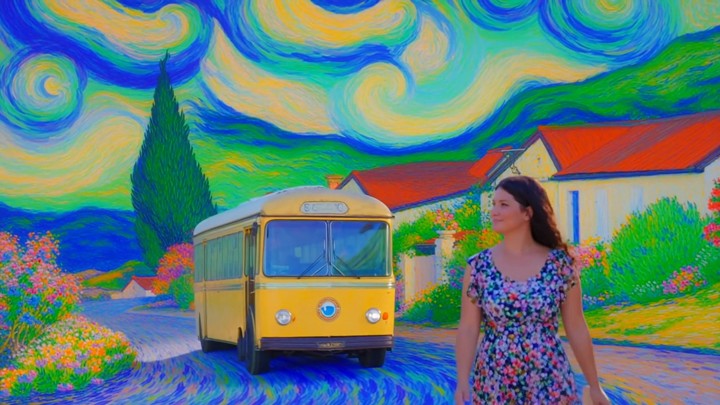}\hspace{-0.0037\textwidth}
\includegraphics[width=0.165\textwidth]{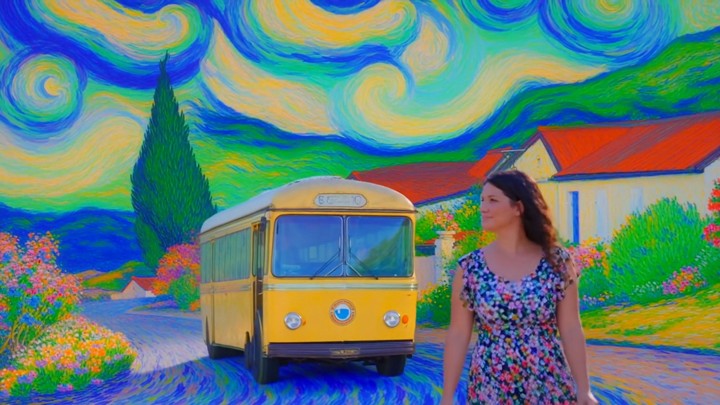}\hspace{-0.0037\textwidth}
\includegraphics[width=0.165\textwidth]{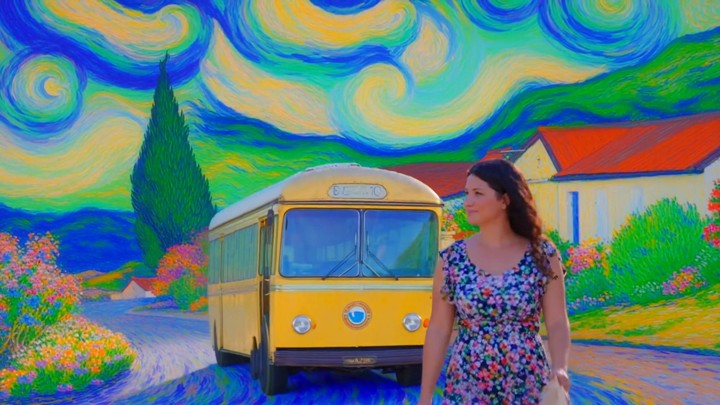}
\vspace{-0.5em}

\end{subfigure}

\vspace{0.2cm}

\begin{subfigure}{\textwidth}
\centering
\textbf{\large TurboDiffusion} ~~\textit{\large Latency: \bf \red{24s}}\\
\vspace{0.1cm}

\includegraphics[width=0.165\textwidth]{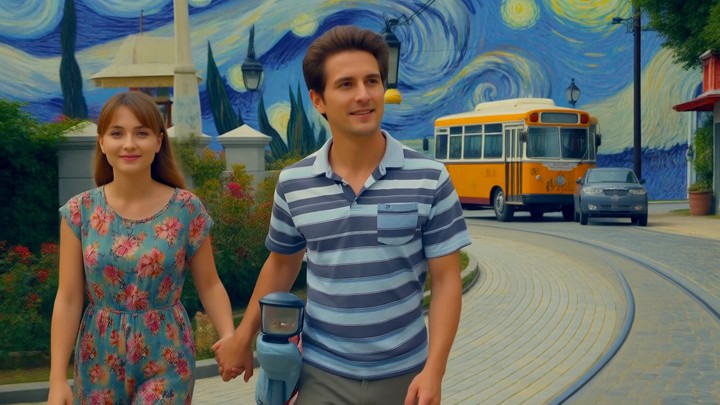}\hspace{-0.0037\textwidth}
\includegraphics[width=0.165\textwidth]{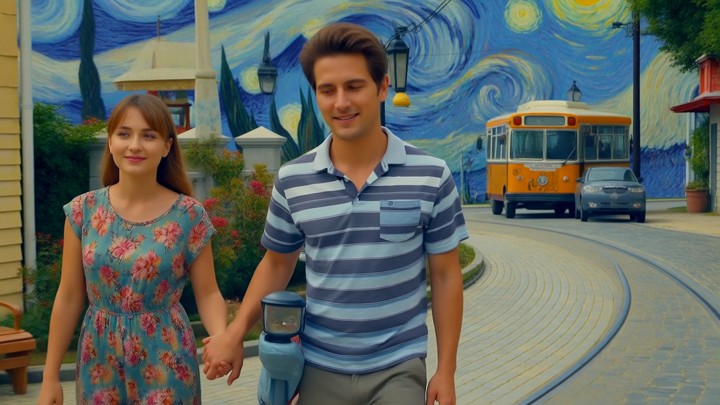}\hspace{-0.0037\textwidth}
\includegraphics[width=0.165\textwidth]{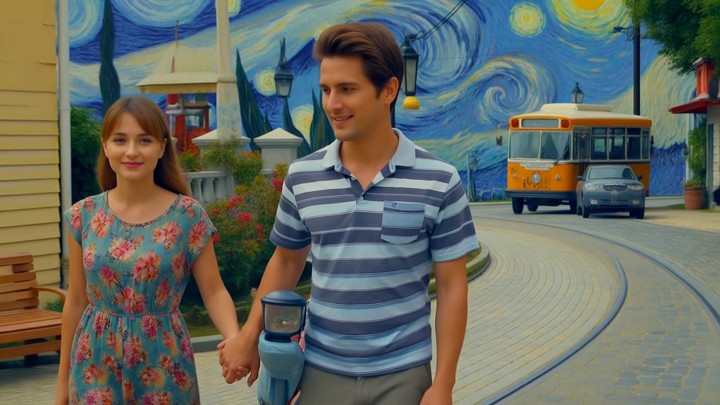}\hspace{-0.0037\textwidth}
\includegraphics[width=0.165\textwidth]{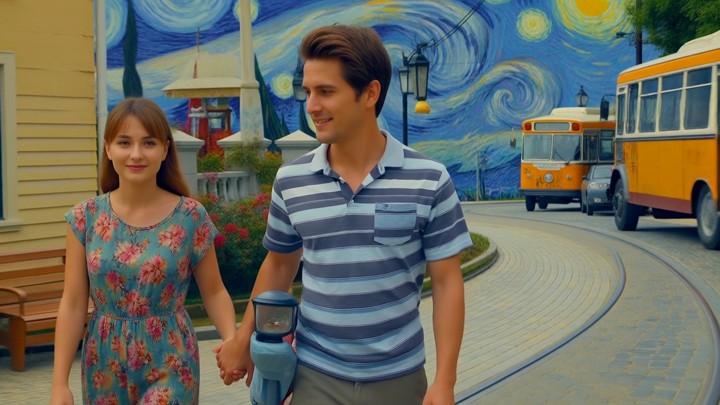}\hspace{-0.0037\textwidth}
\includegraphics[width=0.165\textwidth]{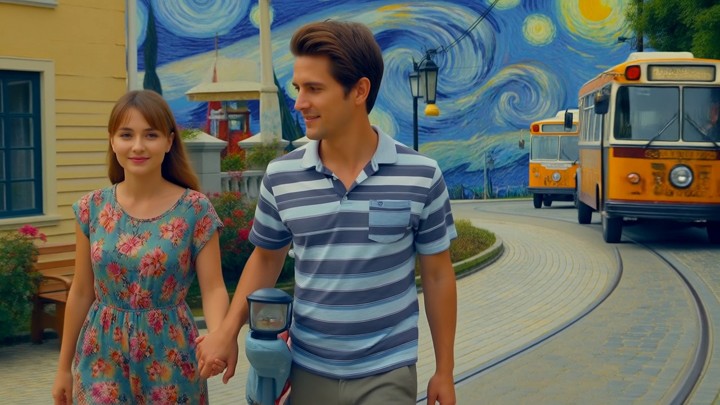}\hspace{-0.0037\textwidth}
\includegraphics[width=0.165\textwidth]{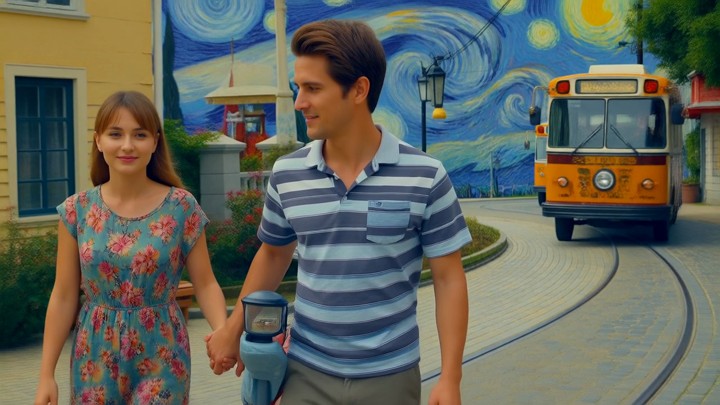}
\vspace{-0.5em}

\end{subfigure}

\vspace{-1em} \caption{5-second video generation on \texttt{Wan2.1-T2V-14B-720P} \textbf{\red{using a single RTX 5090}}.\\\textit{Prompt = "\red{A man and woman}, walking hand in hand down a vibrant, winding street, their figures illuminated against a backdrop reminiscent of a Van Gogh painting. They are strolling leisurely, with the man wearing a casual, striped shirt and the woman in a floral dress, both with warm, content expressions. Behind them, a vintage bus passes slowly, adding to the lively atmosphere of the scene. The background features swirling brushstrokes of blues, yellows, and greens, typical of Van Gogh\u2019s style, with a charming village street filled with blooming flowers and quaint buildings. Medium shot, capturing the couple from mid-torso up, with the bus visible in the background."}}
\label{fig:comparison_14b_720p_video_4}
\end{figure}

\begin{figure}[H]
\centering
\begin{subfigure}{\textwidth}
\centering
\textbf{\large Original} ~~\textit{\large Latency: 4767s}\\
\vspace{0.1cm}

\includegraphics[width=0.165\textwidth]{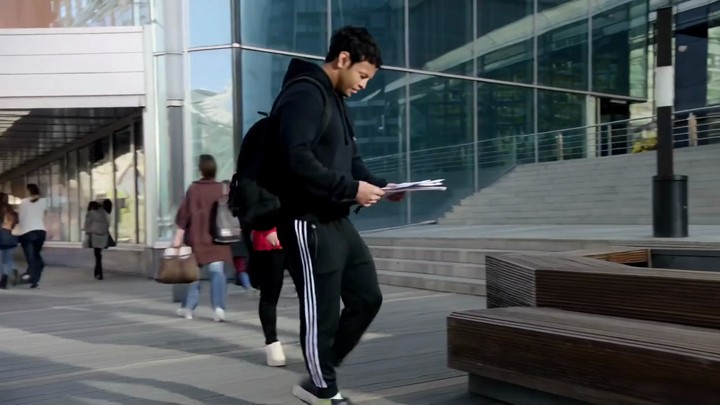}\hspace{-0.0037\textwidth}
\includegraphics[width=0.165\textwidth]{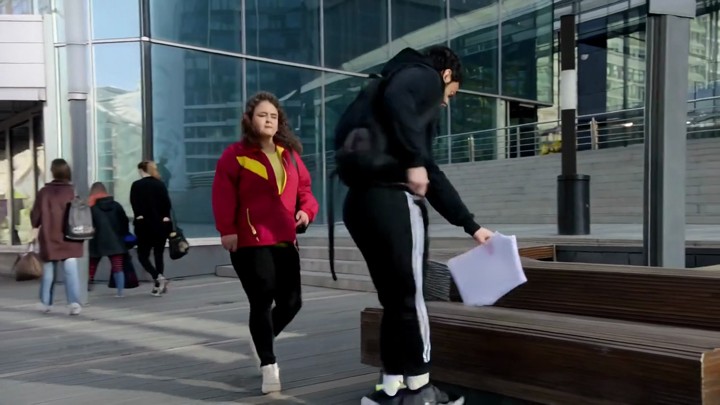}\hspace{-0.0037\textwidth}
\includegraphics[width=0.165\textwidth]{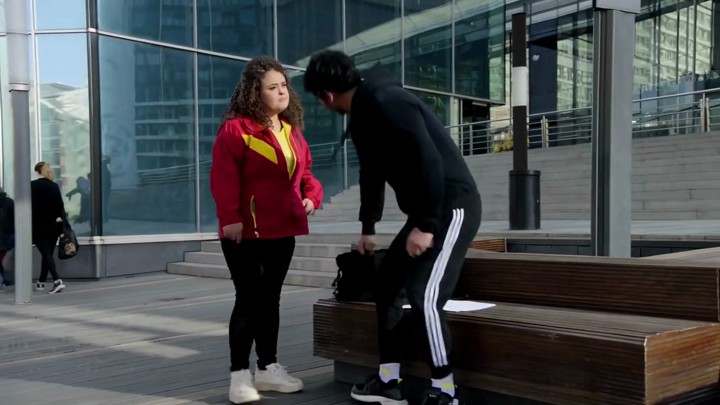}\hspace{-0.0037\textwidth}
\includegraphics[width=0.165\textwidth]{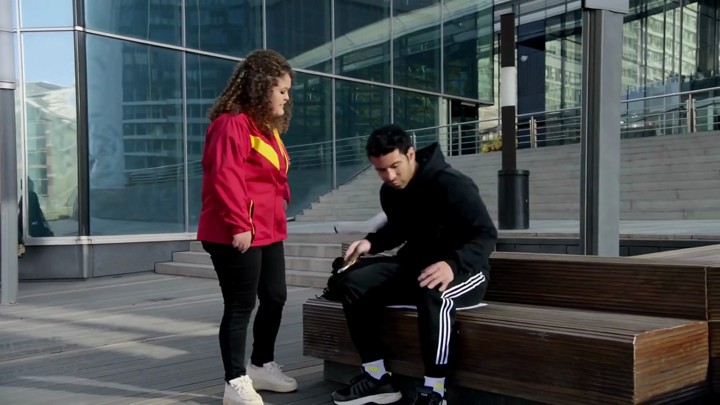}\hspace{-0.0037\textwidth}
\includegraphics[width=0.165\textwidth]{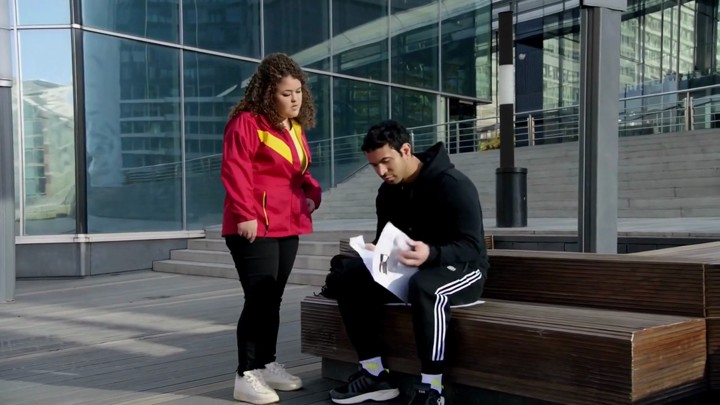}\hspace{-0.0037\textwidth}
\includegraphics[width=0.165\textwidth]{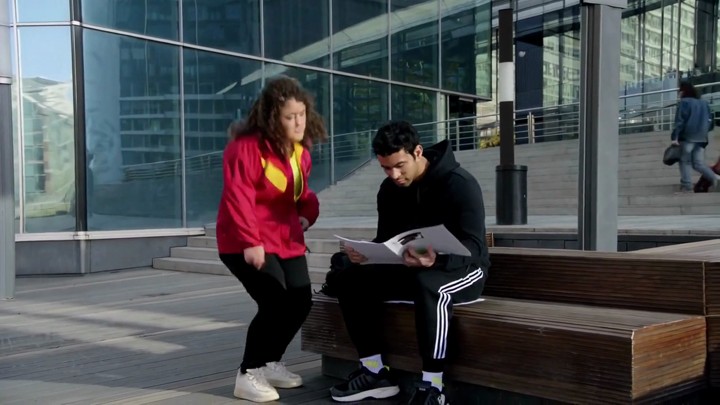}
\vspace{-0.5em}

\end{subfigure}

\vspace{0.2cm}

\begin{subfigure}{\textwidth}
\centering
\textbf{\large FastVideo} ~~\textit{\large Latency: 72.6s}\\
\vspace{0.1cm}

\includegraphics[width=0.165\textwidth]{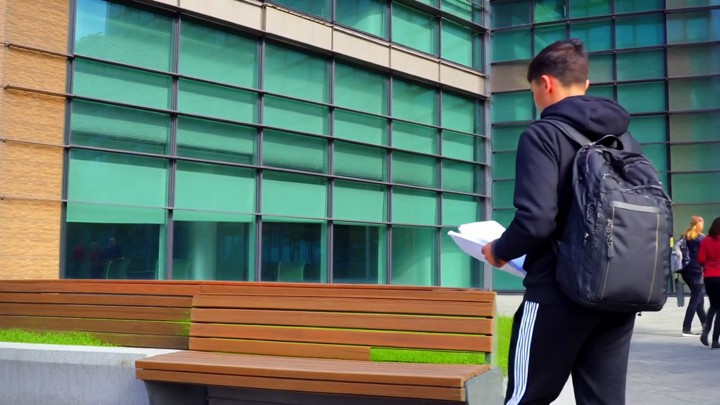}\hspace{-0.0037\textwidth}
\includegraphics[width=0.165\textwidth]{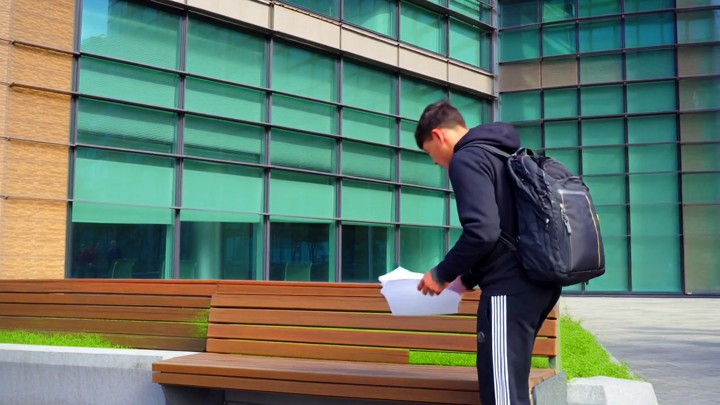}\hspace{-0.0037\textwidth}
\includegraphics[width=0.165\textwidth]{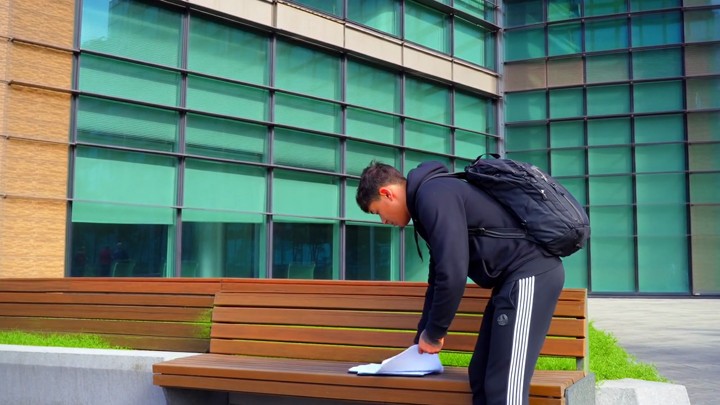}\hspace{-0.0037\textwidth}
\includegraphics[width=0.165\textwidth]{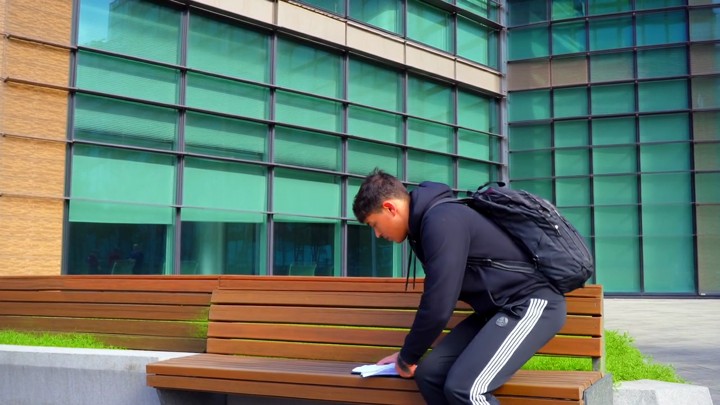}\hspace{-0.0037\textwidth}
\includegraphics[width=0.165\textwidth]{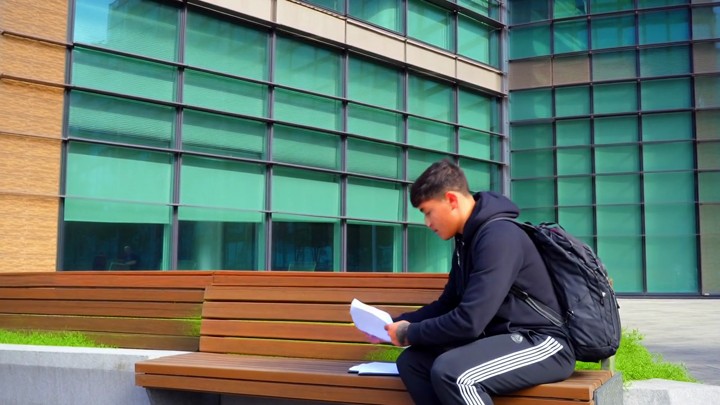}\hspace{-0.0037\textwidth}
\includegraphics[width=0.165\textwidth]{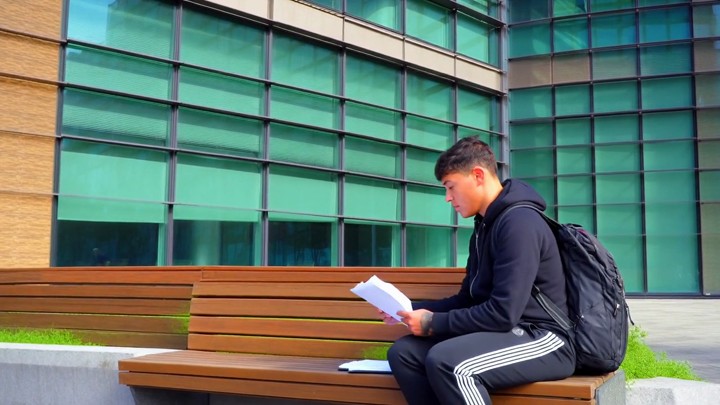}
\vspace{-0.5em}

\end{subfigure}

\vspace{0.2cm}

\begin{subfigure}{\textwidth}
\centering
\textbf{\large TurboDiffusion} ~~\textit{\large Latency: \bf \red{24s}}\\
\vspace{0.1cm}

\includegraphics[width=0.165\textwidth]{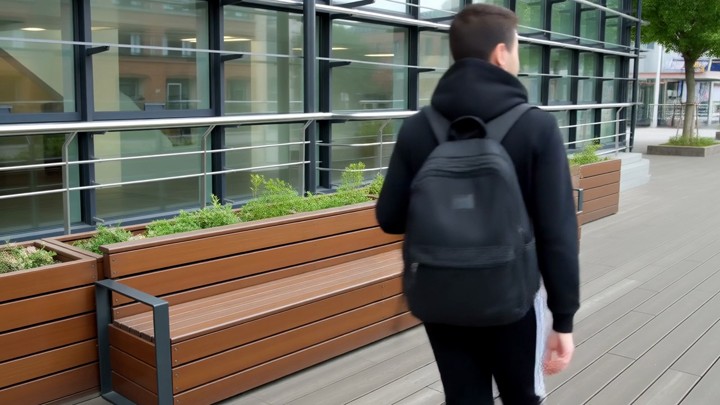}\hspace{-0.0037\textwidth}
\includegraphics[width=0.165\textwidth]{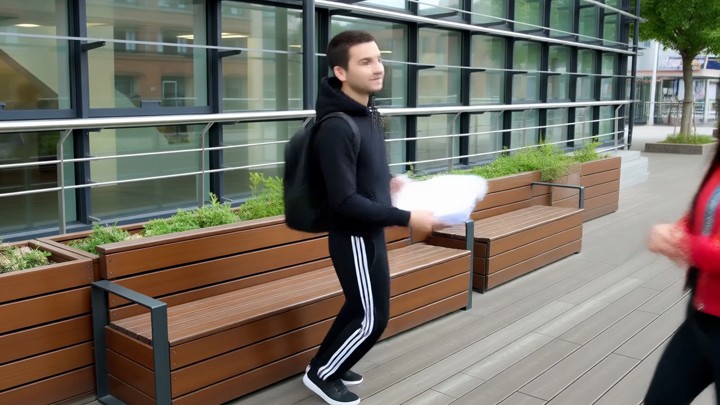}\hspace{-0.0037\textwidth}
\includegraphics[width=0.165\textwidth]{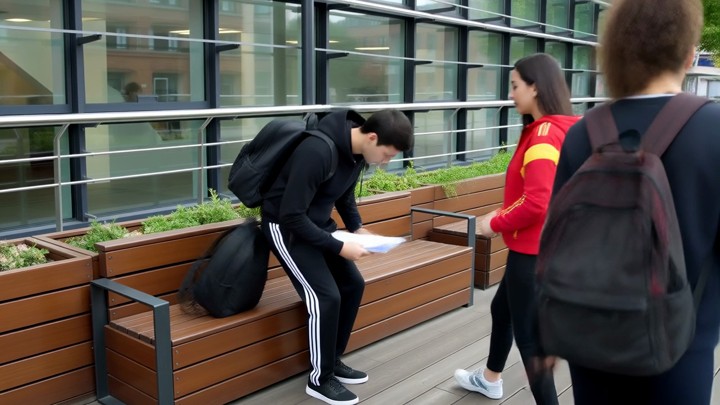}\hspace{-0.0037\textwidth}
\includegraphics[width=0.165\textwidth]{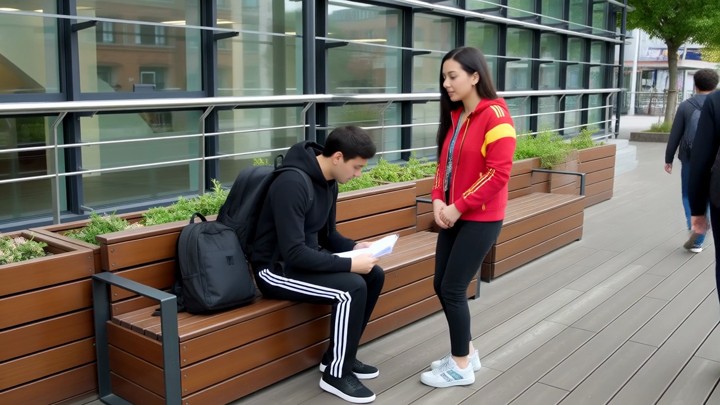}\hspace{-0.0037\textwidth}
\includegraphics[width=0.165\textwidth]{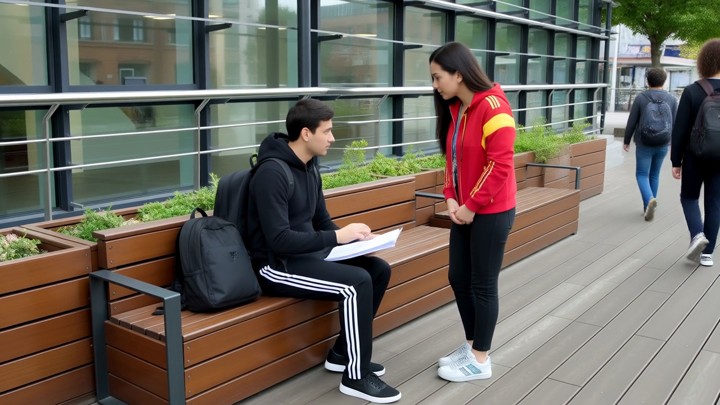}\hspace{-0.0037\textwidth}
\includegraphics[width=0.165\textwidth]{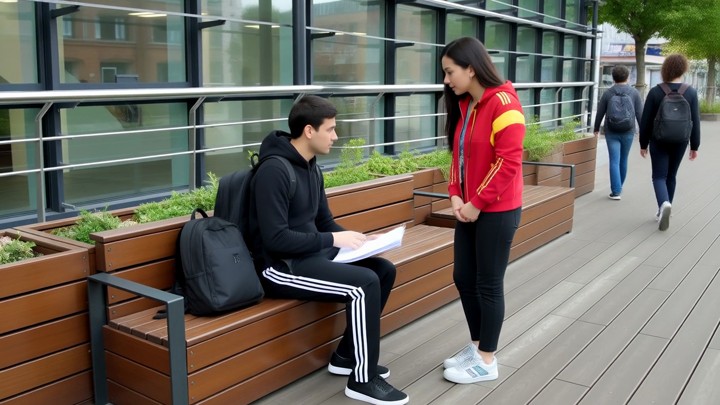}
\vspace{-0.5em}

\end{subfigure}

\vspace{-1em} \caption{5-second video generation on \texttt{Wan2.1-T2V-14B-720P} \textbf{\red{using a single RTX 5090}}.\\\textit{Prompt = "In an urban outdoor setting, a man dressed in a black hoodie and black track pants with white stripes walks toward a wooden bench situated near a modern building with large glass windows. He carries a black backpack slung over one shoulder and holds a stack of papers in his hand. As he approaches the bench, he bends down, places the papers on it, and then sits down. \red{Shortly after, a woman wearing a red jacket with yellow accents and black pants joins him.} She stands beside the bench, facing him, and appears to engage in a conversation. The man continues to review the papers while the woman listens attentively. In the background, other individuals can be seen walking by, some carrying bags, adding to the bustling yet casual atmosphere of the scene. The overall mood suggests a moment of focused discussion or preparation amidst a busy environment."}}
\label{fig:comparison_14b_720p_video_7}
\end{figure}

\begin{figure}[H]
\centering
\begin{subfigure}{\textwidth}
\centering
\textbf{\large Original} ~~\textit{\large Latency: 4767s}\\
\vspace{0.1cm}

\includegraphics[width=0.165\textwidth]{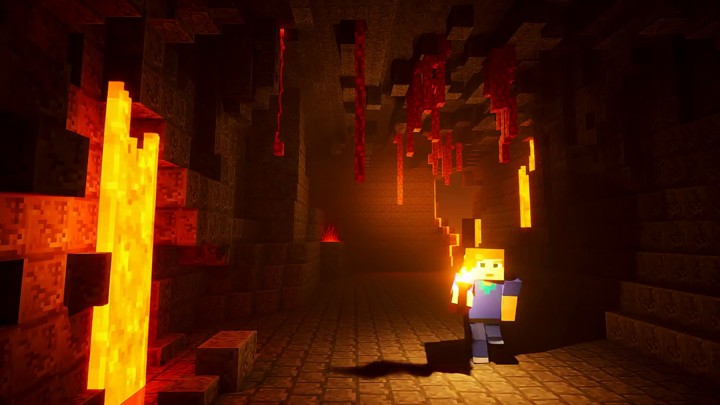}\hspace{-0.0037\textwidth}
\includegraphics[width=0.165\textwidth]{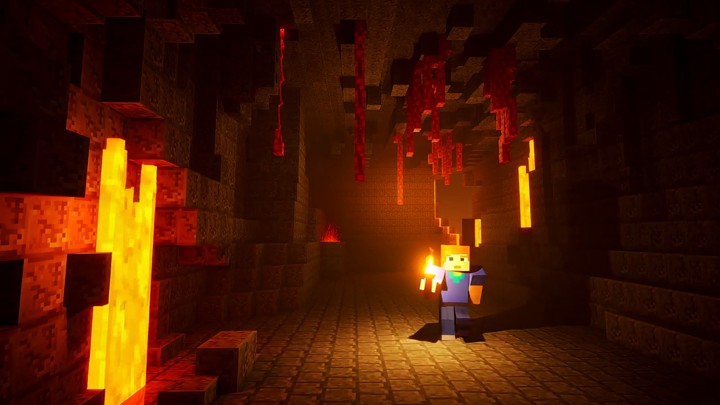}\hspace{-0.0037\textwidth}
\includegraphics[width=0.165\textwidth]{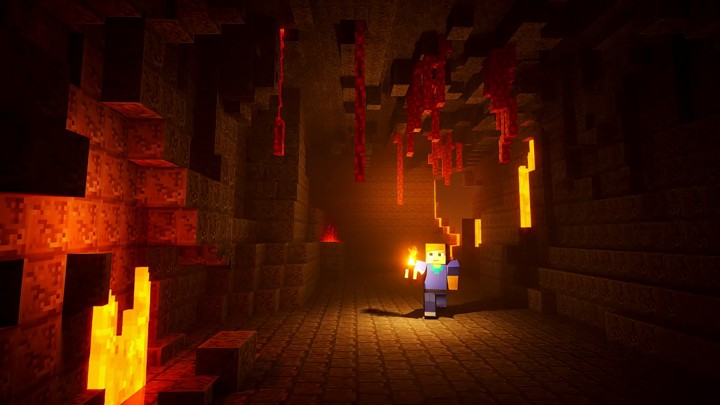}\hspace{-0.0037\textwidth}
\includegraphics[width=0.165\textwidth]{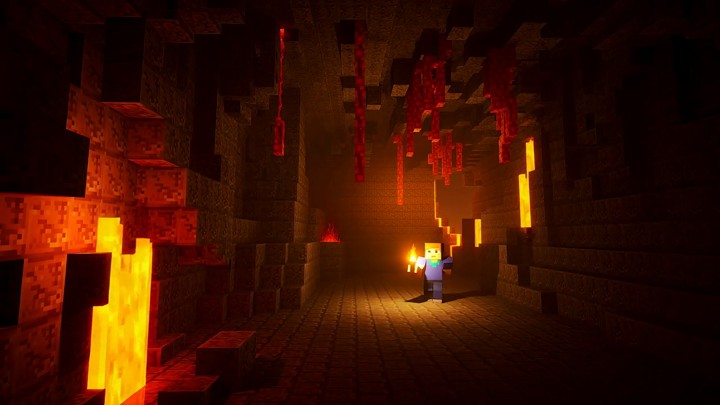}\hspace{-0.0037\textwidth}
\includegraphics[width=0.165\textwidth]{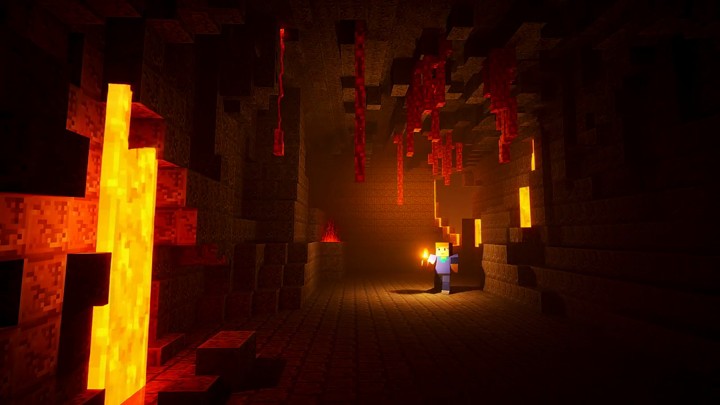}\hspace{-0.0037\textwidth}
\includegraphics[width=0.165\textwidth]{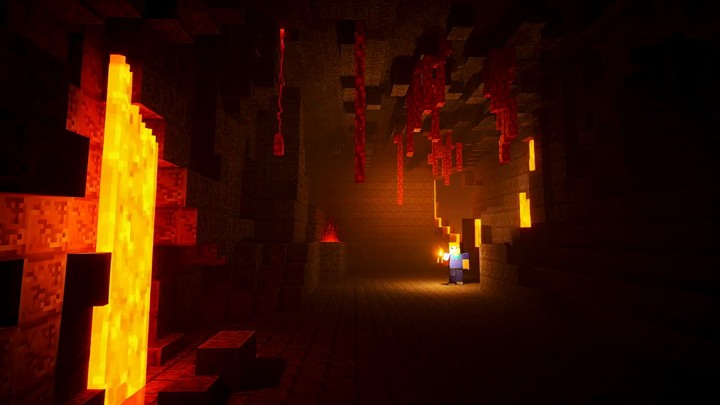}
\vspace{-0.5em}

\end{subfigure}

\vspace{0.2cm}

\begin{subfigure}{\textwidth}
\centering
\textbf{\large FastVideo} ~~\textit{\large Latency: 72.6s}\\
\vspace{0.1cm}

\includegraphics[width=0.165\textwidth]{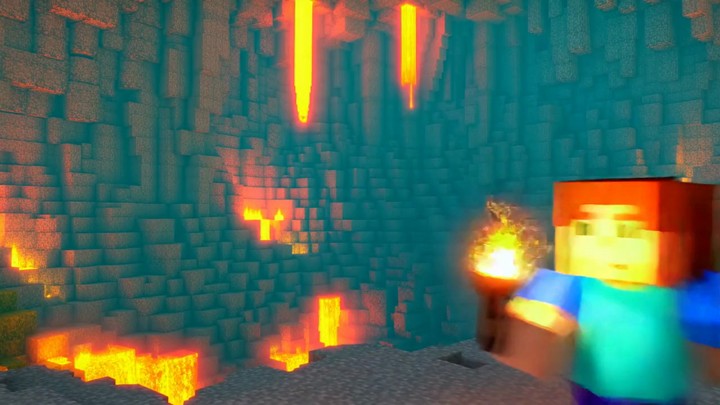}\hspace{-0.0037\textwidth}
\includegraphics[width=0.165\textwidth]{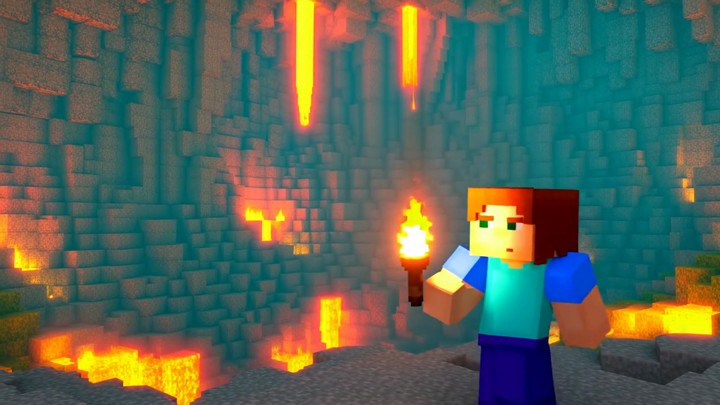}\hspace{-0.0037\textwidth}
\includegraphics[width=0.165\textwidth]{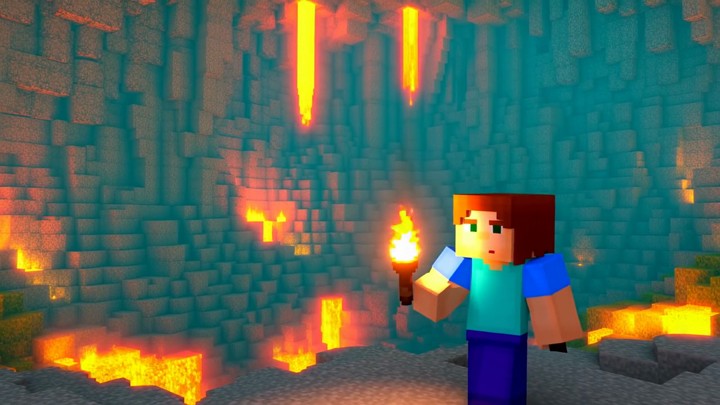}\hspace{-0.0037\textwidth}
\includegraphics[width=0.165\textwidth]{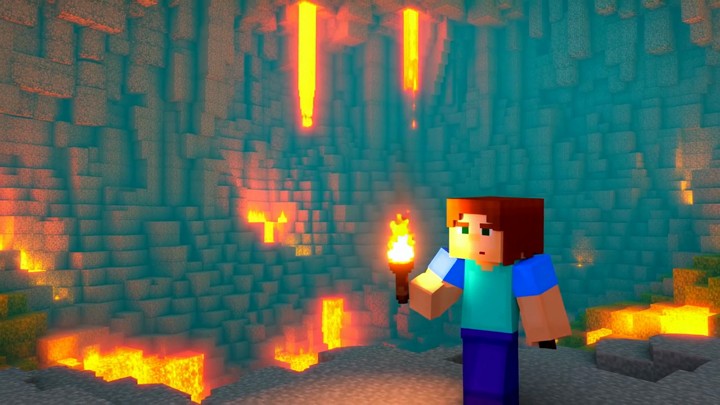}\hspace{-0.0037\textwidth}
\includegraphics[width=0.165\textwidth]{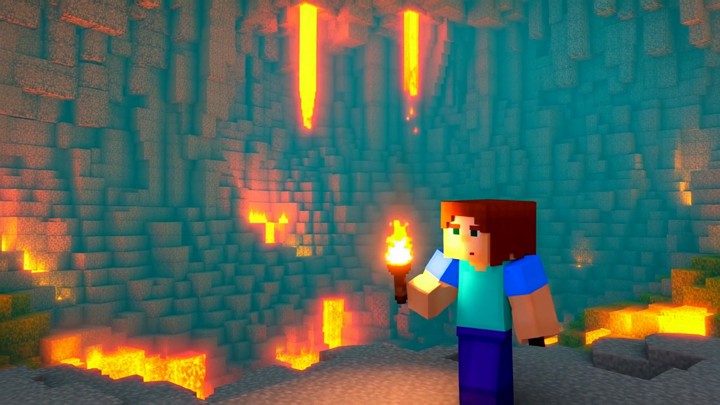}\hspace{-0.0037\textwidth}
\includegraphics[width=0.165\textwidth]{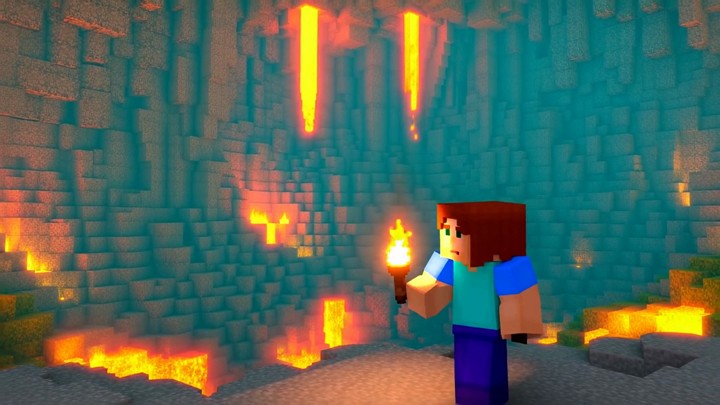}
\vspace{-0.5em}

\end{subfigure}

\vspace{0.2cm}

\begin{subfigure}{\textwidth}
\centering
\textbf{\large TurboDiffusion} ~~\textit{\large Latency: \bf \red{24s}}\\
\vspace{0.1cm}

\includegraphics[width=0.165\textwidth]{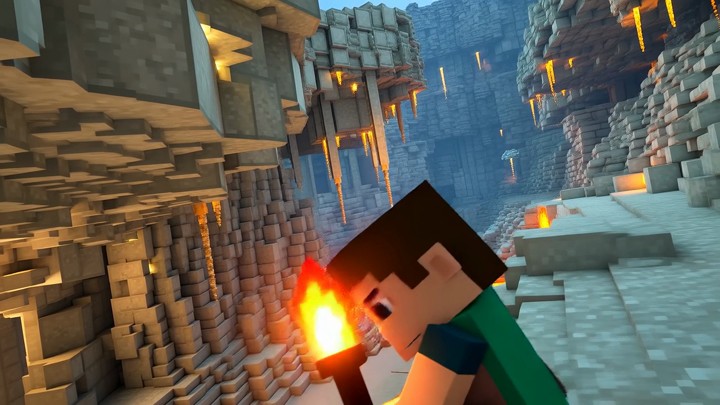}\hspace{-0.0037\textwidth}
\includegraphics[width=0.165\textwidth]{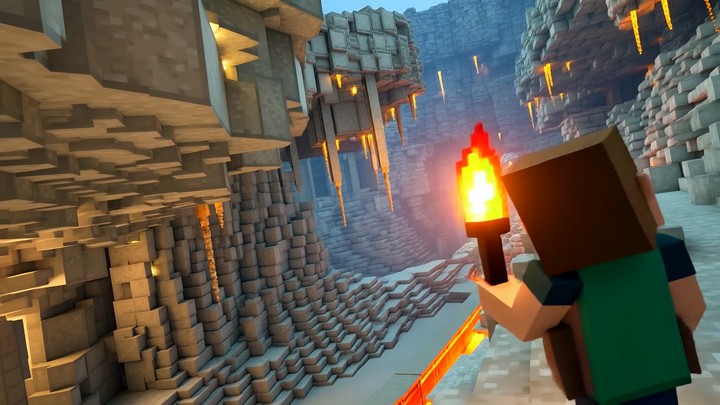}\hspace{-0.0037\textwidth}
\includegraphics[width=0.165\textwidth]{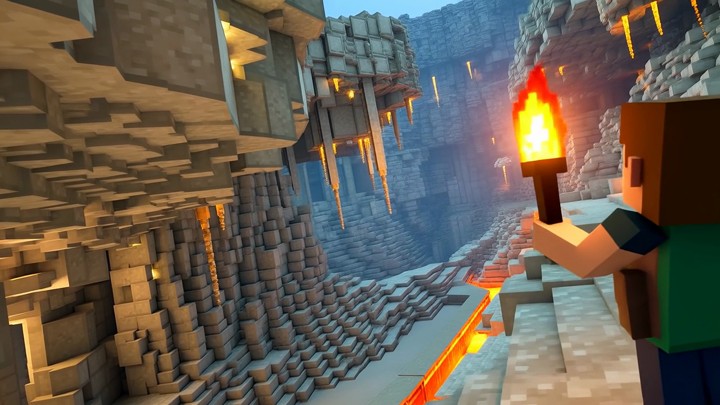}\hspace{-0.0037\textwidth}
\includegraphics[width=0.165\textwidth]{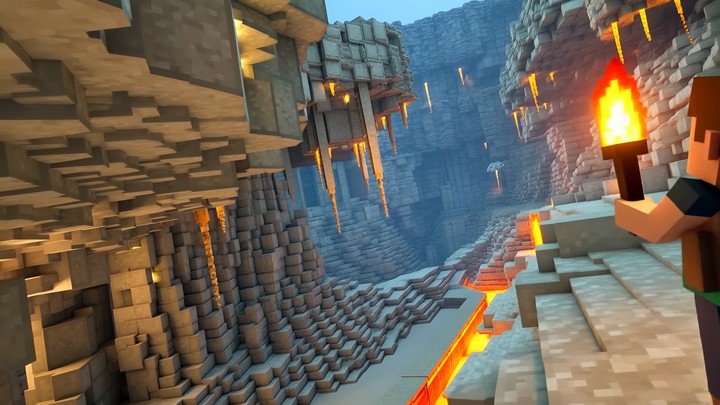}\hspace{-0.0037\textwidth}
\includegraphics[width=0.165\textwidth]{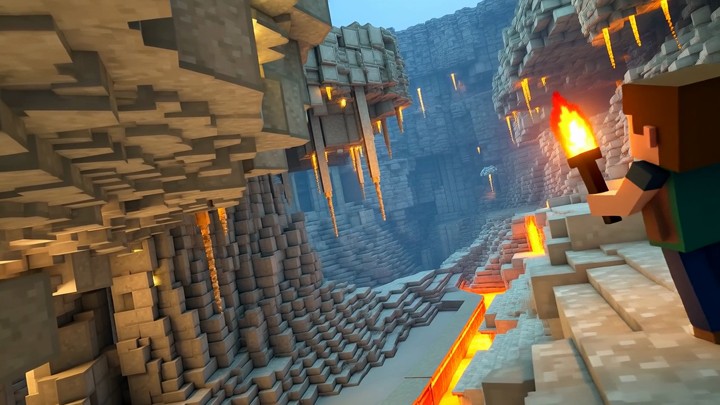}\hspace{-0.0037\textwidth}
\includegraphics[width=0.165\textwidth]{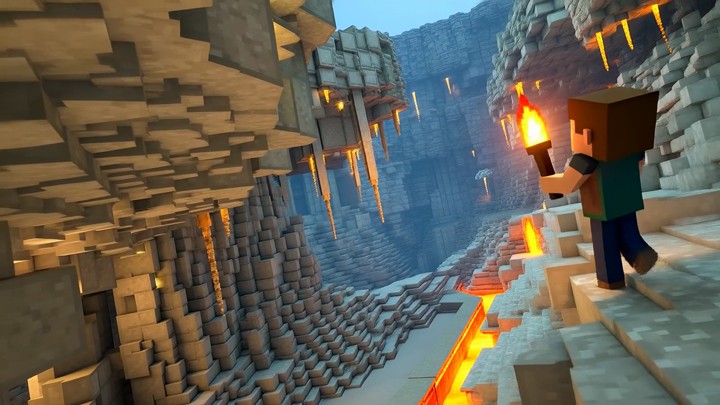}
\vspace{-0.5em}

\end{subfigure}

\vspace{-1em} \caption{5-second video generation on \texttt{Wan2.1-T2V-14B-720P} \textbf{\red{using a single RTX 5090}}.\\\textit{Prompt = "A Minecraft player character holding a torch enters a massive underground cave. The torchlight flickers against jagged stone walls, illuminating patches of iron and diamond ores embedded in the rock. Stalactites hang from the ceiling, lava flows in glowing streams nearby, and the faint sound of water dripping echoes through the cavern."}}
\label{fig:comparison_14b_720p_video_1}
\end{figure}

\begin{figure}[H]
\centering
\begin{subfigure}{\textwidth}
\centering
\textbf{\large Original} ~~\textit{\large Latency: 4767s}\\
\vspace{0.1cm}

\includegraphics[width=0.165\textwidth]{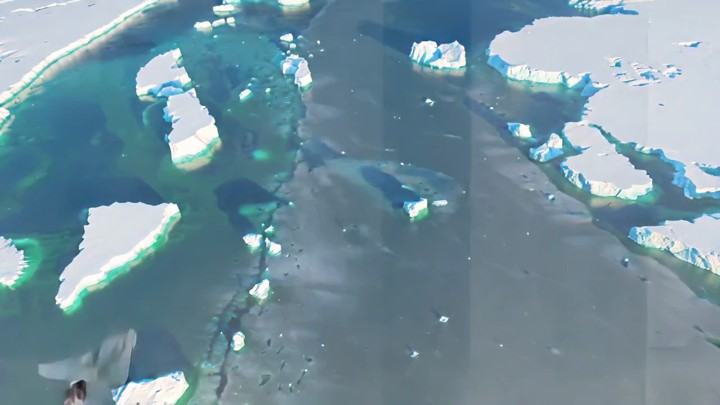}\hspace{-0.0037\textwidth}
\includegraphics[width=0.165\textwidth]{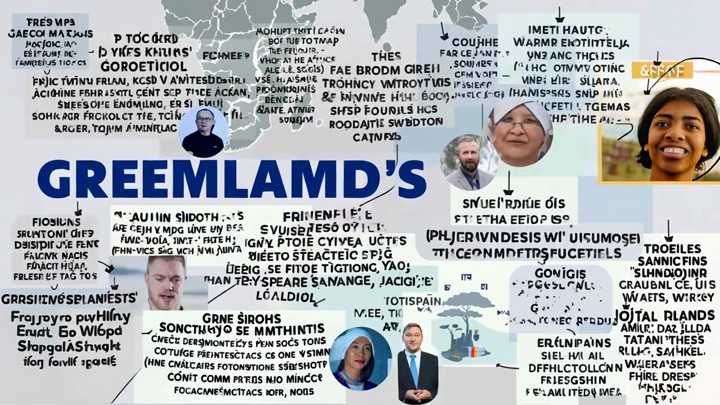}\hspace{-0.0037\textwidth}
\includegraphics[width=0.165\textwidth]{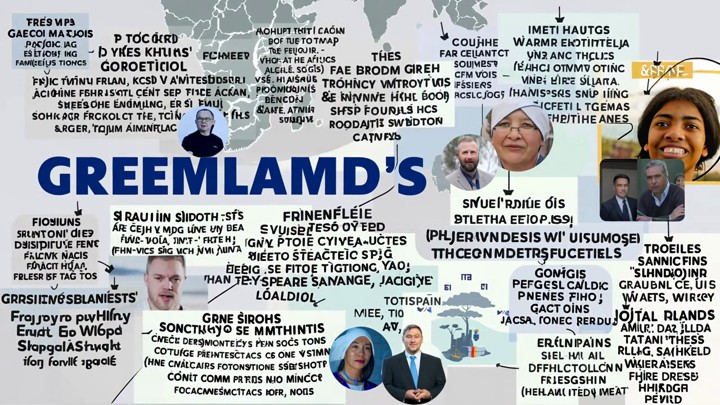}\hspace{-0.0037\textwidth}
\includegraphics[width=0.165\textwidth]{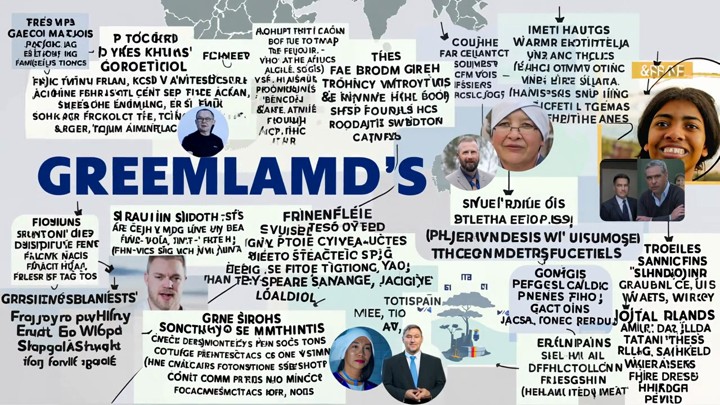}\hspace{-0.0037\textwidth}
\includegraphics[width=0.165\textwidth]{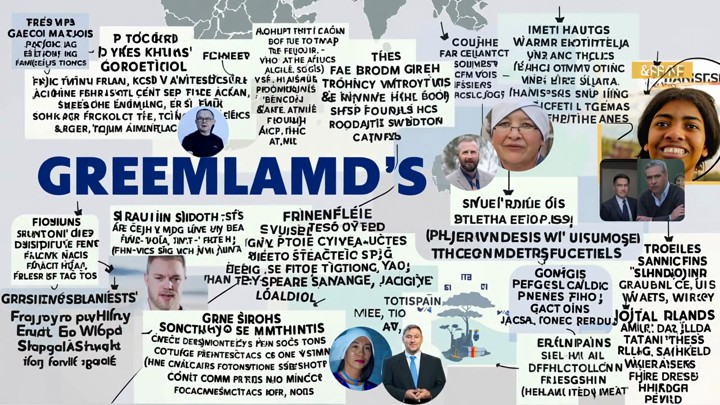}\hspace{-0.0037\textwidth}
\includegraphics[width=0.165\textwidth]{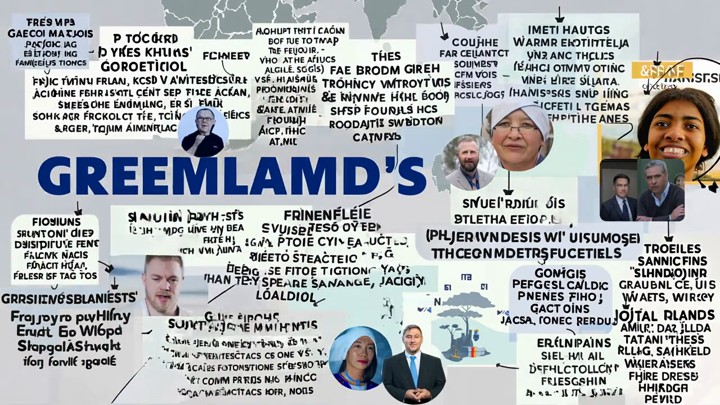}
\vspace{-0.5em}

\end{subfigure}

\vspace{0.2cm}

\begin{subfigure}{\textwidth}
\centering
\textbf{\large FastVideo} ~~\textit{\large Latency: 72.6s}\\
\vspace{0.1cm}

\includegraphics[width=0.165\textwidth]{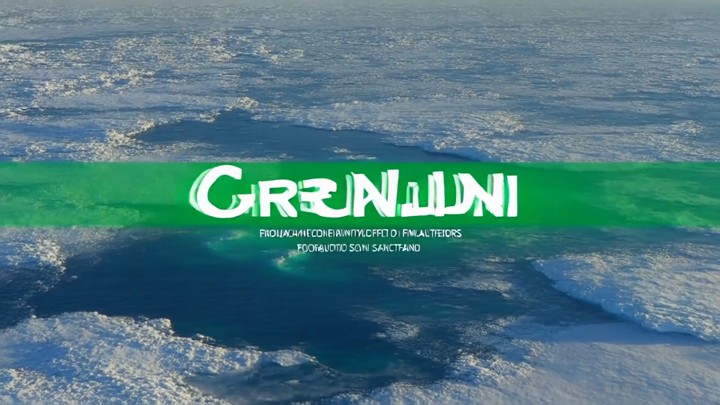}\hspace{-0.0037\textwidth}
\includegraphics[width=0.165\textwidth]{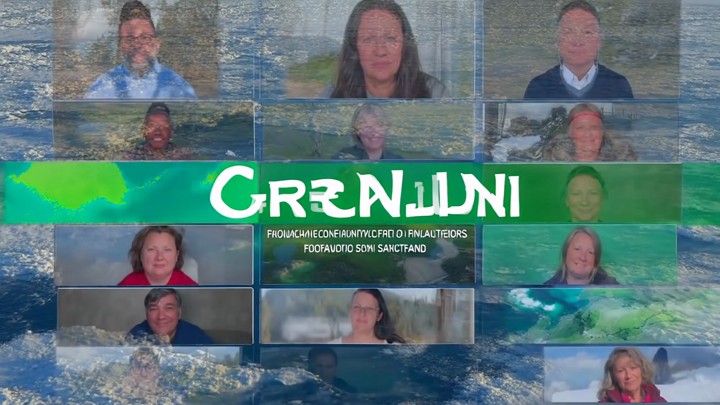}\hspace{-0.0037\textwidth}
\includegraphics[width=0.165\textwidth]{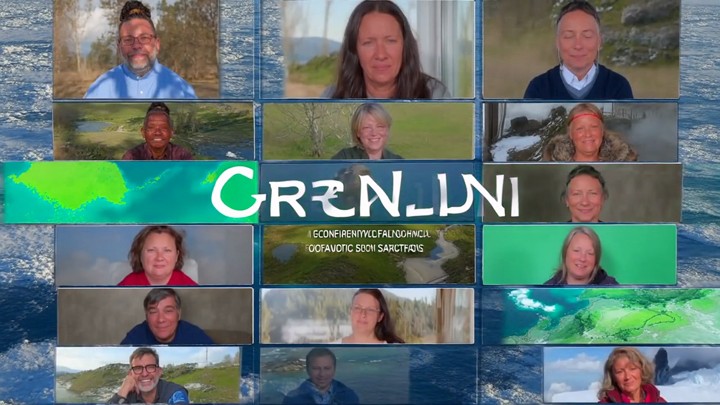}\hspace{-0.0037\textwidth}
\includegraphics[width=0.165\textwidth]{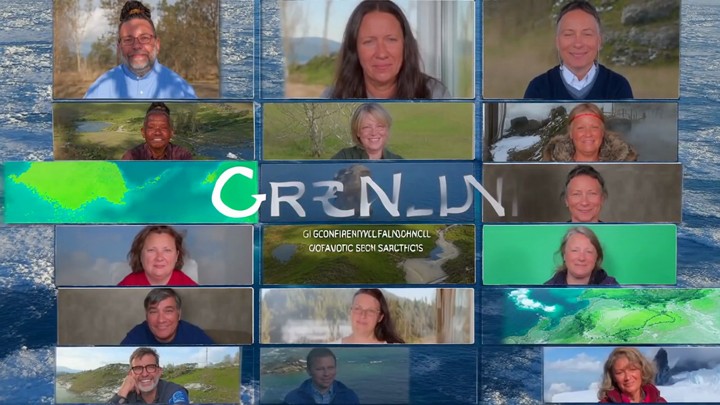}\hspace{-0.0037\textwidth}
\includegraphics[width=0.165\textwidth]{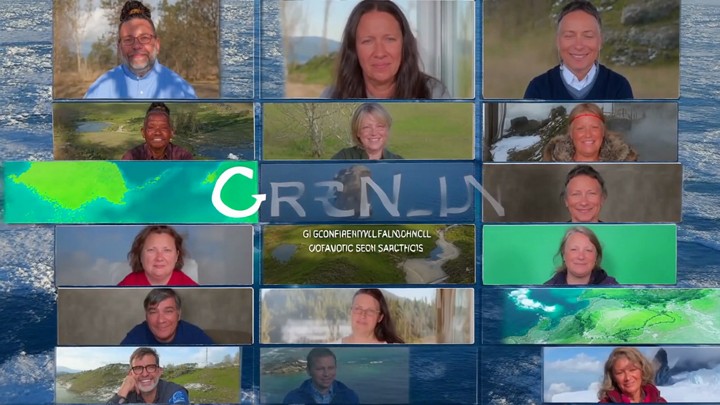}\hspace{-0.0037\textwidth}
\includegraphics[width=0.165\textwidth]{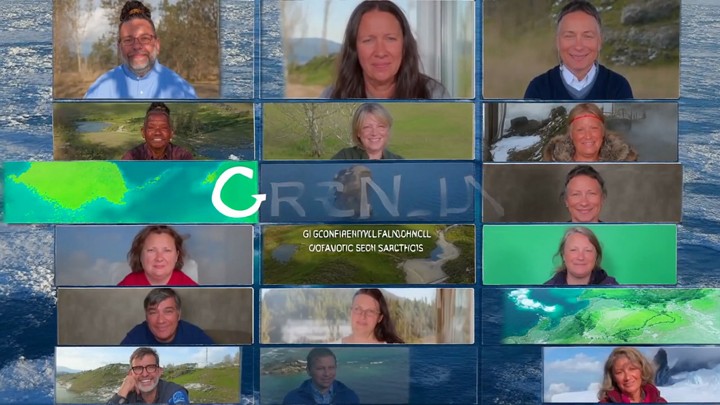}
\vspace{-0.5em}

\end{subfigure}

\vspace{0.2cm}

\begin{subfigure}{\textwidth}
\centering
\textbf{\large TurboDiffusion} ~~\textit{\large Latency: \bf \red{24s}}\\
\vspace{0.1cm}

\includegraphics[width=0.165\textwidth]{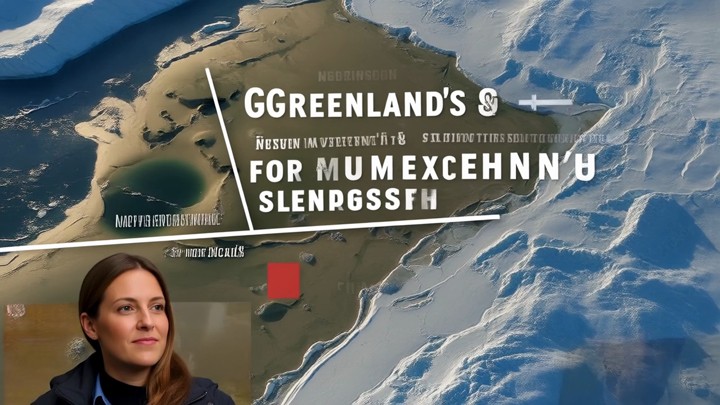}\hspace{-0.0037\textwidth}
\includegraphics[width=0.165\textwidth]{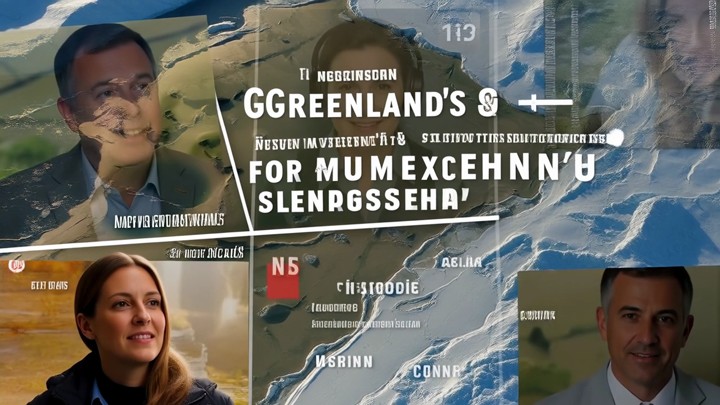}\hspace{-0.0037\textwidth}
\includegraphics[width=0.165\textwidth]{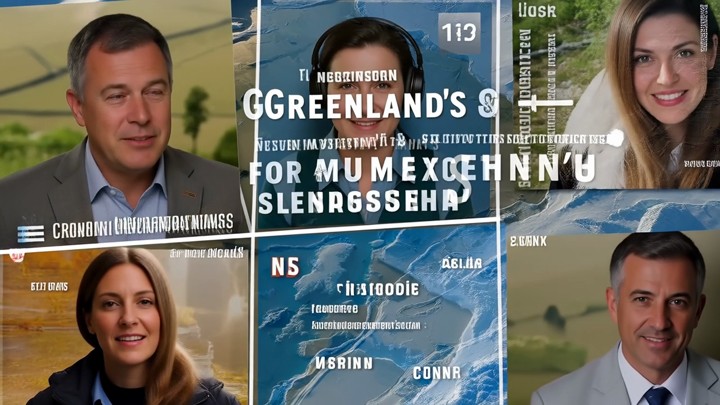}\hspace{-0.0037\textwidth}
\includegraphics[width=0.165\textwidth]{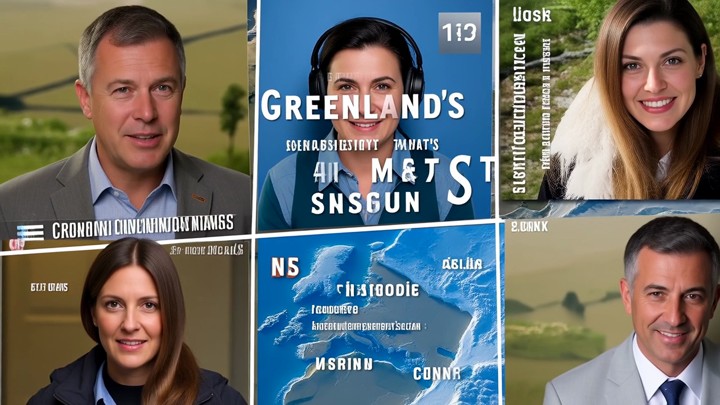}\hspace{-0.0037\textwidth}
\includegraphics[width=0.165\textwidth]{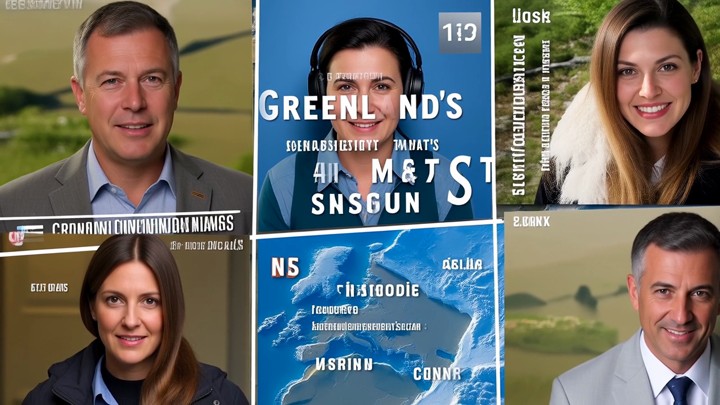}\hspace{-0.0037\textwidth}
\includegraphics[width=0.165\textwidth]{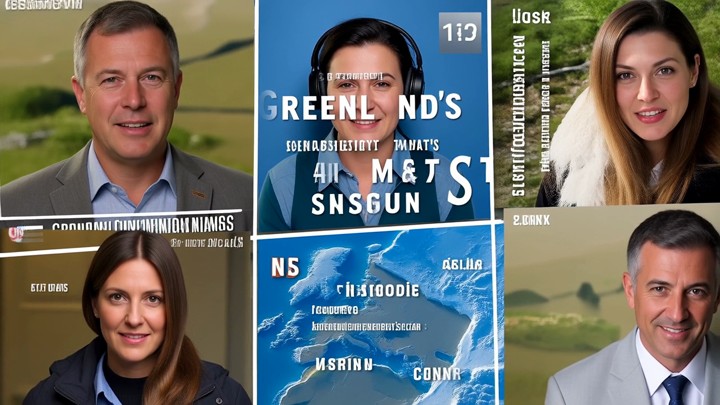}
\vspace{-0.5em}

\end{subfigure}

\vspace{-1em} \caption{5-second video generation on \texttt{Wan2.1-T2V-14B-720P} \textbf{\red{using a single RTX 5090}}.\\\textit{Prompt = "A documentary-style video focusing on \red{Greenland}'s geopolitical landscape. The scene opens with aerial shots of \red{Greenland's vast icy terrain}, emphasizing its strategic location. Transition to close-ups of Greenlandic Inuit people discussing their aspirations for increased autonomy and recognition, conveying a sense of determination and hope. Interspersed with these are interviews with political figures and experts, highlighting the complexities and challenges posed by foreign interests. \red{The background includes maps and satellite imagery illustrating Greenland's unique position.} The overall tone is informative and balanced, presenting both opportunities and risks. Wide and medium shots throughout, with occasional zoom-ins on key points during interviews."}}
\label{fig:comparison_14b_720p_video_2}
\end{figure}

\subsubsection{Wan2.1-T2V-14B-480P}

\begin{figure}[H]
\centering
\begin{subfigure}{\textwidth}
\centering
\textbf{\large Original} ~~\textit{\large Latency: 1676s}\\
\vspace{0.1cm}

\includegraphics[width=0.165\textwidth]{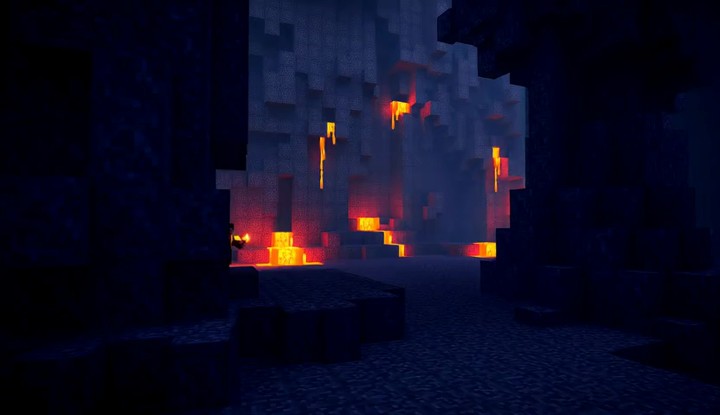}\hspace{-0.0037\textwidth}
\includegraphics[width=0.165\textwidth]{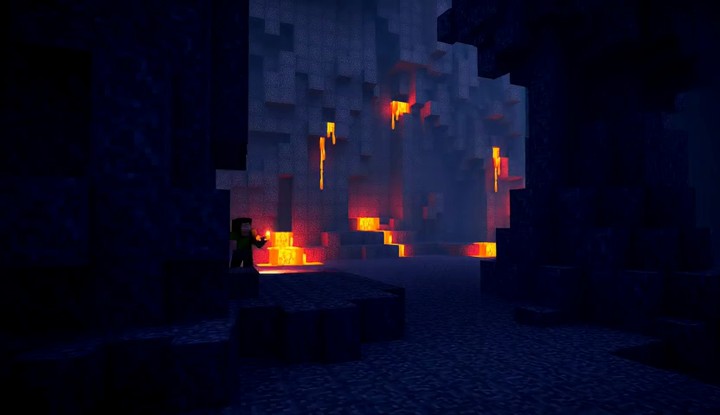}\hspace{-0.0037\textwidth}
\includegraphics[width=0.165\textwidth]{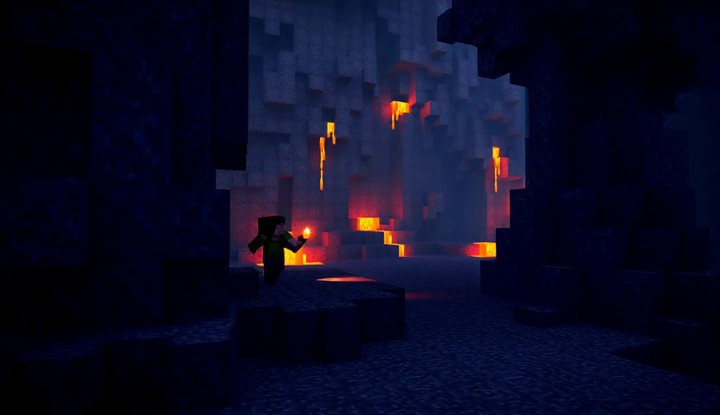}\hspace{-0.0037\textwidth}
\includegraphics[width=0.165\textwidth]{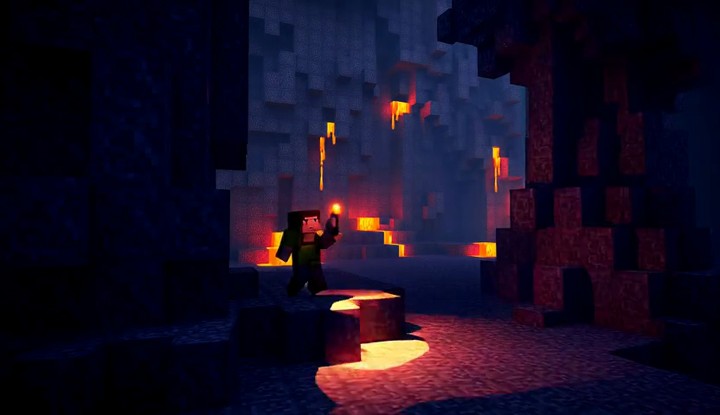}\hspace{-0.0037\textwidth}
\includegraphics[width=0.165\textwidth]{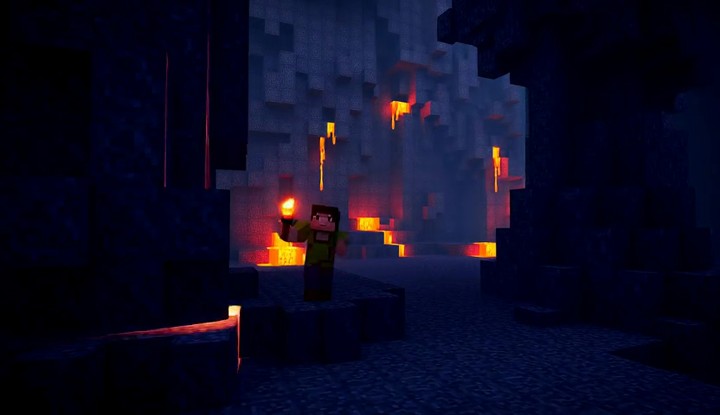}\hspace{-0.0037\textwidth}
\includegraphics[width=0.165\textwidth]{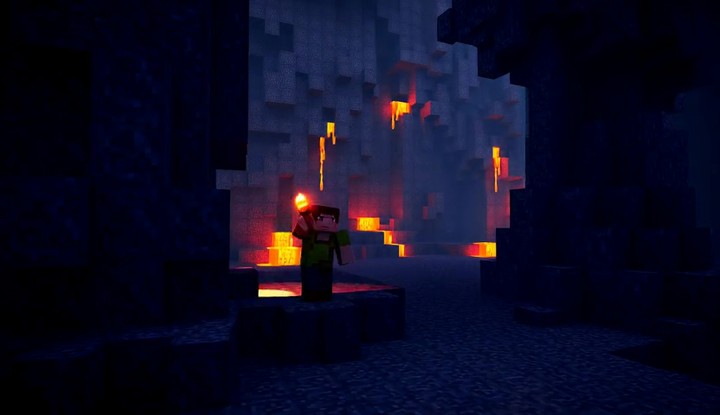}
\vspace{-0.5em}

\end{subfigure}

\vspace{0.2cm}

\begin{subfigure}{\textwidth}
\centering
\textbf{\large FastVideo} ~~\textit{\large Latency: 26.3s}\\
\vspace{0.1cm}

\includegraphics[width=0.165\textwidth]{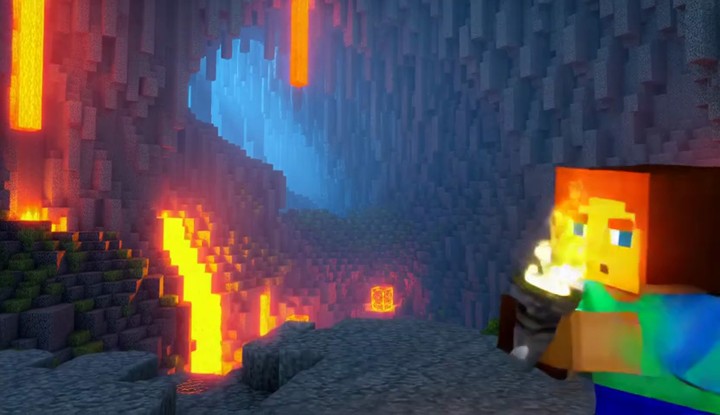}\hspace{-0.0037\textwidth}
\includegraphics[width=0.165\textwidth]{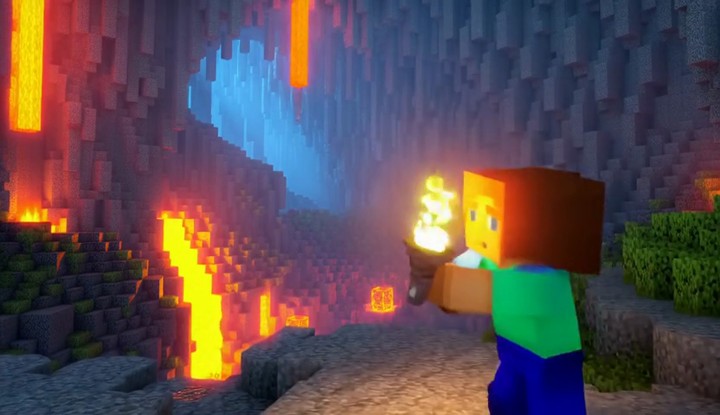}\hspace{-0.0037\textwidth}
\includegraphics[width=0.165\textwidth]{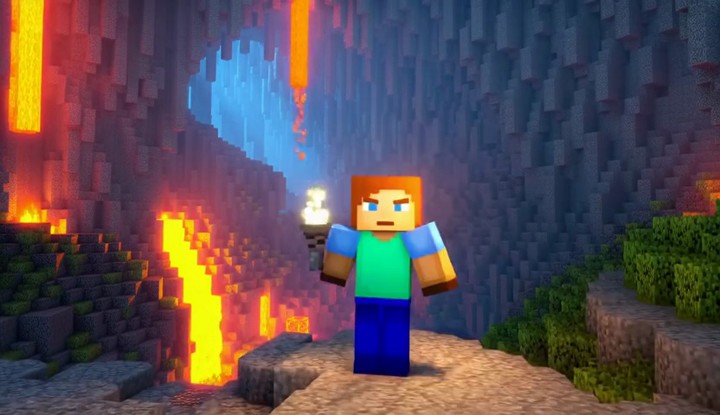}\hspace{-0.0037\textwidth}
\includegraphics[width=0.165\textwidth]{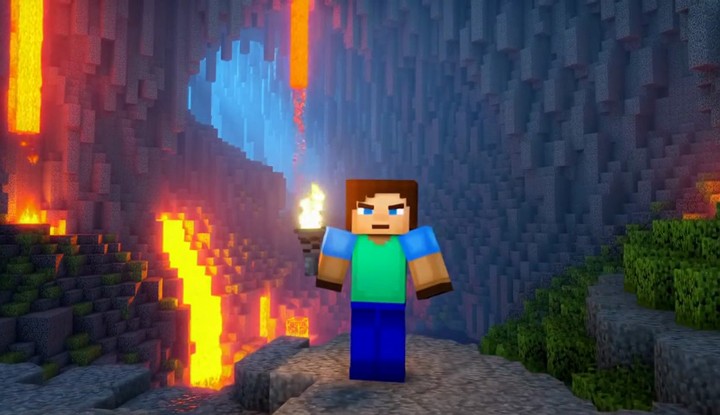}\hspace{-0.0037\textwidth}
\includegraphics[width=0.165\textwidth]{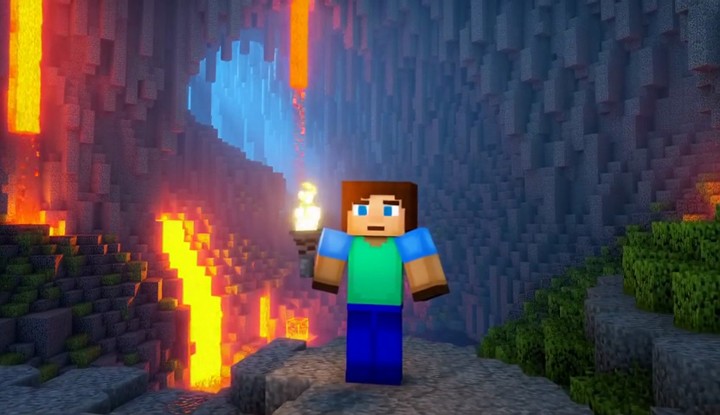}\hspace{-0.0037\textwidth}
\includegraphics[width=0.165\textwidth]{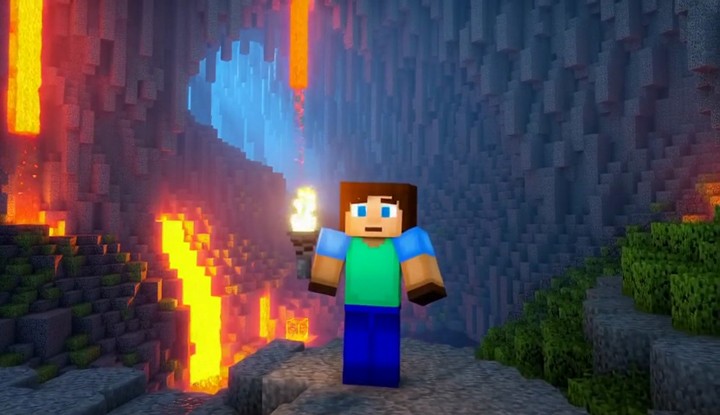}
\vspace{-0.5em}

\end{subfigure}

\vspace{0.2cm}

\begin{subfigure}{\textwidth}
\centering
\textbf{\large TurboDiffusion} ~~\textit{\large Latency: \bf \red{9.9s}}\\
\vspace{0.1cm}

\includegraphics[width=0.165\textwidth]{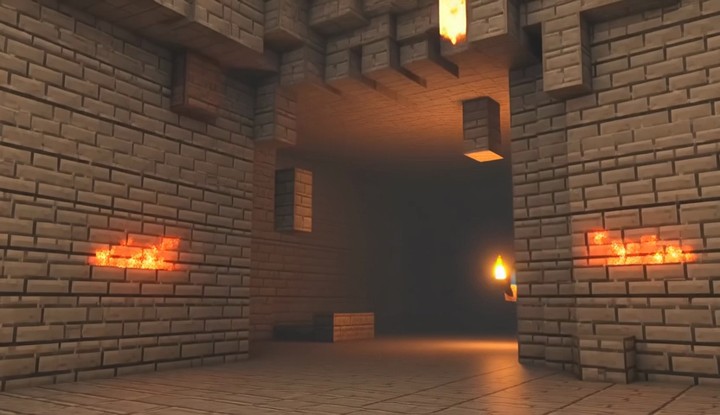}\hspace{-0.0037\textwidth}
\includegraphics[width=0.165\textwidth]{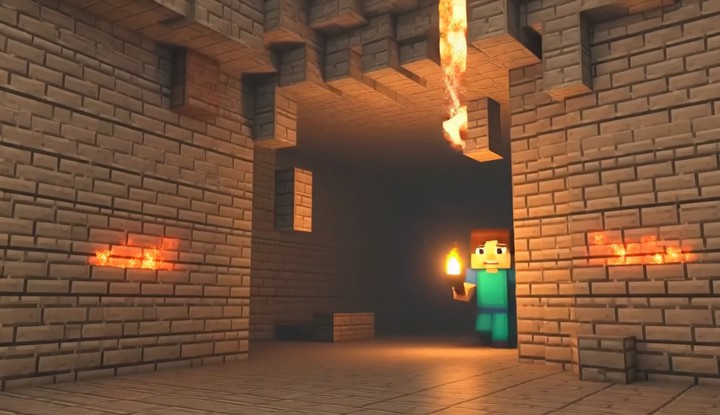}\hspace{-0.0037\textwidth}
\includegraphics[width=0.165\textwidth]{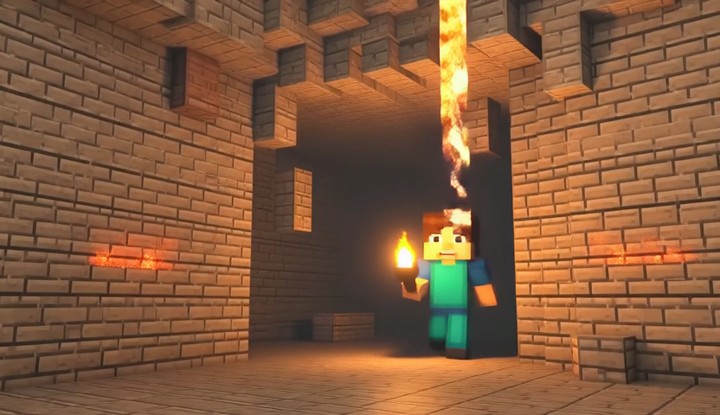}\hspace{-0.0037\textwidth}
\includegraphics[width=0.165\textwidth]{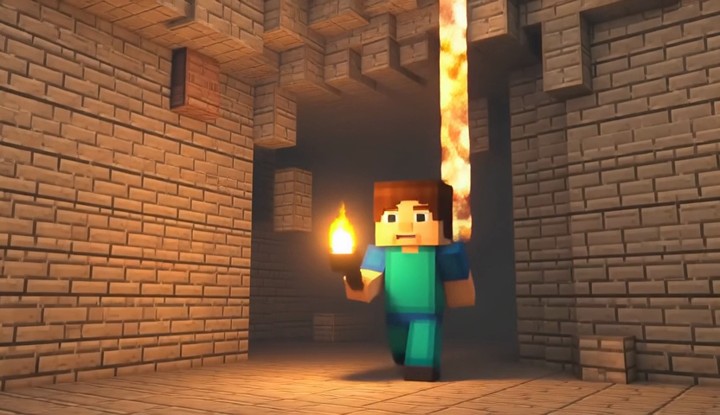}\hspace{-0.0037\textwidth}
\includegraphics[width=0.165\textwidth]{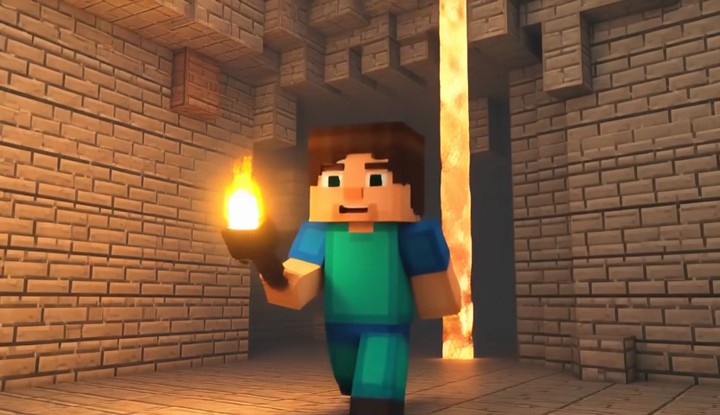}\hspace{-0.0037\textwidth}
\includegraphics[width=0.165\textwidth]{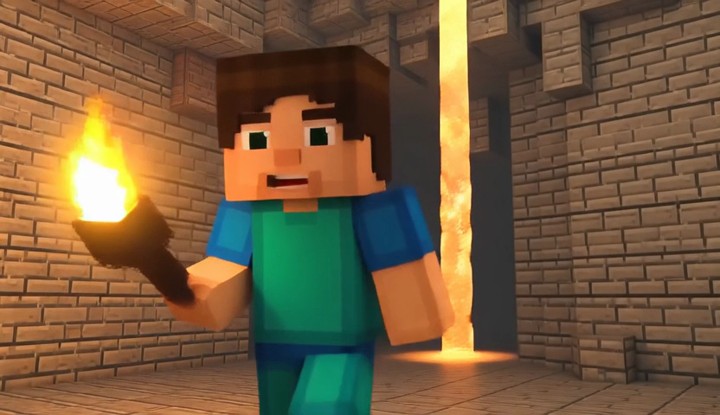}
\vspace{-0.5em}

\end{subfigure}

\vspace{-1em} \caption{5-second video generation on \texttt{Wan2.1-T2V-14B-480P} \textbf{\red{using a single RTX 5090}}.\\\textit{Prompt = "A Minecraft player character holding a torch enters a massive underground cave. The torchlight flickers against jagged stone walls, illuminating patches of iron and diamond ores embedded in the rock. Stalactites hang from the ceiling, lava flows in glowing streams nearby, and the faint sound of water dripping echoes through the cavern."}}
\label{fig:comparison_14b_480p_video_1}
\end{figure}

\begin{figure}[H]
\centering
\begin{subfigure}{\textwidth}
\centering
\textbf{\large Original} ~~\textit{\large Latency: 1676s}\\
\vspace{0.1cm}

\includegraphics[width=0.165\textwidth]{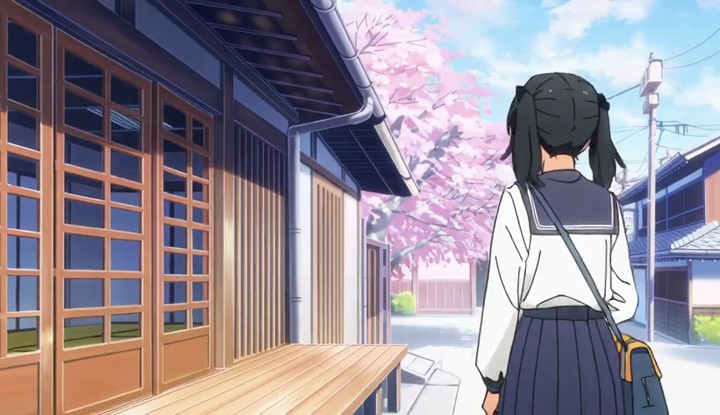}\hspace{-0.0037\textwidth}
\includegraphics[width=0.165\textwidth]{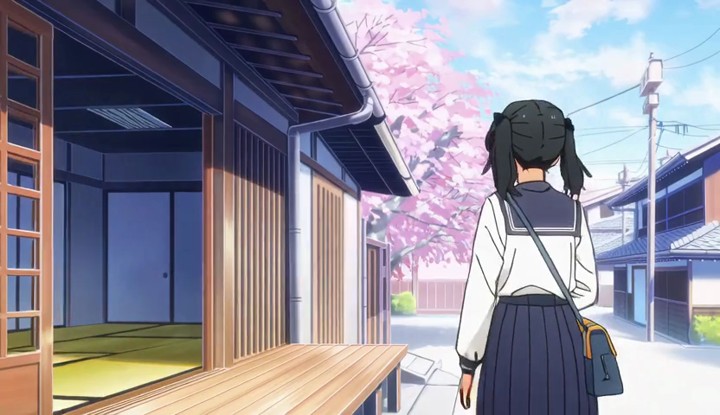}\hspace{-0.0037\textwidth}
\includegraphics[width=0.165\textwidth]{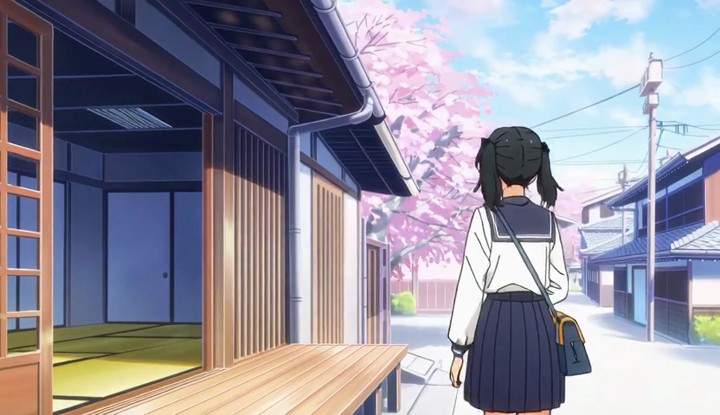}\hspace{-0.0037\textwidth}
\includegraphics[width=0.165\textwidth]{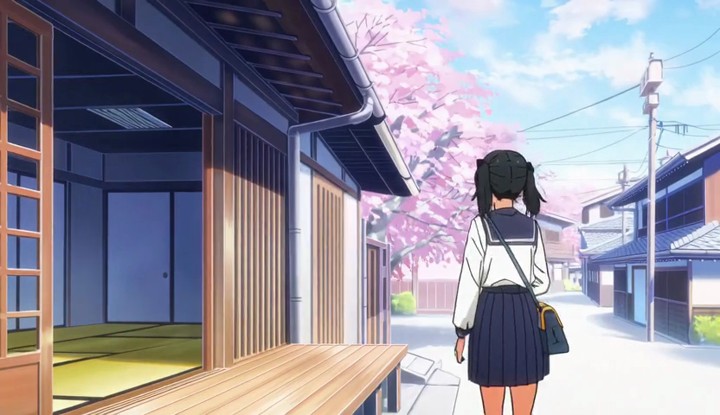}\hspace{-0.0037\textwidth}
\includegraphics[width=0.165\textwidth]{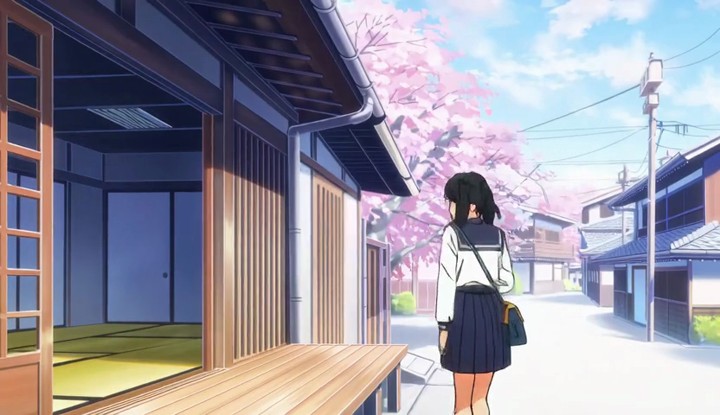}\hspace{-0.0037\textwidth}
\includegraphics[width=0.165\textwidth]{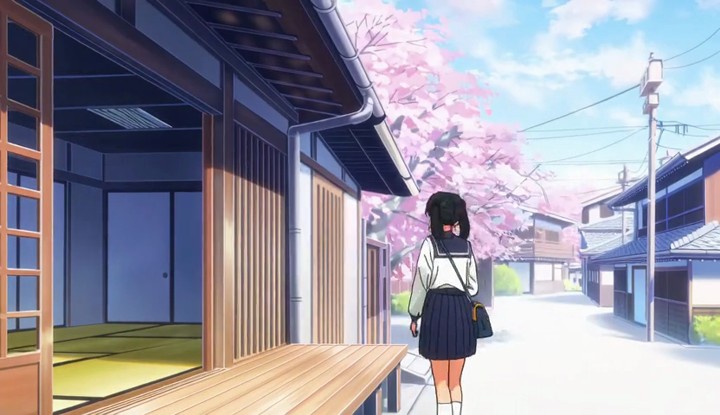}
\vspace{-0.5em}

\end{subfigure}

\vspace{0.2cm}

\begin{subfigure}{\textwidth}
\centering
\textbf{\large FastVideo} ~~\textit{\large Latency: 26.3s}\\
\vspace{0.1cm}

\includegraphics[width=0.165\textwidth]{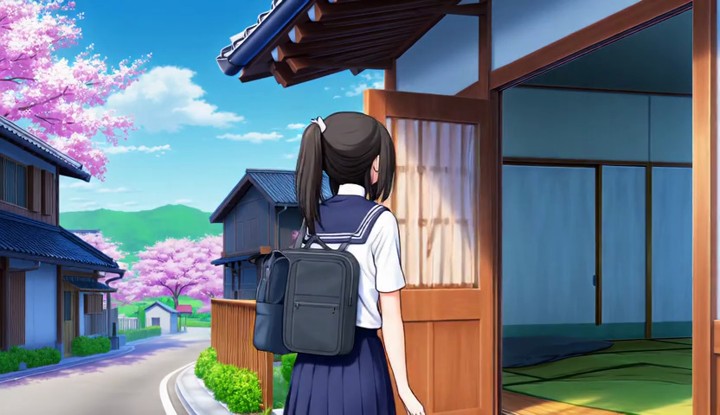}\hspace{-0.0037\textwidth}
\includegraphics[width=0.165\textwidth]{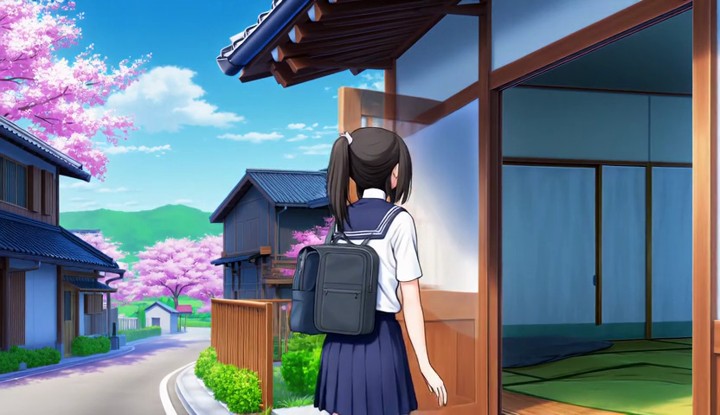}\hspace{-0.0037\textwidth}
\includegraphics[width=0.165\textwidth]{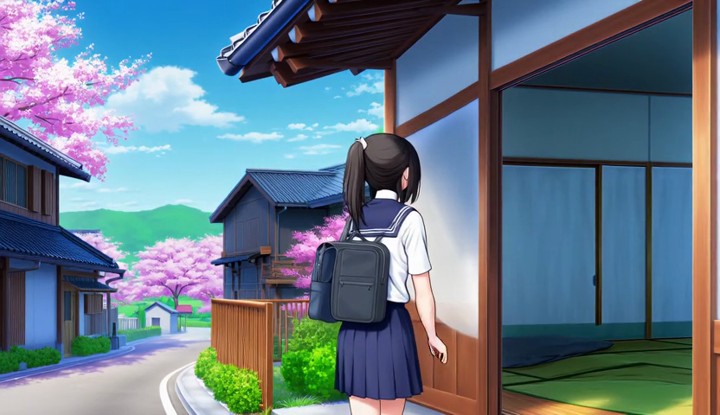}\hspace{-0.0037\textwidth}
\includegraphics[width=0.165\textwidth]{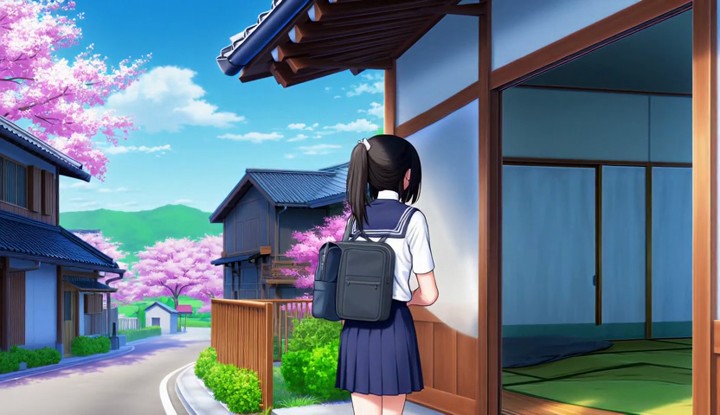}\hspace{-0.0037\textwidth}
\includegraphics[width=0.165\textwidth]{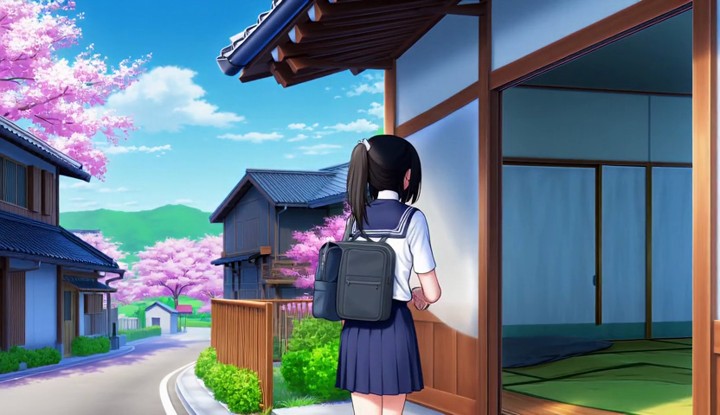}\hspace{-0.0037\textwidth}
\includegraphics[width=0.165\textwidth]{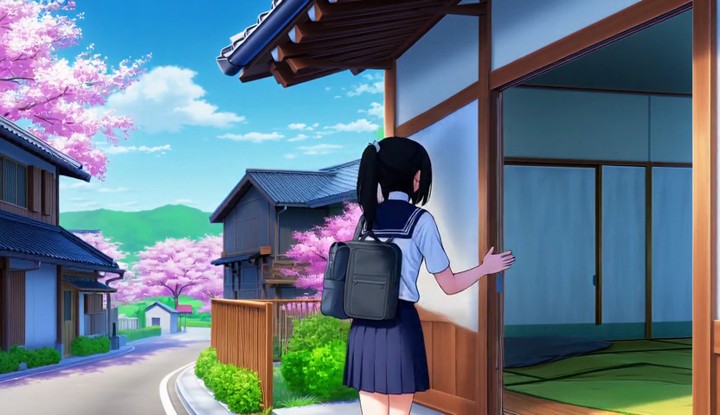}
\vspace{-0.5em}

\end{subfigure}

\vspace{0.2cm}

\begin{subfigure}{\textwidth}
\centering
\textbf{\large TurboDiffusion} ~~\textit{\large Latency: \bf \red{9.9s}}\\
\vspace{0.1cm}

\includegraphics[width=0.165\textwidth]{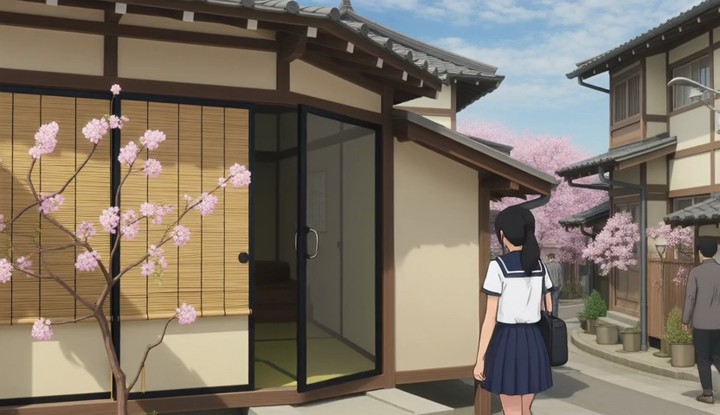}\hspace{-0.0037\textwidth}
\includegraphics[width=0.165\textwidth]{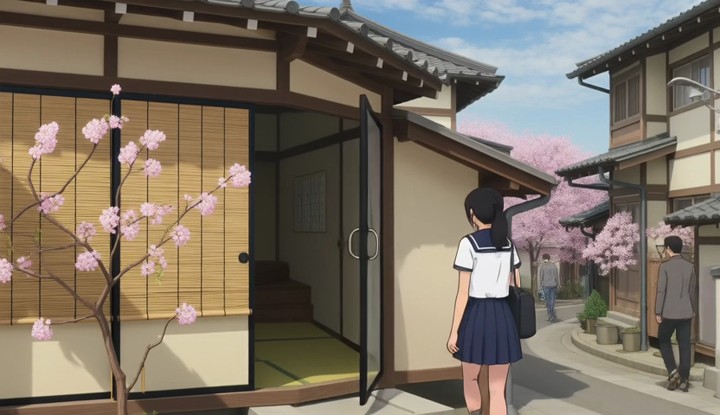}\hspace{-0.0037\textwidth}
\includegraphics[width=0.165\textwidth]{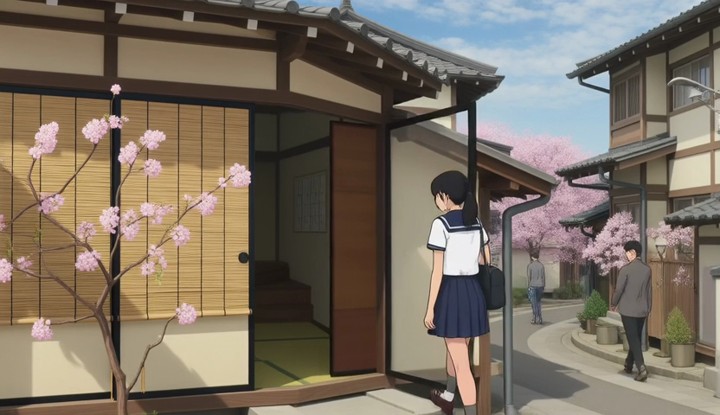}\hspace{-0.0037\textwidth}
\includegraphics[width=0.165\textwidth]{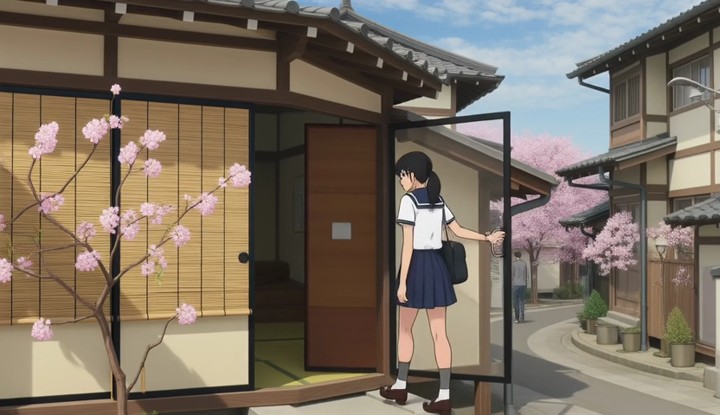}\hspace{-0.0037\textwidth}
\includegraphics[width=0.165\textwidth]{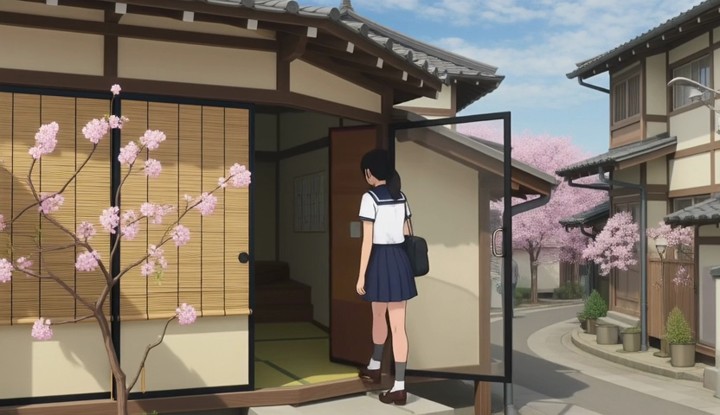}\hspace{-0.0037\textwidth}
\includegraphics[width=0.165\textwidth]{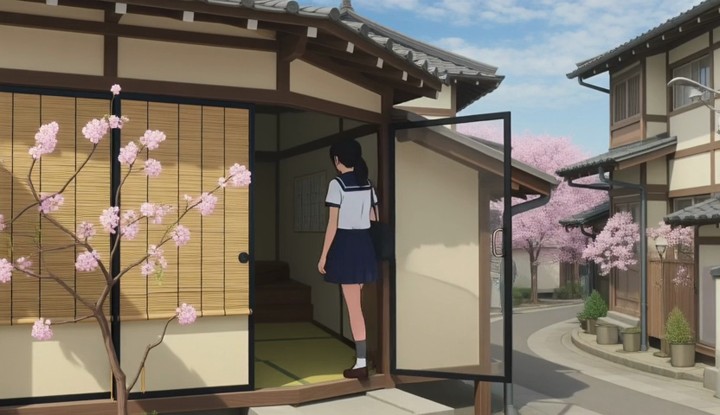}
\vspace{-0.5em}

\end{subfigure}

\vspace{-1em} \caption{5-second video generation on \texttt{Wan2.1-T2V-14B-480P} \textbf{\red{using a single RTX 5090}}.\\\textit{Prompt = "Anime-style illustration, a young Japanese woman walking into a traditional house in a quaint Japanese town. She has shoulder-length black hair tied in a ponytail, wears a simple school uniform consisting of a white blouse and a navy blue pleated skirt, and carries a small backpack. The house features a typical Japanese entrance with a wooden door and a tatami mat inside visible through the partially opened door. Surrounding the house are traditional Japanese buildings and narrow streets lined with cherry blossom trees in full bloom. The sky is clear with soft pastel colors. \red{Medium shot focusing on the woman as she steps into the house, showing her interaction with the traditional door.}"}}
\label{fig:comparison_14b_480p_video_6}
\end{figure}

\section{Conclusion and Future Work}

We present \ours, a video generation acceleration framework that achieves $100\text{–}200\times$ end-to-end diffusion speedup with negligible quality degradation. \ours combines low-bit attention (SageAttention), Sparse-Linear Attention (SLA), step distillation via rCM, and W8A8 quantization, together with several additional engineering optimizations. Experiments on \texttt{Wan2.2-I2V-A14B-720P}, \texttt{Wan2.1-T2V-1.3B-480P}, \texttt{Wan2.1-T2V-14B-720P}, and \texttt{Wan2.1-T2V-14B-480P} demonstrate that \ours reduces the generation time of a single video to <1 minute on a single RTX 5090 GPU, making high-quality video generation substantially more efficient and practical.

For future work, we plan to extend this framework to support more video generation paradigms, such as autoregressive video diffusion.

\section*{Acknowledgments}

We thank Jia Wei, Pengle Zhang, and Kaichao You from Tsinghua University, and Haocheng Xi, Shuo Yang, Roger Wang, and Simon Mo from Sky Lab, UC Berkeley, for their help.

\bibliographystyle{unsrt}
\bibliography{main}

\end{document}